\documentclass[journal]{IEEEtran}

\usepackage{dsfont}
\ifCLASSINFOpdf
\else
\fi
\usepackage{mathtools,booktabs,tabu,graphicx,mystyle,epstopdf,url,rotating}

\usepackage{algorithm}
\usepackage{cite}
\usepackage{algorithmic}
\usepackage{fancyref}

\newcommand{\vertiii}[1]{{\left\vert\kern-0.25ex\left\vert\kern-0.25ex\left\vert #1 
		\right\vert\kern-0.25ex\right\vert\kern-0.25ex\right\vert}}

\def\MCMRF{SGB-Lasso\,}

\newcommand{\mestar}{\meas^\star}

\begin{document}

\title{An off-the-grid approach to multi-compartment magnetic resonance fingerprinting}

\author{Mohammad~Golbabaee~and~Clarice Poon\vspace{-.4cm}
\thanks{MG and CP are with the Computer Science and Mathematics departments, University of Bath, United Kingdom: \{m.golbabaee, cmhsp20\}@bath.ac.uk. 
}}

\newcommand{\m}{m}
\newcommand{\mo}{m_0}
\renewcommand{\t}{t}
\newcommand{\x}{x}
\renewcommand{\a}{a}

\newcommand{\D}{\mathcal{D}}

\newcommand{\RED}[1]{{\color{red} #1}}
\newcommand{\meas}{\mathbf{m}}
\newcommand{\tnorm}[1]{\VERT #1 \VERT }

\maketitle

\begin{abstract}

%
%

We propose a novel numerical approach to separate multiple tissue compartments in image voxels and to estimate quantitatively their nuclear magnetic resonance (NMR) properties and mixture fractions, 
given 
magnetic resonance fingerprinting (MRF) measurements. 
The number of tissues, their types or quantitative properties are not a-priori known, but the image is assumed to be composed of \emph{sparse} compartments with linearly mixed Bloch magnetisation responses within voxels. 
Fine-grid discretisation of the multi-dimensional NMR properties creates large and highly coherent MRF dictionaries 
that can challenge scalability and precision of the numerical methods for (discrete) sparse approximation. 
To overcome these issues, we propose an \emph{off-the-grid} 
approach equipped with an extended notion of the \emph{sparse group lasso} regularisation for sparse approximation using \emph{continuous} (non-discretised) Bloch response models. 
%
Further, the nonlinear and  non-analytical Bloch responses are approximated by a neural network, enabling efficient back-propagation of the gradients through the proposed algorithm.  
Tested on simulated and \emph{in-vivo} healthy brain MRF data, we demonstrate effectiveness of the proposed scheme compared to the baseline multi-compartment MRF methods. 
\end{abstract}


%
\IEEEpeerreviewmaketitle


\newcommand{\vc}{\mathrm{Vec}}
\newcommand{\rc}{{\vc}^{-1}}
\newcommand{\Tspace}{\Tt}
\newcommand{\btheta}{{\Theta}}
\section{Introduction}

\label{sec:intro}
Multi-Compartment (MC) effects, also known as partial volume effects, occur when more than one tissue type
occupies a single image voxel. This effect is common in medical images e.g. in MRI~\cite{Brainweb-collins, PVMRI-tohka, PVMRI-van,PVMRI-cuadra,PVMRI-manjon}
due to the images' finite spatial resolution. 
Estimating MC effects is crucial for obtaining accurate segmentation and estimation of the tissue volumes and their contents e.g. for studies related to  several brain disorders such as Alzheimer’s disease, multiple sclerosis, or Schizophrenia
~\cite{jack2004comparison, llado2012segmentation,shenton2001review}. 

Magnetic resonance fingerprinting (MRF)~\cite{MRF, FISP} is an emerging technology that enables quantitative mapping of several  tissues' physical properties in short and clinically feasible scan times. The MC effects also occur in MRF and if left unmodelled they can produce false and blurry mappings e.g. at the tissue boundaries~\cite{BayesianPVMRF,MCMRF-tang}. On the other hand, a multi‐component MRF (MC-MRF) analysis can potentially help to distinguish more tissues more reliably than e.g. contrast-weighted MRI, 
because multiple quantitative tissue parameters are measured in (naturally) co-registered mapped images. 

Current numerical methods for solving MC-MRF rely on sparse approximation of linear mixtures in a large discretised dictionary of simulated Bloch responses (fingerprints). 
This can lead to several numerical issues: 
the accuracy of the estimated compartments depends on fine-grid discretisation of the tissue properties that amounts to exponentially-large dictionaries in multi-parametric MRF applications and creates storage bottleneck. Fine-grid discretisation also increases the coherence of dictionary atoms (fingerprints) which fundamentally limits the precision of sparse approximation. Further, the precision of fast (first-order) shrinkage solvers such as FISTA~\cite{FISTA} was observed inadequate (despite long iterations) to tackle over-redundancy of the MRF dictionary~\cite{BayesianPVMRF, MCMRF-tang}, demanding instead higher precision and more computation-involved sparse solvers.

%
%

In this study we propose the first numerical approach for off-the-grid MC-MRF estimation. Our approach adopts continuous (non-discretised) models of Bloch responses for sparse approximation, in order to address the (storage) non-scalability issue of the state-of-the-art as well as offering more precise MC estimation with continuous mapping of tissues'  multi-dimensional quantitative properties. 
Our approach consists of a new regularisation, backed with theoretical analysis, for promoting a hybrid notion of group-and-pixel sparsity in the MC-MRF solutions. This regularisation extends the sparse group lasso model of~\cite{SPGLasso} to non-discretised dictionaries. A new shrinkage algorithm based on Frank-Wolfe iterations is proposed to solve this regularisation, and the nonlinear and non-analytical Bloch magnetic responses are approximated by a neural network to enable efficient back-propagation of the gradients through this algorithm.

\section{related works}
MC analysis based on 
regularised  statistical models were proposed for the conventional contrast-weighted MRI~\cite{PVMRI-tohka, PVMRI-van,PVMRI-cuadra,PVMRI-manjon}, 
as well as quantitative MRI relaxometry for splitting multi-exponential Bloch responses e.g. for an individual~\cite{MCMRF-whittall, MCMRF-T1} 
or joint mapping~\cite{MCMRF-west, andica2018automated, MCMRF-despot} 
of the T1 and T2 relaxations times, for measuring 
the white/grey matters volumes, 
the myelin content in the brain 
or the cerebral blood flow~\cite{bouhrara2017rapid, andica2019gray, kim2017quantification, MCMRF-ASL2}, 
besides other applications. 
MC-MRF analysis were also proposed based on multi-parametric tissue mappings e.g. T1/T2 relaxometry~\cite{BayesianPVMRF, MCMRF-tang, SPIJN, PVMRF}. 
MC-MRF baselines use 
large dictionaries of simulated Bloch responses over fine-grid discretisation of the multi-dimensional parameter space. 
Methods~\cite{BayesianPVMRF}~and~\cite{MCMRF-tang, SPIJN}  
iteratively solve (nonconvex) reweighted $\ell_2$ and reweighted nonnegative (squared) $\ell_1$ regularisations, respectively, for promoting sparsity in the mixture weights. 
High-precision 
(e.g. second-order~\cite{MCMRF-tang}) optimisation methods were employed to solve each reweighted iteration, besides empirical dictionary pruning heuristics~\cite{SPIJN,BayesianPVMRF}  to reduce long runtimes. 
Method~\cite{SPIJN} promotes group-sparsity to cluster the whole image into sparse compartments which is shown (in-vitro and in-vivo) more accurate and easy-to-visualise than the pixel-wise sparsity~\cite{BayesianPVMRF, MCMRF-tang}. Other group-sparse models~\cite{PVMRF, MCMRF-Yves} based on k-means clustering assume a-priori known number of compartments and that most image pixels are 100\% pure (single-compartment). For instance~\cite{PVMRF} applies k-means on single-compartment mappings obtained via MRF dictionary matching~\cite{SVDMRF} to cluster tissues' T1/T2s, followed by an additional (combinatorially large) dictionary matching step for estimating mixture fractions. 
Our approach differs from the MC-MRF baselines: for improved numerical precision and scalability, we use continuous Bloch response models 
for sparse approximation rather than discretised (gridded) dictionaries. 
Single-compartment continuous MRF mapping were proposed in~\cite{sbrizzi2017dictionary, dong2019quantitative} without sparsity penalties. 
This work however proposes a different numerical approach 
rooted in the off-the-grid sparse approximation literature~\cite{candes2014towards,duval2015exact} 
for promoting sparsity in the MC-MRF solutions. 

\newcommand{\fv}{f_{\btheta^\star}}
\newcommand{\qv}{Q_{\btheta^\star}}

\newcommand{\gv}{g_{\btheta^\star}}
\newcommand{\etav}{\eta_{\btheta^\star}}

\newcommand{\proj}{{\mathcal{P}}}

\section{Multi-compartment Quantitative MRI model}

We are interested in quantifying tissues NMR properties given a Time-Series of Magnetisation Images (TSMI) $X=[x_1,x_2,\ldots,x_v]\in\RR^{T \times v}$ with $v$ voxels and $T$ timeframes. 
Per-voxel magnetisation signal $x\in \RR^T$ resulted by multiple compartments follows a linear mixture model~\cite{BayesianPVMRF}:
\eql{\label{eq:model1}
x = \sum_s c_s \phi(\theta_s), \vspace{-.1cm}
}
where $ c_s\geq 0$ 
are the mixture weights and $\phi:\Tspace \rightarrow \RR^T$ is the \emph{Bloch} magnetisation response model that maps $d$-dimensional NMR properties $\theta_s \in \Tspace \subseteq \RR^d$ to time signals. For instance, $d=2$ for T1 and T2 relaxometry with constraint $\Tspace=\{\mathrm{T1}>\mathrm{T2}\}$. We can extend this formulation to all TSMI voxels:
\eql{\label{eq:model2}
X = \sum_s \phi(\theta_s) C_s^\top
}
where $C_s\in\RR_+^v$  correspond to the $\theta_s$-dependent mixture weights for all voxels i.e. the mixture map images.

For factorising the TSMI and estimating tissue properties and mixture maps $\ens{\theta_s, C_s}$,  the MC-MRF baselines proposed to quantise 
the space of NMR properties by a dense grid $\Theta\subset \Tt$, and form an exponentially-large MRF dictionary $$
D_\Theta \eqdef\pa{\phi(\theta)}_{ \theta \in \Theta}
$$ of $n=\Card(\Theta)=O(\mathrm{e}^d)$ fingerprints ($\mathrm e>1$). This has lead to the following discretised or gridded formulation:
\eql{\label{eq:model4}
X = D_\Theta C^\top,
}
where one aims to find a sparse (or a column-sparse) representation for the very large-sized matrix $C\in\RR_+^{v\times n}$ containing the mixture maps for every possible fingerprint. Besides sparsity, the non-negativity of the mixture maps was observed an important constraint for rejecting spurious solutions to this ill-posed inverse problem~\cite{PVMRF, SPIJN}.

\section{Sparse Group Beurling Lasso (SGB-Lasso)}
We introduce our off-the-grid approach by rewriting \eqref{eq:model2} as a \emph{continuous linear model} over the space of vector-valued measures. 
Our measure $\meas \in \Mm(\Tspace;\RR^v)$
\eql{\label{eq:meas}
\meas = \sum_s C^\top_s \delta(\theta-\theta_s)
}
is characterised jointly by the mixture maps $C_s\in \RR_{+}^{v}$ and the tissue properties 
through the weighted sum of Dirac's mass function $\delta(\theta-\theta_s)$ at positions $\theta_s$. Now by defining a linear operator $\Phi$ over the space of vector-valued measure $\Mm(\Tspace;\RR^v)$ as
\eql{\label{eq:model3}
\Phi \meas \eqdef \int \phi(\theta) \mathrm{d}\meas(\theta),
}
equation \eqref{eq:model2} becomes $X = \Phi \meas$. We therefore consider the infinite dimensional linear inverse problem of estimating the measure $\meas$ from the TSMI $X$. We regularise the underlying measure~\eqref{eq:meas} to be sparse i.e. composed of few compartments in the sum and \emph{additionally}, each mixture map $C_s$ itself to be a sparse image. For this we propose to solve \eqref{eq:model3} by the following variational formulation, coined as the Sparse Group Beurling Lasso:
\eql{\label{eq:spglasso}
\argmin_{\meas\in\Mm(\Tspace;\RR_{+}^v)} \frac12 \norm{X - \Phi \meas}_{F}^2 + \alpha \norm{\meas}_{\beta} \tag{SGB-Lasso}
}
Where $\|.\|_F$ denotes the matrix Frobenius norm. The regularisation parameter $\alpha > 0$ balances between a TSMI fidelity term and a sparsity-promoting norm. 
We define the Sparse Group Total Variation (SGTV) norm $\|.\|_{\beta}$ though the following composition of variation norms for vector-valued measures: 
Given a vector-valued measure $\meas\in \Mm(\Tspace; \Yy)$ taking values in a Banach space $\Yy$ endowed with a norm $\norm{\cdot}_\Yy$, its variation 
is defined to be:
$$
\abs{\meas}_Y(\Tspace)  \eqdef \sup_{\Aa_i} \sum_{i=1}^n \norm{\meas(\Aa_i)}_Y
$$
where the supremum is over all partitions $\ens{\Aa_i}_{i=1}^n$ of $\Tspace$. In our setting, $\Yy = \RR^v_+$ and by endowing this space with norms $\norm{y}_1 \eqdef \sum_{i=1}^v \abs{y_i}$ or $\norm{y}_2 \eqdef \sqrt{\sum_{i=1}^v \abs{y_i}^2}$ and respectively, their variations $\abs{\meas}_1(\Tspace)$ and $\abs{\meas}_2(\Tspace)$, we can define the SGTV norm as:
$$
\norm{\meas}_{\beta} \eqdef (1-\beta ) \abs{\meas}_1(\Tspace) + \beta \sqrt{v} \abs{\meas}_2(\Tspace).
$$
With a slight abuse of notation, we also define a matrix norm: 
$$
\norm{C}_\beta \eqdef \sum_{s} (1-\beta) \norm{C_s}_{1} + \beta \sqrt{v} \norm{C_s}_{2},
$$
where $C_s$ denotes the $s^\text{th}$ column of a matrix $C$ containing multi-compartment mixture maps. 
With respect to~\eqref{eq:meas}, we have the identities $\abs{\meas}_1 = \sum_s \norm{C_s}_1$, $\abs{\meas}_2 = \sum_s\norm{C_s}_2$, and therefore $\norm{\meas}_\beta = \norm{C}_\beta$.


Note that $\norm{\meas}_{\beta}$ is a continuous extension of the  sparse group regularisation introduced in~\cite{SPGLasso} for discretised dictionaries. 
The term  $\abs{\meas}_{2}(\Tspace)$ promotes \emph{group sparsity} in a sense that few compartments  ($\theta_s$)  should 
contribute to approximate the entire TSMI across all voxels \eqref{eq:model2}.  The term $\abs{\meas}_{1}(\Tspace)$ in addition promotes \emph{spatial sparsity} within the mixture map/image ($C_s$) of each contributing compartment i.e., a voxel should usually receive contributions from fewer compartments than those composing the entire TSMI, which is related to having some level (not necessarily 100\%) of voxel purity that helps cross-compartment spatial separability of the mixture maps.
The parameter $0<\beta<1$ provides a degree of freedom to balance between these forms of sparsity. 

\section{Identifiability of the mixtures via \MCMRF}
We begin with some definitions and notations. 
The \emph{adjoint} of $\Phi$ operator is denoted by $\Phi^*:\RR^{T\times v}\to \Cc(\Tspace;\RR^v)$ which maps matrices into the space of  vector-valued continuous functions. Given a (e.g. TSMI) matrix $X \in\RR^{T\times v}$ with columns $x_i$:
\eql{ \label{eq:adj}
\Phi^* X \eqdef (\eta_i)_{i=1}^v \quad\text{where} \quad \eta_i(\theta) =
 \dotp{\phi(\theta)}{x_i}.
 }
Further, $\btheta = (\theta_s)_{s\in[k]} \in (\RR^d)^k\subset \Tspace$ denotes a set of $k$ vectors of multi-parametric NMR properties. For a $k\in \NN$, $[k]\eqdef \ens{1,\ldots,k}$. For a vector $v\in\RR^d$, $v^\top$ is its transpose, and $v_I$ is a sub-vector of $v$ with entries indexed by a set $I\subseteq [d]$, and $\mathbf{0}_d$ denotes an all-zero vector of length $d$. 
Given $q\in\RR^v$, let $q_+ \eqdef \max(q,0)$ denote its positive part.

Suppose now that $X = \Phi \mestar + W$ where $\mestar = \sum_{s=1}^k C^\star_s \delta(\theta-\theta^\star_s)$ is a $k$-sparse ($k$-compartment) measure  with the ground-truth mixture weights $C^\star\in \RR^{k\times v}_+$ and tissue properties $\Theta^\star = \pa{\theta^\star_s}_{s\in [k]} \subset \Tt$, and $W\in \RR^{T\times v}$ is some bounded additive noise $\norm{W}_F \leq \epsilon$.  In this section we describe a condition (certificate) that guarantees stable estimation of $\mestar$ i.e., stable demixing of $\{\Theta^\star, C^\star\}$, given a noisy TSMI $X$. 



%
%

\subsection{Theoretical guarantees}
We first define a matrix used for evaluating our certificate
$$
\qv \eqdef \mathop{\argmin}_{Q\in \RR^{T\times v}} \enscond{\norm{Q}_F}{  f \eqdef \frac{ (\Phi^* Q +\beta-1)}{\sqrt{v}\beta} \in \Kk  }
$$
where $\Kk\subset  \Cc(\Tspace; \RR^v) $ is
$$
\Kk \eqdef \enscond{f}{\forall s\in [k], \; [f(\theta_s^\star) ]_{I_s}=  \frac{ [C_s^\star]_{I_s}}{\norm{C_s^\star}_2}, \nabla \norm{f(\theta_{s}^\star)_{I_s}}_2^2 = \mathbf{0}_d  },
$$
and $I_s \eqdef \Supp(C_s^\star)$ is the support/position of the non-zero elements of $C_s^\star$. Note that the constraint set $\Kk$ consists of $\sum_s \abs{I_s} + kd$ linear equations and hence, $\qv$ can be computed by solving a linear system. In particular, we can write (the vectorized version of) $\qv$ as the least squares solution to $$ \Gamma^\top Q = 
\begin{pmatrix}
\pa{\beta\sqrt{v} + (1-\beta) \frac{[C_s]_{I_s}}{\norm{C_s}}}_{s\in[k]}\\
\mathbf{0}_{kd}
\end{pmatrix}
$$
for some full-rank matrix $\Gamma \in \RR^{Tv \times (\sum_{s=1}^k \abs{I_s} +kd)}$.
See \cite{spglasso2020} for the precise formulation of this linear system.

\begin{defn}\label{def:precert}
Define $\fv\in\Cc(\Theta;\RR^v)$ and $\gv\in\Cc(\Theta;\RR)$ by 
\begin{equation}
\begin{aligned}
\fv(\theta) &\eqdef \frac{\left(\Phi^* \qv +\beta-1\right)_+ }{\beta\sqrt{v}} ,
\\
 \gv(\theta)  &\eqdef \norm{ \fv(\theta) }^2_2.
\end{aligned}
\end{equation}
By definition $\gv(\theta) =1$, $\forall \theta\in\btheta^\star$. We call $\gv$ a nondegenerate certificate if it satisfies:
\begin{enumerate}
\item (non-saturation) $\gv(\theta) <1$, $\forall \theta\notin\btheta^\star$.
\item (curvature)   $\nabla^2 \gv(\theta)$ is negative definite, $\forall \theta\in\btheta^\star$.
\end{enumerate}


\end{defn}


The following result, whose proof can be found in \cite{spglasso2020}, shows that these conditions are sufficient for stable demixing. Furthermore,  following similar arguments in \cite[Proposition 8]{duval2015exact} one could expect that these conditions are tight so that if $\gv(\theta)>1$, then demixing is necessarily unstable:  

\begin{thm}\label{thm:stability1}
Let $X = \Phi \mestar + W$ where $\mestar = \sum_{s=1}^k C^\star_s \delta(\theta-\theta^\star_s)$ and $\norm{W}_F \leq \epsilon$. If $\Gamma$ is full-rank and the certificate $\gv$ associated to $\mestar$ is nondegenerate, then there exists a constant $\gamma >0$ such that by setting $\alpha\leq \gamma$ and $\epsilon/\alpha \leq \gamma$,  \eqref{eq:spglasso} recovers a unique solution of the form $ \sum_{s=1}^k  C_s \delta(\theta-\theta_s)$ with bounded errors $\norm{ C - C^\star}_F =\Oo( \epsilon)$ and $\norm{ \btheta - \btheta^\star}_F =\Oo( \epsilon)$.
\end{thm}

\subsection{Numerical illustration of the certificate}
\begin{figure}
\centering
\begin{tabular}{ccc}
\includegraphics[width=.32\linewidth]{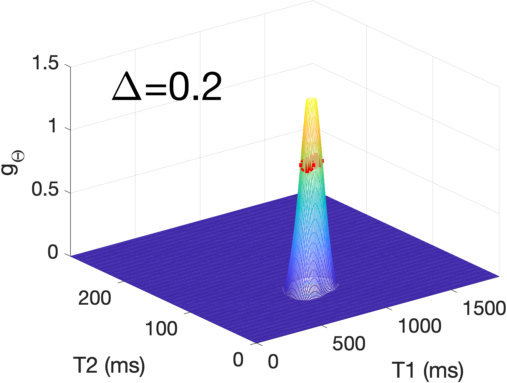}
\includegraphics[width=.32\linewidth]{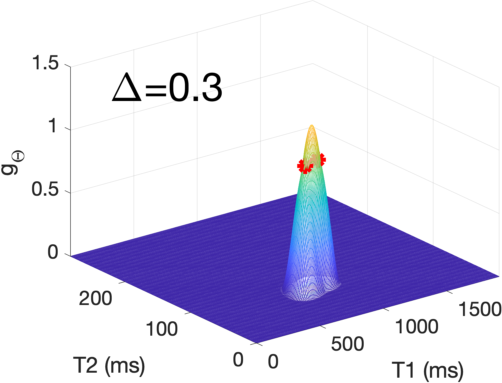}
\includegraphics[width=.32\linewidth]{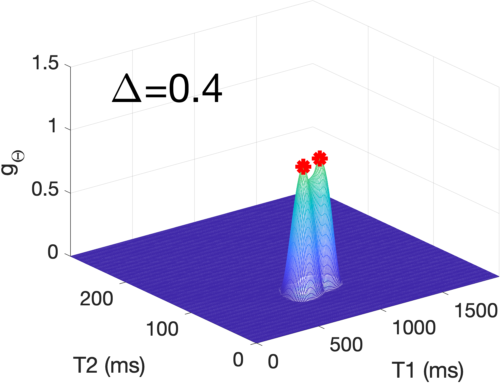}
\\
\includegraphics[width=.32\linewidth]{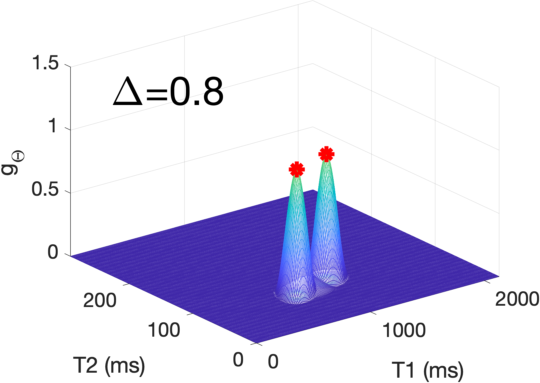}
\includegraphics[width=.32\linewidth]{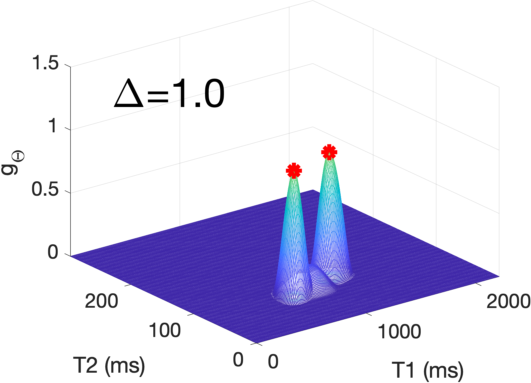}
\includegraphics[width=.32\linewidth]{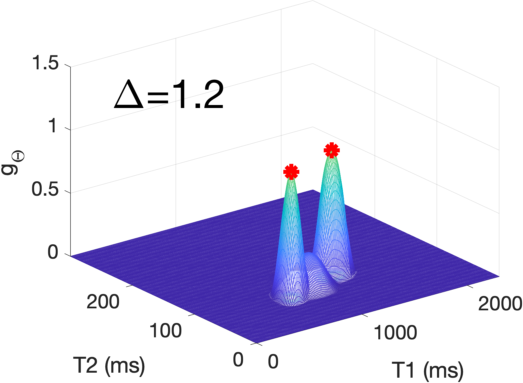}
\end{tabular}
\caption{\footnotesize{The certificate $\gv(\theta)$ values across the $\theta=(\mathrm T1,\mathrm T2)$ plane, shown for six  two-compartment mixture examples where the compartments' T1/T2 values (red points) are separated by distance $\Delta$. A certificates is nondegenerate if $\gv<1$ elsewhere than red points, which holds for cases $\Delta\geq 0.4$.}
 \label{fig:separation}}
\end{figure}

Illustrations below adopt a neural network embedded Bloch response model $\phi(\cdot)$ used in our experiments for encoding $\theta = (\mathrm{T1},\mathrm{T2})$ NMR relaxation properties (see sections~\ref{sec:algo-dnn} and \ref{sec:expe} for model details). 
We examine the certificate nondegeneracy in Theorem~\ref{thm:stability1}, particularly the non-saturation condition, on several two-compartment examples  $\mestar = C_1^{\star\top} \delta(\theta-\theta^{\star}_1) + C_2^{\star\top} \delta(\theta-\theta^{\star}_2)$ in order to highlight 
the following points:

\paragraph{Stable demixing requires a minimum separation between the compartments' T1/T2 values}
Consider simulating mixtures with  $\theta^\star_1 = (784,  77)$ ms and $\theta^\star_2 = \theta_1 + \Delta \hat \theta$ where $ \hat \theta = (1216, 96) - \theta^\star_1$, associated to  the two left-most mixture maps of the brain phantom in Figure~\ref{fig:brainweb_GTmaps}.
For $\beta = 10^{-3}$, we plot the certificate $\gv$ for different values of $\Delta$ in Figure \ref{fig:separation}. We can observe that $\gv$ becomes degenerate when $\Delta$ is too small (for the plots shown, we have nondegeneracy when $\Delta\geq 0.4$). This means, there is a minimum separation distance below which demixing becomes unstable.

\begin{figure}
\centering
\includegraphics[width=0.38\textwidth]{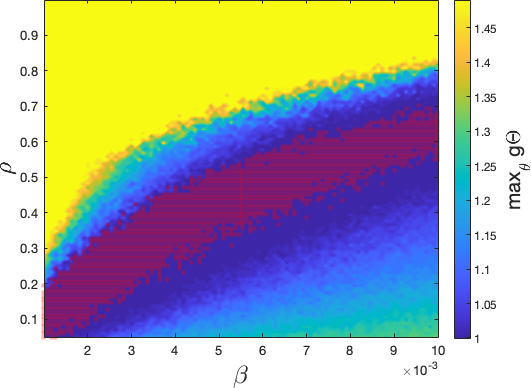}
\caption{\footnotesize{The colour-coded values of  $\max_{\theta}{\gv(\theta)}$ for different \MCMRF parameters $\beta $ and mixture map sparsity levels $\rho$. The red region satisfies  nondegeneracy $\max_{\theta}{\gv(\theta)}\leq 1$, and guarantees stable demixing by \MCMRF. We see that for small levels of sparsity, $\beta$ should be chosen small.}  
\label{fig:nondegen_graph} }
\end{figure}

\paragraph{Choice of $\beta$ depends on the sparsity of the mixture maps}
We simulate several mixtures where $\theta^\star_1 = (719,  80)$, $\theta^\star_2 = ( 1190, 98)$ ms, and the mixture weights $C^\star_1,C^\star_2\in \RR^{v=1024}$ are randomly generated from i.i.d. normal distribution with at most $\lceil\rho v\rceil$ nonzero entries. In figure \ref{fig:nondegen_graph}, we display the values of parameter $\beta$ and sparsity level ratios $\rho \in [0,1]$ for which the nondegeneracy condition is satisfied.  As expected, $\beta$ should be taken smaller for sparser mixture maps/weights. 

%
%


\begin{figure}

\centering
\includegraphics[width=0.12\textwidth]{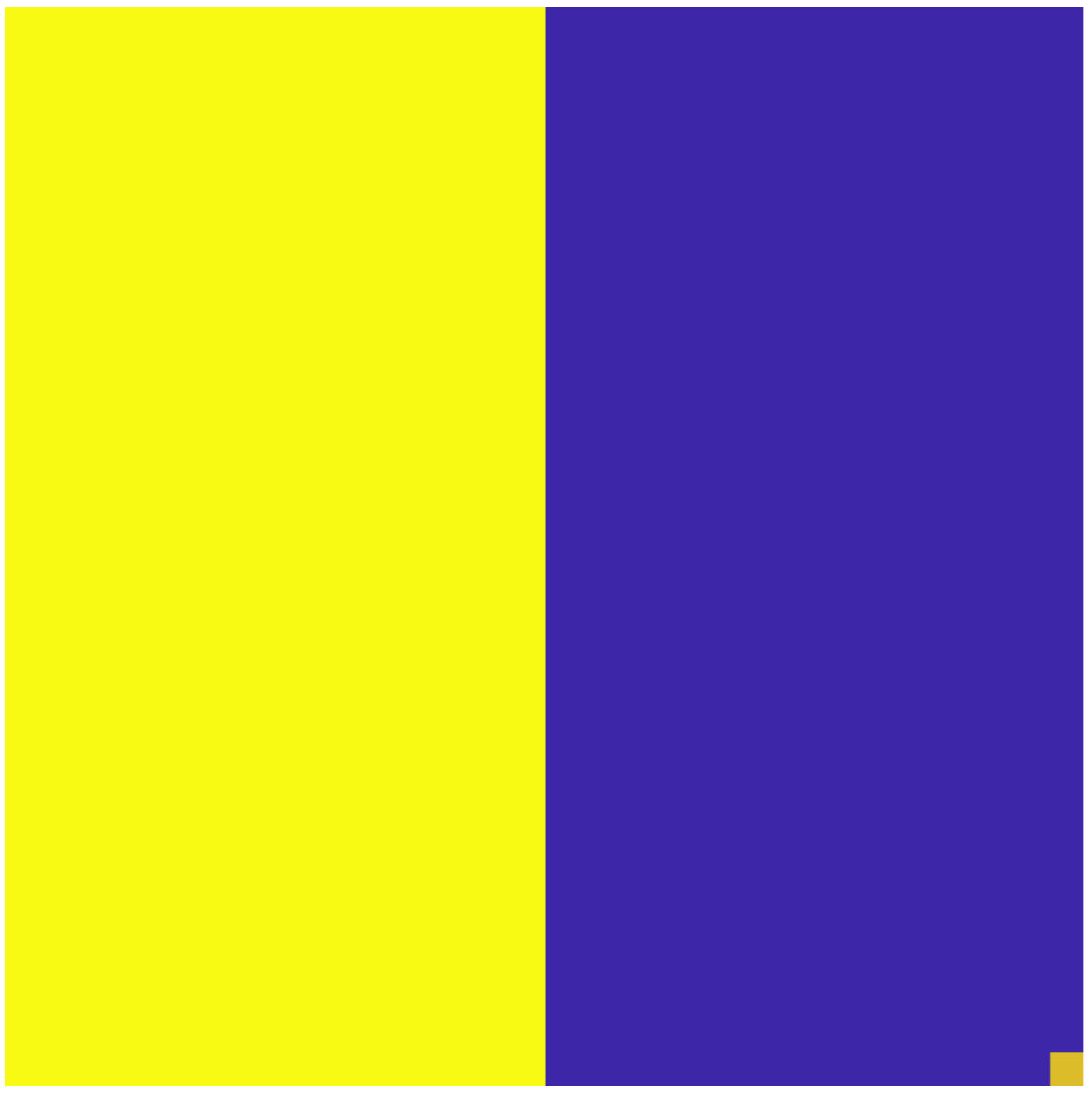}\qquad
\includegraphics[width=0.12\textwidth]{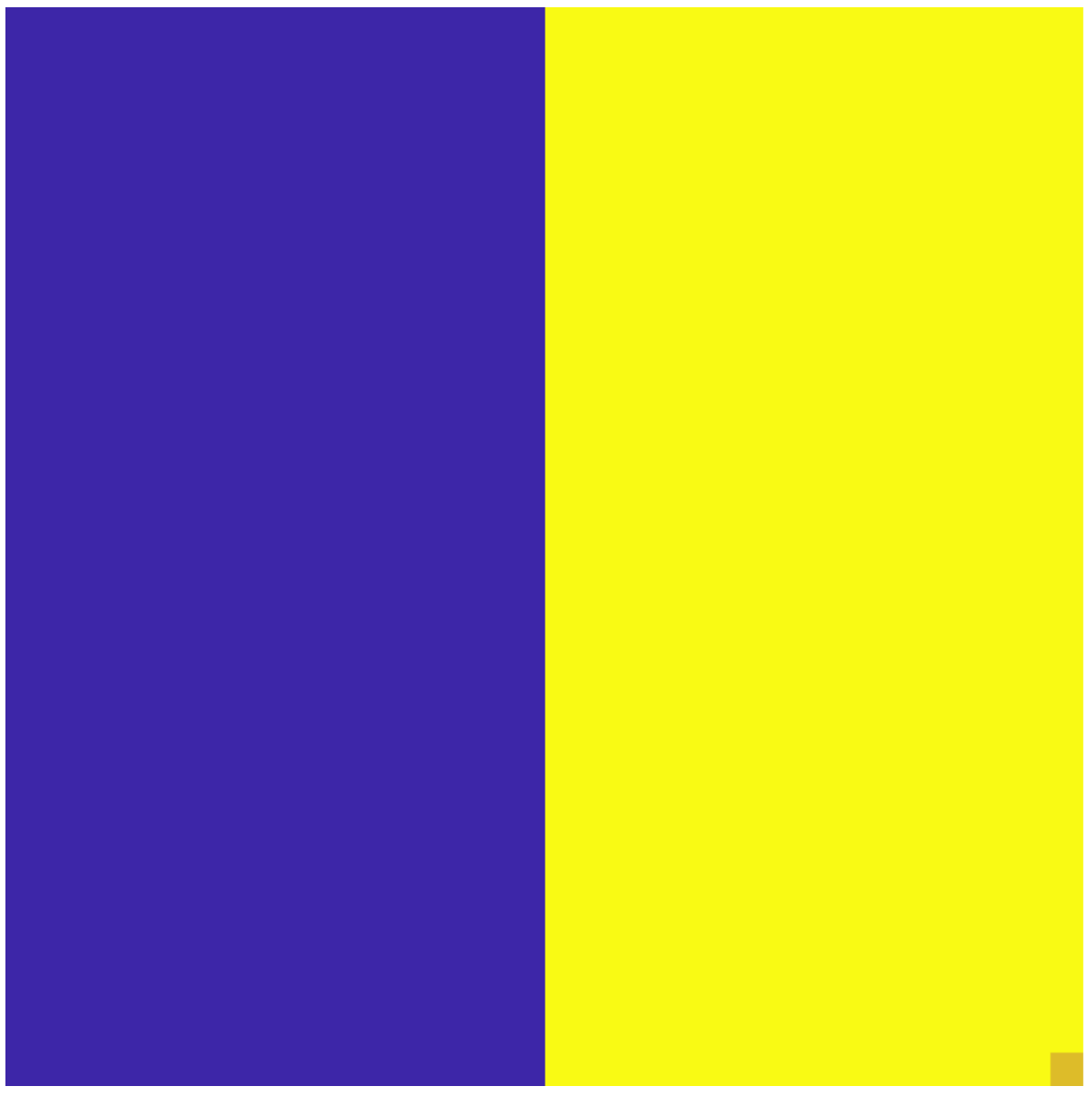}
\begin{turn}{90} \includegraphics[height=.42cm]{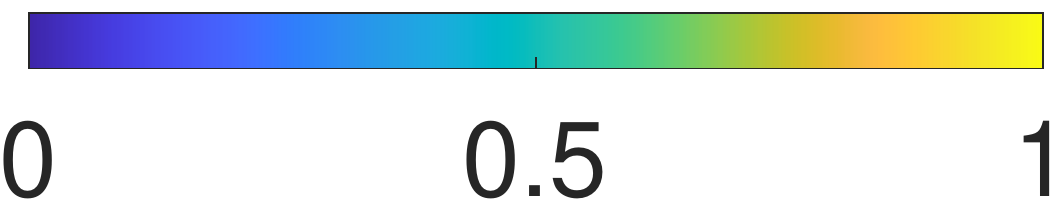}\end{turn}
\\
\footnotesize{(a) Mixture with two \emph{almost} pure component maps, except one corner pixel.}
\\
\includegraphics[width=0.24\textwidth]{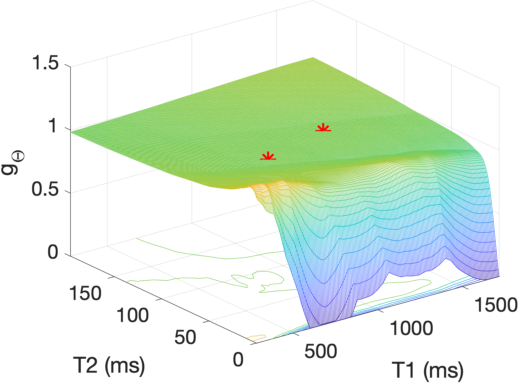}
\includegraphics[width=0.24\textwidth]{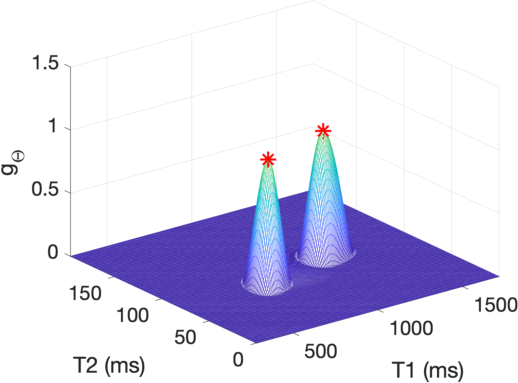}
\\
\footnotesize{ (b) Two certificates corresponding to (left) $\beta=0$ which is degenerate as $\max_\theta \gv(\theta) = 1.2$, and (right) $\beta=10^{-3}$ which is nondegenerate.} 
\caption{
\footnotesize{(a) A two-compartment mixture and (b) its certificate $\gv(\theta)$ values for $\beta=0$ and $\beta=10^{-3}$ plotted across the $\theta=(\mathrm T1,\mathrm T2)$ plane.} 
 \label{fig:comp_maps_for_lasso} }
\end{figure}

\paragraph{Group sparsity is essential} 
We finally highlight the issue that, the case of $\beta =0$ (which relates to pure pixel sparsity framework e.g.~\cite{MCMRF-tang}) is numerically unstable, and taking $\beta>0$ is necessary to ensure stable mixture separation. Consider a mixture example 
with $\theta^\star_1 = (784,  77)$, $\theta^\star_2 = ( 1216, 96)$ ms, and almost pure (except one pixel) mixtures maps $C^\star_1,C^\star_2$ in Figure \ref{fig:comp_maps_for_lasso}(a). 
We compute and plot in Figure \ref{fig:comp_maps_for_lasso}(b) two certificates $\gv$ for when \MCMRF would use $\beta=0$ and $\beta = 10^{-3}$. 
We observe a discontinuity in the \MCMRF's behaviour as per removing or adding the non-smooth group-sparsity penalty through changing $\beta=0$  to  $\beta> 0$.\footnote{When $\beta=0$ and we regularise with only $\abs{\meas}_1$, the problem becomes separable, equivalent to solving  per voxel $i\in [v]$, 
$\min_{\meas \in \Mm(\Tspace,\RR_+)} \frac12 \norm{X_i - \Phi \meas}^2 + \alpha \abs{\meas}_1$.  
This case was studied in \cite{duval2015exact}: 
per-voxel support is $J_i \eqdef \enscond{s\in [k]}{(C^\star_s)_i \neq 0}$ and  the corresponding certificate is 
$g^{(i)} = \mathop{\argmin}_{g = \Phi^* q}\enscond{\norm{q}_2}{   \forall s \in J_i, \; g(\theta^\star_s) = 1, \\ \nabla g(\theta^\star_s) = 0}.$
This certificate is nondegenerate (and hence leads to stable recovery) if for all voxels $i$ the non-saturation and curvature conditions hold i.e. $\forall \theta\not\in \ens{\theta^\star_s}_{s\in J_i}$ $g^{(i)}(\theta) <1$, and  $\nabla^2 g^{(i)}(\theta^\star_s) \neq 0$  is negative definite, $\forall s\in J_i$.
The total number of constraints in the definition of each $g^{(i)}$ is $\abs{J_i}(d+1)$ and hence, the total number of constraints across all $v$ certificates is $(d+1)\sum_s \abs{I_s}$. There is therefore a jump in the number of constraints when
 we switch from $\beta>0$ to $\beta =0$ (due to the non-smoothness of the regularisation terms), leading to a discontinuous behaviour of \MCMRF for these cases.} 
As expected $\gv(\theta^\star_1)=\gv(\theta^\star_2)=1$ in both cases (red points), but for certificate $\gv(\theta)$ to be nondegenerate its value should not exceed 1 at $\theta\notin\{\theta^\star_1,\theta^\star_2\}$. In this case, switching from $\beta = 0$ to $\beta>0$ results in a drastic difference: the certificate is nondegenerate only when $\beta>0$.

\section{Algorithms}
\subsection{TSMI reconstruction} Prior to mixture separation, the TSMI is computed from MRF's undersampled k-space measurements using the LRTV algorithm~\cite[Eq(10)]{MRFLRTV}. This method is dictionary-matching-free and does not limit the reconstruction accuracy to the finite resolution of a discretised MRF dictionary. By exploiting the TSMI's spatiotemporal structures, LRTV can efficiency remove aliasing artefacts 
and is shown more accurate than the Fourier backprojection scheme SVD-MRF~\cite{SVDMRF} used by most MC-MRF baselines. 
LRTV (also SVD-MRF) exploits a low-rank subspace \emph{dimensionality reduction} for accelerated reconstructions. For many MRF sequences, including FISP~\cite{FISP} in our experiments, the Bloch responses and TSMIs can be factorised to a low-dimensional subspace: 
\eq{
\phi(.) \approx VV^T\phi(.)\,  \qandq X \approx VV^TX, 
}
where $V$ is a $T\times \tau$ tall matrix representing the subspace of dimension $\tau\ll T$. 
The raw TSMIs (complex-valued) are then \emph{phase-corrected} and mapped to real-valued images before being fed to the mixture separation step. This is particularly important for imposing the non-negativity constraint in~\eqref{eq:model2} and \MCMRF. For FISP sequence with constant TE per-voxel signal evolution has a constant complex-valued phase~\cite{SPIJN,FISP}. This  phase can be estimated from the first (principal) image component of the dimension-reduced TSMI and removed from the image~\cite{AIRMRF, coverblip_iop, MRFLRTV}.

\subsection{Solving \MCMRF by Frank-Wolfe iterations}

To solve \eqref{eq:spglasso}, we apply  Algorithm \ref{alg:FW} which generalises the sliding Frank-Wolfe algorithm of \cite{denoyelle2019sliding} beyond $\beta = 0$, where the authors guaranteed convergence in finite number of iterations $t$ at the rate $\Oo(1/t)$, under nondegeneracy assumptions. Algorithm~\ref{alg:FW} recovers one compartment per iteration (steps 5 and 6) and undergoes refinement steps 7 and 8 to update the current solution $\{\btheta^{t}, C^{t}\}$.
Below we highlight these key steps (for more details about the algorithm derivation see the supplementary section~\ref{sec:deriv}): 

\subsubsection{Nonconvex steps}
Algorithm~\ref{alg:FW} (lines 5 and 8) optimises nonconvex objectives (due to the nonlinearity of the Bloch responses) with respect to the NMR parameters $\theta$ or a set of them $\btheta=\pa{\theta_s}$. We use the L-BFGS algorithm
\footnote{We used implementation \url{https://github.com/stephenbeckr/L-BFGS-B-C}} to locally solve these steps, 
constrained that the mixture maps $C$ are non-negative and the parameters e.g. $\theta = (\mathrm{T1},\mathrm{T2})$ satisfy $ \Tspace=\{{\mathrm{T1}} > {\mathrm{T2}}\}$.
 Line 5 is initialised by a coarse \emph{grid-search} over a small subset of discretised NMR parameters in $\Tspace$ i.e. a coarsely gridded dictionary used for enumerating $\eta(\theta)$ on this subset. Step 8 is initialised by the outcomes of lines 6 and 7  for the NMR parameters and mixture weights, correspondingly. 

\subsubsection{Convex step}  Algorithm~\ref{alg:FW} recovers one compartment $\theta^t$ per iteration $t$, 
simulates its Bloch response and adds it to a dictionary $D_{\btheta^{t+1/2}} = D_{\btheta^{t}}\cup \phi(\theta^t)$. In line 7, this $D$ is used for solving discrete sparse group lasso~\cite{SPGLasso} by the (restarted) fast shrinkage algorithm FISTA \cite{o2015adaptive} and the shrinkage operator:
\eql{
\begin{aligned}
\text{Prox}_{\norm{\cdot}_\beta}(C) &=
\argmin_{z\geq 0} \frac12 \norm{Z-C}_F^2 + \norm{Z}_\beta
\\
&=
\mathsf G_{\beta \sqrt{v} }\circ \mathsf S_{ (1-\beta)}(C).
\end{aligned}
}
Where for a $\lambda>0$ and matrix $C$, operator $\mathsf S_\lambda(C) = (C-\lambda)_+$ is the element-wise positive soft thresholding, and $\mathsf G_{\lambda}(C)_s = \frac{C_s}{\norm{C_s}_2}\pa{\norm{C_s}_2 - \lambda}_+$ is the group soft thresholding for every column $s$ of $C$.
%
%
Note that $D$ has $t+1$ columns at each Frank-Wolfe iteration. In practice/our numerical results, Algorithm~\ref{alg:FW} is convergent before a maximum 30 iterations. Hence $D$ is small, leading to fast and accurate FISTA updates without facing numerical issues of dictionary over-redundancy. 

\subsubsection{Neural network approximation}
\label{sec:algo-dnn}
Solving lines 5 and 8   requires computing derivatives of the Bloch responses with respect to the NMR parameters i.e. the jacobian matrix $\partial \phi(\theta)/\partial \theta$. 
While this can be analytically computed for simple forms of Bloch responses (e.g. MRF sequences in~\cite{dong2019quantitative,sbrizzi2017dictionary}, or the exponential models in 
classical quantitative MRI), such approach would not extent to more general non-analytical response models e.g. those like FISP simulated by the Extended Phase Graph (EPG) formalism~\cite{EPGWeigel}.
To circumvent this issue, we leverage on universal approximation property of neural networks~\cite{cybenko1989,lecun2015deep} that enables embedding complicated functions in conveniently differentiable surrogates via back-propagation mechanism. 
We train a neural network $\widetilde{\phi}: \Tspace \rightarrow \RR^{\tau} $ to approximate the dimension-reduced Bloch responses (see section~\ref{sec:expe_neural} for details, also~\cite{PGDNET-MRF} which applied this idea for single-compartment MRF reconstruction). 
This idea greatly accelerates Algorithm~\ref{alg:FW} in steps requiring function evaluation and differentiating Bloch responses: for solving 
\eqref{eq:spglasso}, $X$ and $\phi(.)$ are replaced by 
\eq{
\widetilde{X}\approx V^TX \qandq \widetilde{\phi}(\theta)\approx V^T\phi(\theta)
}
i.e. 
the dimension-reduced TSMI and the neural approximation of the (compressed) Bloch responses, respectively.


\begin{algorithm}[t!]
\begin{algorithmic}[1]
\STATE \textbf{Inputs:} TSMI $X$, Bloch model $\phi(.)$, params $\alpha,\beta$
\STATE \textbf{Outputs:} NMR parameters $\btheta$, mixture weights $C$ \vspace{.05cm}
\STATE \textbf{Initialise:} $t=0$,  $(\btheta^0,C^0)=\{\}$, $\eta(\theta)  \eqdef\frac{1}{\alpha} \Phi^* X$ as in~\eqref{eq:adj}

\REPEAT {
     \STATE  $\theta^t =  \argmax_{\theta\in\Tspace} 
     \norm{\left(\eta(\theta) + \beta-1\right)_+}_2^2$ \vspace{.05cm}
 \STATE $\btheta^{t+\frac12} = \btheta^t \cup \ens{\theta^t}$ \vspace{.05cm}
\STATE $C^{t+\frac12} = \argmin_{C \geq 0}\frac12 \norm{X- D_{\btheta^{t+\frac12}}  C }_F^2 + \alpha  \norm{C}_\beta$ \vspace{.14cm}
\STATE Initialised by $C^{t+1/2}$ and $\btheta^{t+1/2}$,  solve 
$$(C^{t+1},\btheta^{t+1}) = \mathop{\argmin}_{\btheta\in \Tspace, C\geq0} \frac12 \norm{X- \sum_{s=1}^{t+1} \phi(\theta_s) C_s^\top}_F^2 + \alpha  \norm{C}_\beta$$

\STATE $\eta(\theta) \eqdef \frac{1}{\al} \Phi^* (X- D_{\btheta^{t+1}}(C^{t+1})^\top)$\vspace{.05cm}
\STATE $t = t+1$
}
 \UNTIL{$\max_{\theta\in\Tspace}   \norm{\left(\eta(\theta) +\beta-1\right)_+}_2^2 \leq v\beta^2$}

\end{algorithmic}
\caption{\label{alg:FW} Frank-Wolfe iterations to solve \eqref{eq:spglasso}} 
\end{algorithm}

\section{Experiments}
\label{sec:expe}
Computations were conducted using MATLAB on an Intel Xeon gold CPU core and 32 GB RAM. 
Source codes for the proposed algorithm are available at \url{https://github.com/mgolbabaee/SGB-Lasso-for-partial-volume-quantitative-MRI}. 

All experiments (simulated and \emph{in-vivo}) adopted a joint T1/T2-encoding MRF excitation sequence similar to the Fast Imaging Steady State Precession (FISP) protocol~\cite{FISP} with the same flip angle schedule, fixed repetition/echo times  TR/TE = 10/1.9 ms, and the inversion time 18 ms. This sequence had the length of $T=1000$ timepoints (repetitions). 

\subsection{Tested algorithms}
We compared \MCMRF (Algorithm~\ref{alg:FW}) to the MC-MRF baselines SPIJN~\cite{SPIJN}, BayesianMRF~\cite{BayesianPVMRF} and PVMRF~\cite{PVMRF}. All algorithms used subspace dimensionality reduction $\tau=10$~\cite{SVDMRF,MRFLRTV}. Reconstructed TSMIs were phase-corrected and mapped to real-valued images before applying mixture separation. 
Baselines work with fine-gridded MRF dictionary. 
In Section~\ref{sec:vivo} we also compared a discretised variant of
our algorithm, named SG-Lasso~\cite{SPGLasso}, 
using the same  MRF dictionary as the baselines. Methods' parameters were grid searched and chosen based on scoring low model errors to~\eqref{eq:model4}, visual separability of the mixture maps, and consistency of the estimated T1/T2 values with respect to the ground-truth (simulations) or literature values (in-vivo).
The regularisation parameters of BayesianMRF and SPIJN were $\mu_{\text{B}}=\{0.001,1\}$ and $\la_{\text{SPIJN}}=\{0.03,0.5\}$ for dirichlet phantoms 
and in-vivo experiments, respectively. BayesianMRF used shape parameters $\alpha_{\text{B}}=1.75, \beta_{\text{B}}=0.1$. PVMRF used parameter $k_{\text{PVMRF}}=11$ for the k-means. To stabilise PVMRF, 
 k-means steps were repeated 10 times (randomly initialised) and result scoring lowest model error~\eqref{eq:model4} was selected. \MCMRF used parameters $\alpha=10^{-6},\beta=0.9$ for the dirichlet phantoms experiment and $\alpha=0.8,\beta=10^{-3}$ for the simulated and in-vivo brain experiments. 

\vspace{-.1cm}
\subsection{Embedding Bloch responses by a neural network} 
\label{sec:expe_neural}

A neural network was trained to approximate Bloch responses.  
For training and evaluation 95'143 Bloch responses were simulated using the EPG formalism~\cite{EPGWeigel} over a (T1,T2) $\in [10, 6000]$ ms $\times[4,4000]$ ms grid discretised by logarithmically spaced values of T1 and T2 (400 points each) with T1 $>$ T2 constraint. PCA was applied to compress Bloch responses' temporal dimensions to a $\tau=10$ dimensional subspace following~\cite{SVDMRF}.  
Data was randomly splitted in $80-10-10$ precent ratios for training, validation and testing sets, correspondingly. 
For our application we followed~\cite{MRFLRTV} and used a (convolutional) network with $1\times1$ filters for pixel-wise processing: 2-channel inputs for T1 and T2 values, 10-channel (linear) outputs for the compressed Bloch responses, and one hidden layer of 500 channels with nonlinear ReLU activations.  
For training we minimised the MSE loss between the EPG-generated and network-predicted (dimension-reduced) Bloch responses using ADAM optimiser ran for 100 epochs with the initial learning rate $0.005$, learning rate/gradient decay factors $0.95/0.95$, and minibatch size $100$. Training and validation curves are shown in supplementary  Figure~\ref{fig:trainingcurve}. Training, validation and testing normalised RMSEs for approximating Bloch responses were $\{8.9397, 8.9742, 8.9561\}\times10^{-3}$, correspondingly. 


\MCMRF used only neural network approximations. 
For the initialising grid-search in step 5 of Algorithm~\ref{alg:FW} , a fixed 64-atom dictionary was simulated over 10 logarithmically-spaced points per T1 and T2, and selecting T1$>$T2, whereas the tested baselines used the actual EPG dictionary comprising 8'540 
 fingerprints:  
 120 values per T1 and T2, respecting T1$>$T2. 

\begin{figure}[t!]
\begin{minipage}{\linewidth}
\centering
\includegraphics[width=\linewidth]{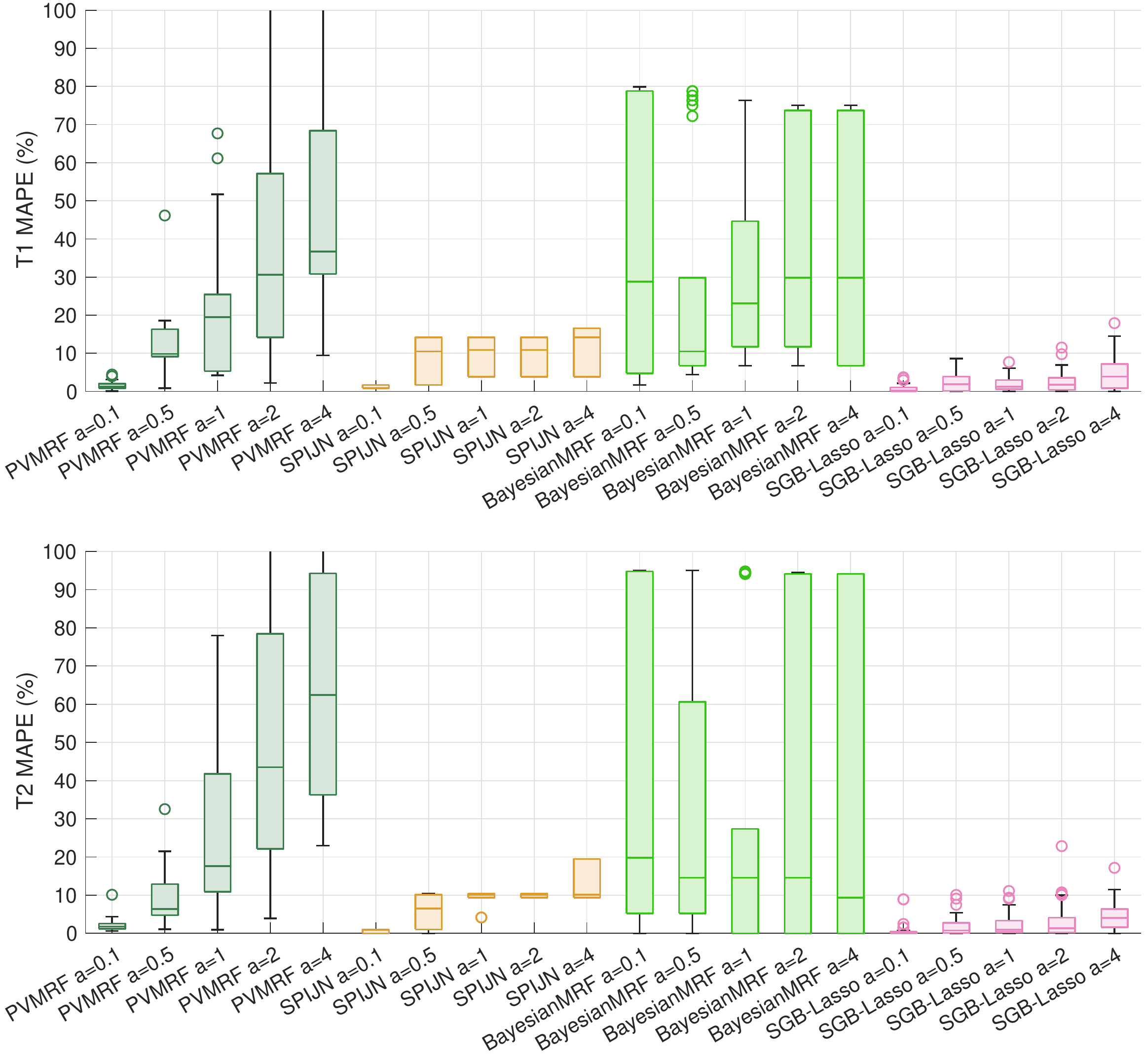}
\caption{\footnotesize{The T1 (top) and T2 (bottom) MAPE errors of the MC-MRF algorithms for estimating dirichlet phantoms' compartments with different mixture parameters $a$. Plots are cropped to 100\% MAPE and colour-coded for different algorithms. 10 independent phantoms were simulated per $a$ value.}\label{fig:synth_diri} }
\end{minipage}
\end{figure}

\subsection{Numerical dirichlet phantoms experiment}
\label{sec:diri_expe}

Several phantoms were simulated according to the mixture model~\eqref{eq:model2} for creating TSMIs 
of $20\times20$ pixels spatial resolution. In these phantoms three compartments were used  with relaxation properties T1/T2 $=\{784/77, 1216/96, 4083/1394\}$ ms. 
Per-pixel mixture weights were drawn randomly from i.i.d. dirichlet distributions parametrised by various values of $a>0$. This parameter governs the mixture levels such that  large $a$ values result in highly mixed pixels that receive similar contributions from all compartments and hence are more difficult for demixing~\cite{bioucas_sisal}. Instead, small $a$ creates sparse (pixel-pure) mixture maps i.e. pixels receive contributions from fewer compartments, making demixing task easier. 

\subsubsection{Different mixture levels}
We created a dataset of such phantoms from different mixture distributions with values $a\in\{0.1,0.5,1,2,4\}$ and for each distribution we simulated 10 phantoms independently at random (see exemplar mixtures in supplementary Figure~\ref{fig:diri_maps}). We used this dataset to compare the MC-MRF baselines to the \MCMRF algorithm. 
The Mean Absolute Percentage Errors (MAPE) of the estimated compartments' T1 and T2 values 
were measured and reported in Figure~\ref{fig:synth_diri} using MATLAB's \texttt{boxchart} tool (the inside box line, box edges, whiskers and circles represent the mean, quartiles, extreme values and outliers, correspondingly). 
The \MCMRF outperforms tested baselines with accurate T1/T2 predictoins i.e. less than \%5 average errors for all tested mixture distributions. For $a=0.1$ where the mixtures have the highest pixel-purity the PVMRF and SPIJN methods perform comparably well. However \MCMRF shows robustness for separating less pixel-pure mixtures i.e. larger $a$ values, where the gap between baselines and \MCMRF in terms of both the mean and variation of the MAPE errors increases.

\subsubsection{Noise stability and grid-search size}
\begin{figure}[t!]
\begin{minipage}{\linewidth}
\centering
\includegraphics[width=\linewidth]{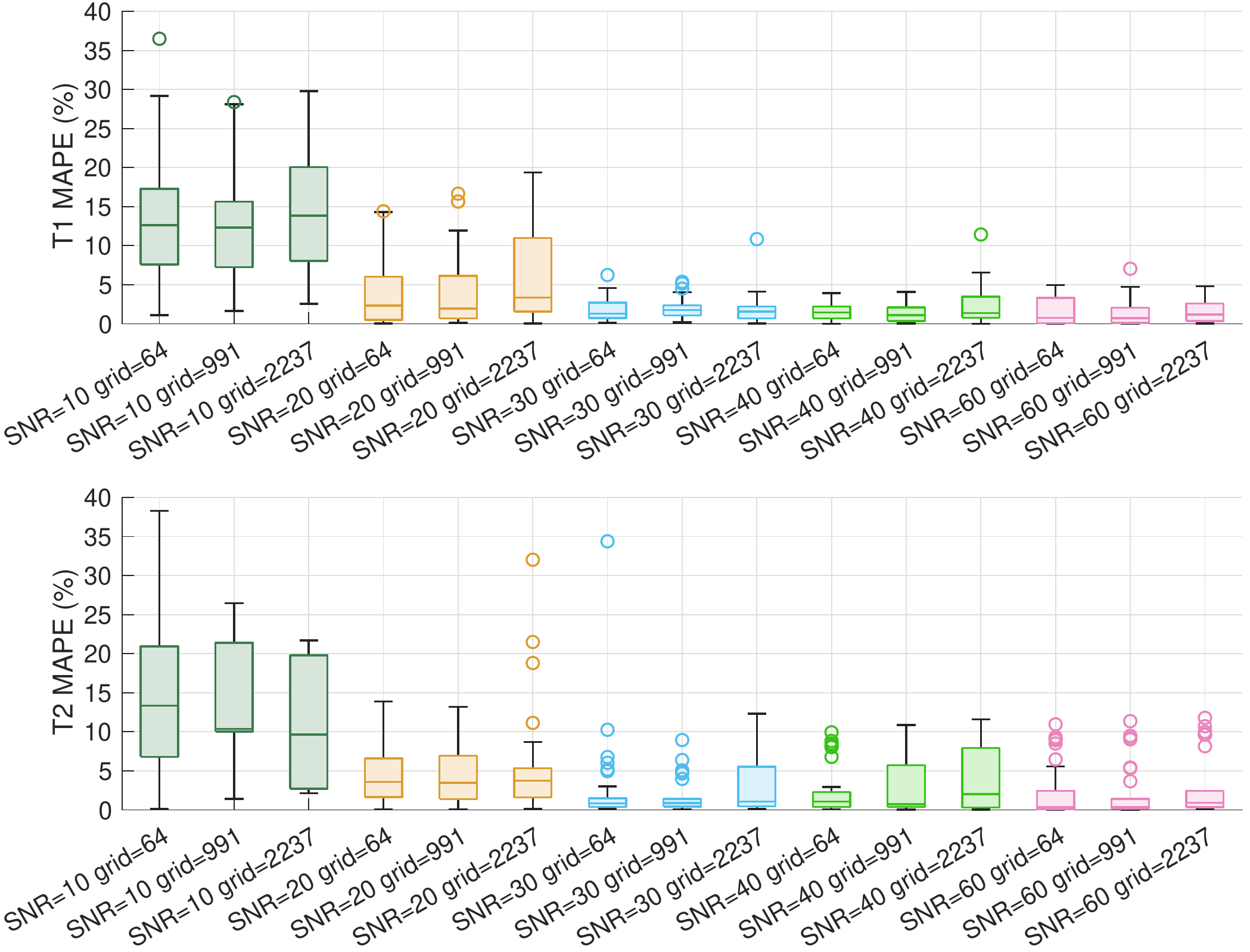}
\caption{\footnotesize{The T1 (top) and T2 (bottom) MAPE errors for estimating dirichlet phantoms' compartments (dirichlet mixture $a=0.5$) using \MCMRF\ with different initialising grid sizes and different SNR (dB) levels. 10 independent phantoms were simulated/demixed per SNR value.}\label{fig:synth_gidnoise} }
\end{minipage}
\end{figure}

In another experiment we created a dataset of phantoms by setting the dirichlet parameter $a=0.5$, and corrupt the TSMIs by additive white gaussian noises of various SNRs $\in\{10, 20, 30, 40, 60\}$ dB. 
We simulated 10 noisy phantoms per SNR value. 
We used this dataset to measure the sensitivity of \MCMRF to noise and also the size of the grid-search used in Algorithm~\ref{alg:FW} (step 5). For this we increased the storage requirement of \MCMRF using grid-sizes  \{64, 991, 2237\} corresponding to sampling  \{10,  40, 60\} logarithmically-spaced points per T1 and T2, T1$>$T2. 
 The estimated T1 and T2 MAPEs are reported in Figure~\ref{fig:synth_gidnoise}. For SNRs $\geq20$ dB the average errors are less than \%5. These then increase to \%10-\%15 in the lowest tested SNR regime 10 dB. Further, it can be observed that \MCMRF's overall performance per SNR regime has small variations with respect to the tested grid-sizes (less than \%3 difference on average MAPEs) which could motivate adoption of small grid-sizes to overcome the memory inefficiency (challenge of dimensionality) of the dictionary-based MC-MRF approaches. Throughout other experiments, \MCMRF used the smallest grid-size 64 that is 2 orders of magnitude smaller and more memory-efficient than the size (8'540 fingerprints) of the MRF dictionary used by the baselines.

\begin{figure}[t]
	\centering
	\scalebox{.9}{
	\begin{minipage}{\linewidth}
		\centering		
		\begin{turn}{90} \,\,{\footnotesize{Ground truth maps}} \end{turn}
\includegraphics[width=.3\linewidth]{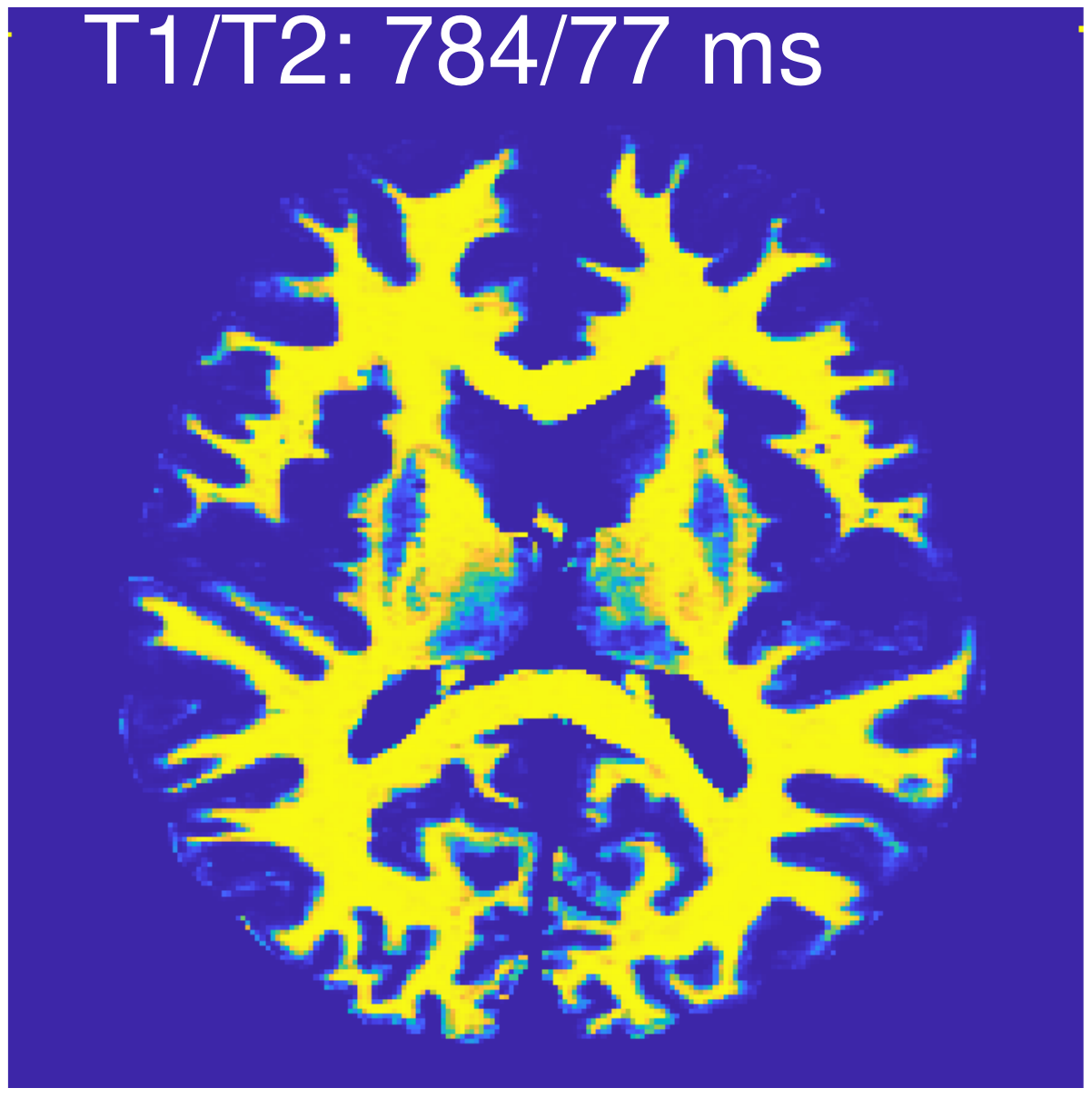}\hspace{-.1cm}
\includegraphics[width=.3\linewidth]{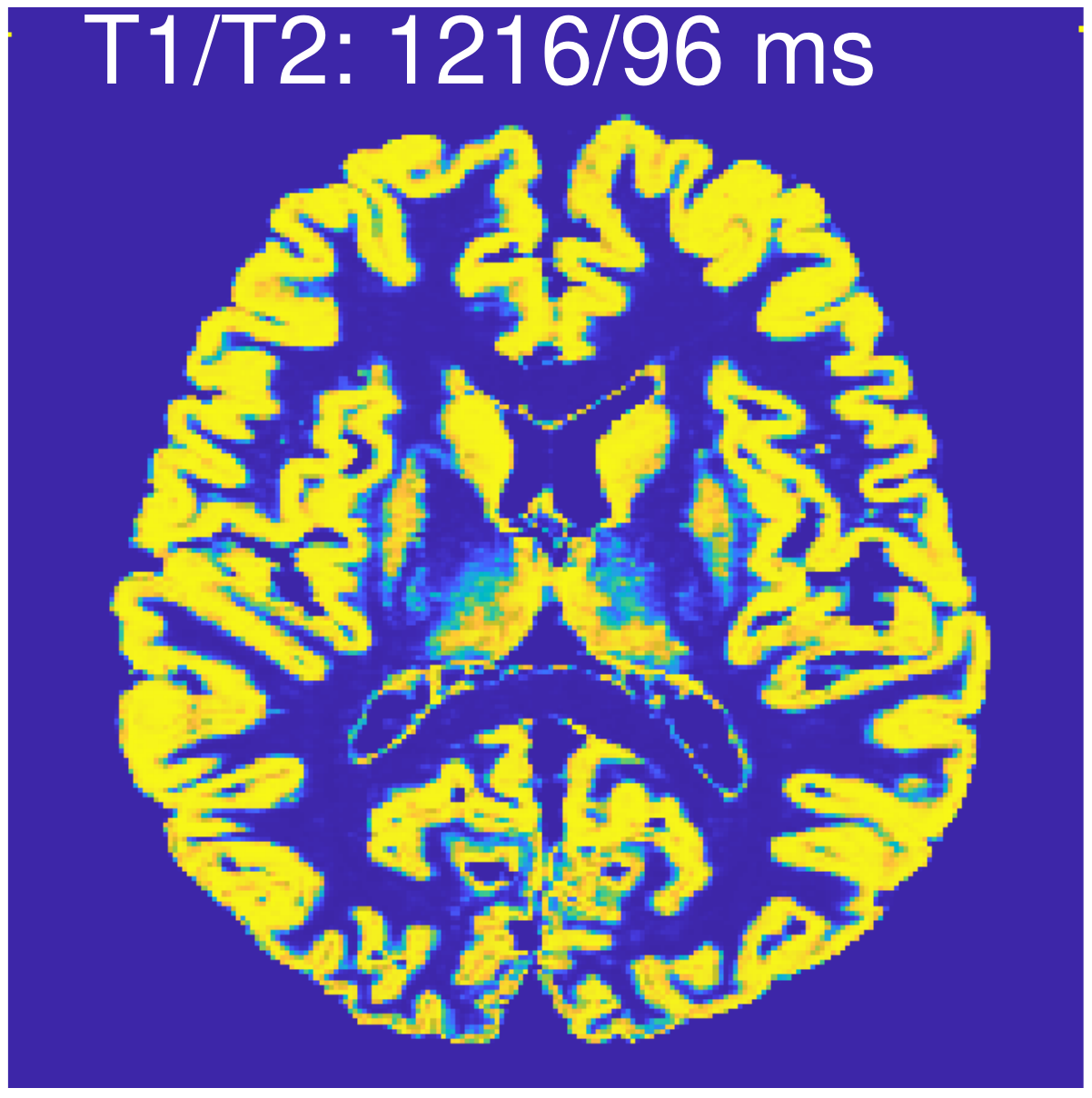}\hspace{-.1cm}
\includegraphics[width=.3\linewidth]{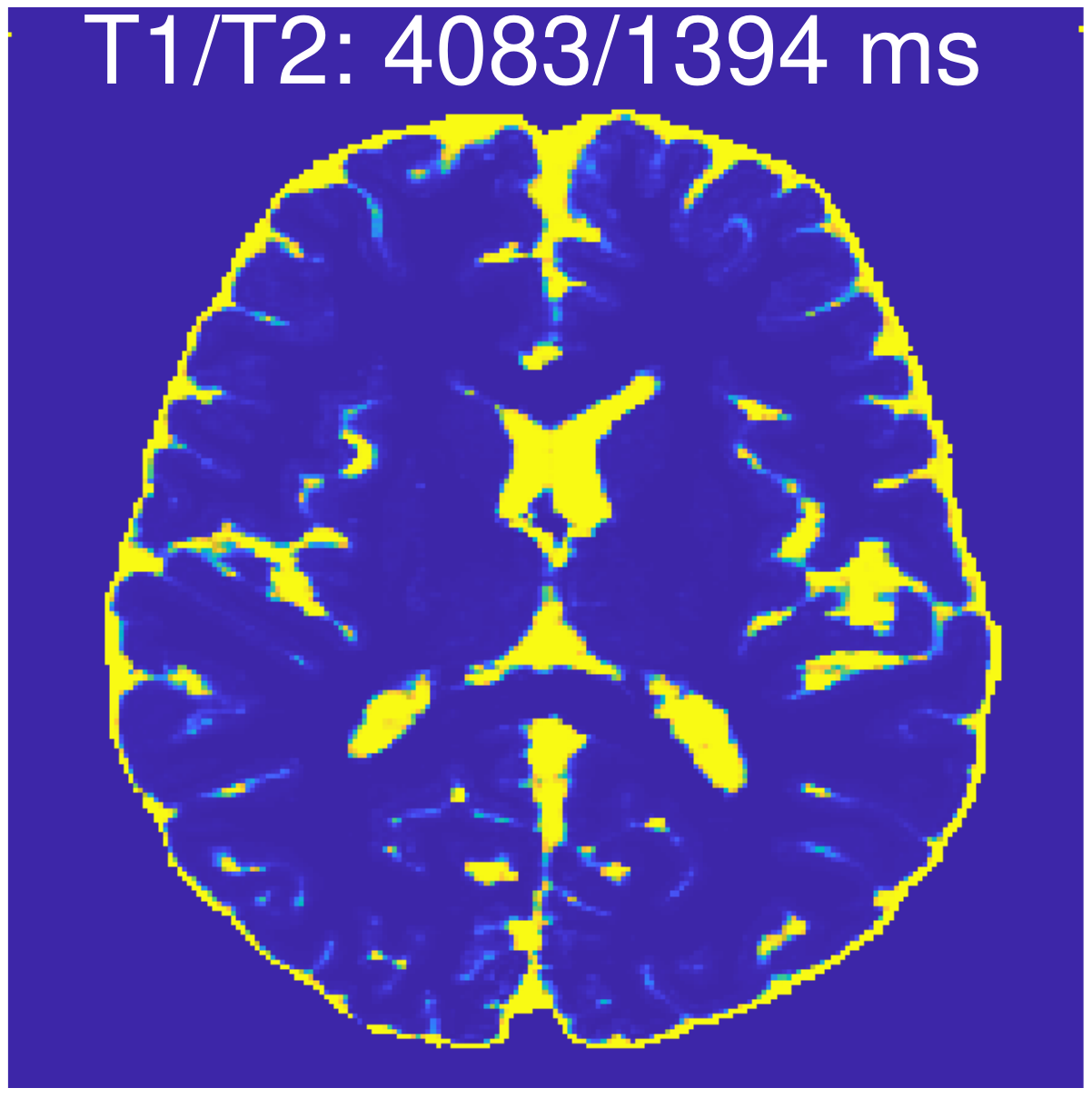}
\begin{turn}{90} \includegraphics[height=.4cm]{./figs/colorbarMaps_c.pdf}\end{turn}
\\
\begin{turn}{90}  {\footnotesize{\quad diff. maps (Full) }}\end{turn}
\includegraphics[width=.3\linewidth]{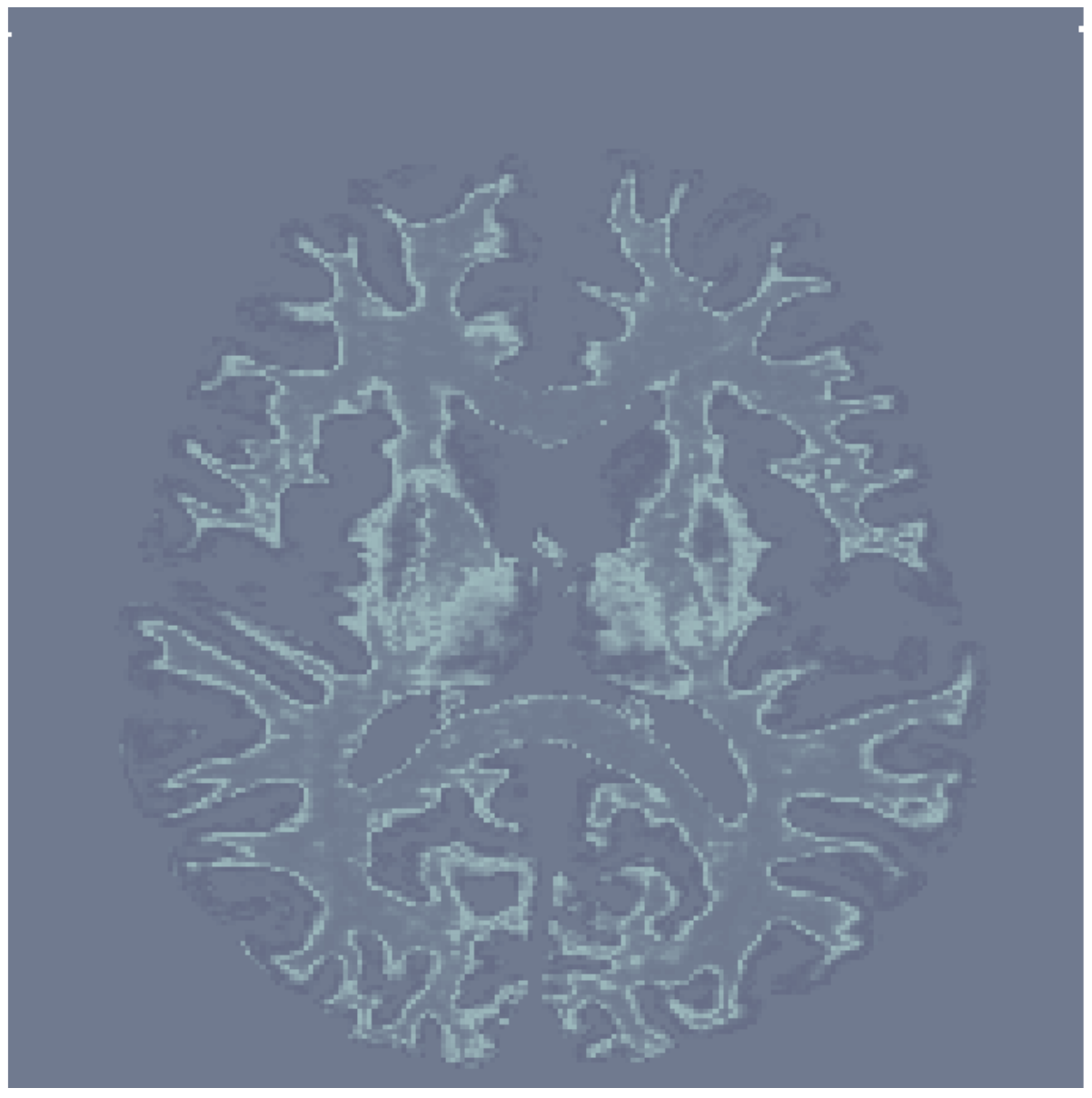}\hspace{-.1cm}
\includegraphics[width=.3\linewidth]{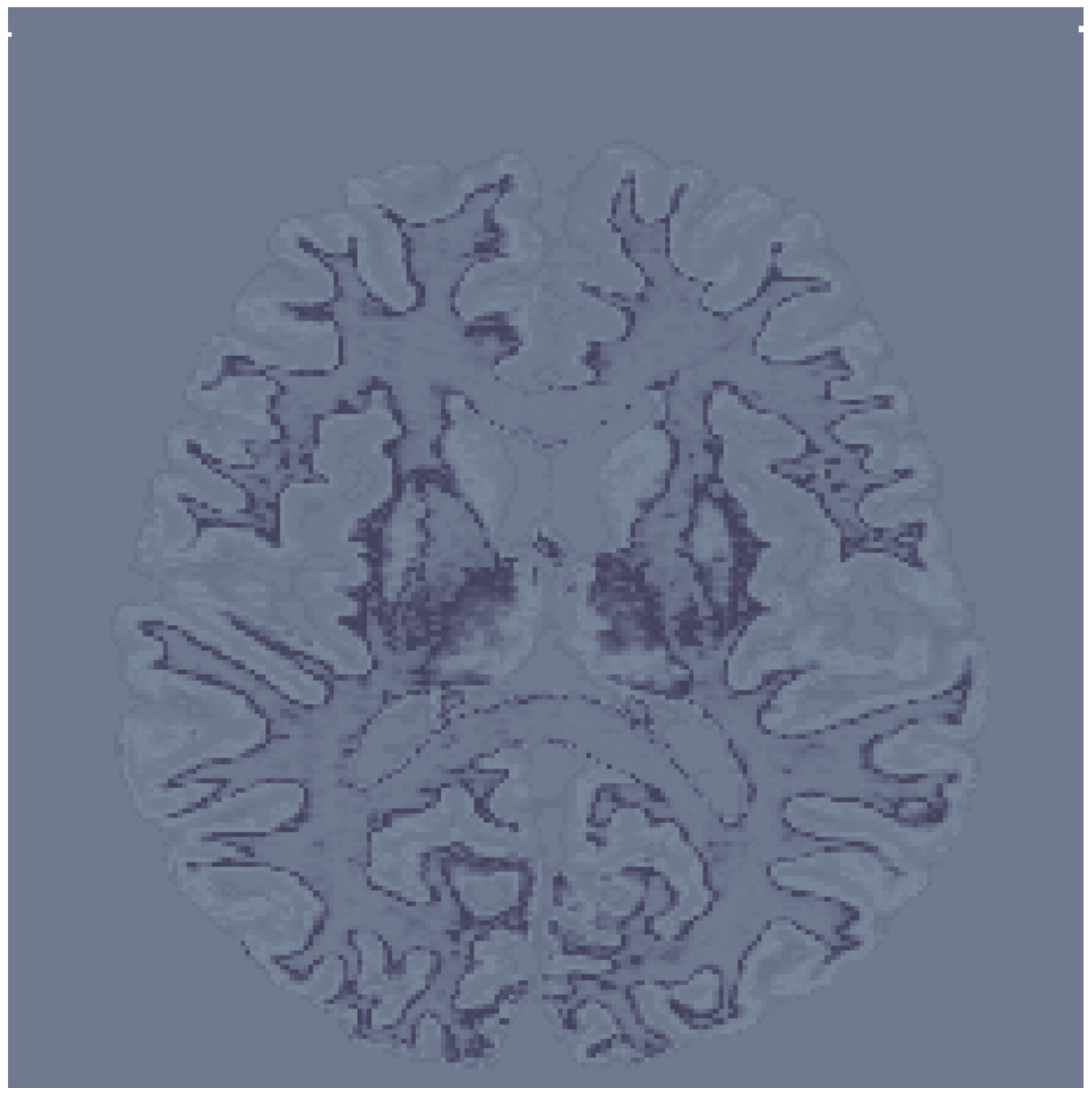}\hspace{-.1cm}
\includegraphics[width=.3\linewidth]{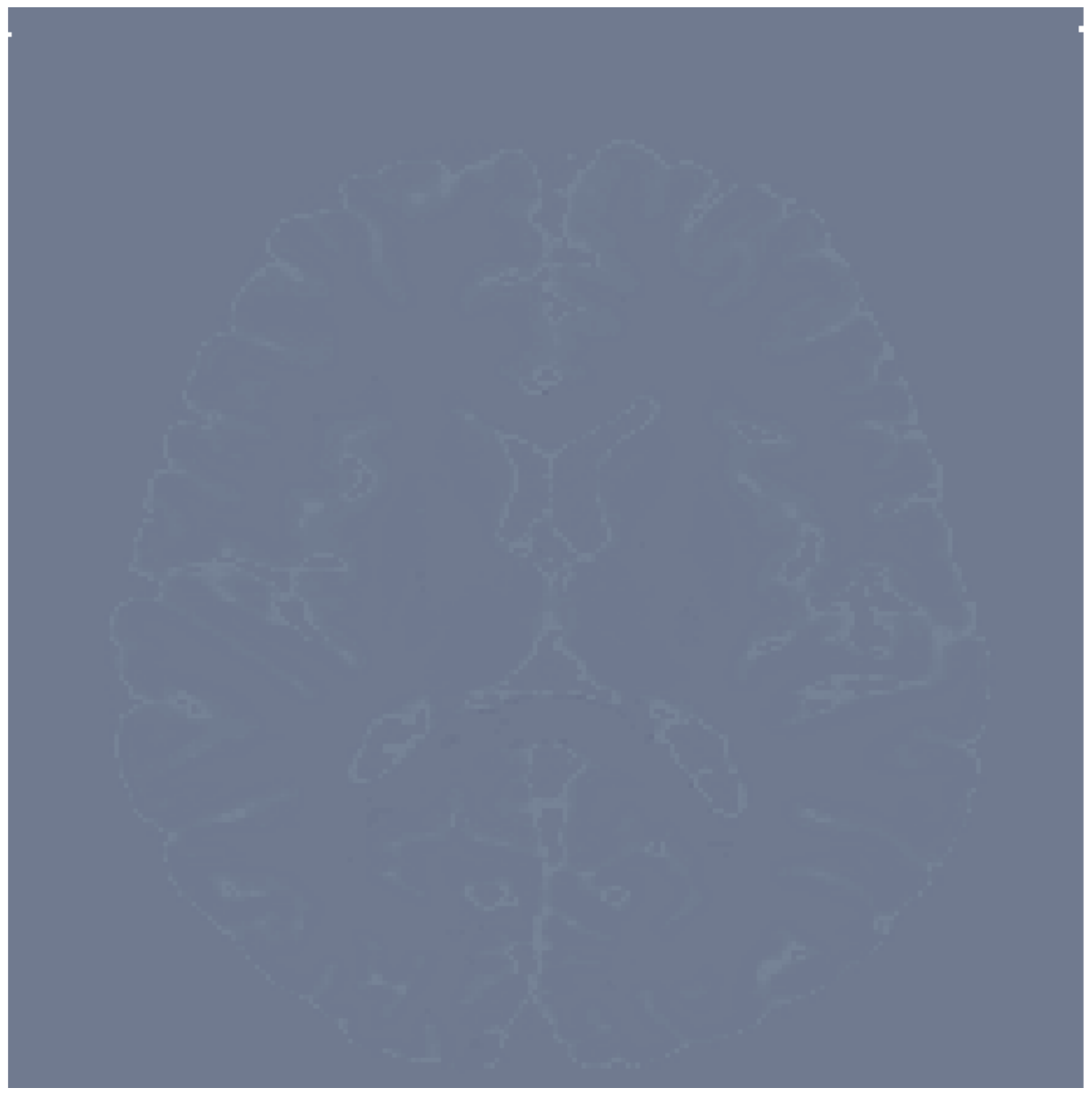}
\begin{turn}{90} \includegraphics[height=.4cm]{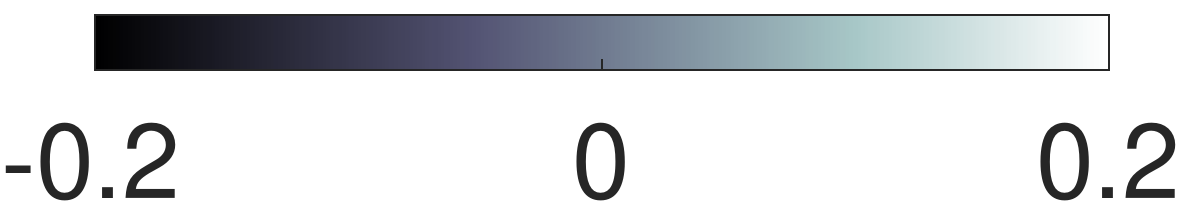}\end{turn}
\\
\begin{turn}{90}  {\footnotesize{\quad diff. maps (LRTV) }}\end{turn}
\includegraphics[width=.3\linewidth]{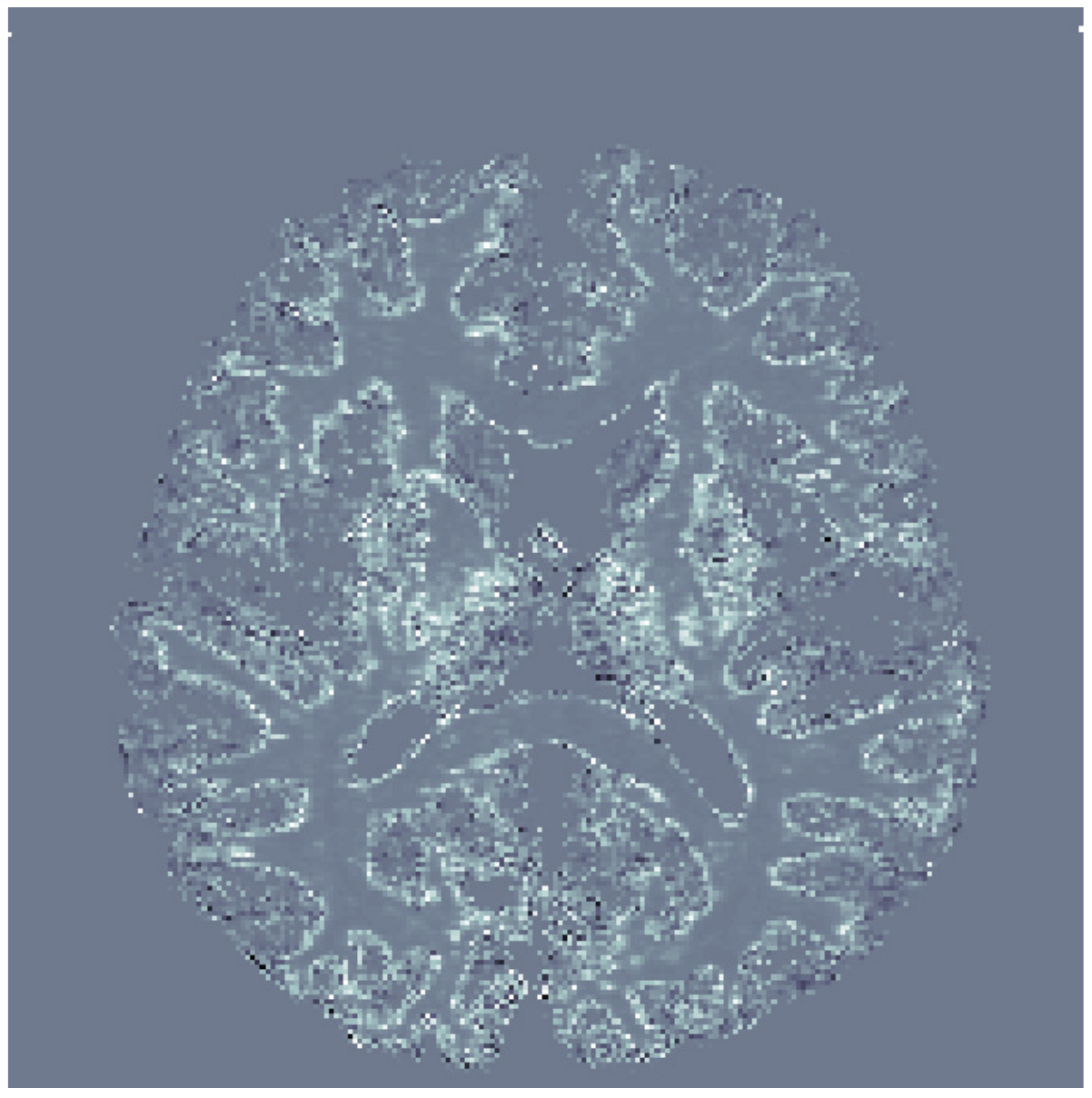}\hspace{-.1cm}
\includegraphics[width=.3\linewidth]{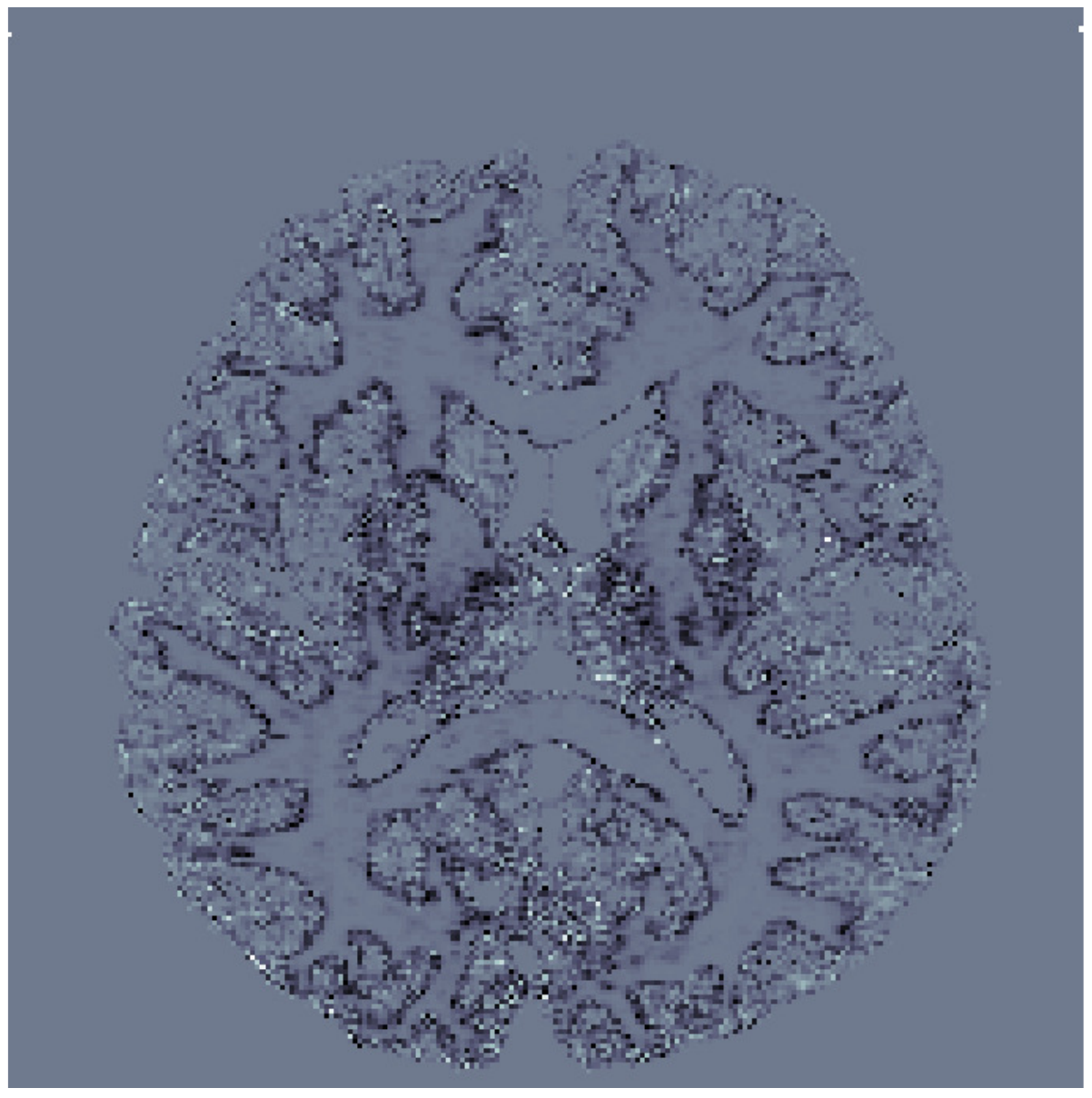}\hspace{-.1cm}
\includegraphics[width=.3\linewidth]{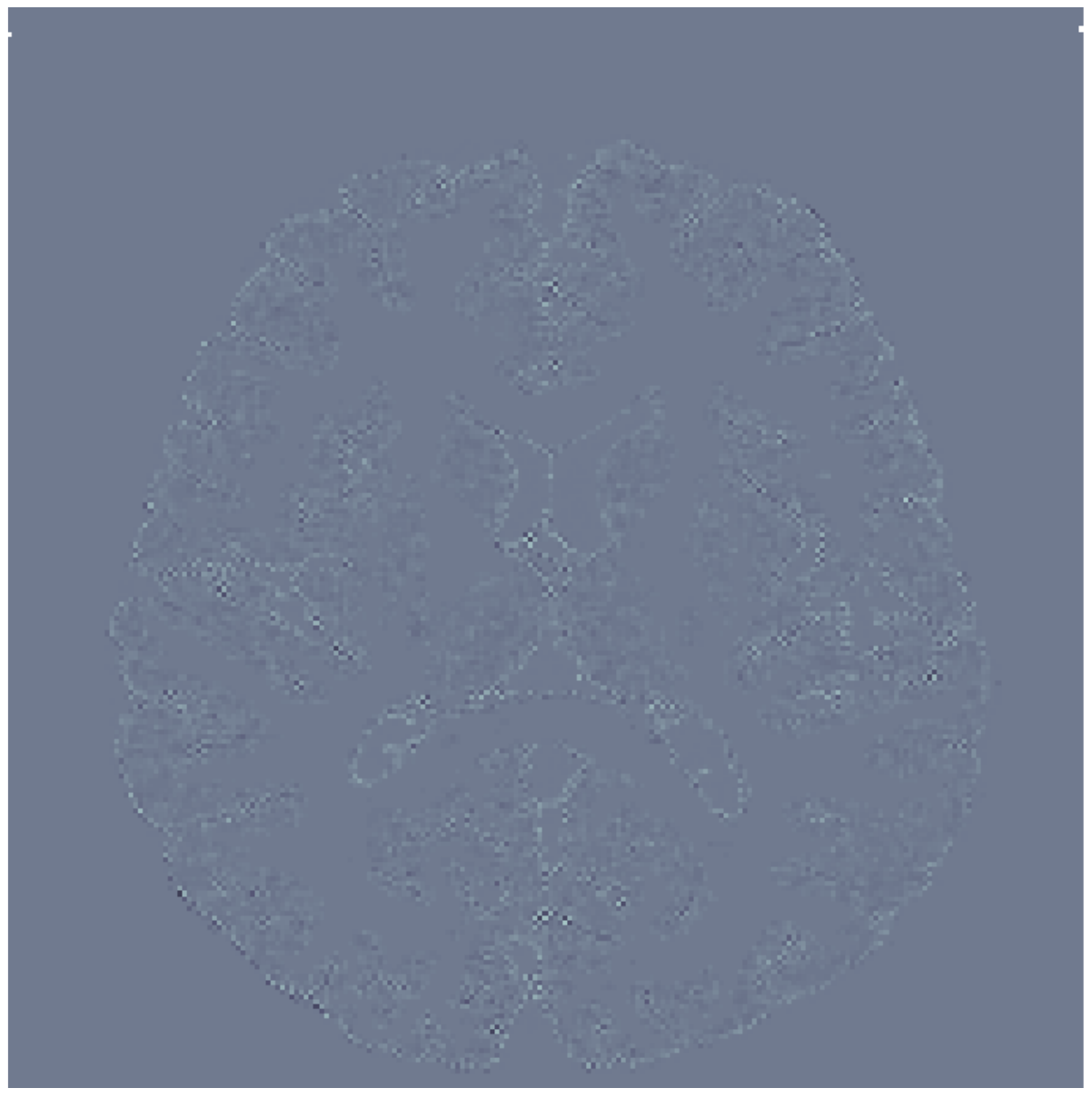}
\begin{turn}{90} \includegraphics[height=.4cm]{./figs/colorbardiffMaps.pdf}\end{turn}
\\
\begin{turn}{90}  {\footnotesize{diff. maps (SVD-MRF) }}\end{turn}
\includegraphics[width=.3\linewidth]{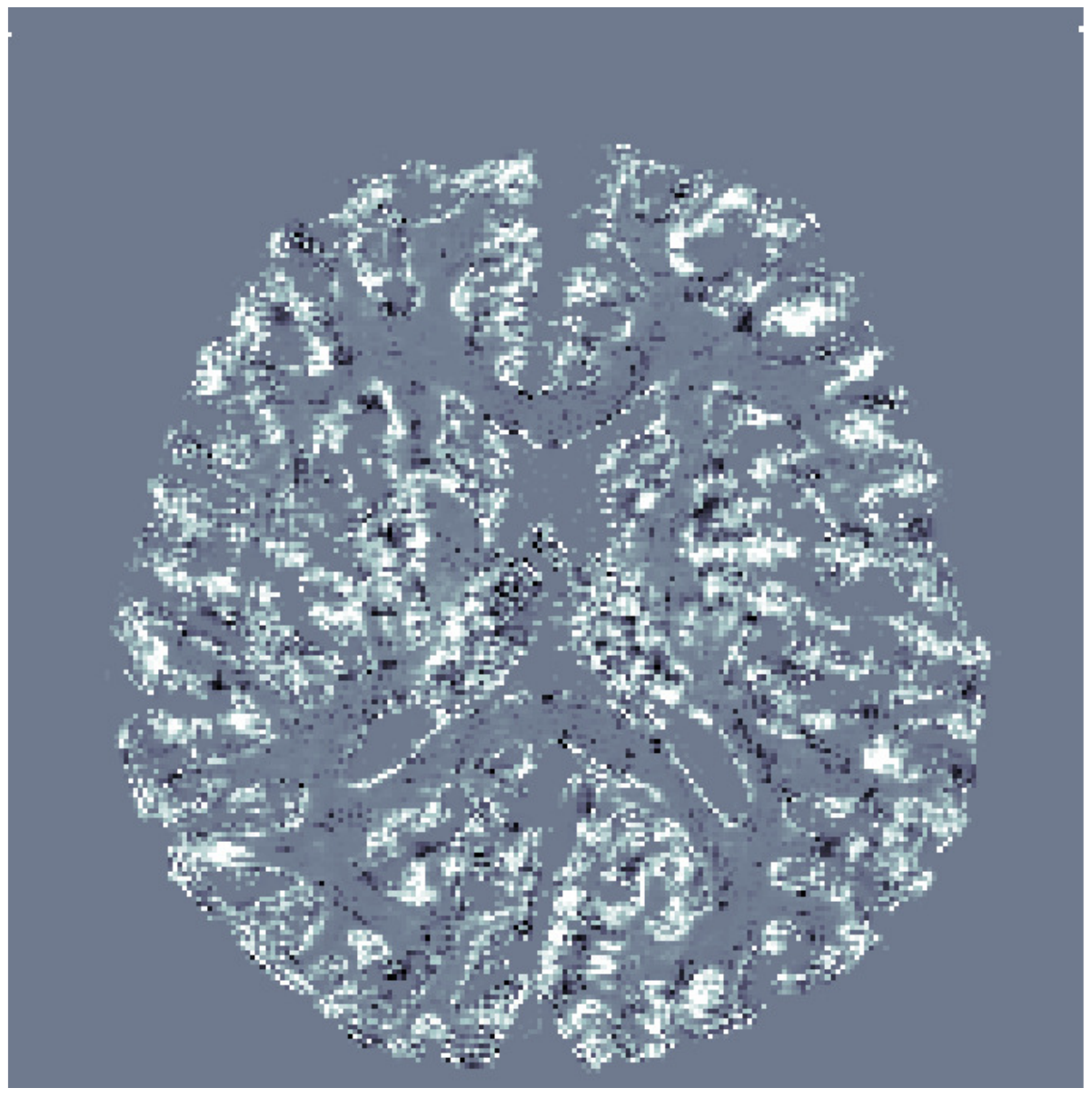}\hspace{-.1cm}
\includegraphics[width=.3\linewidth]{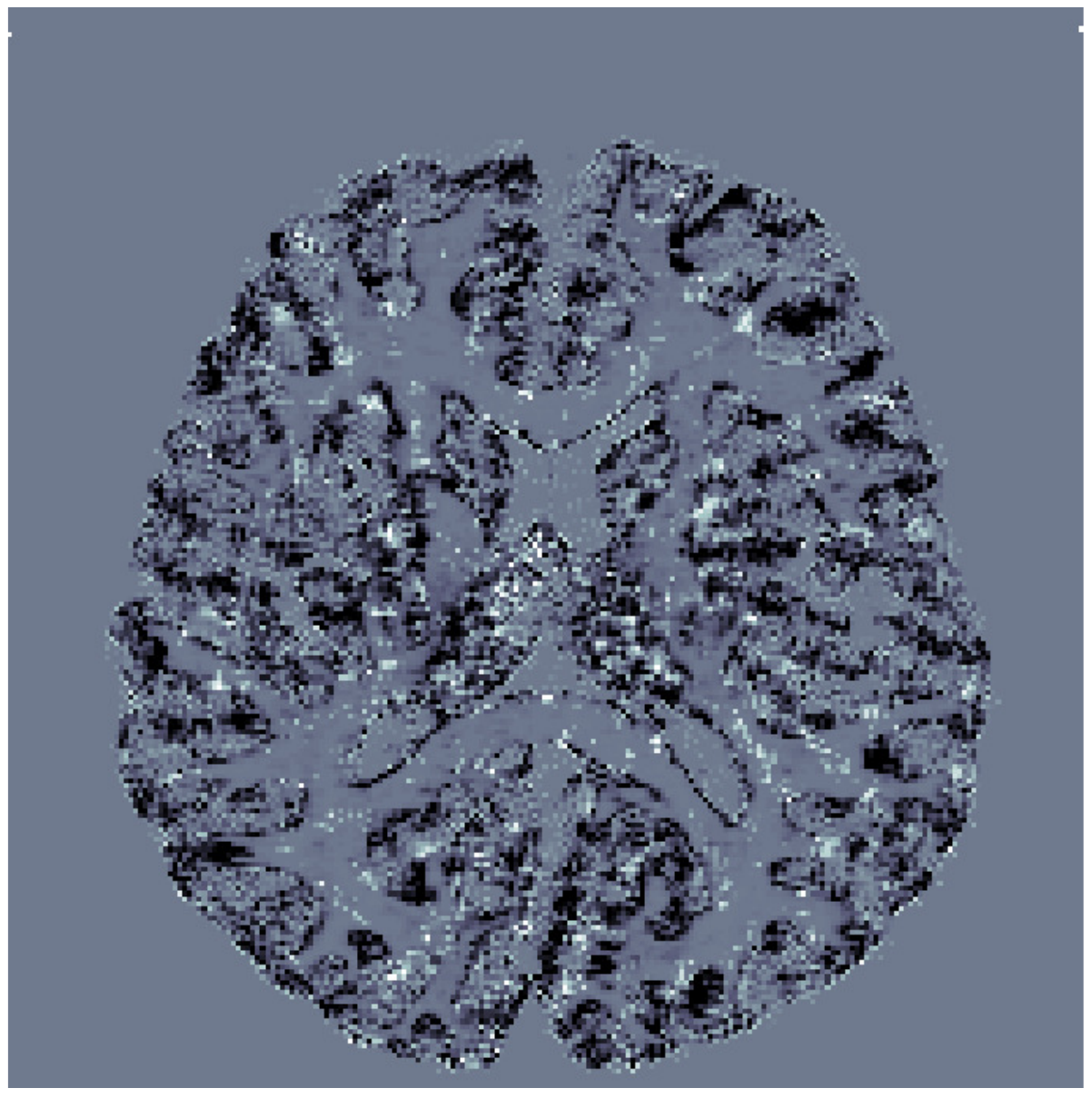}\hspace{-.1cm}
\includegraphics[width=.3\linewidth]{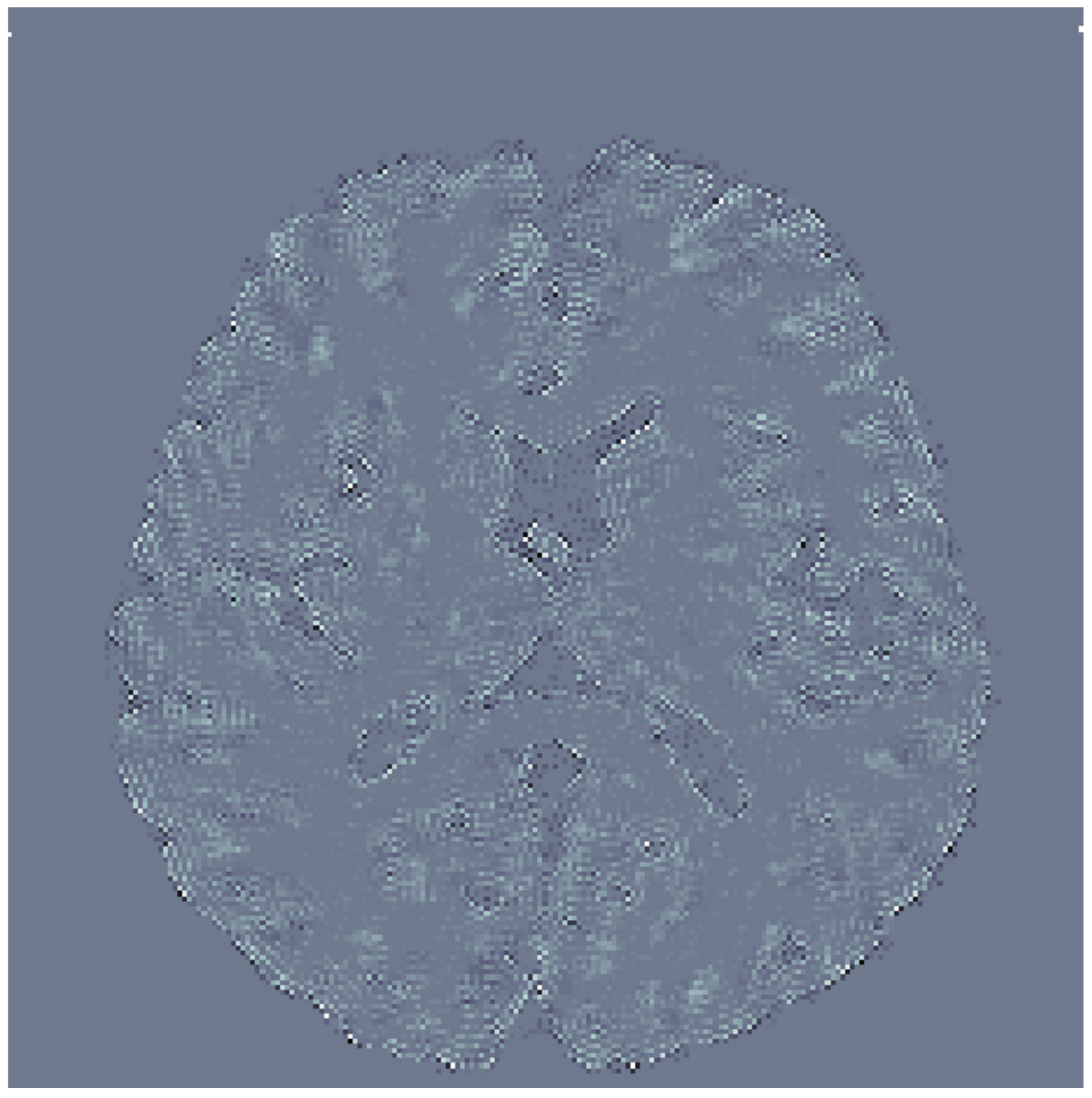}
\begin{turn}{90} \includegraphics[height=.4cm]{./figs/colorbardiffMaps.pdf}\end{turn}
\end{minipage}}
\caption{\footnotesize{The ground truth mixture maps for the simulated Brainweb phantom, 
and the estimated mixture maps errors 
for the WM, GM and CSF compartments 
using \MCMRF. TSMIs were either un-compressed (Full) or subsampled using compressed sensing and reconstructed with the LRTV and SVD-MRF schemes before mixture separation. }
\label{fig:brainweb_GTmaps}}
\end{figure}

\subsection{Numerical Brain phantom experiment}
\label{sec:brainweb}
This experiment used the Brainweb's anatomical model of healthy brain~\cite{Brainweb-collins, webbrain}. This data includes fuzzy segmentations of the white matter (WM), gray matter (GM) and cerebrospinal fluid (CSF) and provides precise control on the ground truth mixture maps (Figure~\ref{fig:brainweb_GTmaps}). The assigned relaxations for these tissues were T1/T2 $=\{784/77, 1216/96, 4083/1394\}$ ms, 
and TSMI was constructed according to the model~\eqref{eq:model2}. 
We used this data to study \MCMRF for the case where TSMIs were un-compressed (fully sampled) and compare it to the case where acquisitions were accelerated using (compressed sensing)
the same subsampled k-space readouts as for our in-vivo data. 
Gaussian noise (50 dB SNR) were added to the measurements. Where compressed sensing applied, we adopted the reconstruction schemes LRTV and SVD-MRF to estimate TSMIs before the demixing step for comparison. 

Figures~\ref{fig:brainweb_GTmaps}~and~\ref{fig:brainweb_SGBlasso_main} illustrate 
the reconstructed mixture maps (weights) and their differences (errors) to the ground truth. 
The joint T1/T2 MAPEs and the mixture maps' reconstruction PNSRs are reported in Table~\ref{tab:expe_simul}. 
We observe 
accurate  T1/T2 estimations 
with less than \%4 MAPE for all compartments using the full and compressed-sampled data. Estimated mixture maps have larger errors using subsampled data, but this error is smaller using LRTV than SVD-MRF that produces subsampling (aliasing) artefacts (Figure~\ref{fig:brainweb_GTmaps}).  

\begin{table}[t!]
	\centering
	\fontsize{9}{9}\selectfont
	\scalebox{.95}{
		\begin{tabular}{cccc|ccc}
			\toprule[0.2em]
			&\multicolumn{3}{c}{T1/T2 MAPE (\%)} & \multicolumn{3}{c}{PSNR (dB)} \\
			\midrule[0.1em]
			&{WM} &{GM} &   {CSF} &{WM} &{GM} &   {CSF}    \\
			\midrule[0.05em]
			\midrule[0.05em]
			Full & $3.09$ & $1.72$ & $1.43$ & $35.05$ & $35.03$ & $53.70$\\
			LRTV & $3.61$ & $1.64$ & $1.71$ & $29.92$  & $29.44$  & $43.34$\\
			SVD-MRF &$3.39$ & $1.62$ & $2.33$ & $24.69$  & $23.69$  & $34.08$\\
			
		\bottomrule[0.2em]
		\end{tabular}}
		\caption{\footnotesize{The (joint) T1/T2 MAPE errors and the mixture maps PNSRs for the estimated compartments of the simulated brain in Figures~\ref{fig:brainweb_GTmaps}~and~\ref{fig:brainweb_SGBlasso_main}.}} 
\vspace{.3cm}\label{tab:expe_simul}
		\end{table}

\begin{figure*}[ht!]
	\centering
	\scalebox{.9}{
	\begin{minipage}{\linewidth}
		\centering		

\begin{turn}{90}  $\beta=0.0001$ \end{turn}
\includegraphics[width=.12\linewidth]{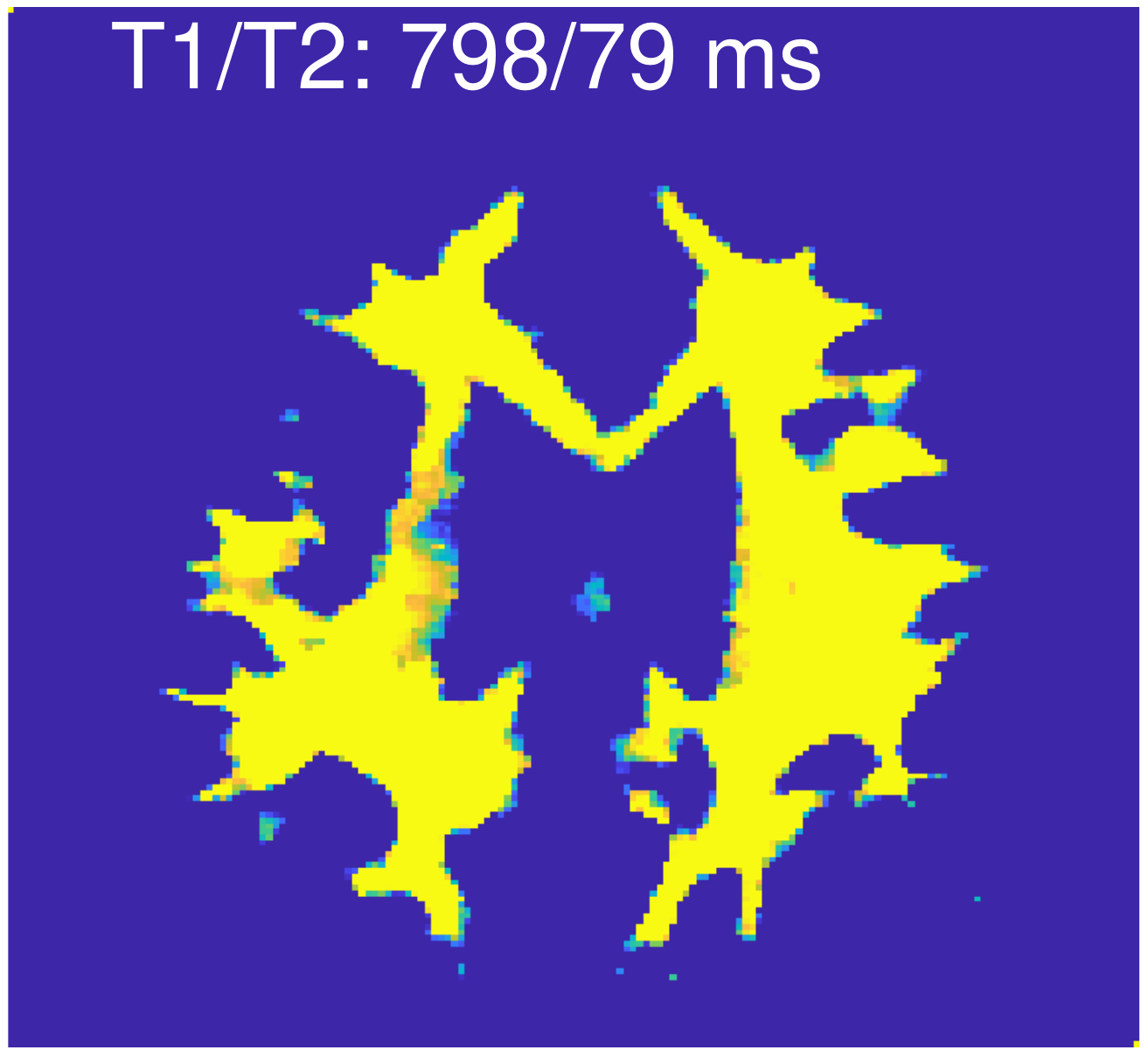}\hspace{-.1cm}
\includegraphics[width=.12\linewidth]{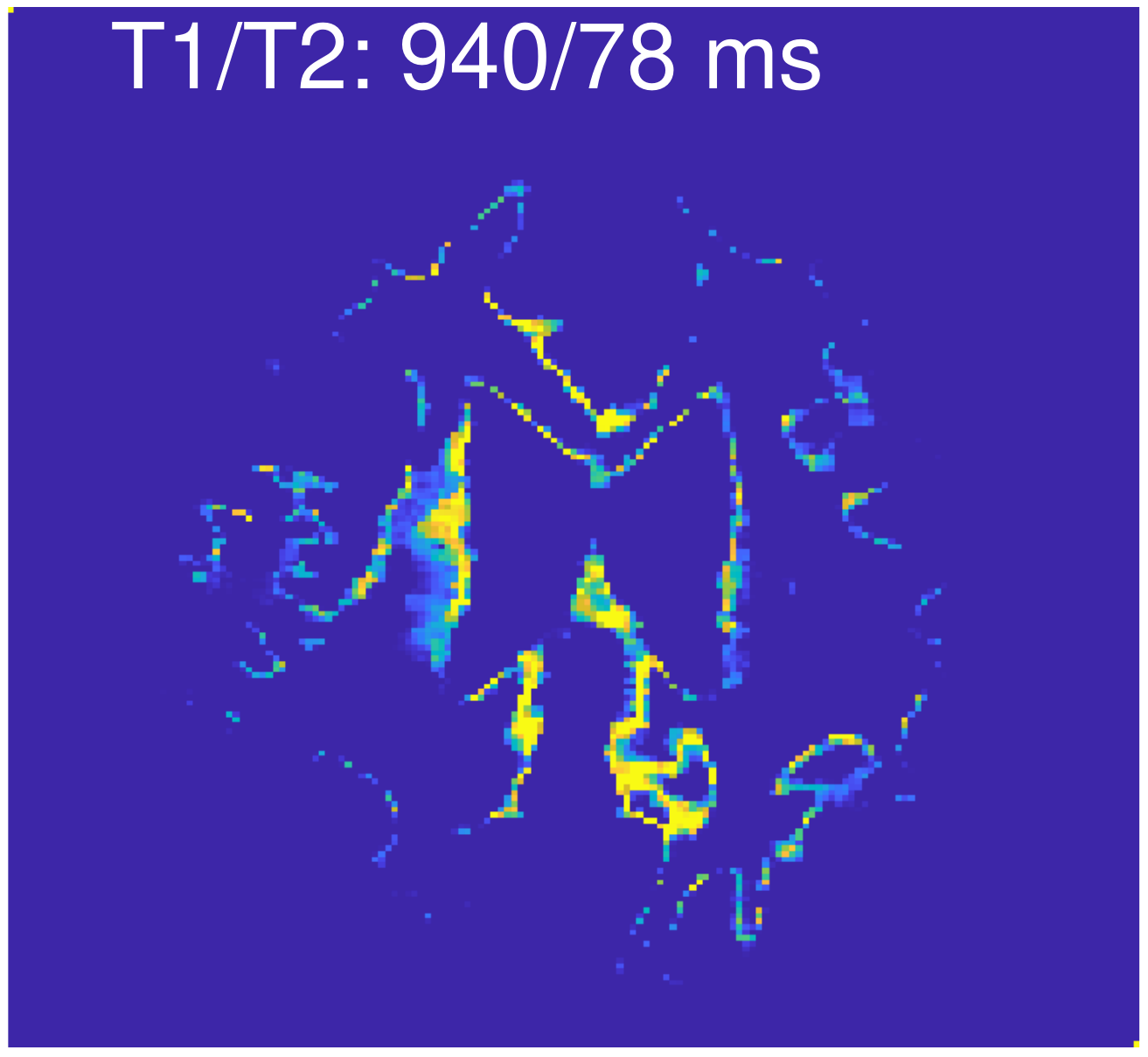}\hspace{-.1cm}
\includegraphics[width=.12\linewidth]{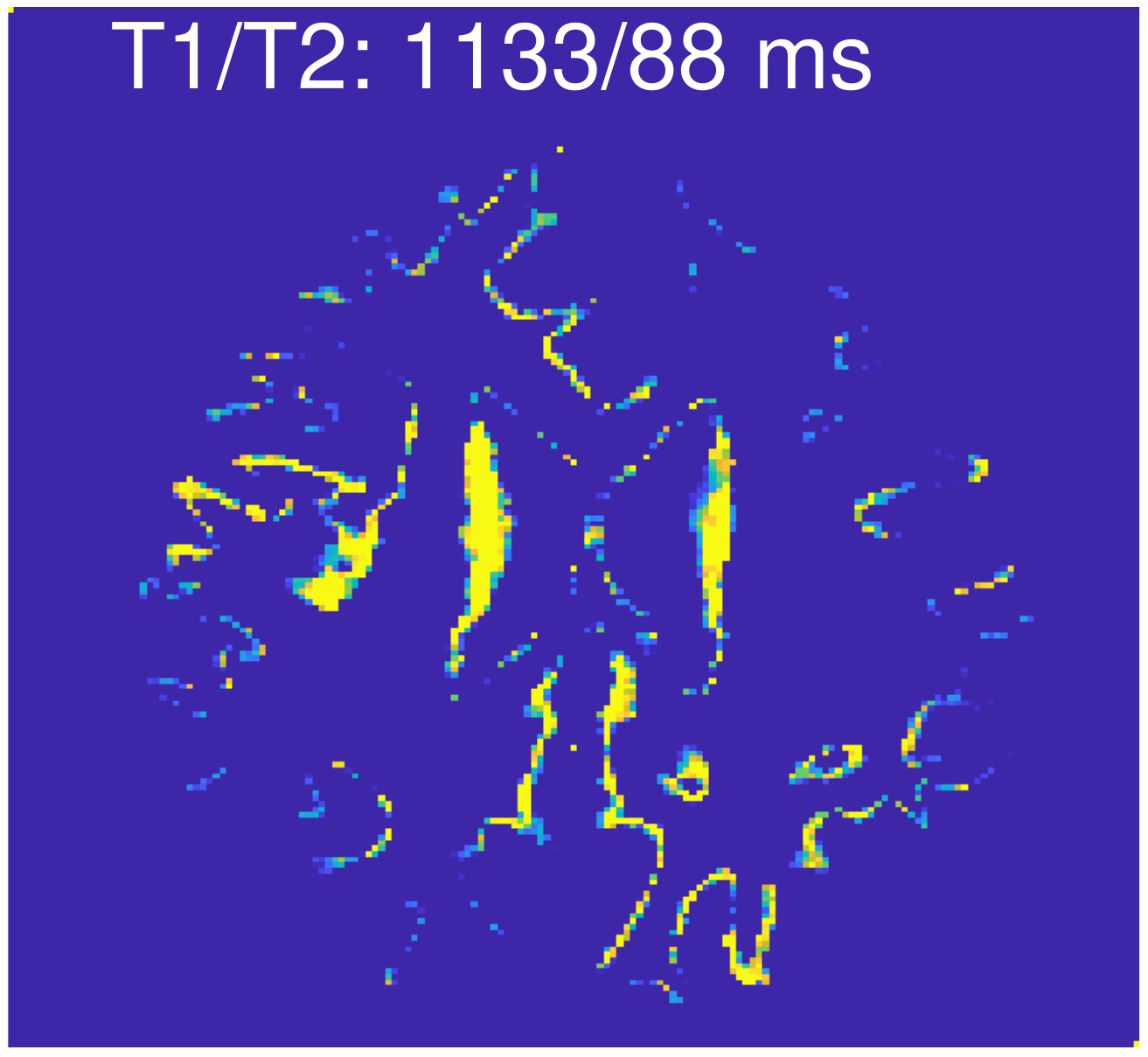}\hspace{-.1cm}
\includegraphics[width=.12\linewidth]{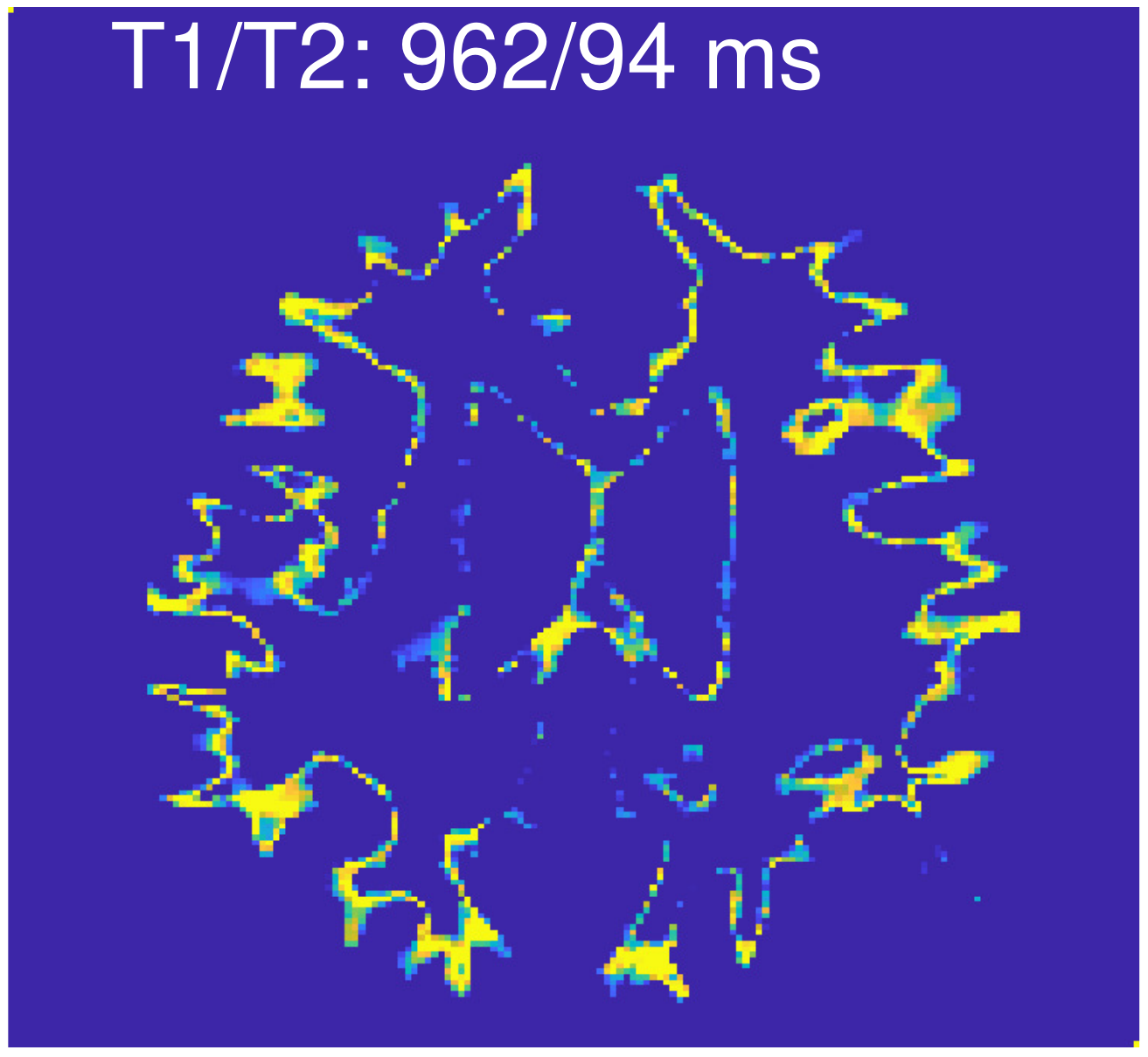}\hspace{-.1cm}
\includegraphics[width=.12\linewidth]{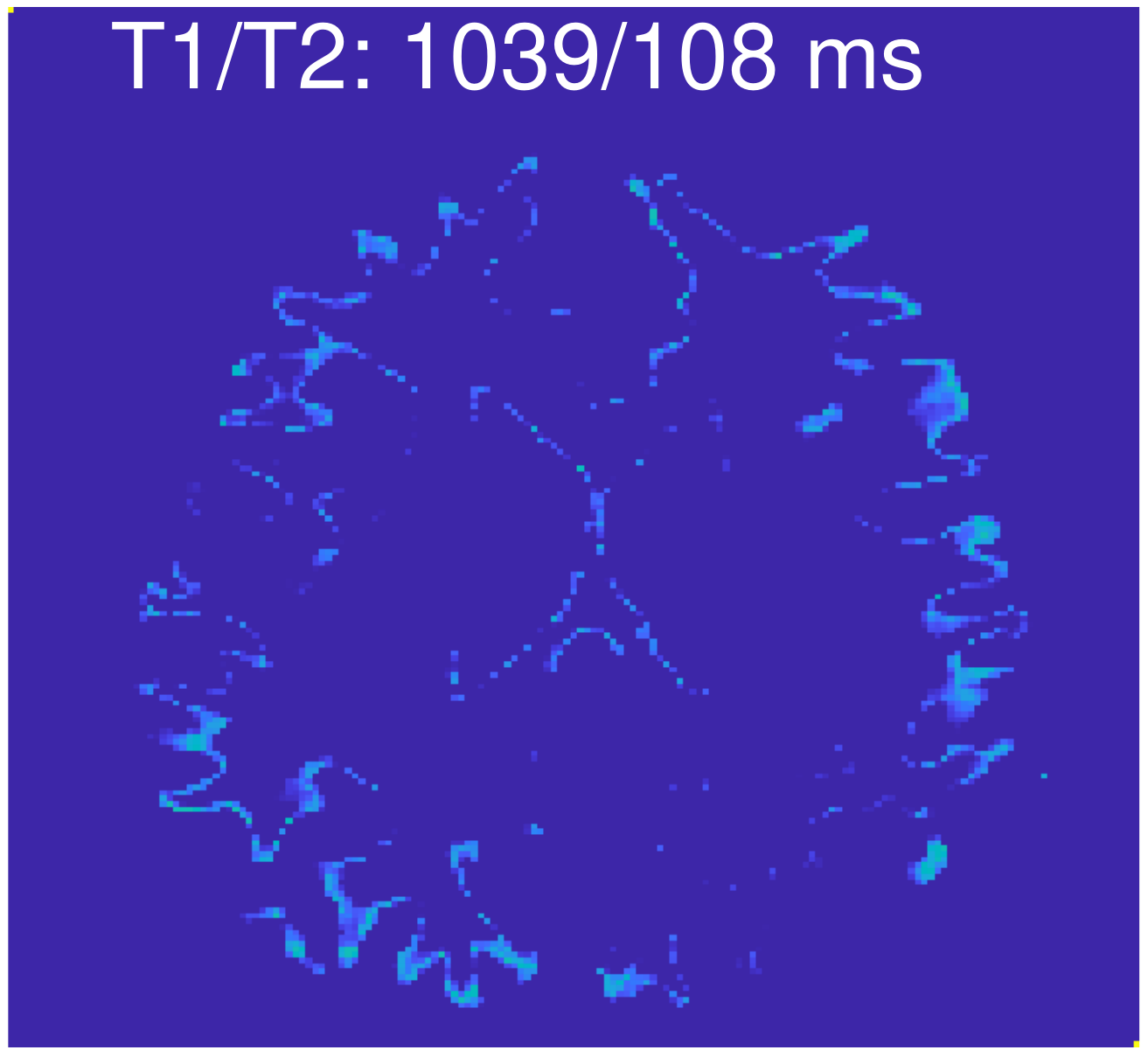}\hspace{-.1cm}
\includegraphics[width=.12\linewidth]{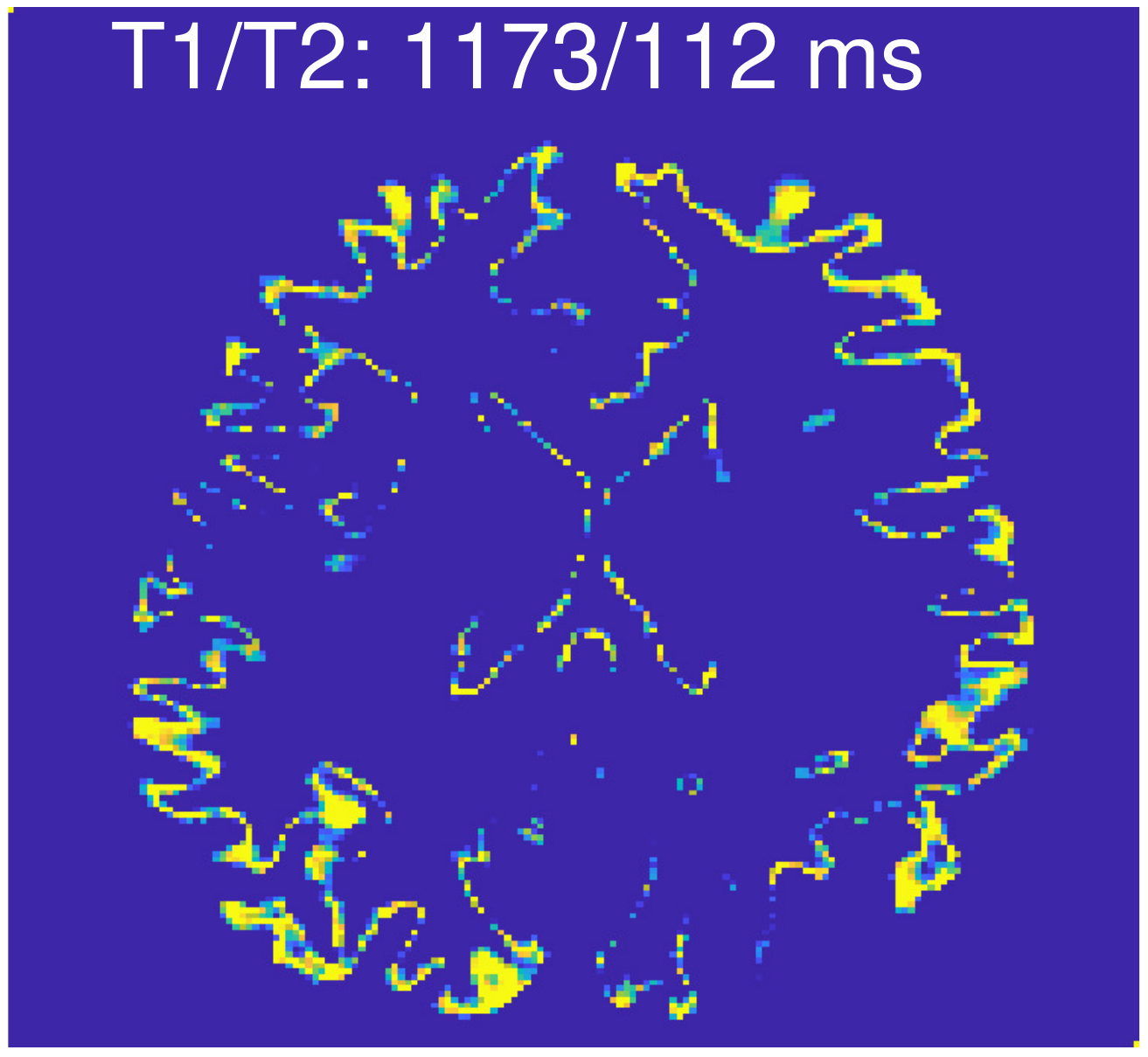}\hspace{-.1cm}
\includegraphics[width=.12\linewidth]{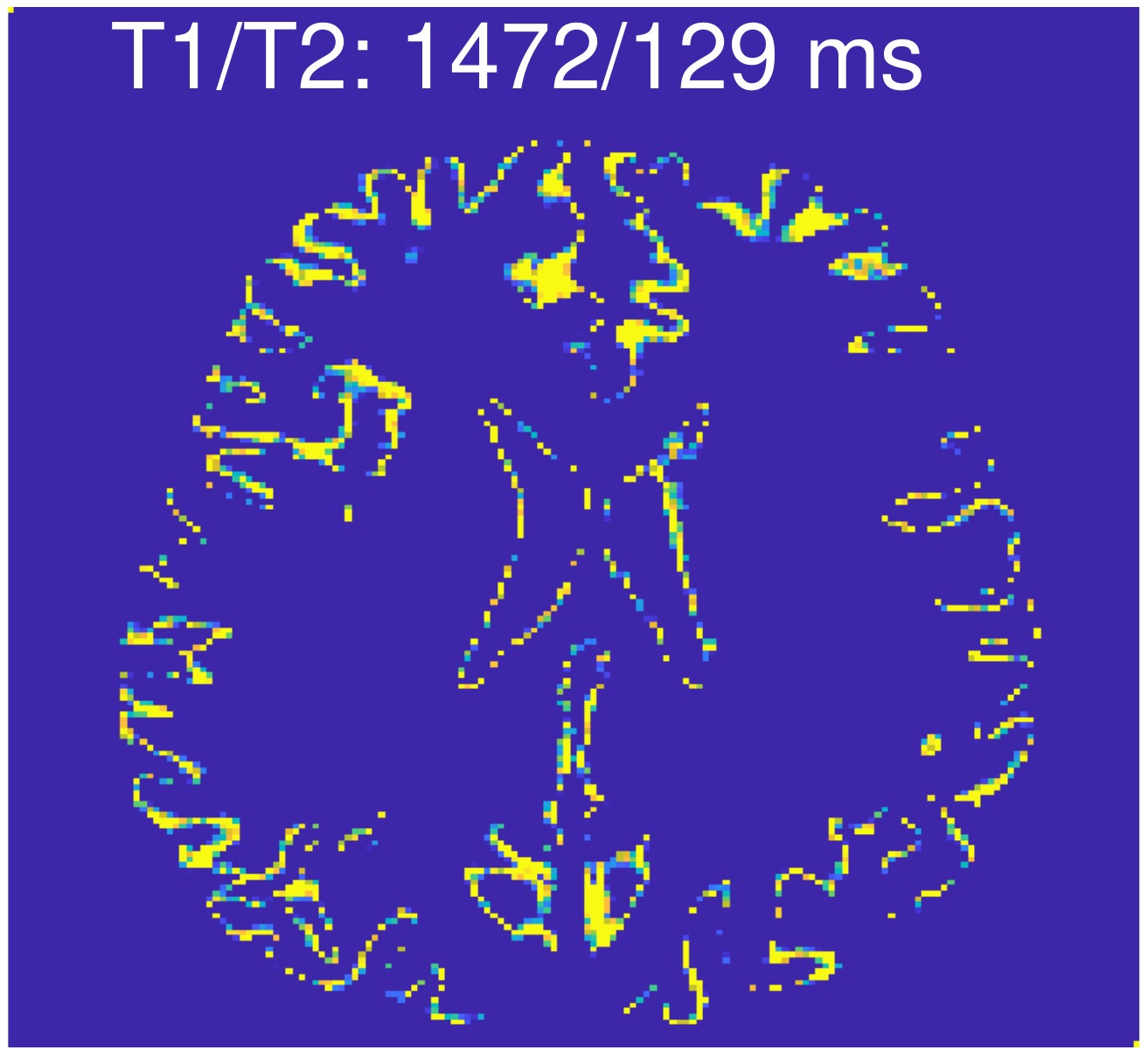}\hspace{-.1cm}
\includegraphics[width=.12\linewidth]{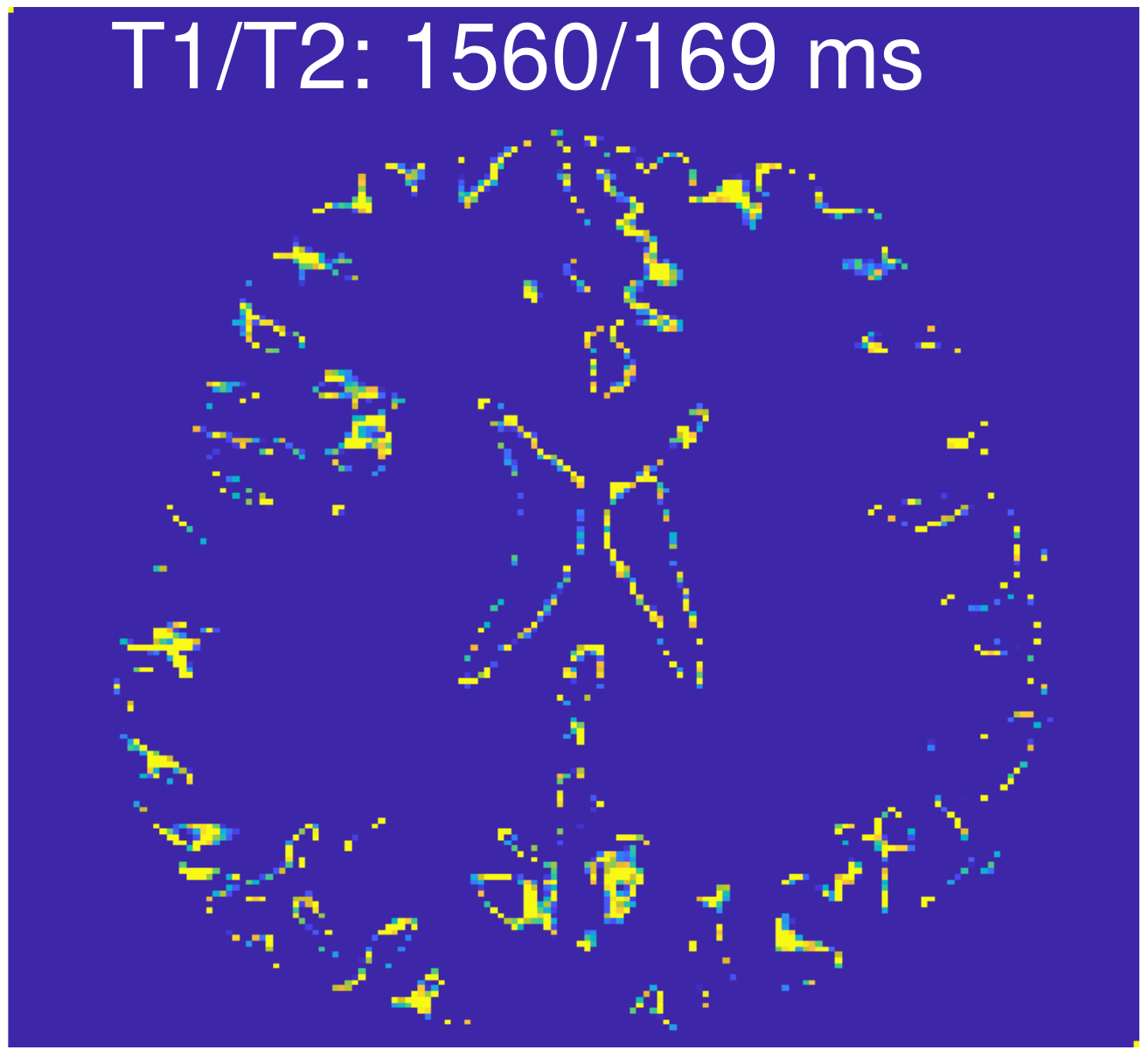}
\\
\hspace{.3cm}
\includegraphics[width=.12\linewidth]{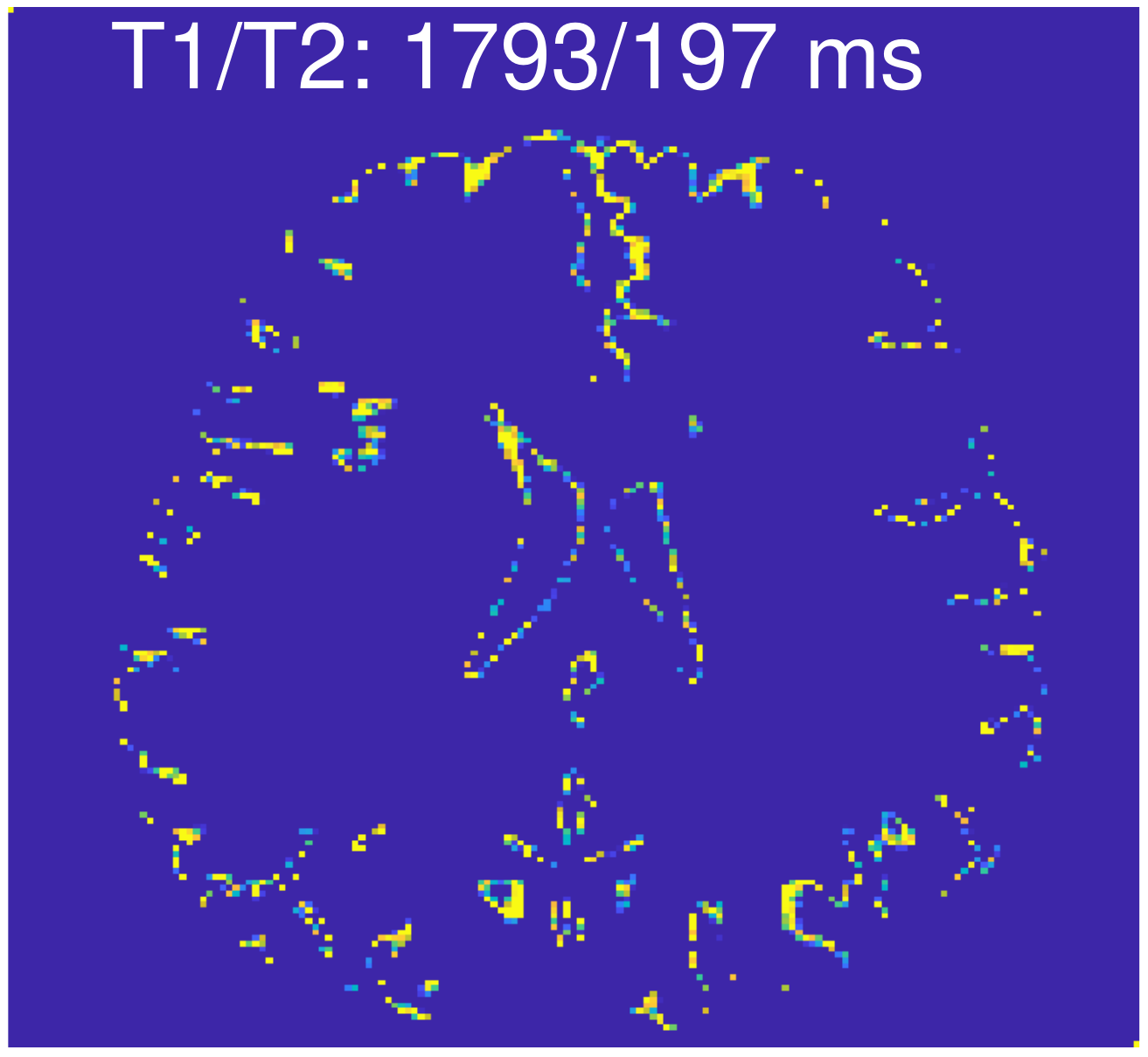}\hspace{-.1cm}
\includegraphics[width=.12\linewidth]{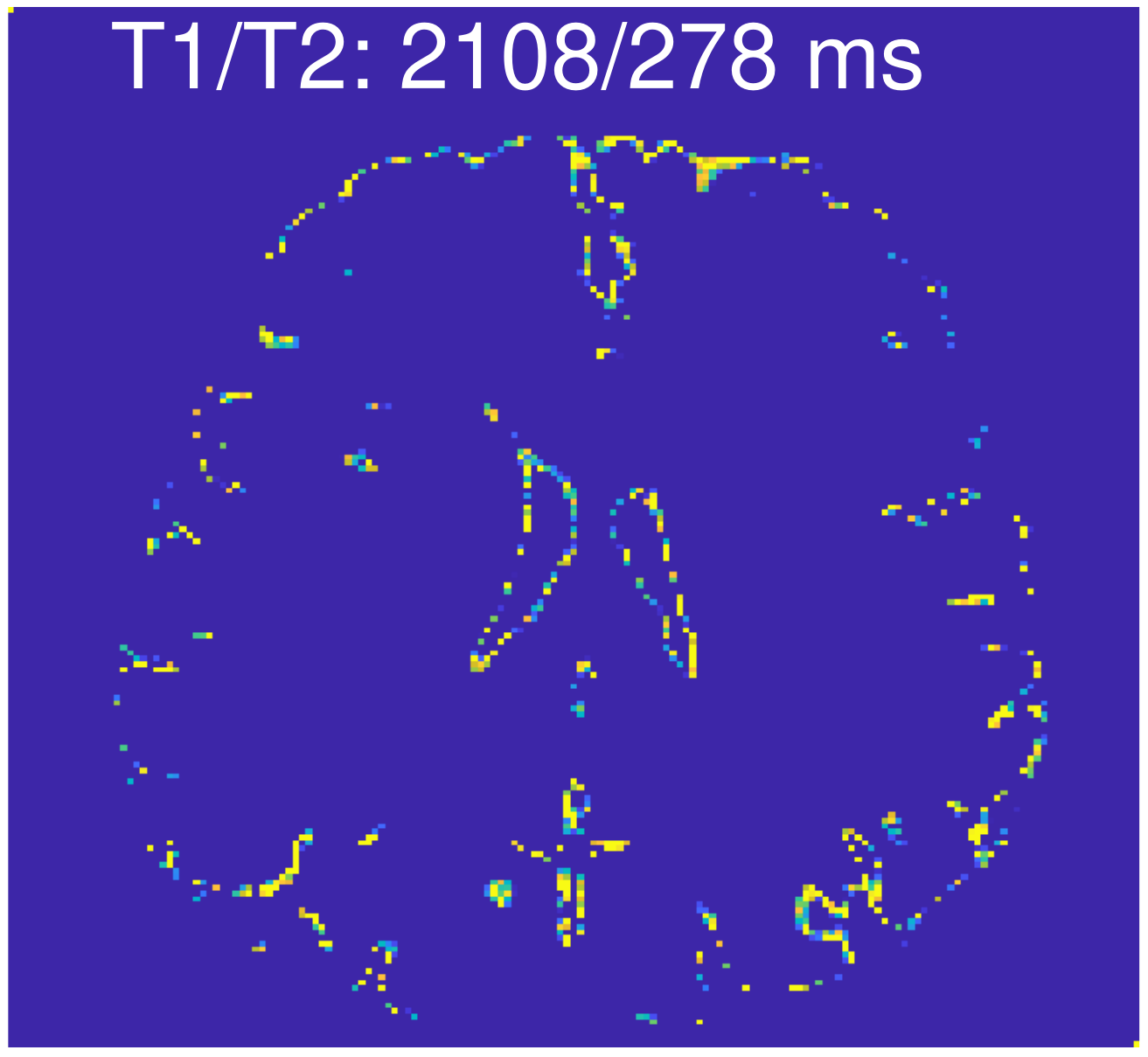}\hspace{-.1cm}
\includegraphics[width=.12\linewidth]{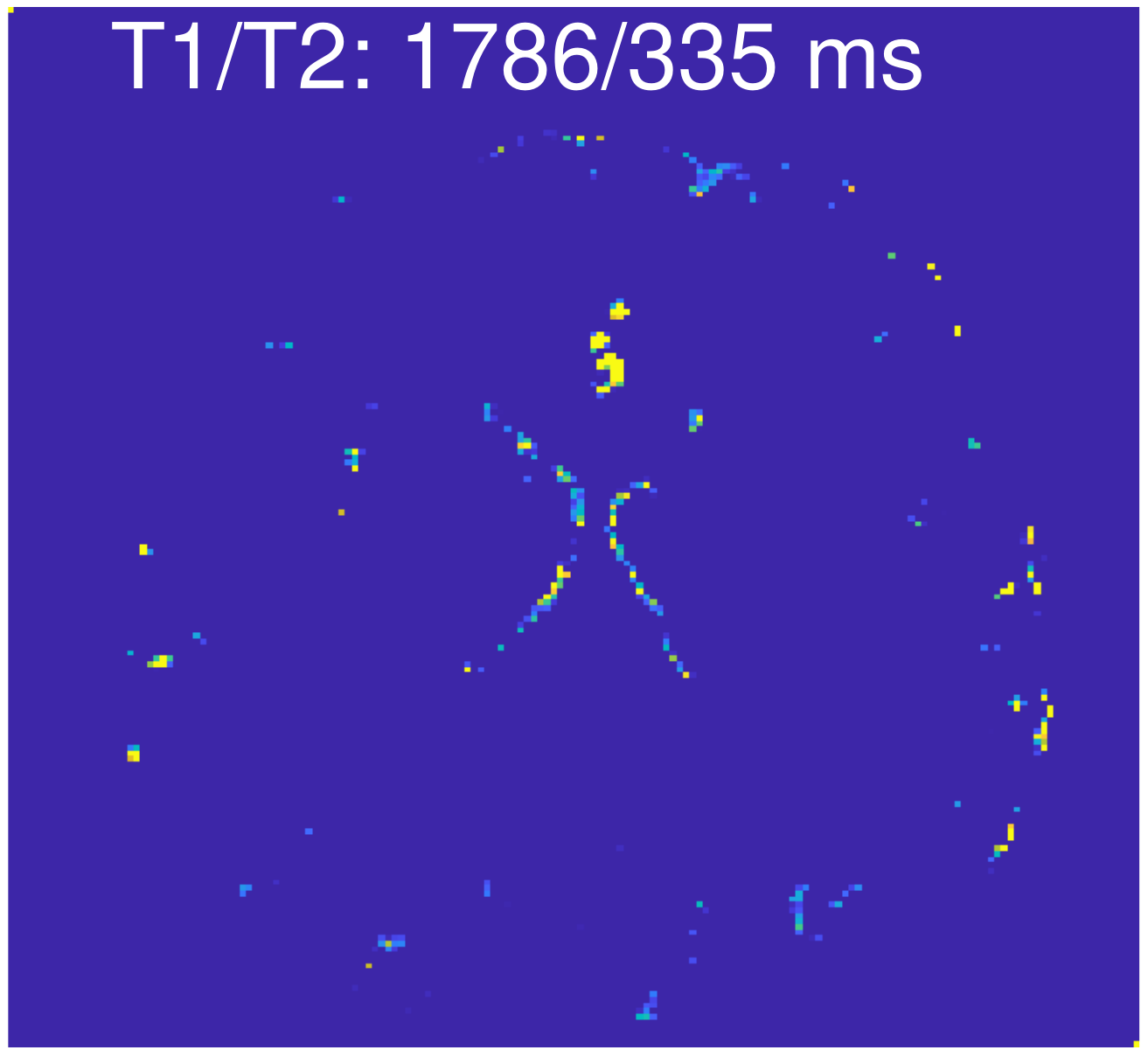}\hspace{-.1cm}
\includegraphics[width=.12\linewidth]{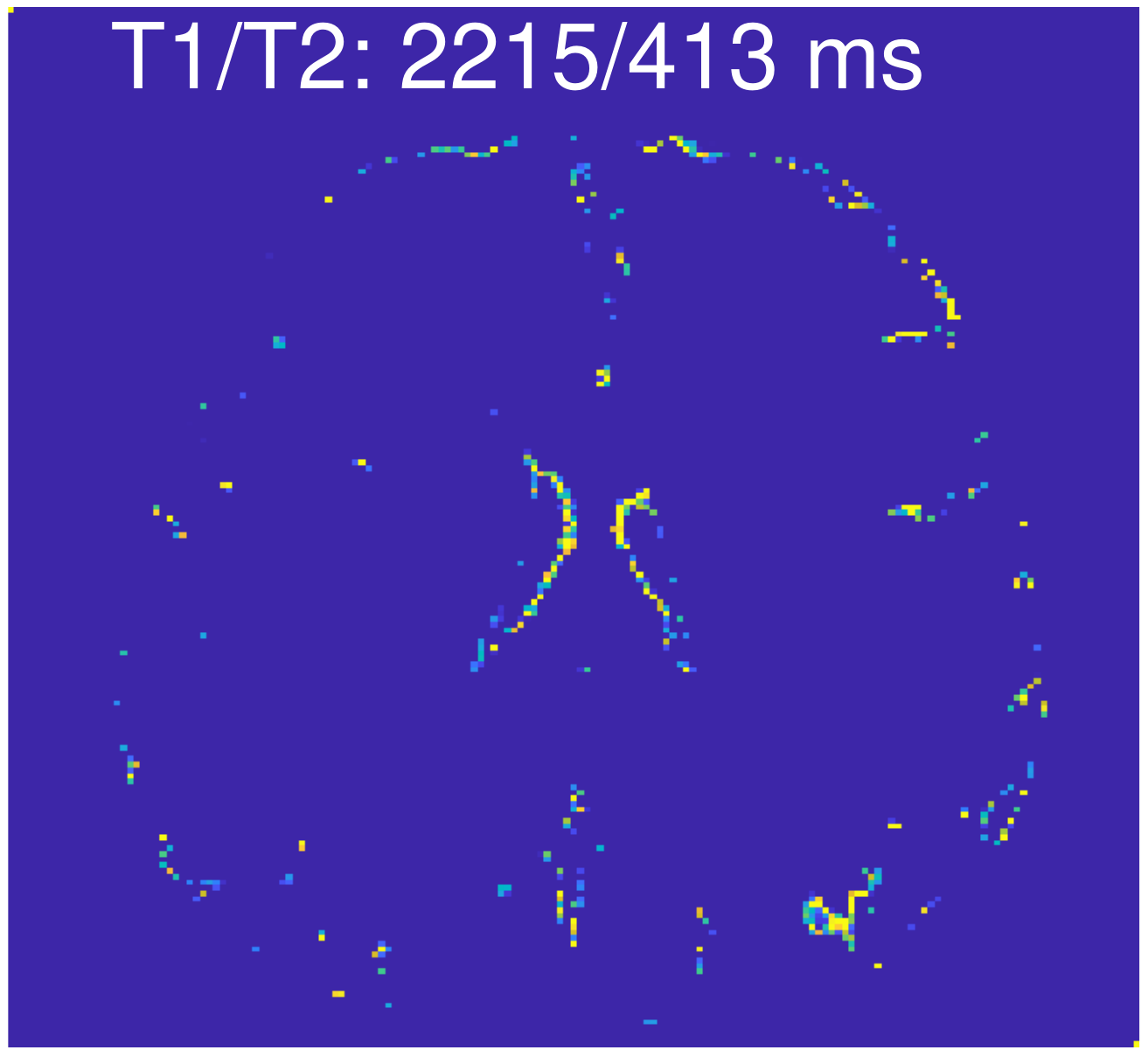}\hspace{-.1cm}
\includegraphics[width=.12\linewidth]{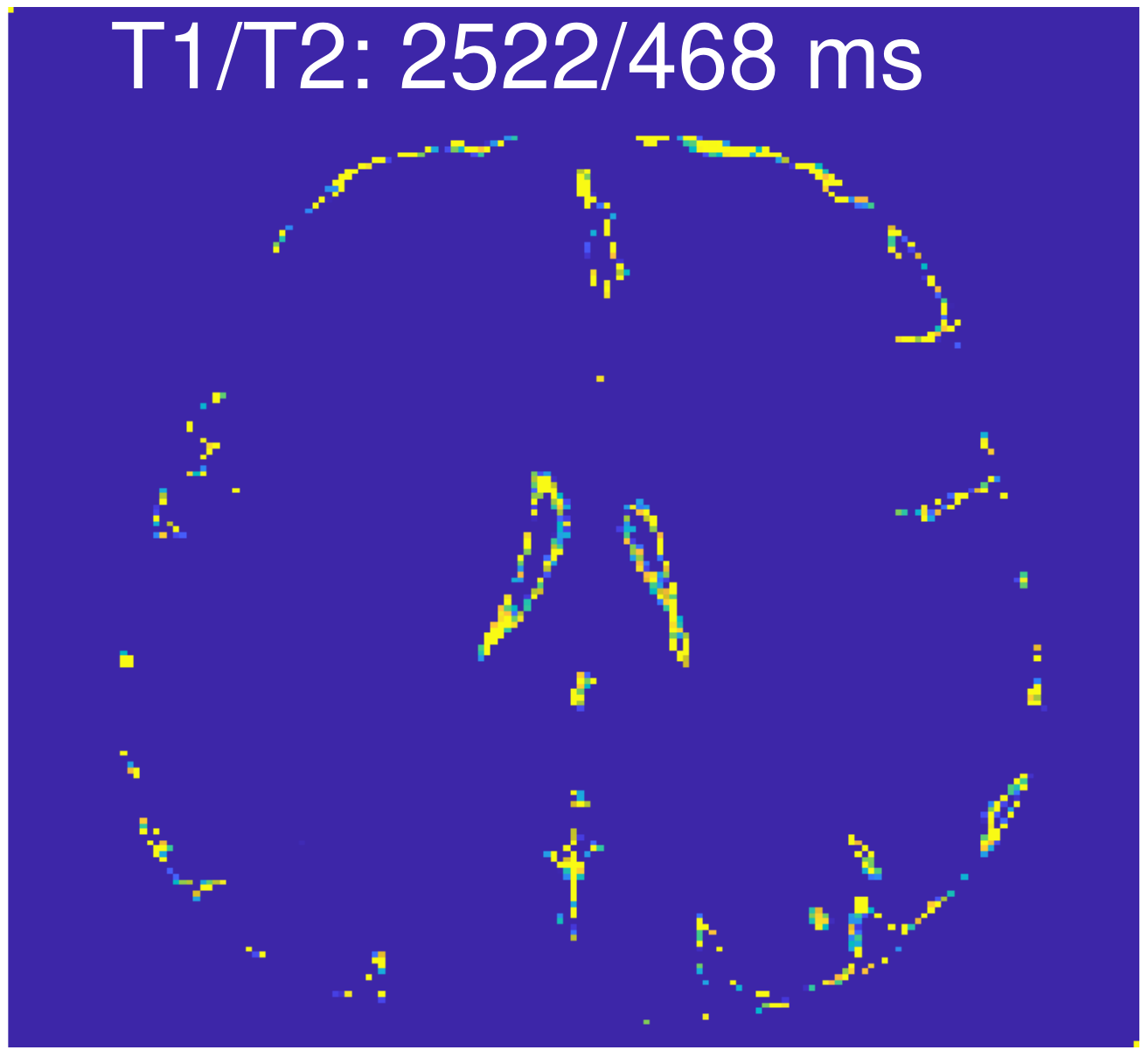}\hspace{-.1cm}
\includegraphics[width=.12\linewidth]{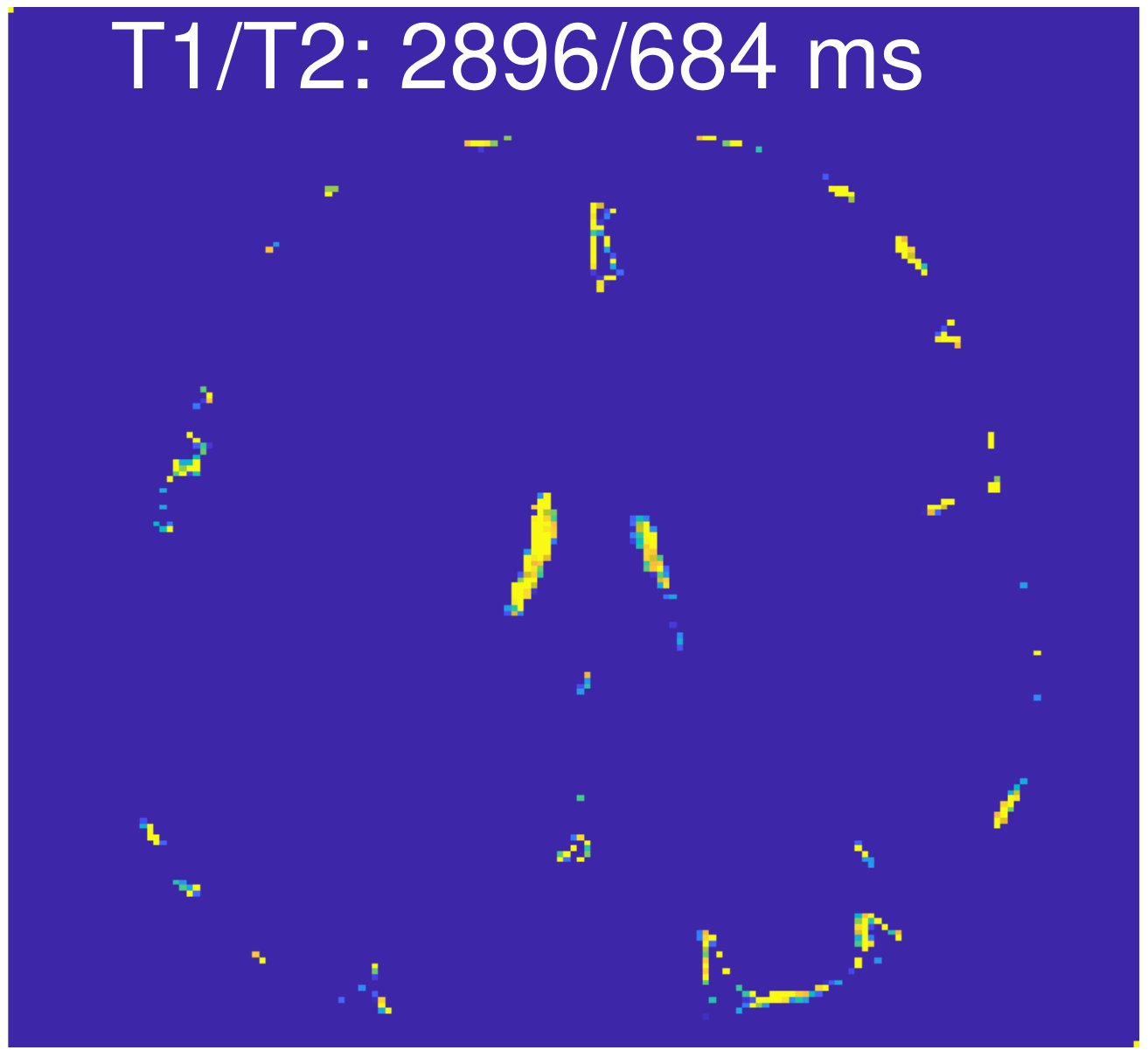}\hspace{-.1cm}
\includegraphics[width=.12\linewidth]{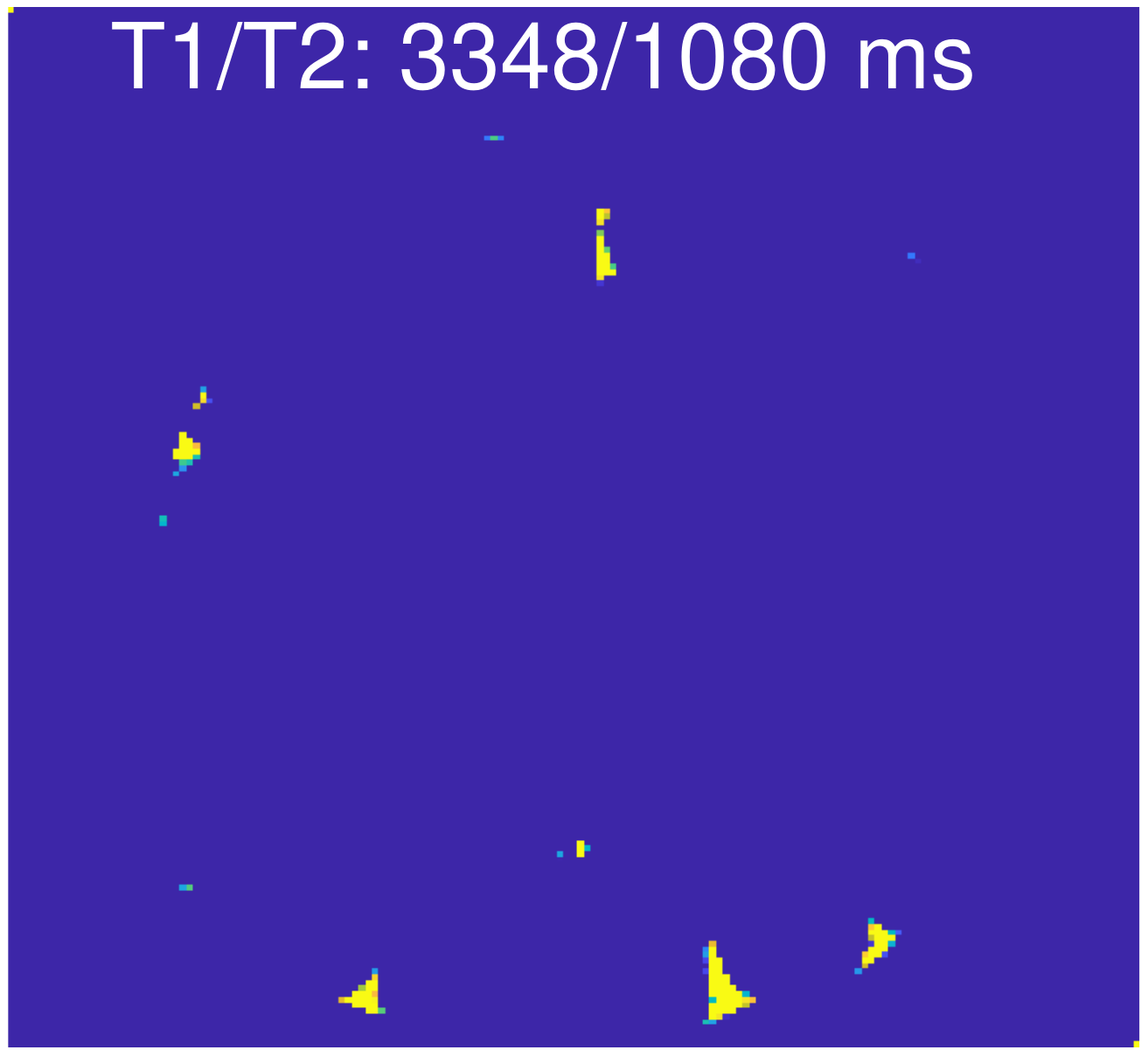}\hspace{-.1cm}
\\
\vspace{.05cm}
\hrule
\hrule
\vspace{.05cm}
\begin{turn}{90} \, $\mathbf{\beta=0.001}$ \end{turn}
\includegraphics[width=.12\linewidth]{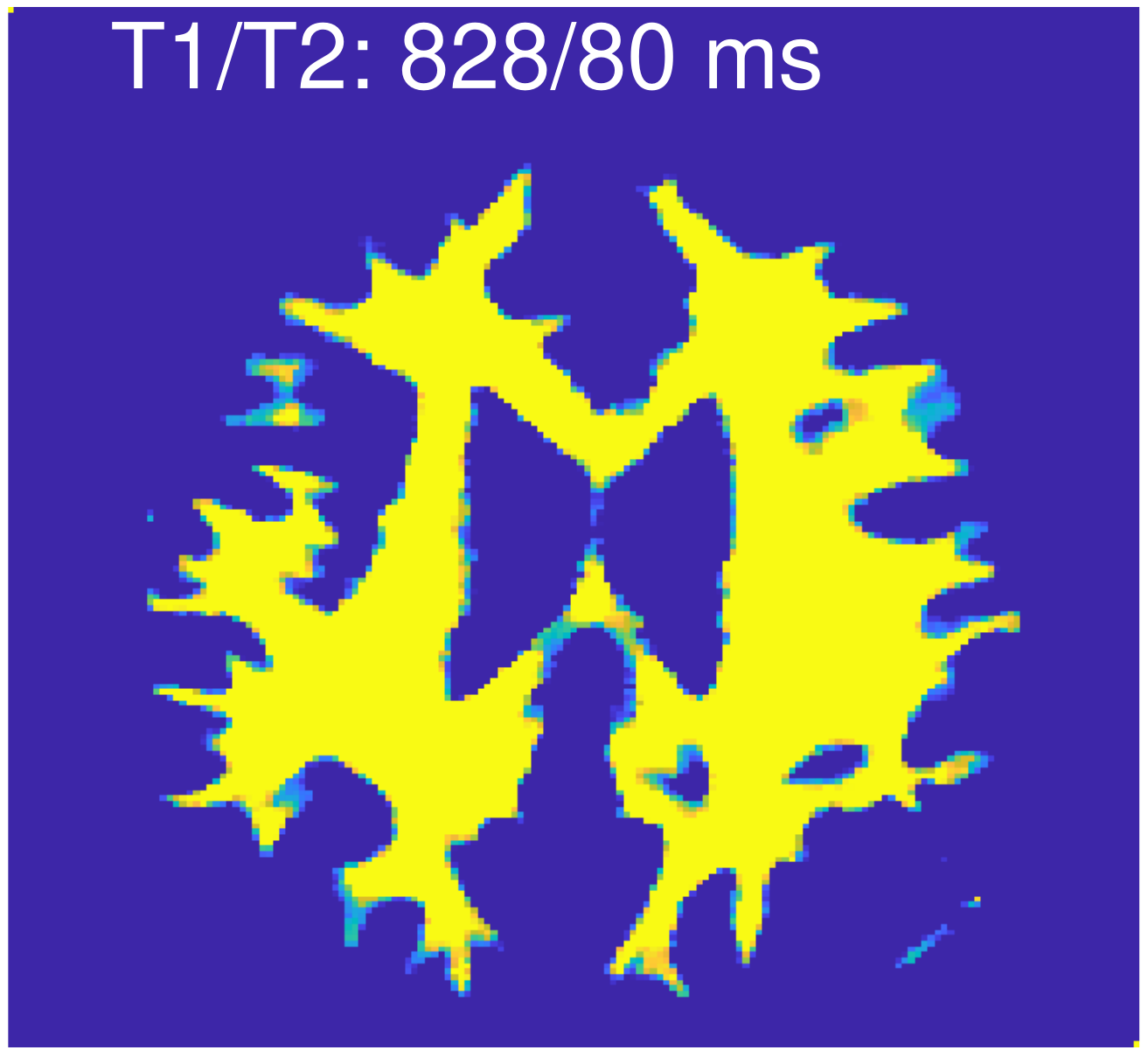}\hspace{-.1cm}
\includegraphics[width=.12\linewidth]{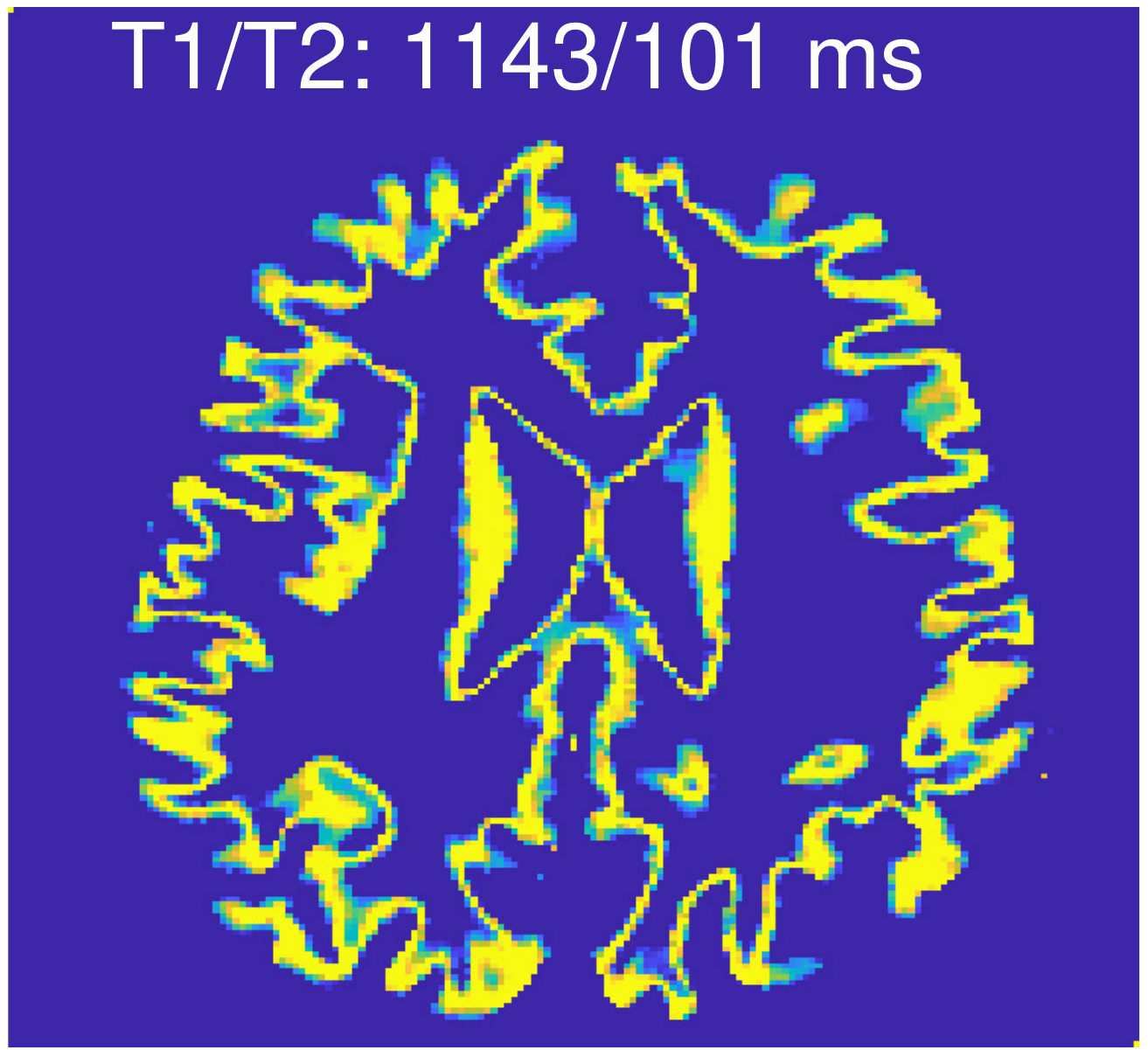}\hspace{-.1cm}
\includegraphics[width=.12\linewidth]{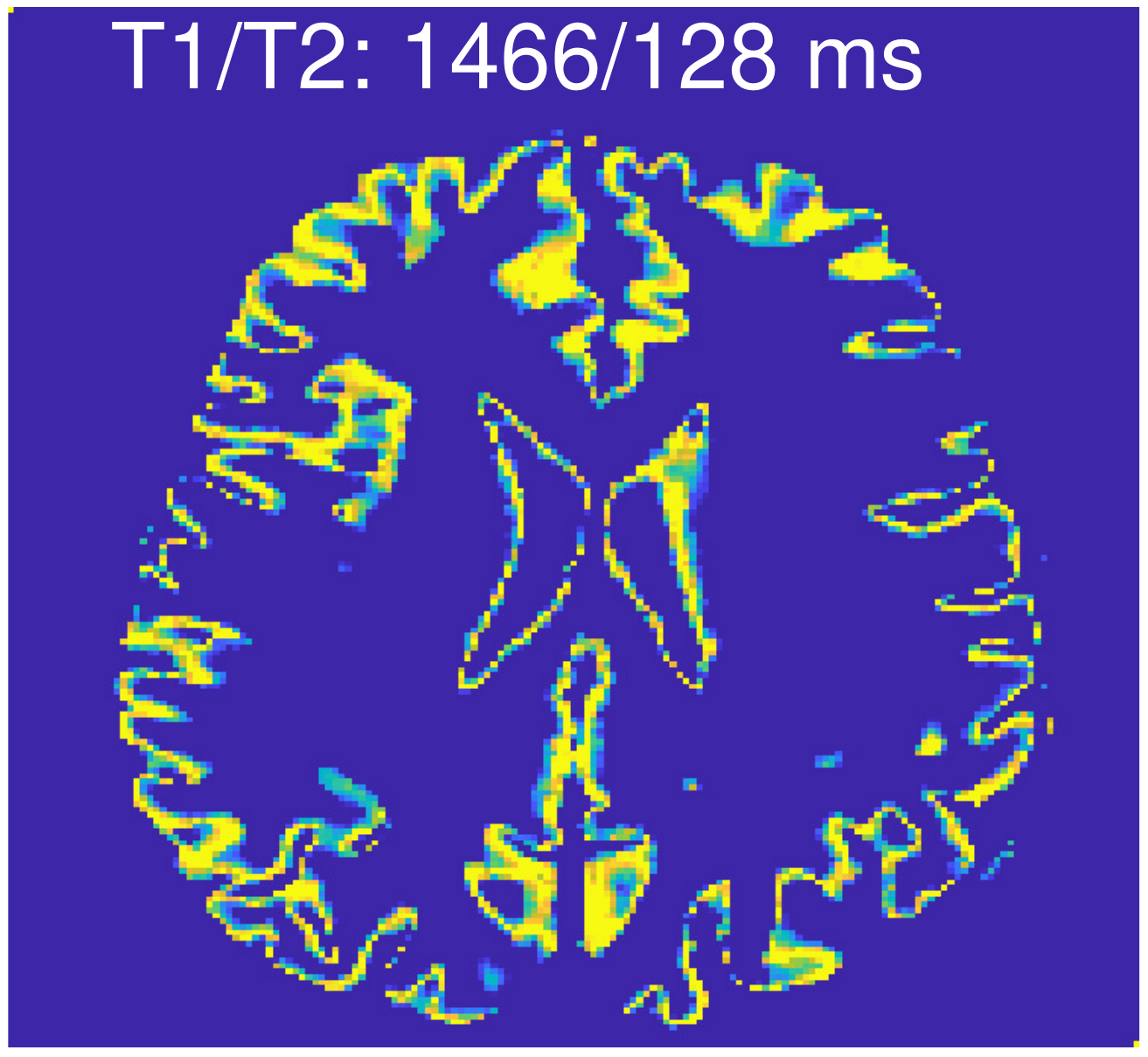}\hspace{-.1cm}
\includegraphics[width=.12\linewidth]{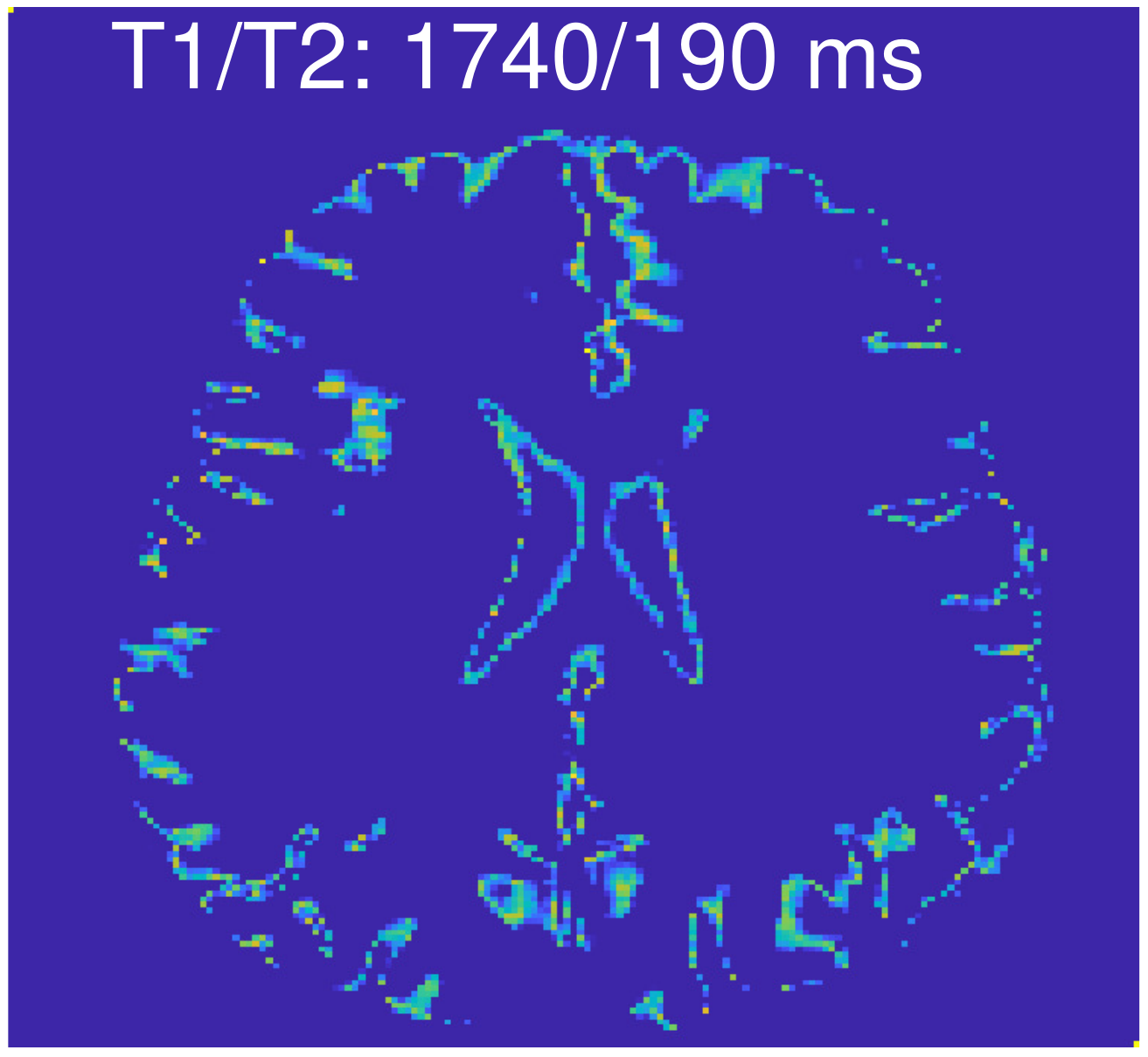}\hspace{-.1cm}
\includegraphics[width=.12\linewidth]{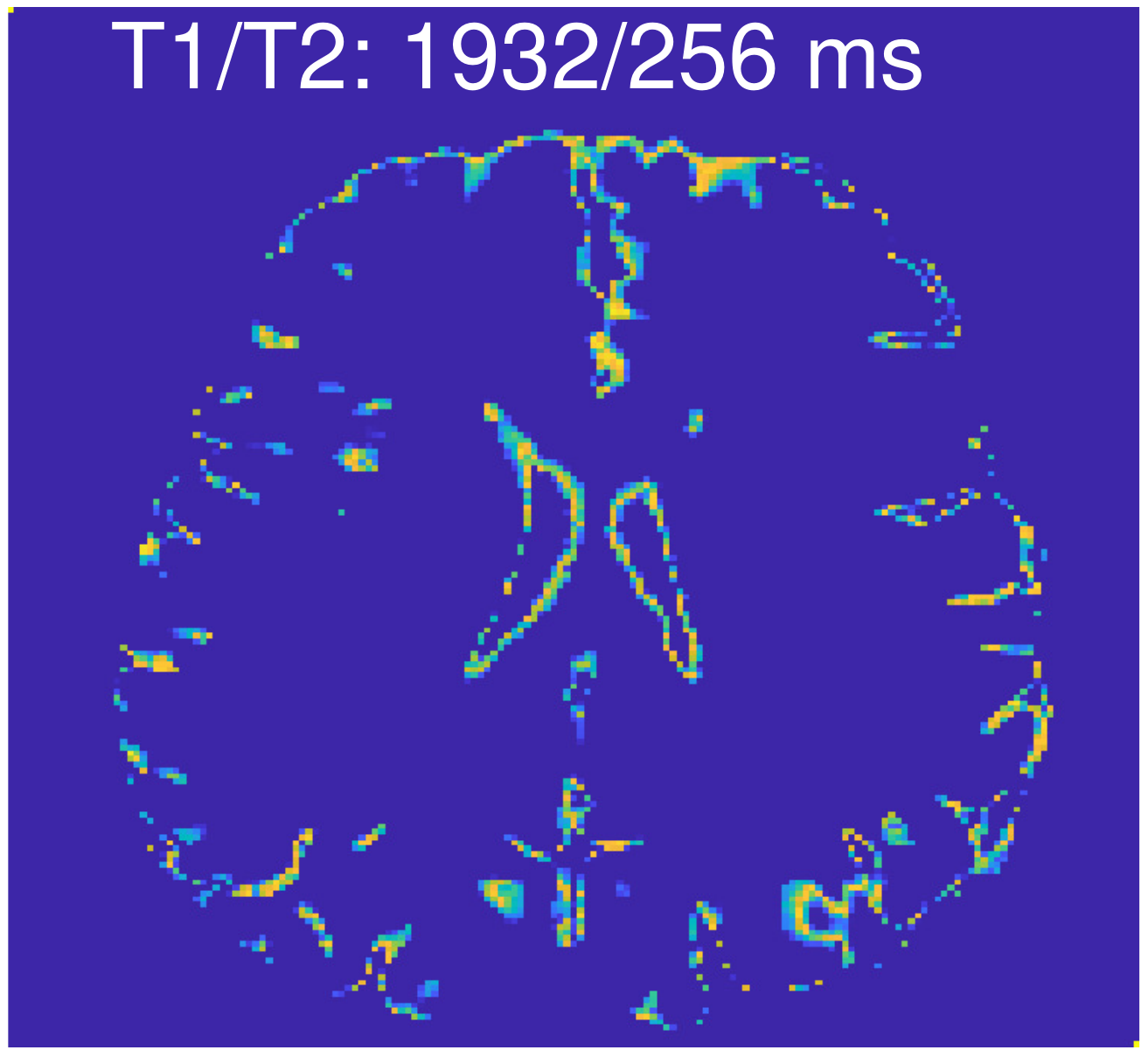}\hspace{-.1cm}
\includegraphics[width=.12\linewidth]{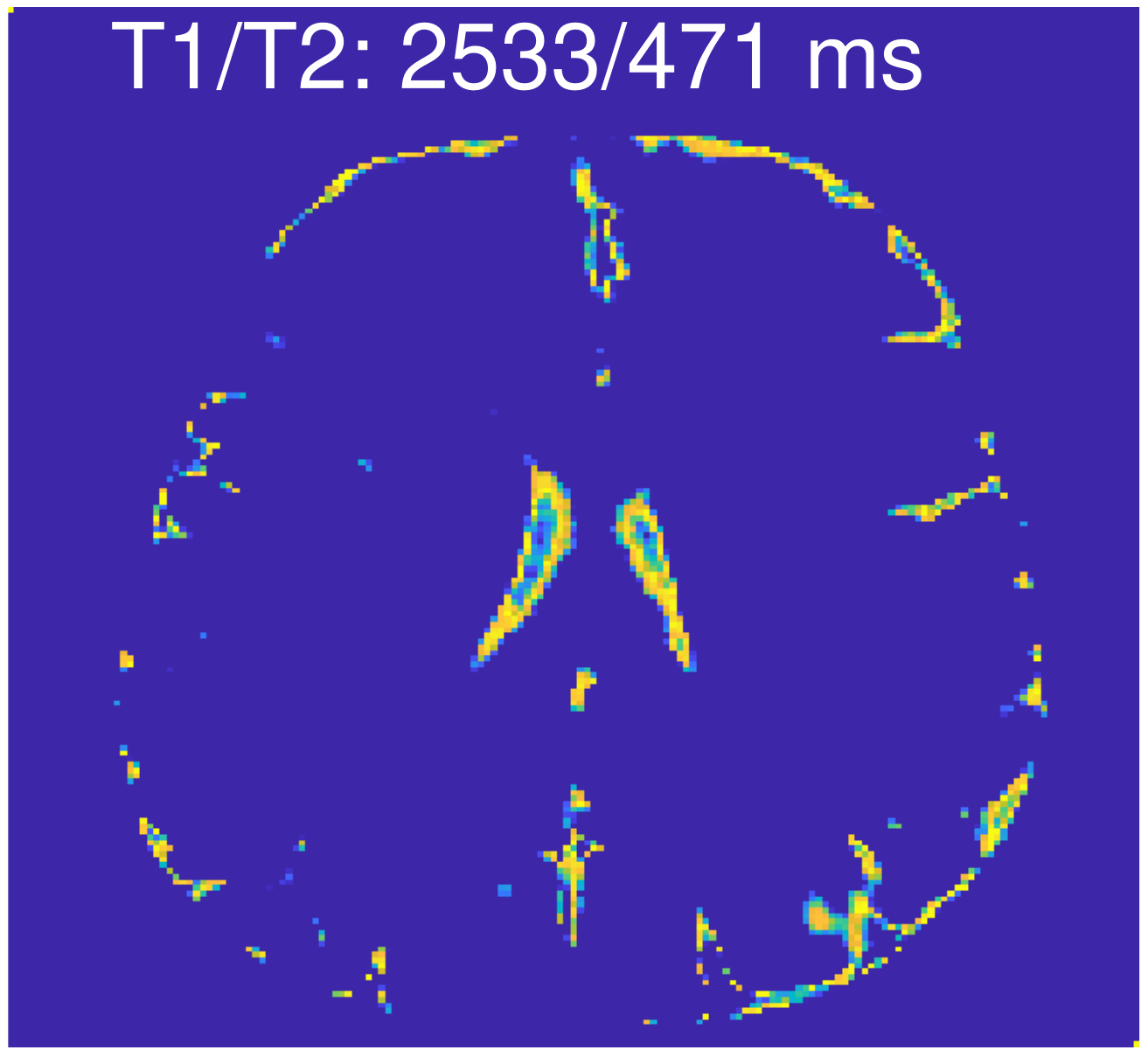}\hspace{-.1cm}
\includegraphics[width=.12\linewidth]{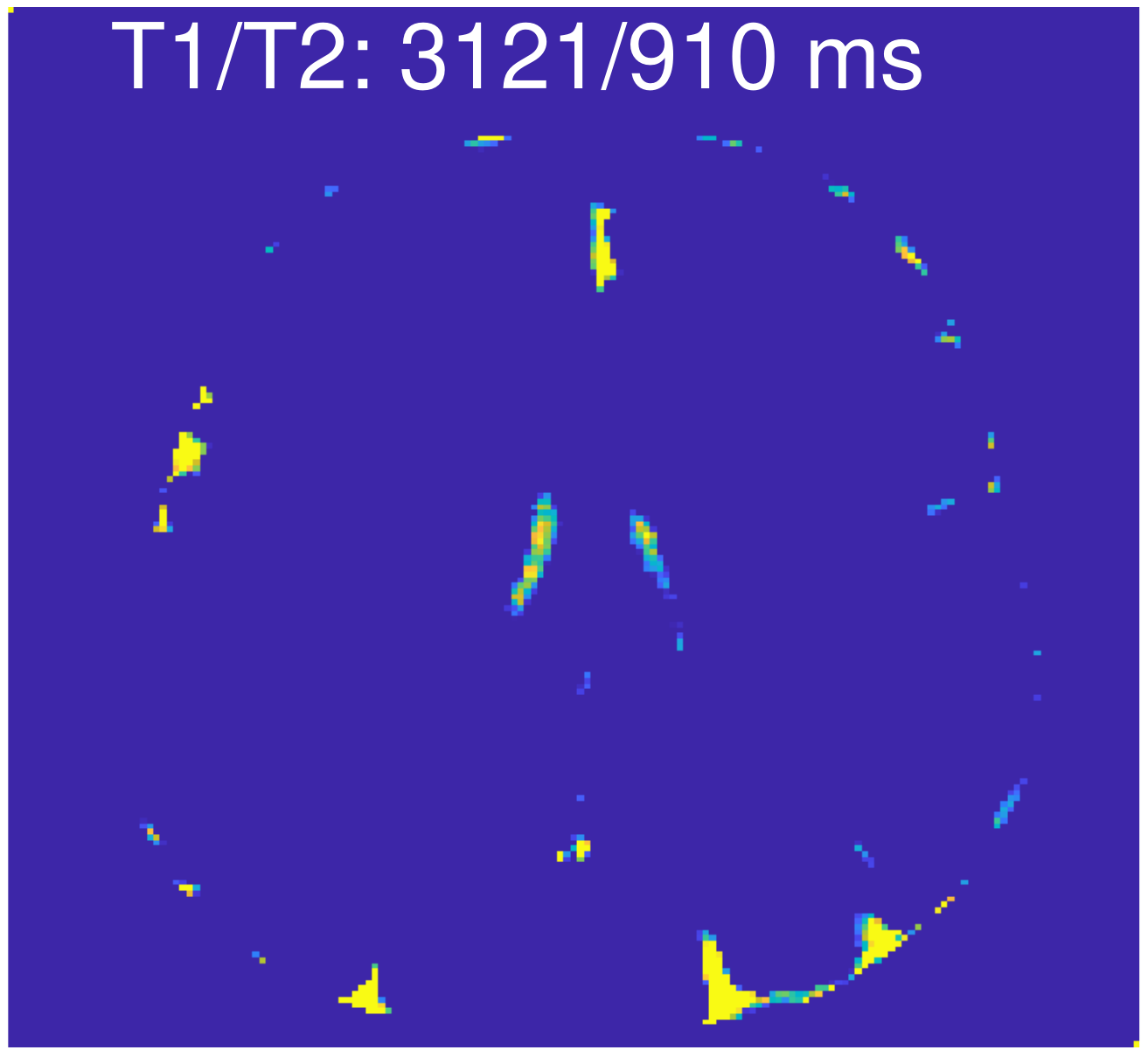}\hspace{-.1cm}
\\
\vspace{.05cm}
\hrule
\hrule
\vspace{.05cm}
\begin{turn}{90} \quad $\beta=0.01$ \end{turn}
\includegraphics[width=.12\linewidth]{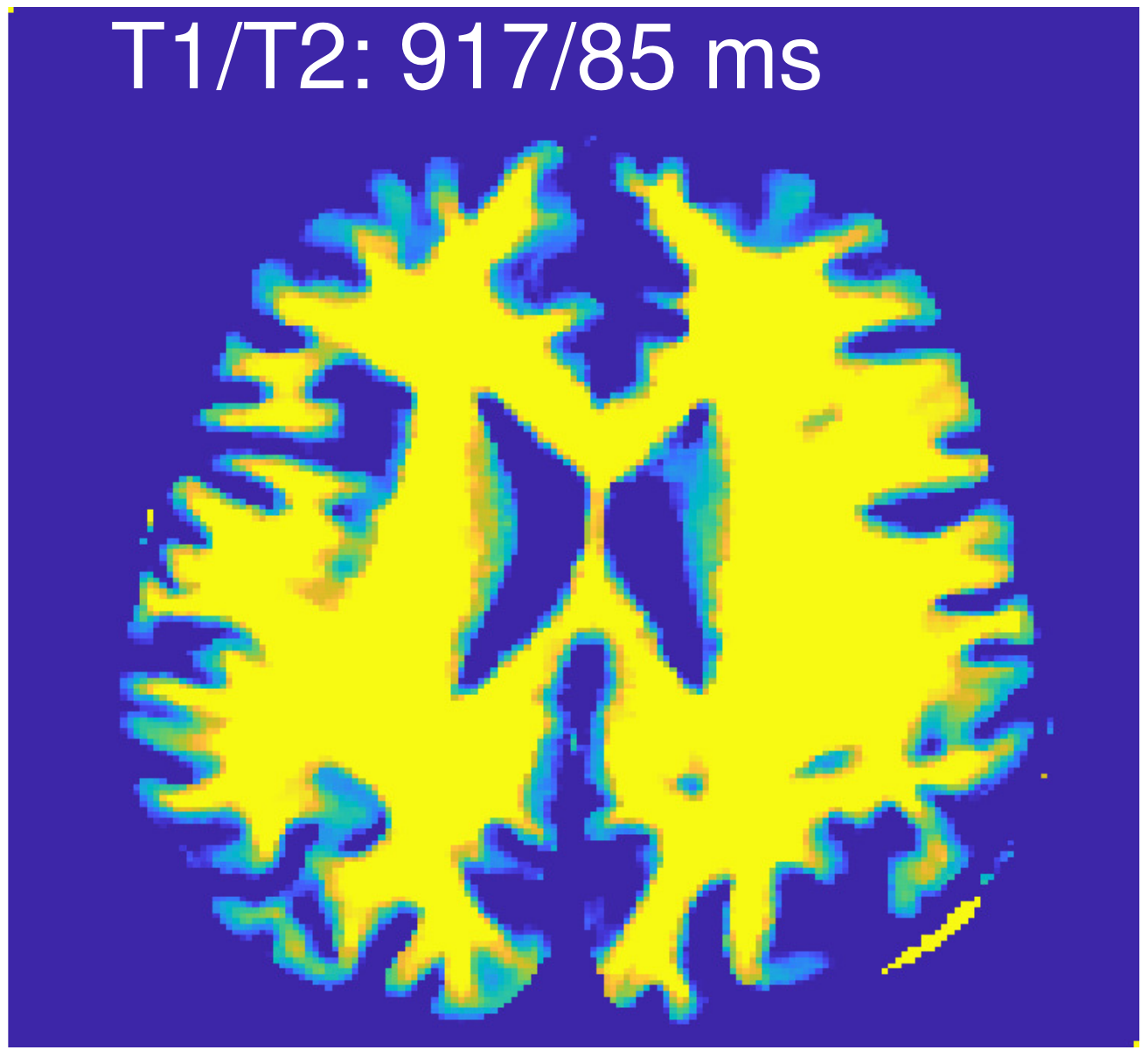}\hspace{-.1cm}
\includegraphics[width=.12\linewidth]{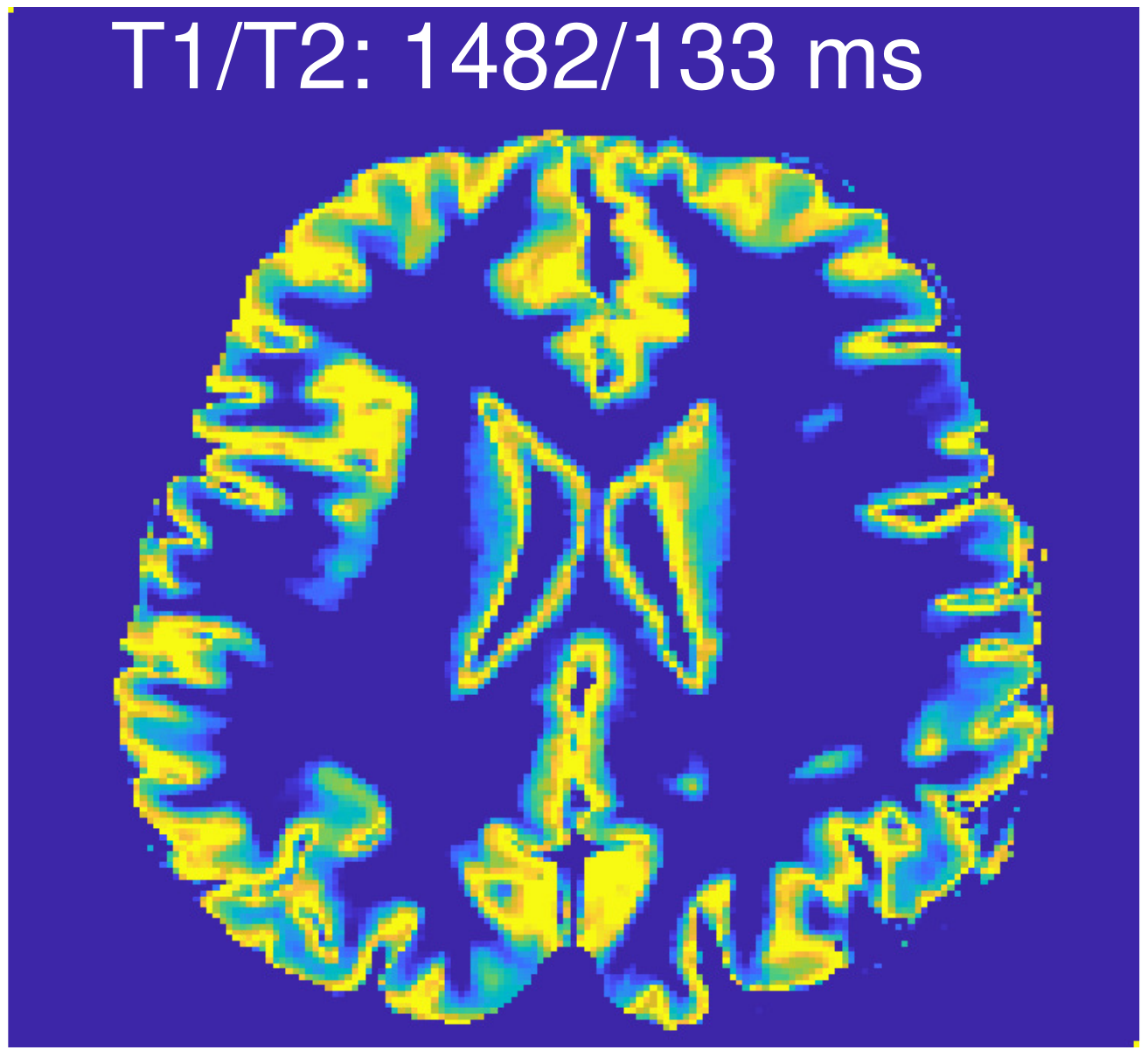}\hspace{-.1cm}
\includegraphics[width=.12\linewidth]{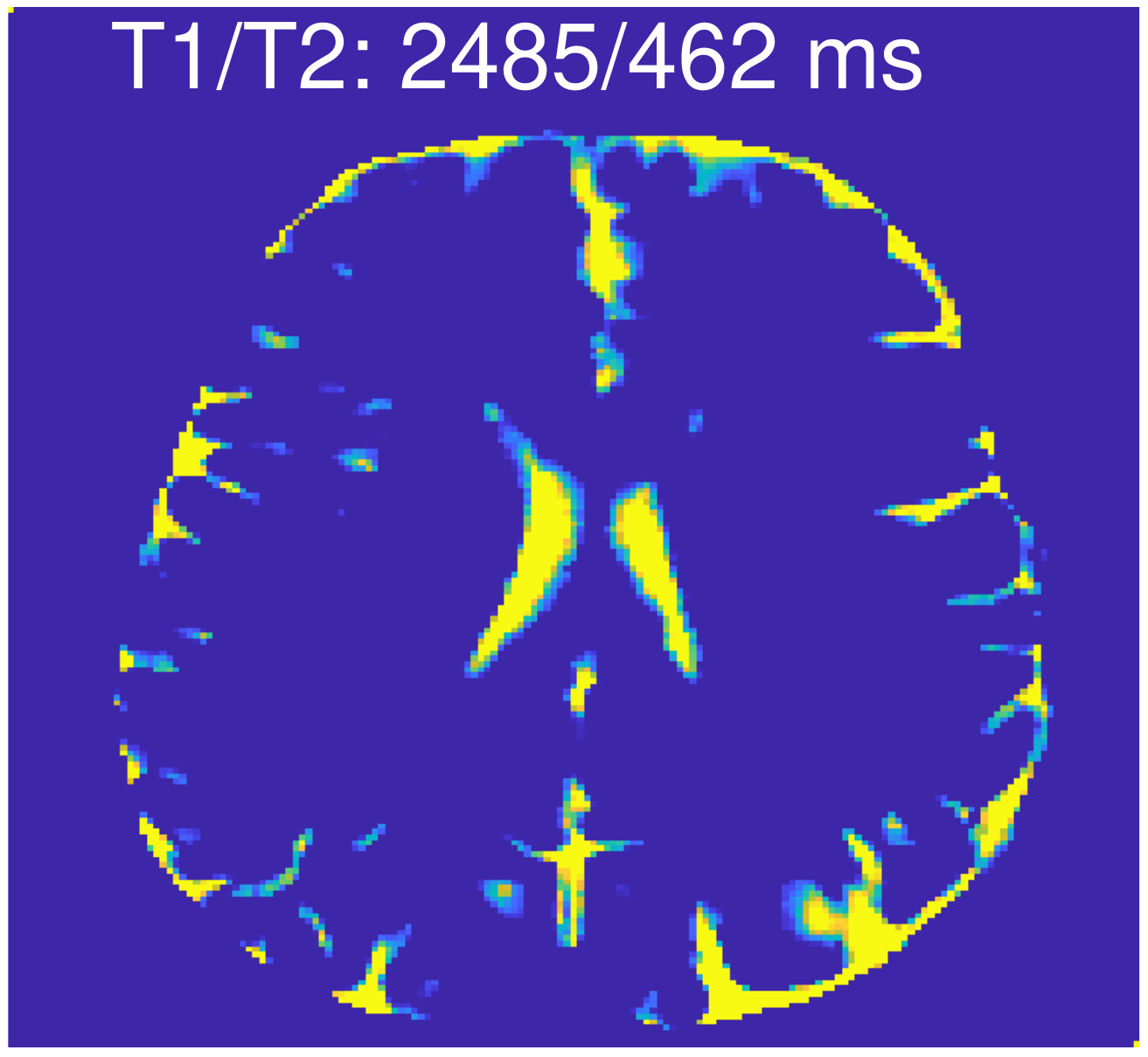}\hspace{-.1cm}
\qquad\quad
\begin{turn}{90} \quad $\beta=0.05$ \end{turn}
\includegraphics[width=.12\linewidth]{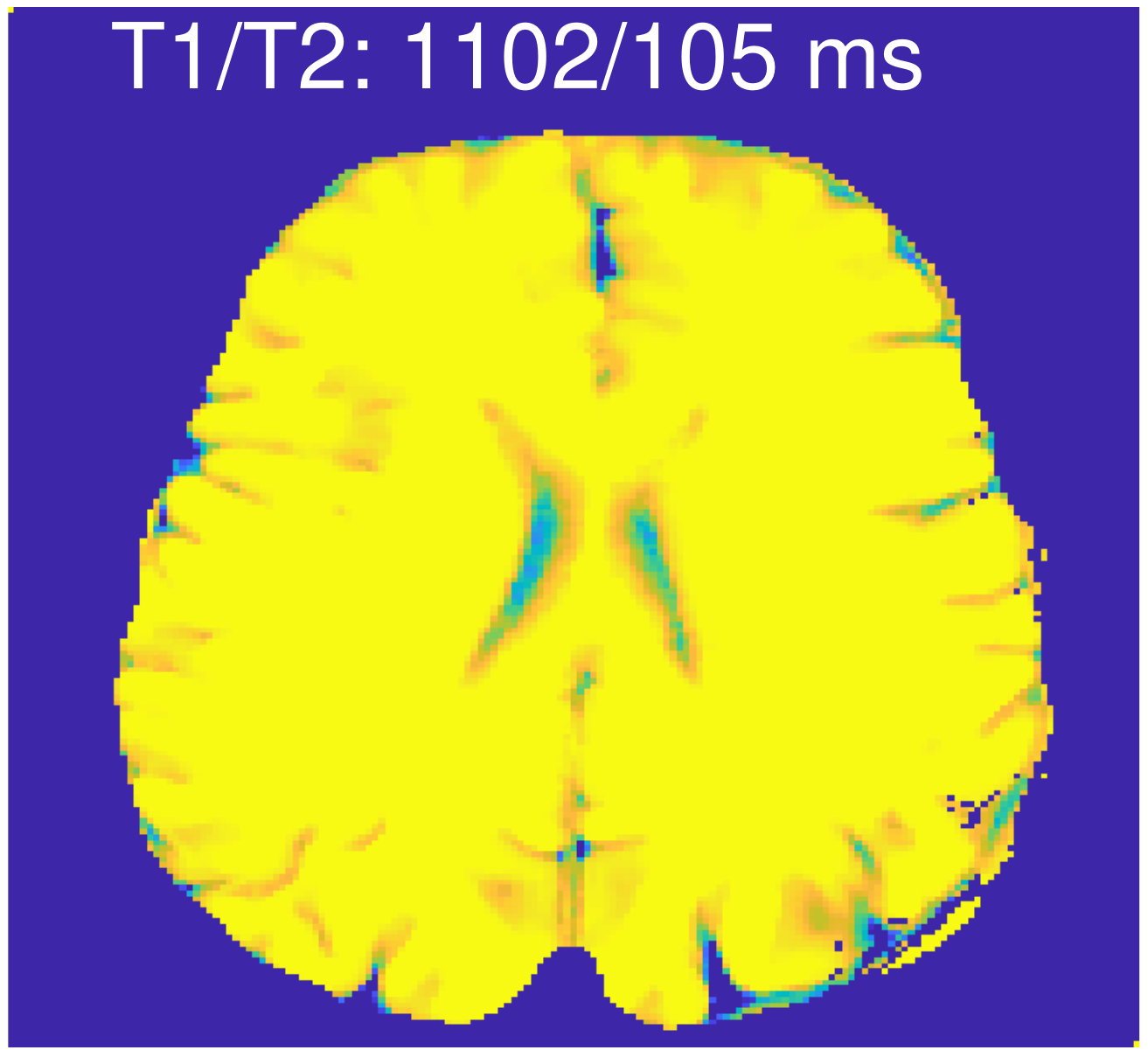}\hspace{-.1cm}
\includegraphics[width=.12\linewidth]{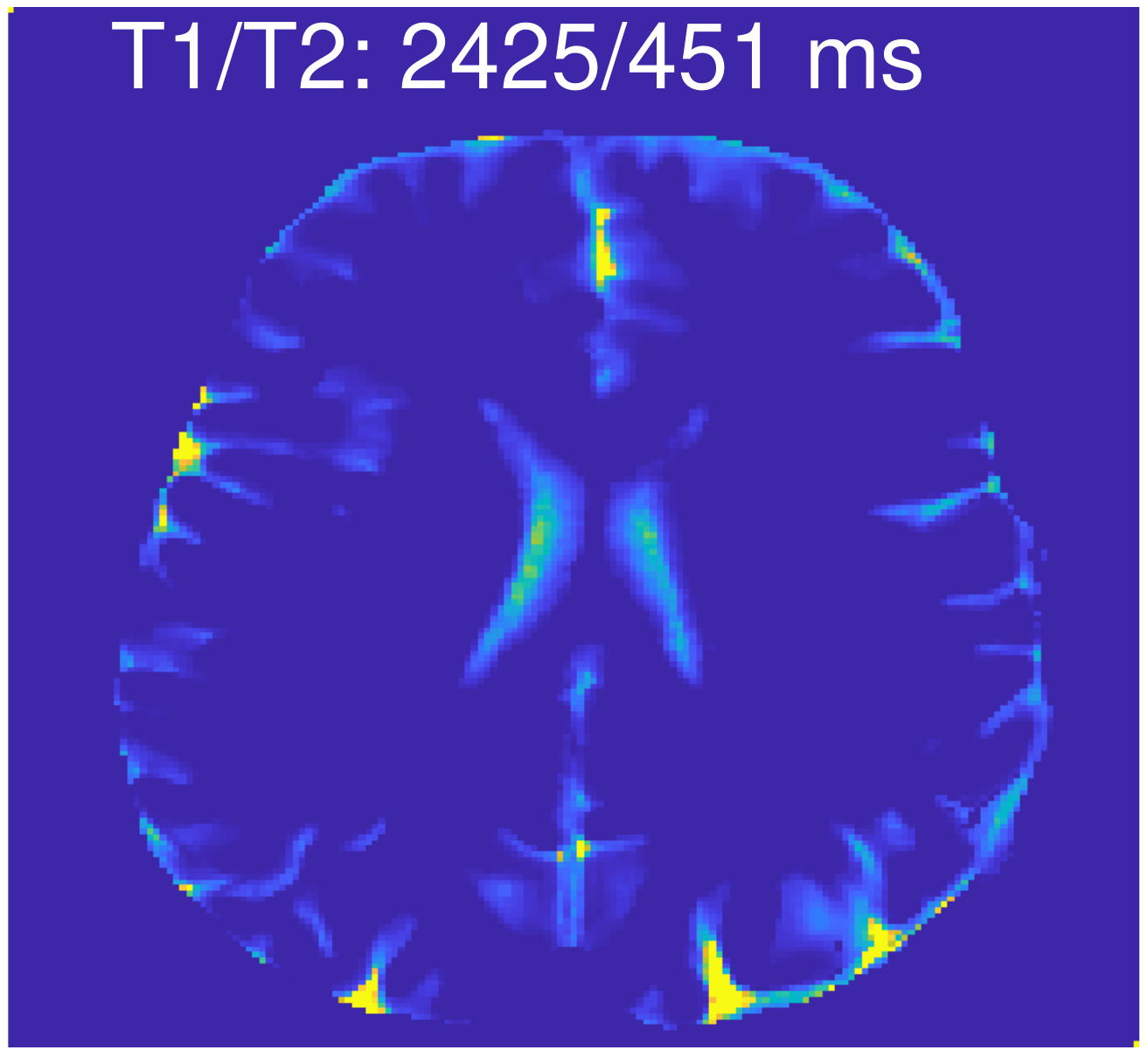}
\begin{turn}{90} \includegraphics[height=.4cm]{./figs/colorbarMaps_c.pdf}\end{turn}
\end{minipage}} 
\caption{\footnotesize{
Estimated in-vivo compartments i.e. T1/T2 values and mixture maps, using  
\MCMRF (un-thresholded) for four values of $\beta$. 15 out of 22 compartments with highest energies are shown for $\beta=10^{-4}$, while for other $\beta$s all compartments are shown. Small $\beta$ values create fine/detailed decompositions whereas larger $\beta$ values hierarchically cluster the compartments together.
}\label{fig:vivo_fullmixtures} }
\end{figure*}

\begin{figure*}[ht!]
	\centering
	\scalebox{1}{
	\begin{minipage}{\linewidth}
		\centering		
\includegraphics[width=.24\linewidth]{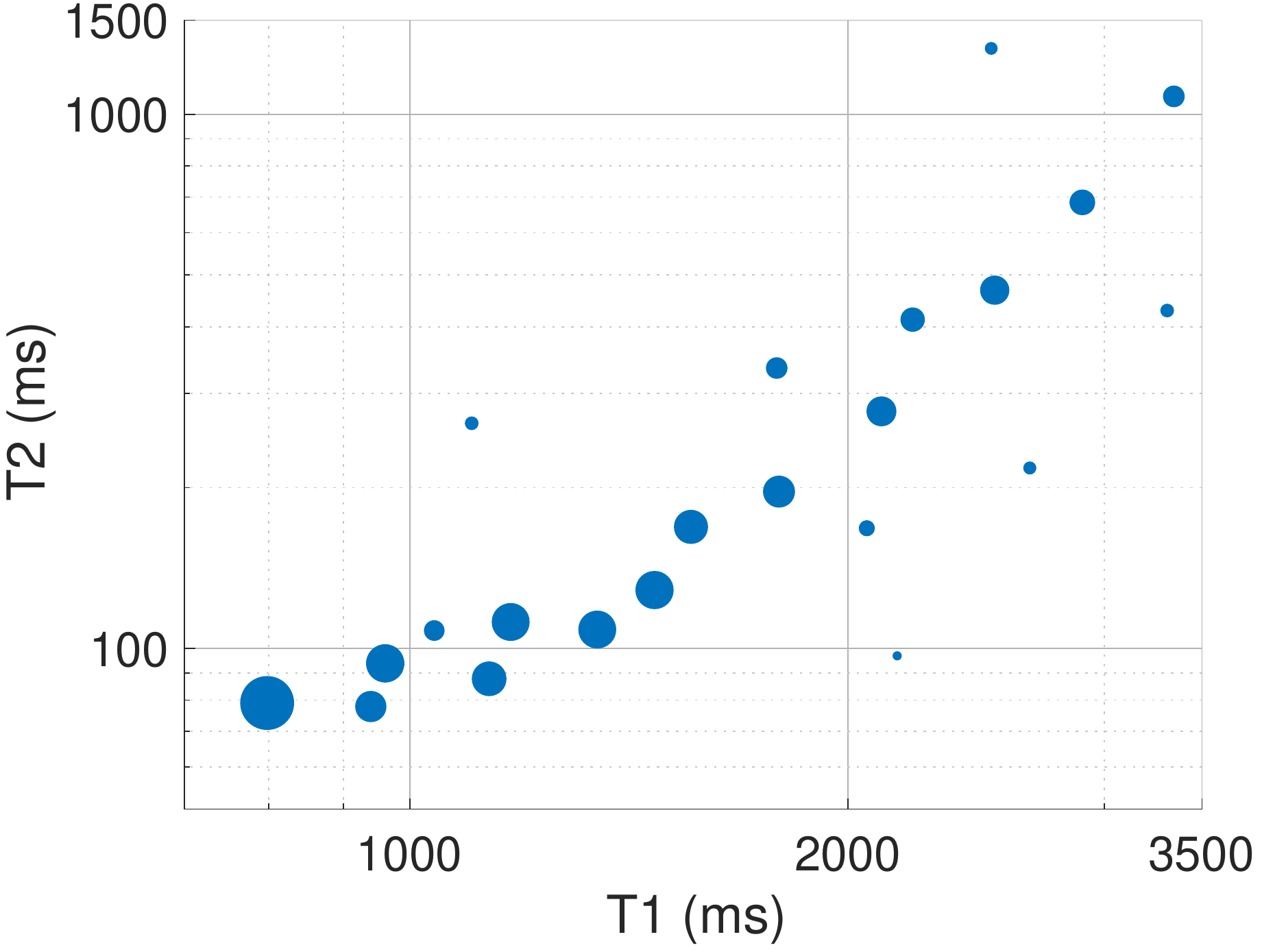}
\includegraphics[width=.24\linewidth]{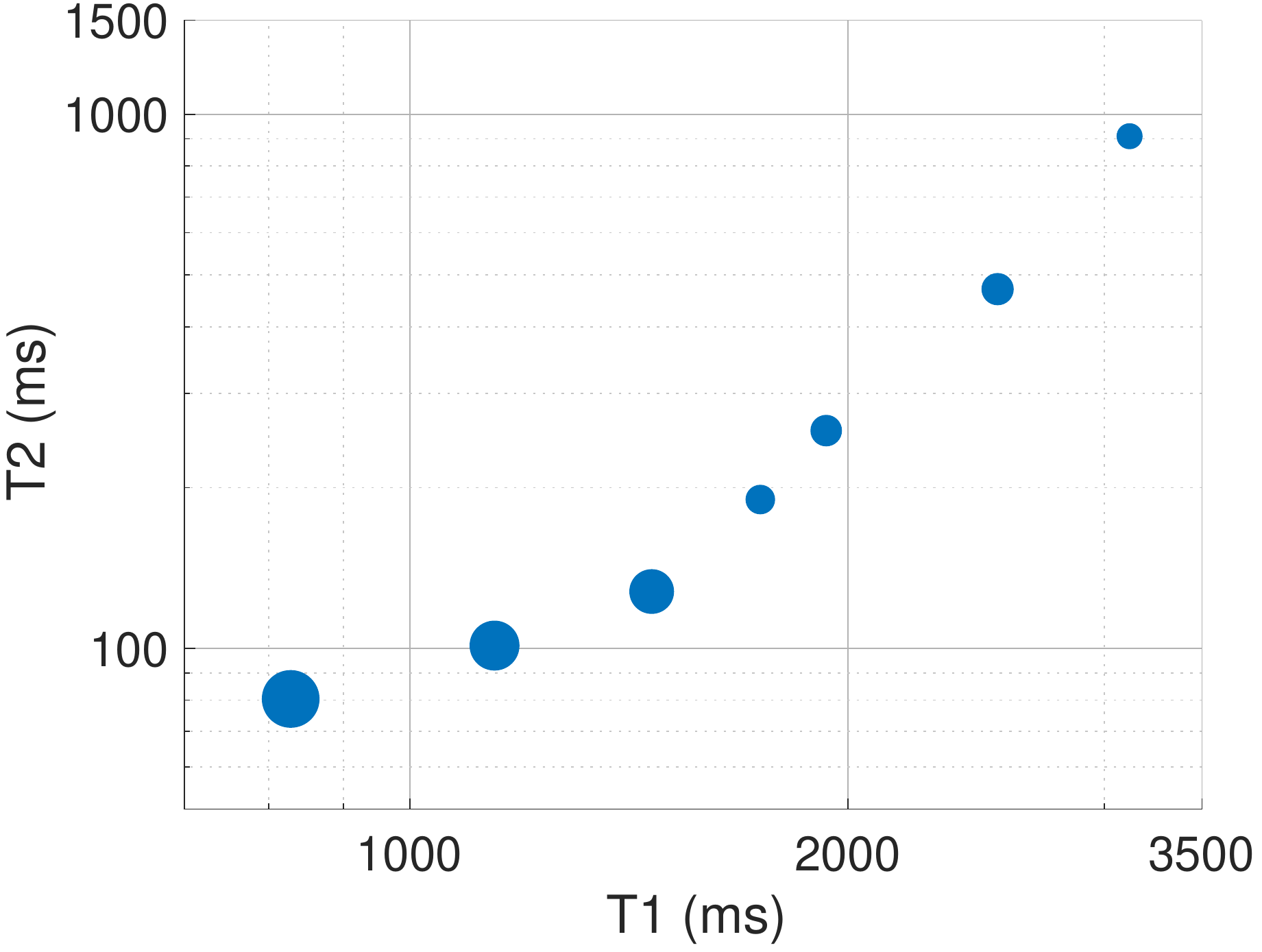}
\includegraphics[width=.24\linewidth]{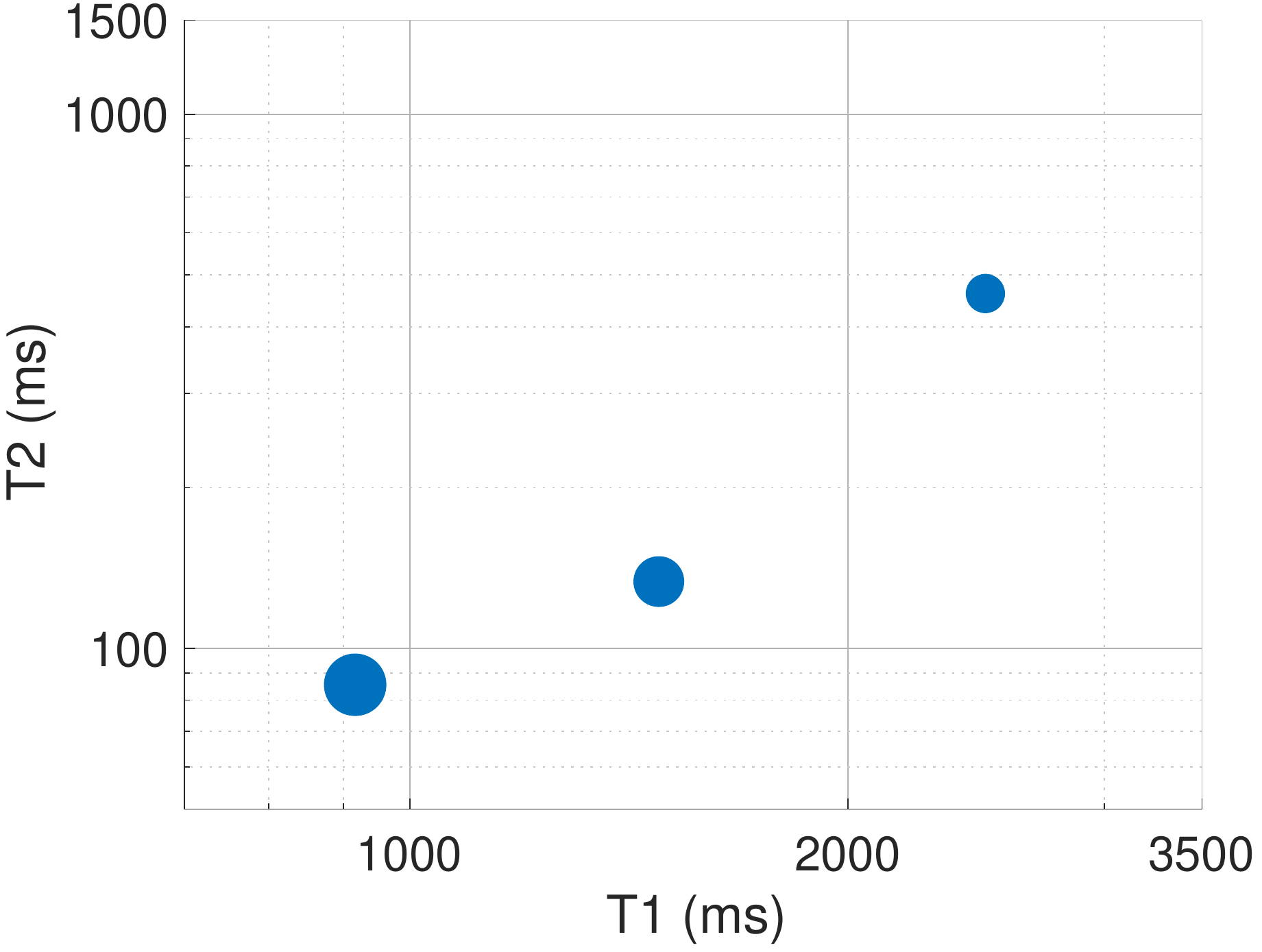}
\includegraphics[width=.24\linewidth]{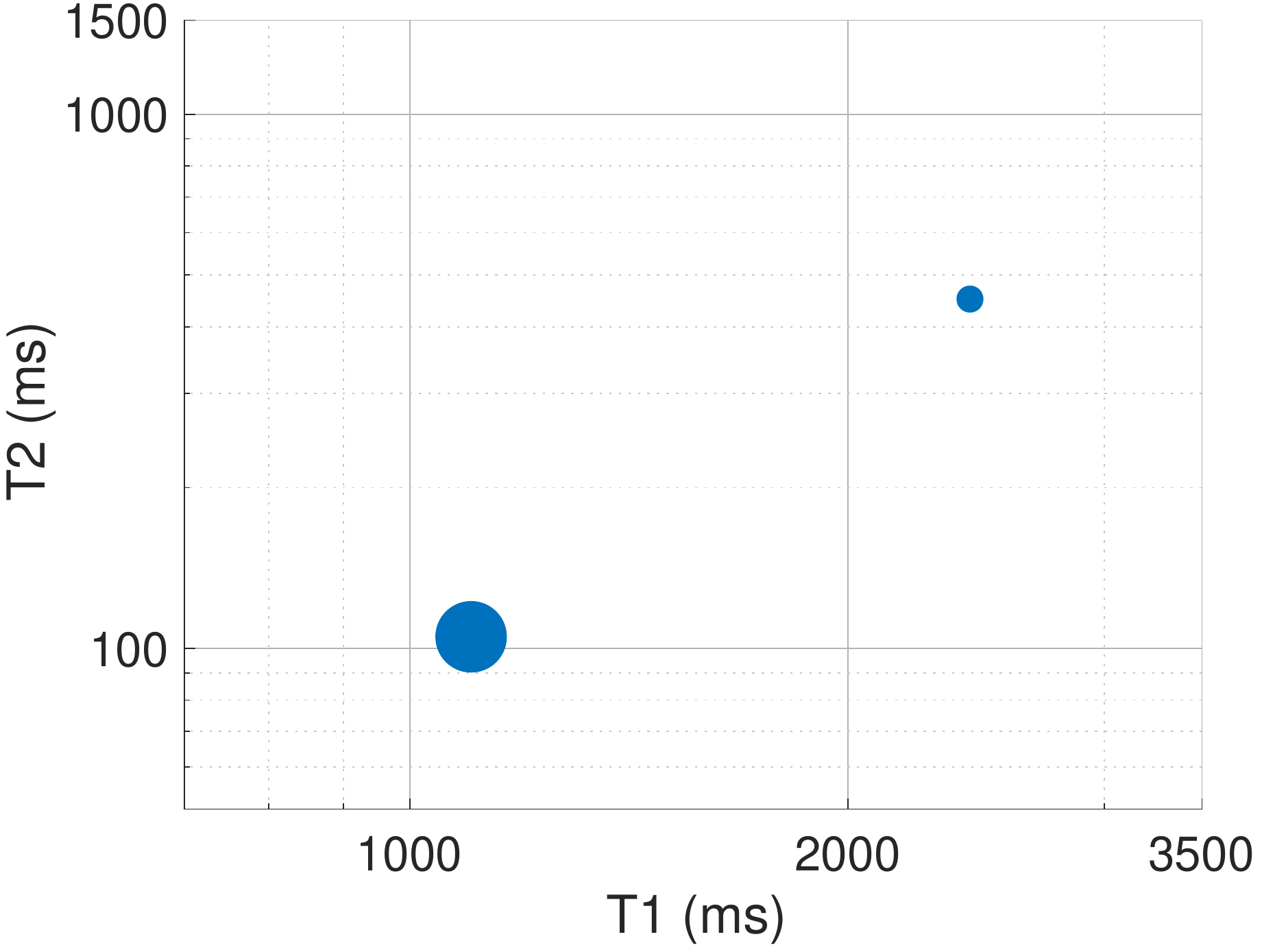}
\\
 \hspace{.5cm}  (a) \MCMRF $\beta=10^{-4}$ \hspace{.7cm} (b) \textbf{\MCMRF $\mathbf{\beta=10^{-3}}$} \hspace{.5cm} (c) \MCMRF $\beta=10^{-2}$ \hspace{.5cm} (d) \MCMRF $\beta=5\times10^{-2}$ \\
\vspace{.2cm}

\includegraphics[width=.24\linewidth]{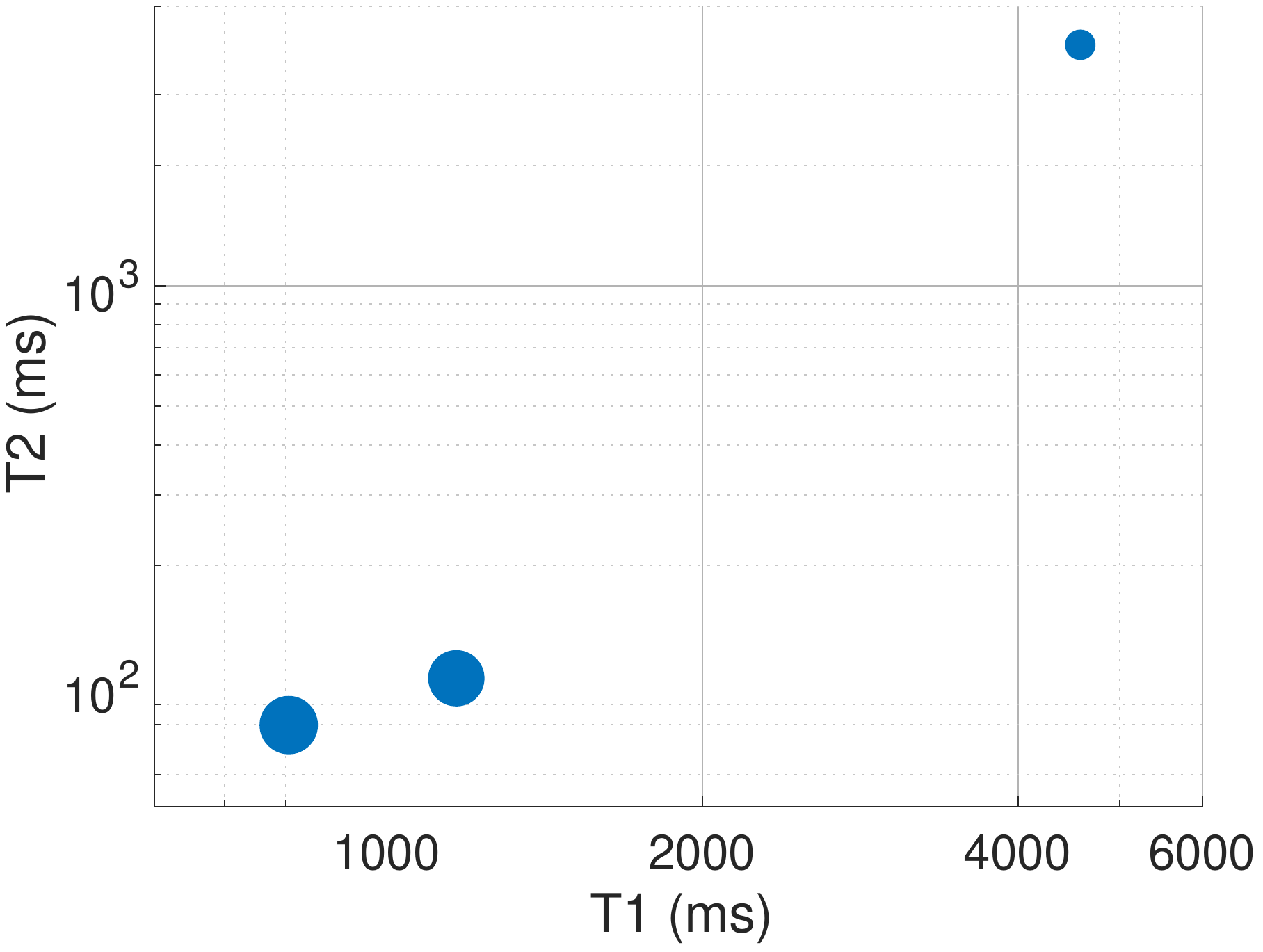}
\includegraphics[width=.24\linewidth]{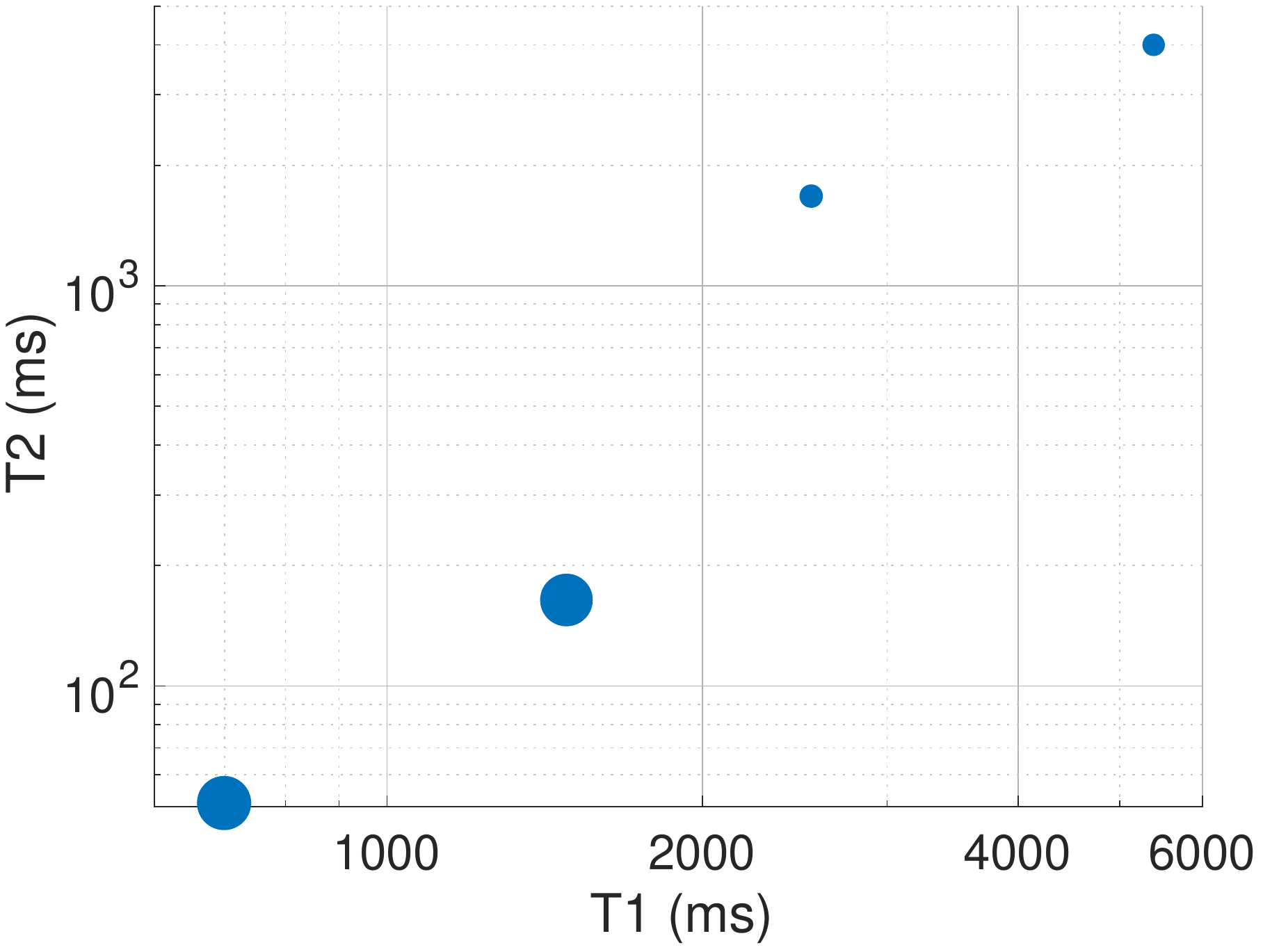}
\includegraphics[width=.24\linewidth]{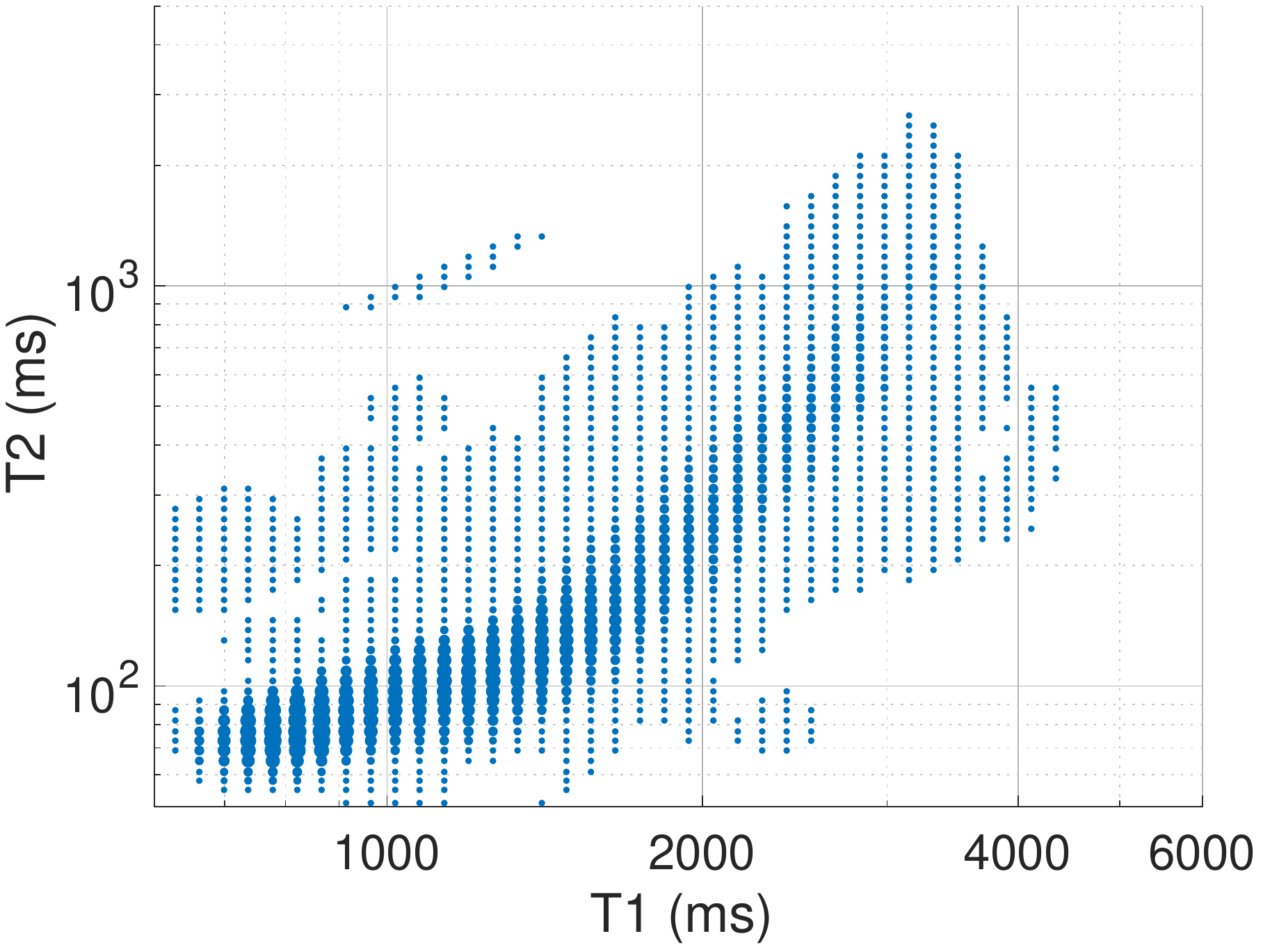}
\includegraphics[width=.24\linewidth]{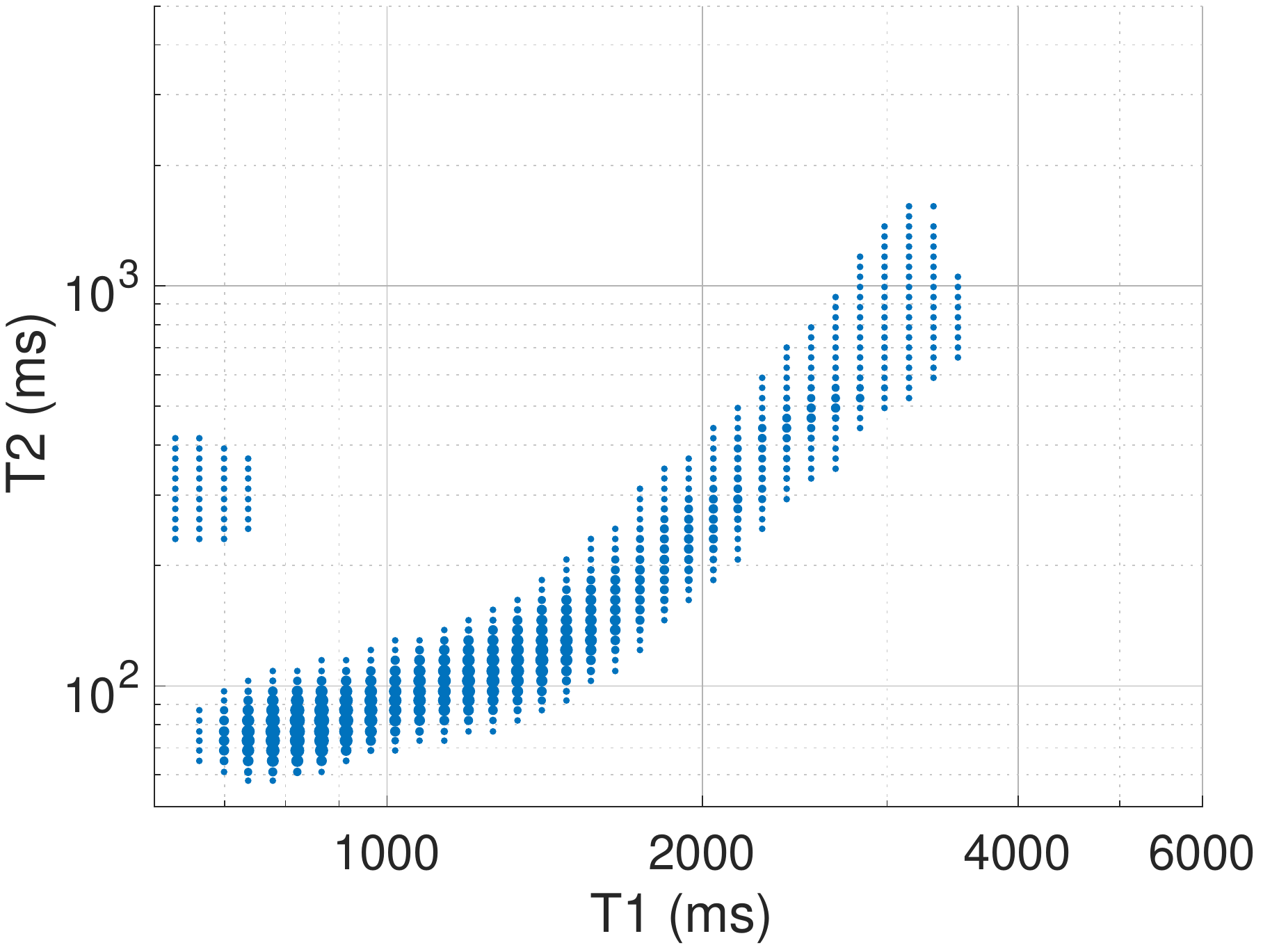}
\\
 \hspace{-1.6cm}(e) PVMRF \hspace{2.7cm} (f) SPIJN \hspace{2.9cm} (g) BayesianMRF \hspace{1.5cm} (h) SG-Lasso\\
\caption{\footnotesize{The T1/T2 values of the estimated in-vivo brain compartments (un-thresholded) using (a-d)
\MCMRF 
for different parameters $\beta$, and (e-h) the MC-MRF baselines. All methods used LRTV reconstruction before mixture separation. Scatter points' radii are scaled by the $\ell_2$ norms of the mixture weights (maps) corresponding to the estimated T1/T2 values. }
\label{fig:vivo_compareScatter} }
\end{minipage}}
\end{figure*}

\begin{figure}[t]
	\centering
	\scalebox{1}{
	\begin{minipage}{\linewidth}
		\centering		
\begin{turn}{90} \qquad \textbf{\MCMRF} \end{turn}
\includegraphics[width=.31\linewidth]{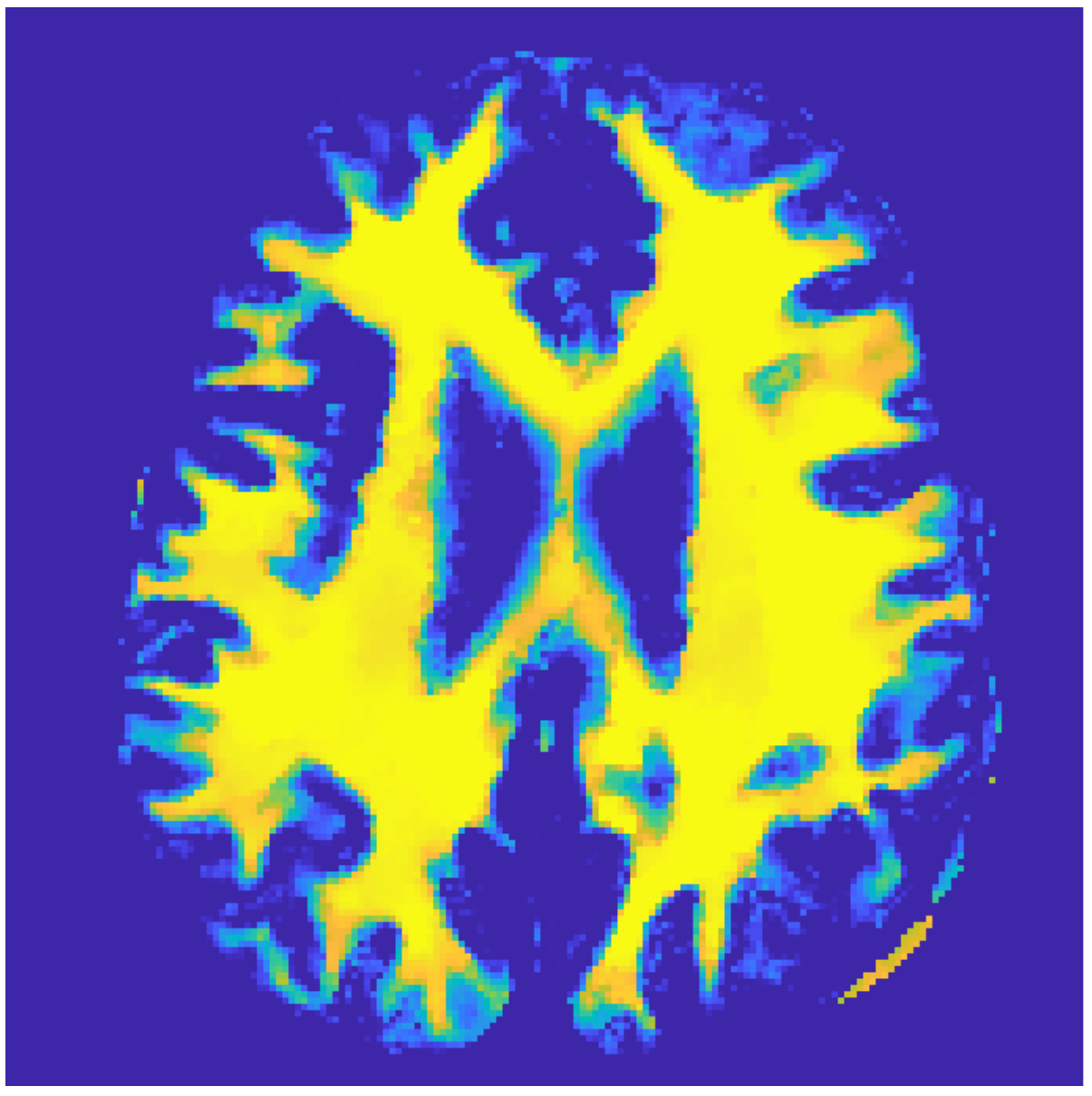}\hspace{-.1cm}
\includegraphics[width=.31\linewidth]{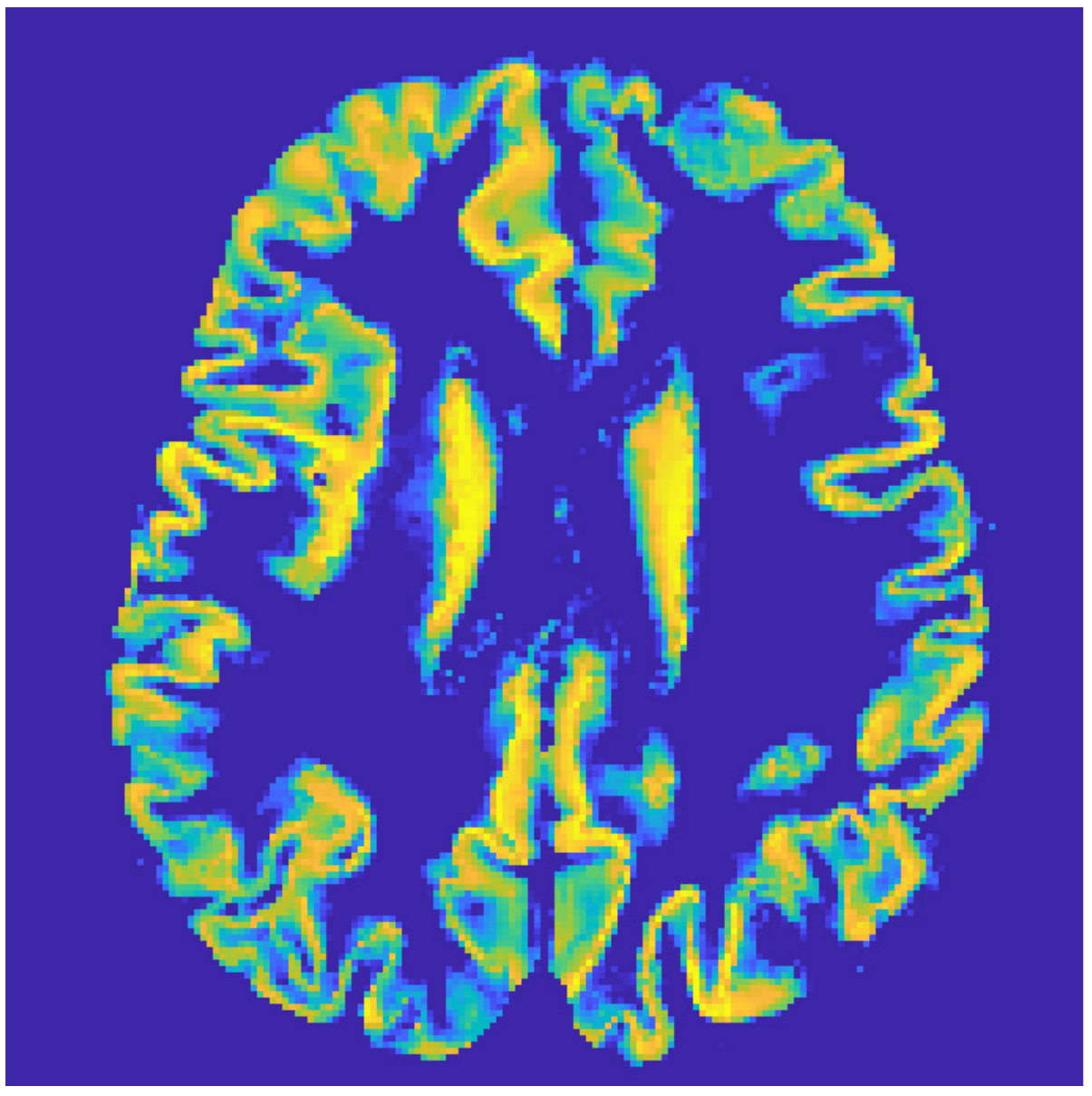}\hspace{-.1cm}
\includegraphics[width=.31\linewidth]{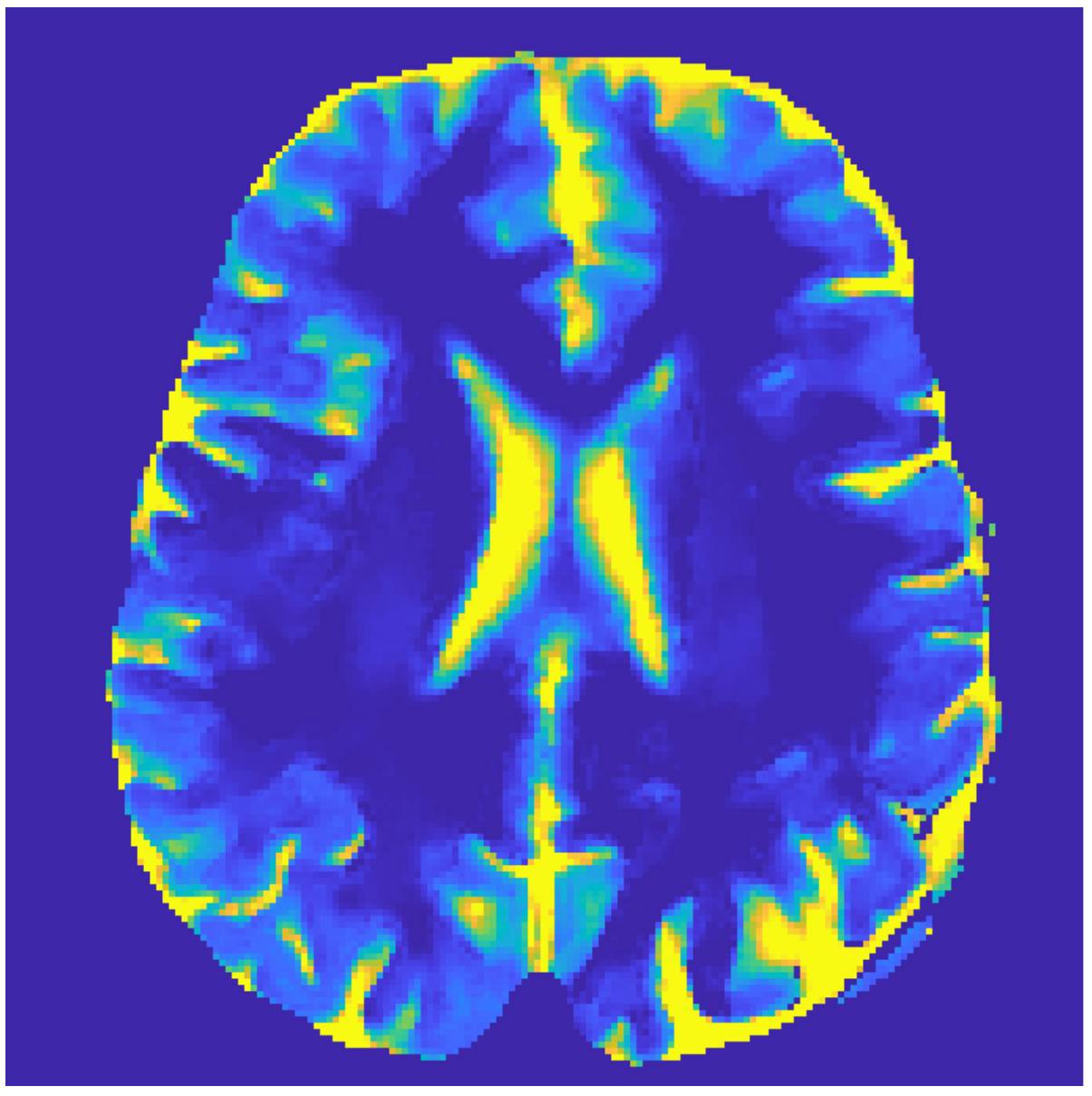}
\\
\begin{turn}{90} \qquad PVMRF \end{turn}
\includegraphics[width=.31\linewidth]{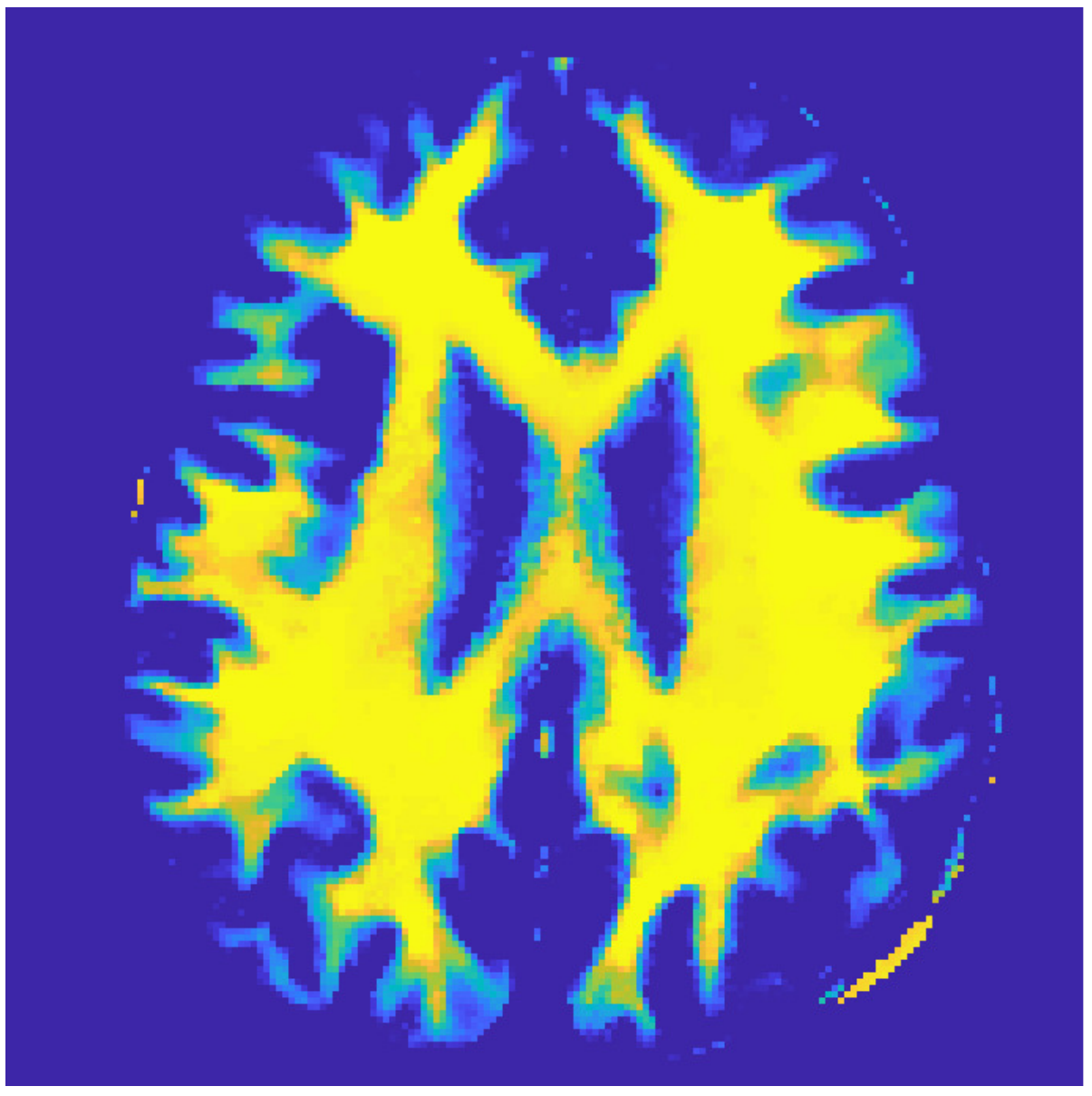}\hspace{-.1cm}
\includegraphics[width=.31\linewidth]{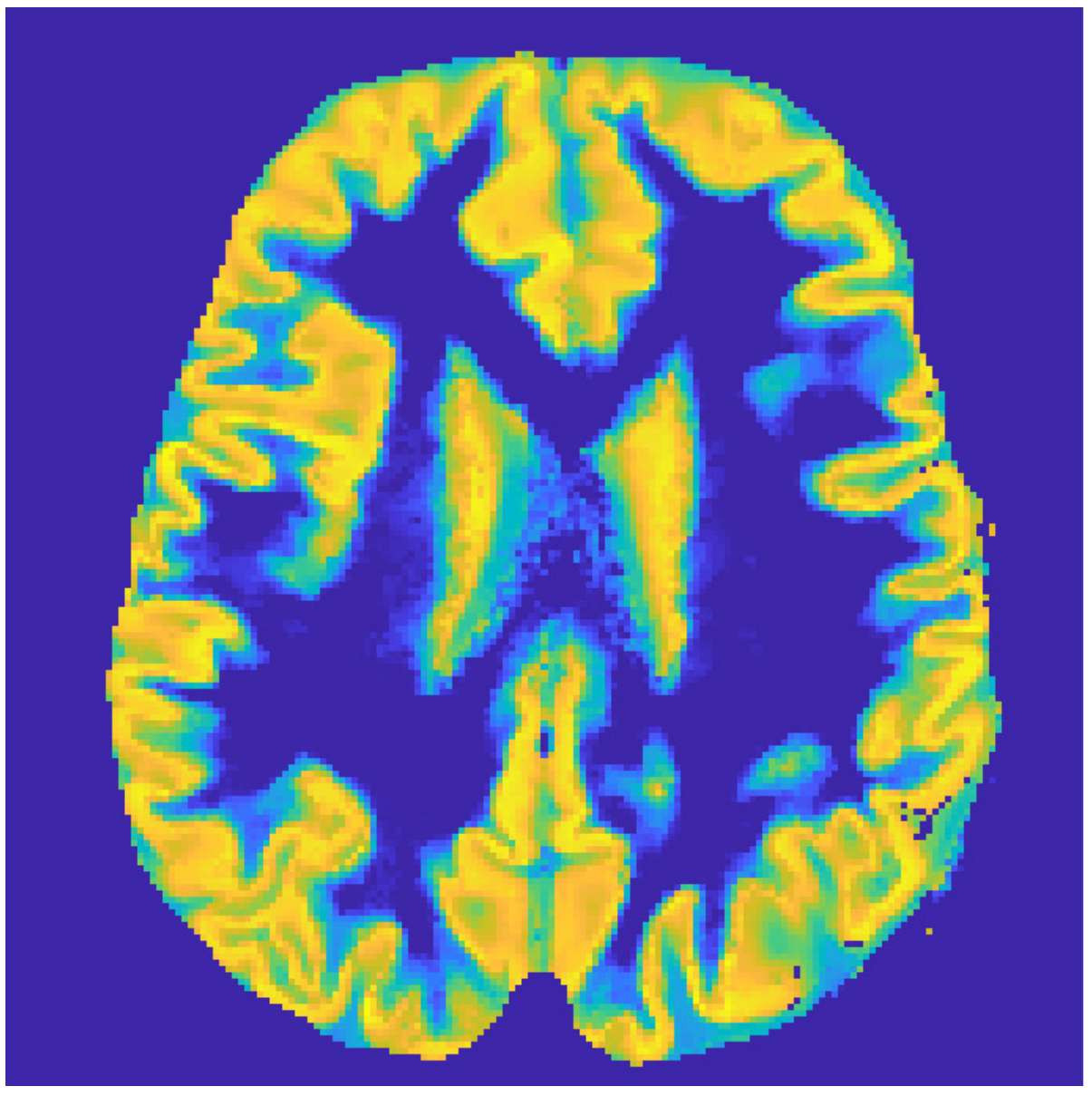}\hspace{-.1cm}
\includegraphics[width=.31\linewidth]{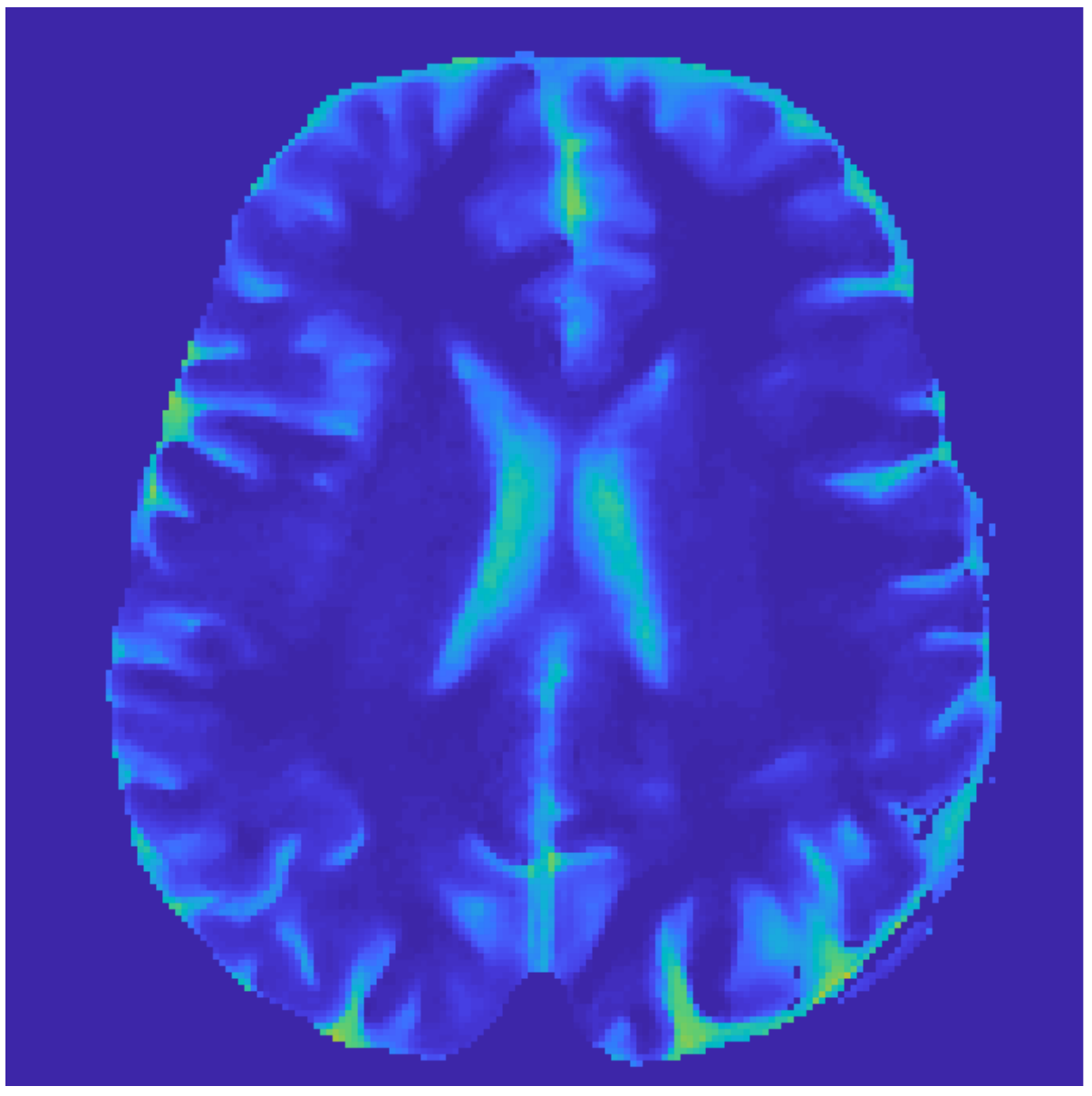}
\\
\begin{turn}{90} \qquad\quad SPIJN \end{turn}
\includegraphics[width=.31\linewidth]{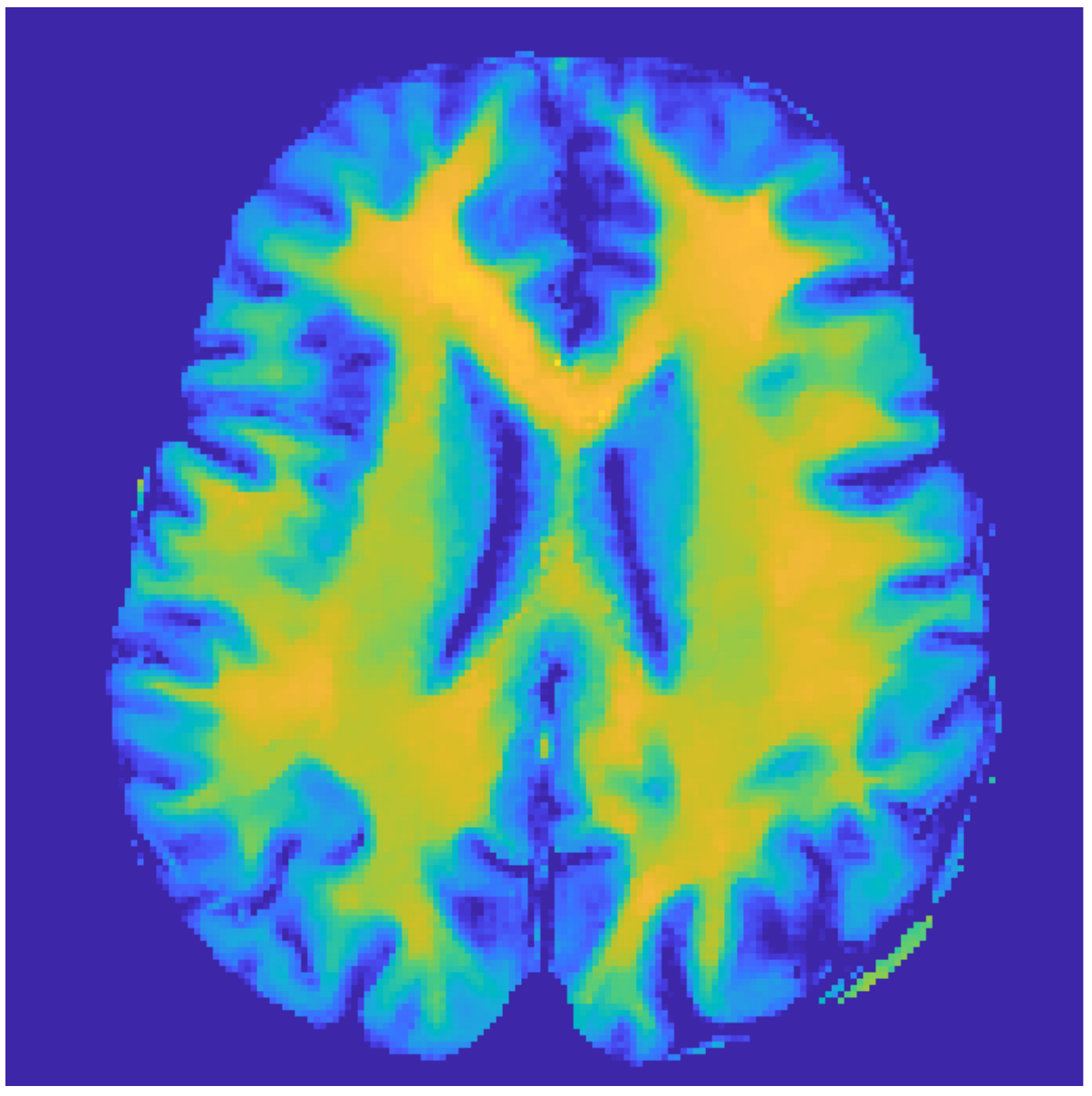}\hspace{-.1cm}
\includegraphics[width=.31\linewidth]{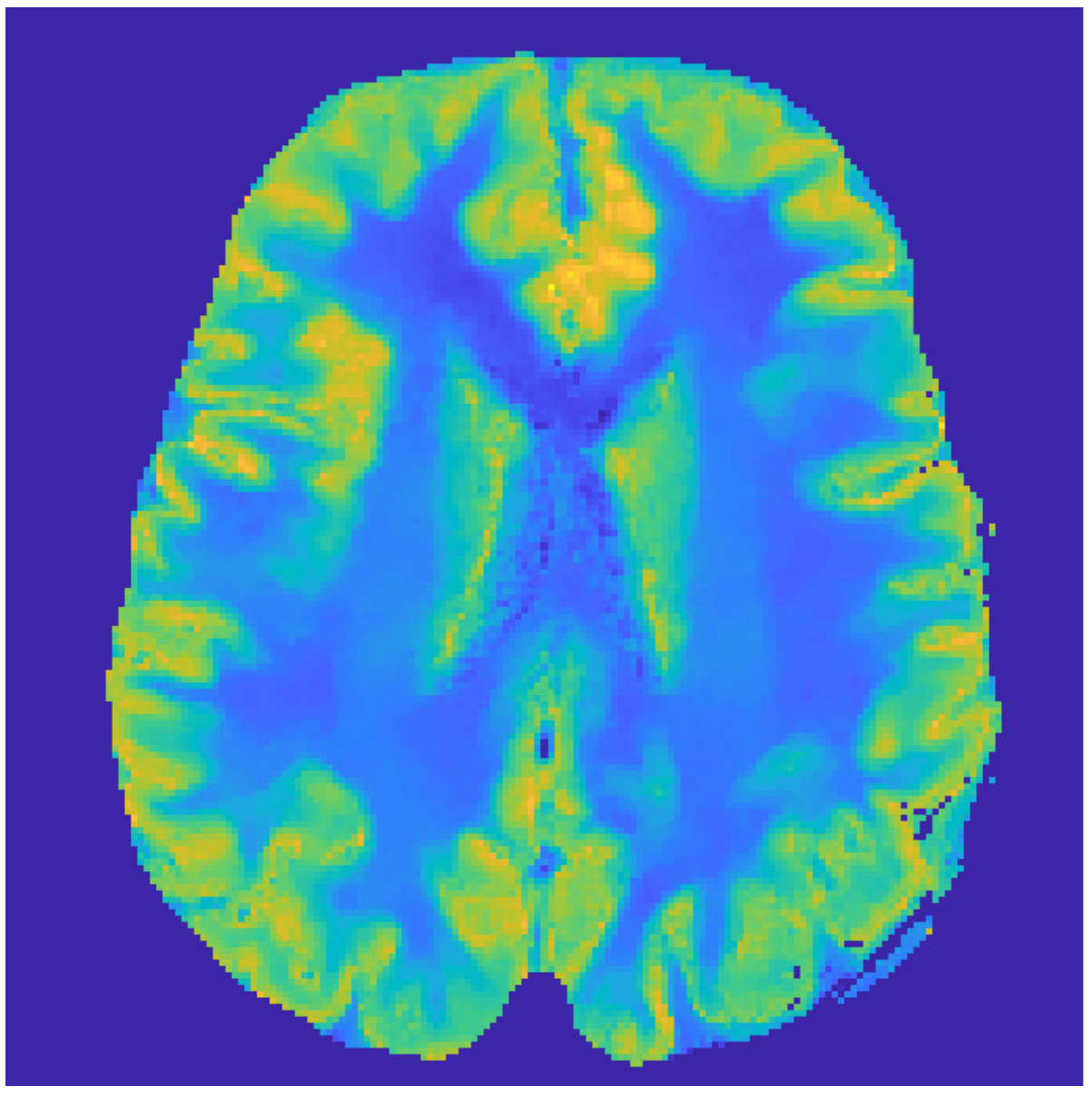}\hspace{-.1cm}
\includegraphics[width=.31\linewidth]{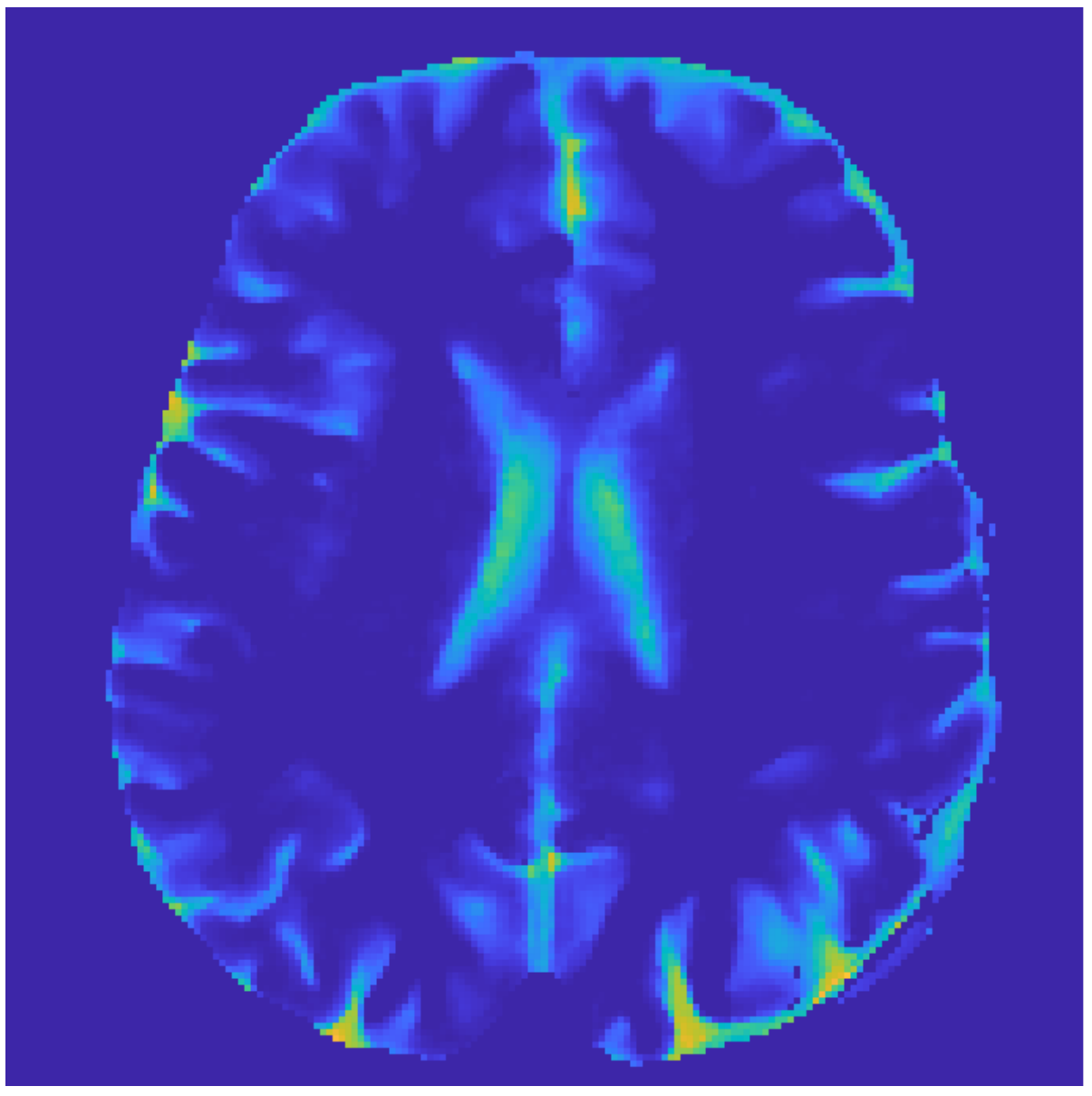}
\\
\begin{turn}{90} \quad BayesianMRF \end{turn}
\includegraphics[width=.31\linewidth]{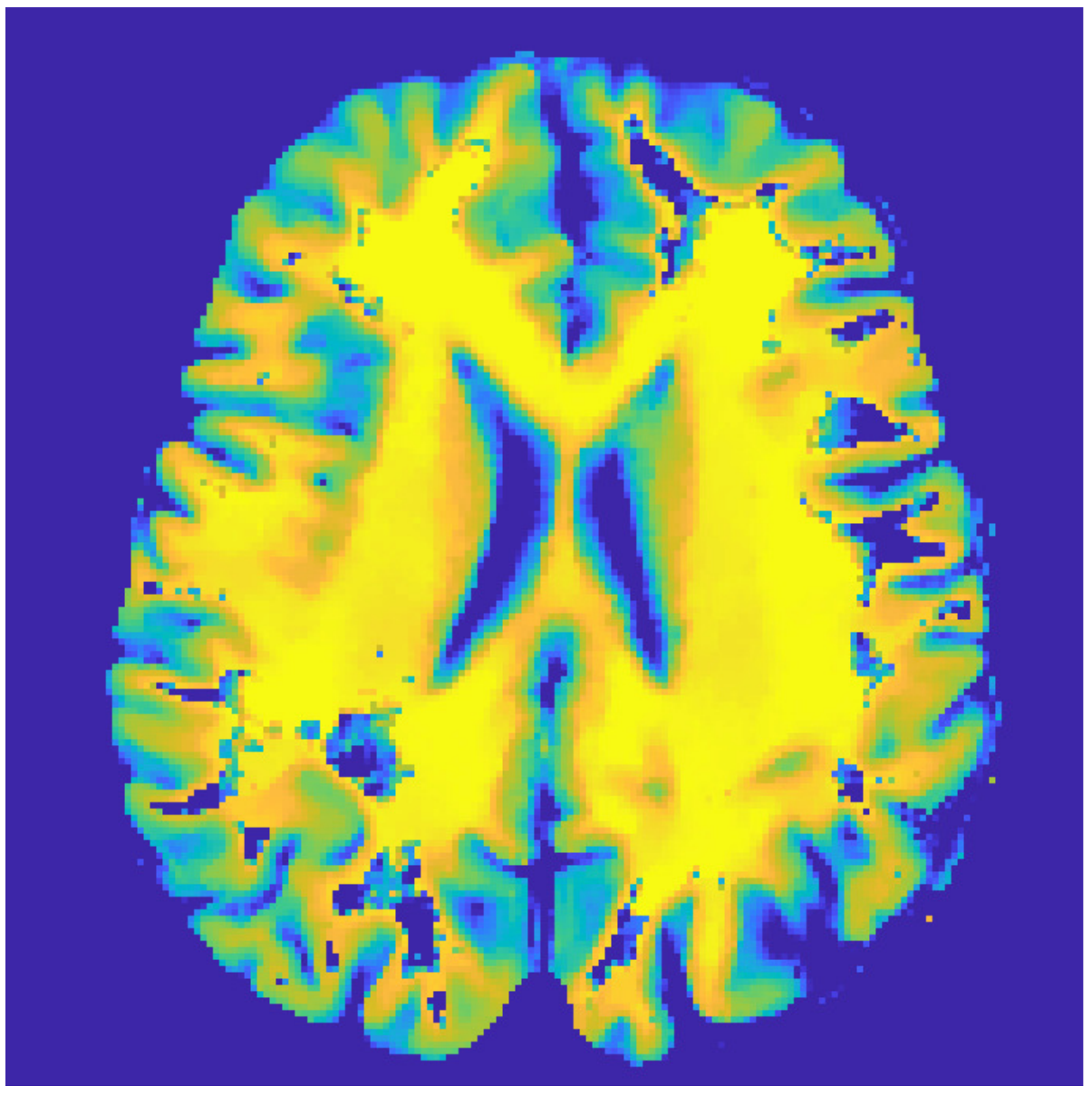}\hspace{-.1cm}
\includegraphics[width=.31\linewidth]{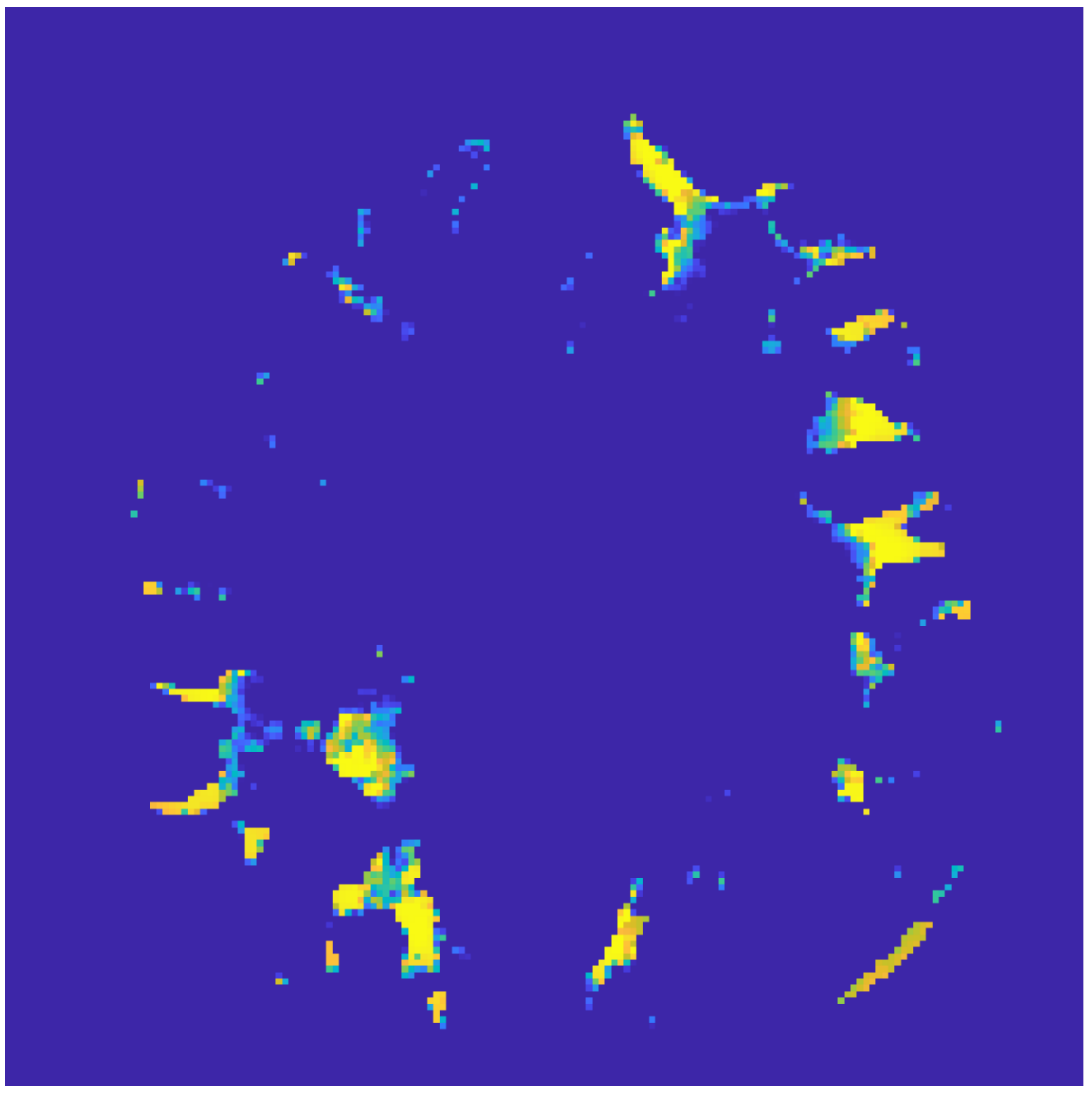}\hspace{-.1cm}
\includegraphics[width=.31\linewidth]{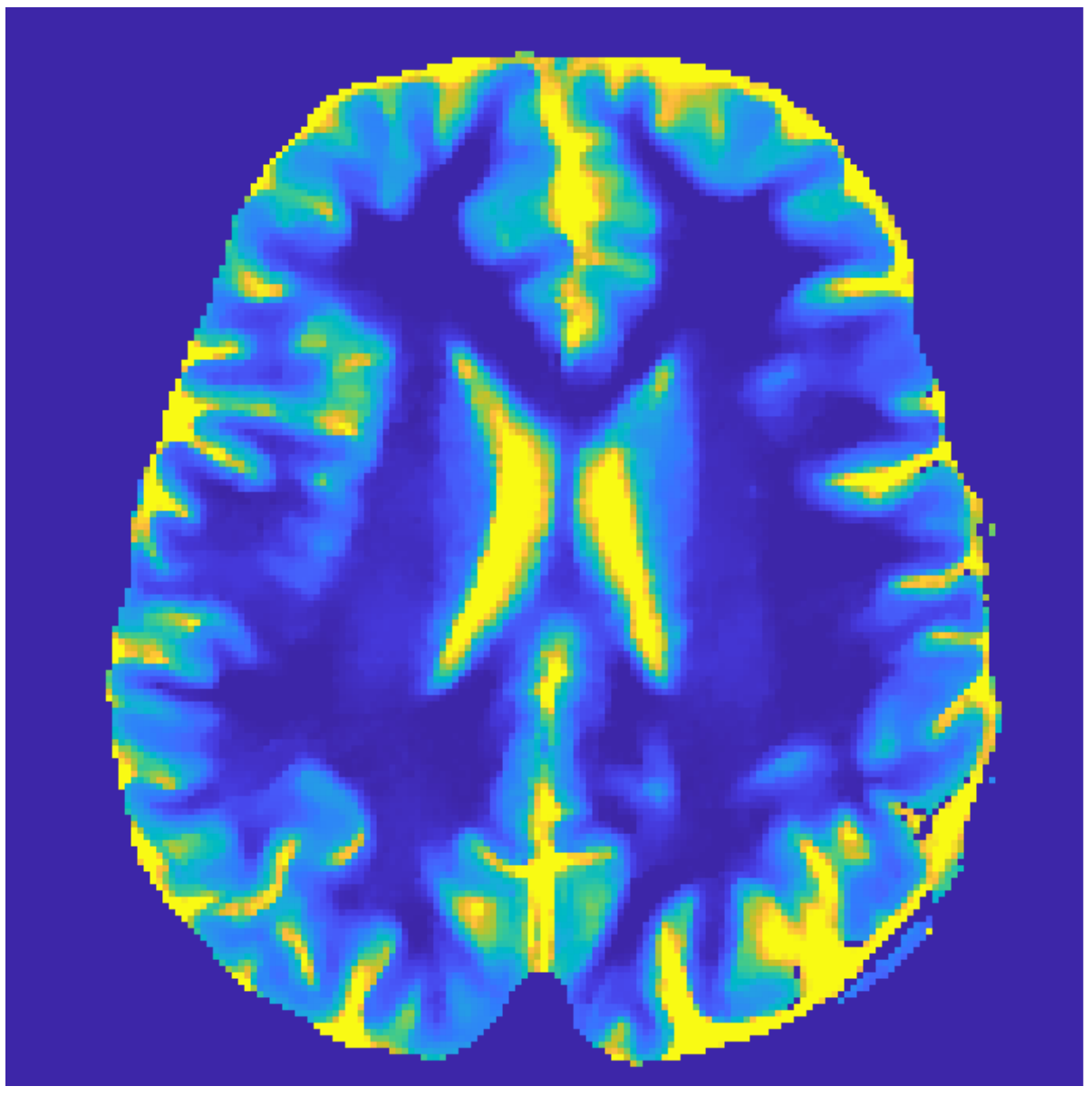}
\\
\begin{turn}{90} \qquad\ SG-Lasso \end{turn}
\includegraphics[width=.31\linewidth]{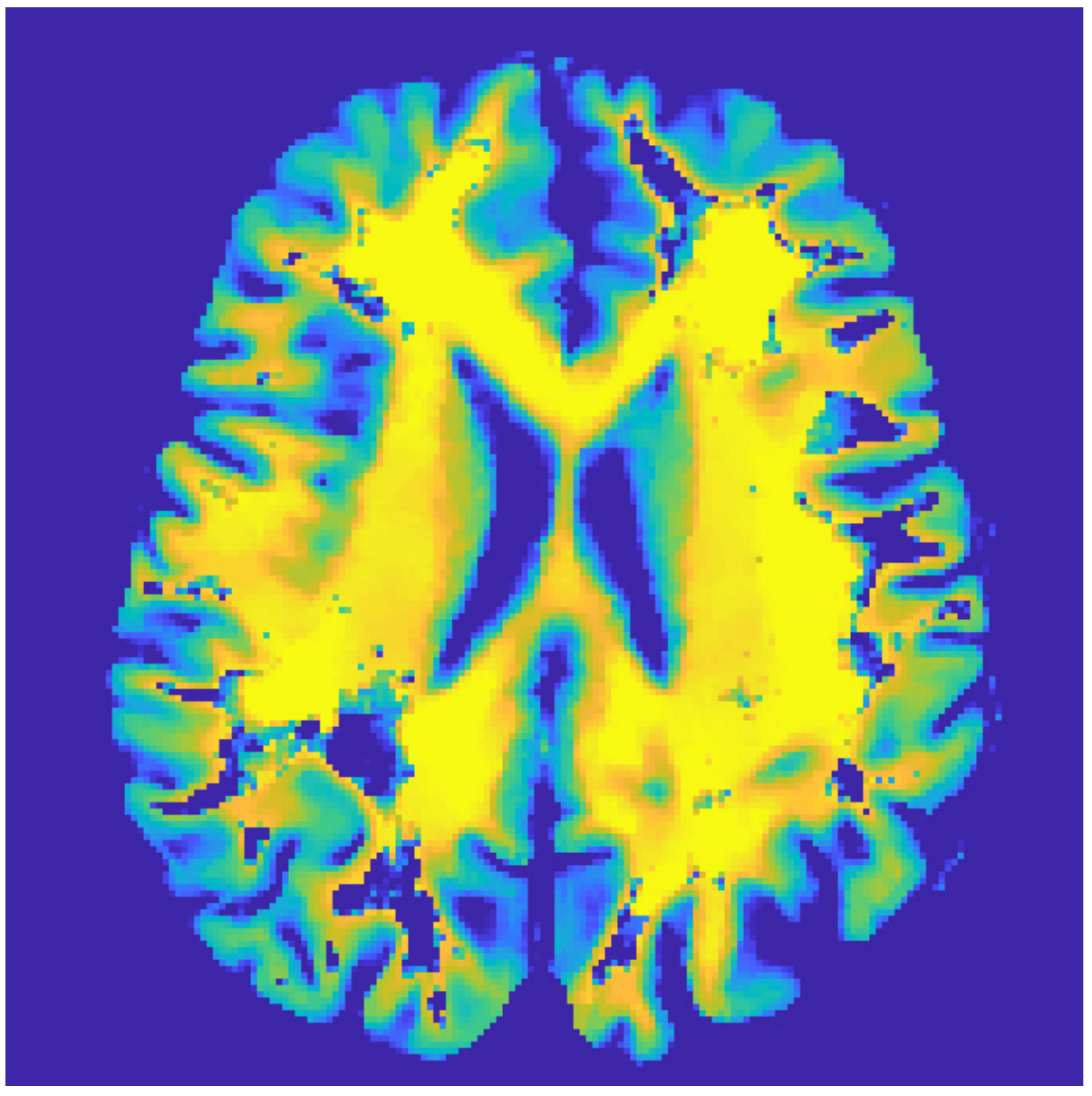}\hspace{-.1cm}
\includegraphics[width=.31\linewidth]{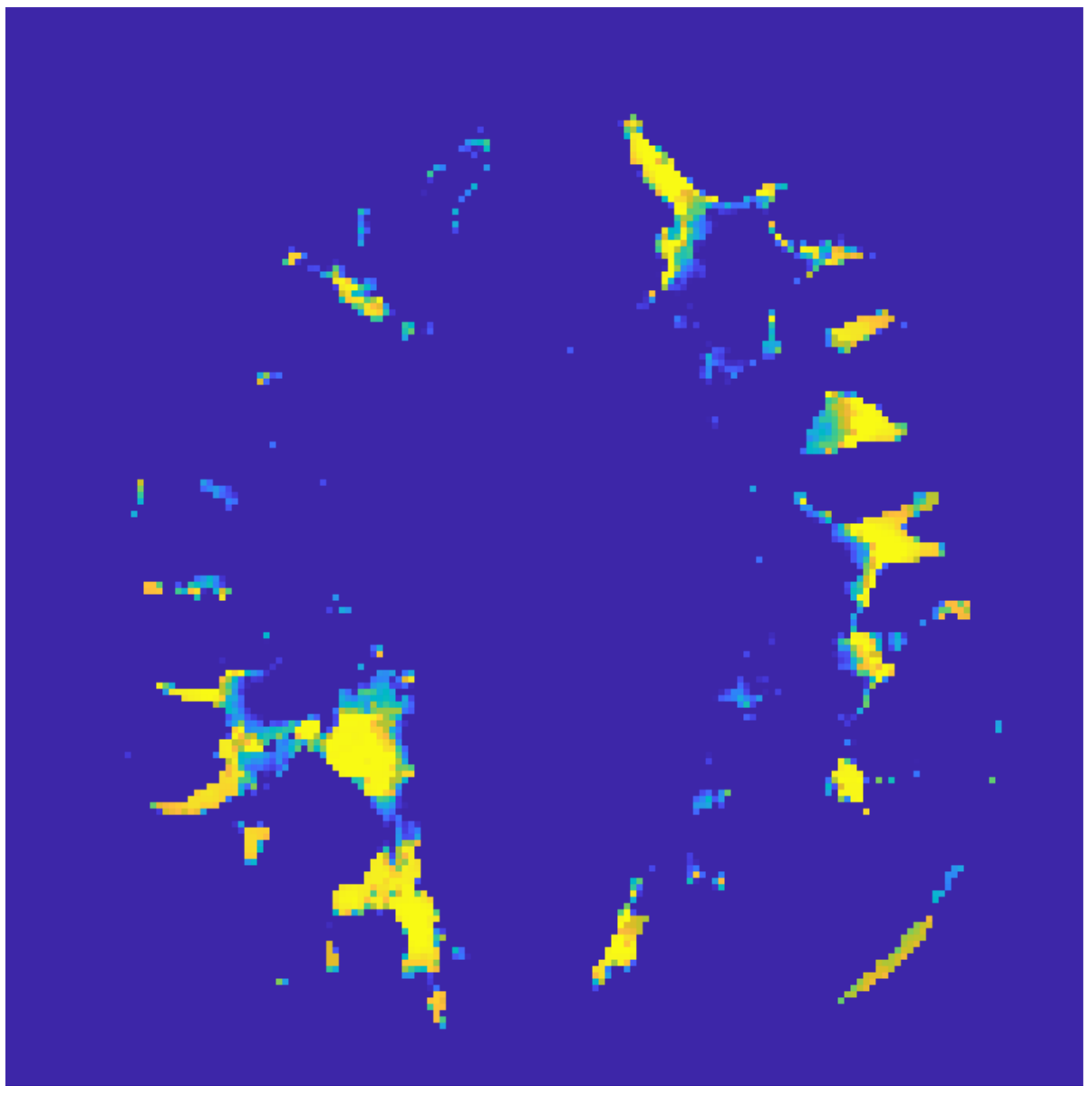}\hspace{-.1cm}
\includegraphics[width=.31\linewidth]{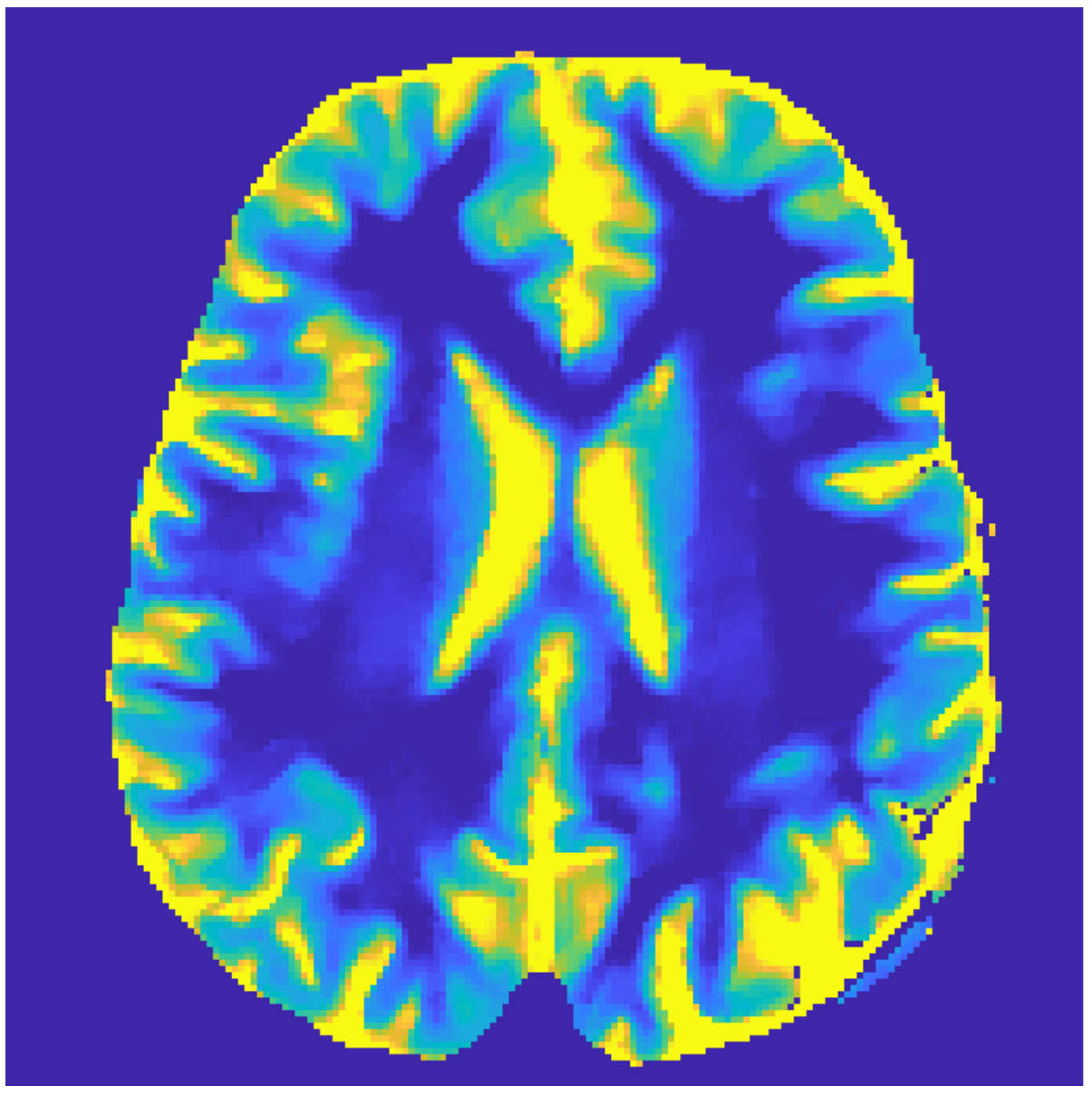}
\end{minipage}}
\caption{\footnotesize{Mixture maps (margins cropped) of the WM, GM and a CSF related compartment for the in-vivo brain using different MC-MRF algorithms. 
}\label{fig:vivo_compareMM} }

\end{figure}

\begin{table*}[t!]
	\centering
	\scalebox{1}{
		\begin{tabular}{c|ccccc|ccccc}
			\toprule[0.2em]
			\multicolumn{1}{c}{}&\multicolumn{5}{c}{T1 (ms)}&\multicolumn{5}{c}{T2 (ms)}  \\	
			\midrule[0.1em]		
			Tissue &{Literature} & \textbf{\MCMRF} & PVMRF & SPIJN & BayesianMRF & {Literature} & \textbf{\MCMRF} & PVMRF & SPIJN & BayesianMRF \\

			\midrule[0.1em]		
				\midrule[0.1em]					
			WM &   $694-862^{\dag}$  & 829 & 806 & 699 & 821  		& $68-87^{\dag,*}$& 81& 80 & 51  & 77\\
			GM &    $1074-1174^*$ & 1114 & 1165 & 1483 & 874 		& $87-103^{\dag,*}$   & 102 & 105 & 164 & 82 \\
			\bottomrule[0.2em]
		\end{tabular}		}
		\caption{\footnotesize{Estimated T1/T2 values for in-vivo WM and GM compartments using MC-MRF algorithms compared to the 1.5T literature values \cite{sled2001quantitative}$^{\dag}$ \cite{stanisz2005t1}$^{*}$.
}}\label{tab:t1t2vivo}
	\end{table*}

\subsection{In-vivo brain experiment}
\label{sec:vivo}
To demonstrate feasibility of the proposed approach in vivo, compressed-sampled MRF data was acquired from a healthy volunteer's brain with an informed consent obtained.~\footnote{We thank GE Healthcare (Munich, Germany) for providing this data.} Acquisition used a 1.5T GE HDxT scanner with 8-channel receive-only head RF coil, $230\times230$ mm$^2$ field-of-view, $230\times230$ image pixels, 5 mm slice thickness, with the above-mentioned FISP MRF protocol. A variable density spiral trajectory was used for k-space sampling. 
The total number of spiral interleaves were 377. 
One spiral arm sampled 920 k-space locations per TR/timeframe and this pattern was rotated for the next TR. 
The overall acquisition window for $T=1000$ timeframes was about 10 seconds. 


\subsubsection{The effects of parameter $\beta$}
We first used this data to examine the effects of parameter $\beta$ in \MCMRF\ 
on mixture separation. 
TSMIs were reconstructed using LRTV ($\la_{\text{LRTV}}=4\times10^{-5}$). We then ran \MCMRF\ for different values of $\beta=\{1, 10, 100, 500\}\times10^{-4}$ and fixed regularisation weight $\alpha=0.8$. The estimated compartments (T1/T2 values) are scatter plotted in Figures~\ref{fig:vivo_compareScatter}(a-d). Mixture maps of these compartments are also shown in Figure~\ref{fig:vivo_fullmixtures}.  
As can be observed, interpolating between the two extreme cases of pure pixel vs. group sparsity (i.e. $\beta \rightarrow 0 \,\,\text{or}\,\, 1$)
creates 
different demixing solutions 
across the T1/T2 values and the spatial mixture maps. For small $\beta$ values tissues are decomposed into spatially finer/detailed compartments (Figure~\ref{fig:vivo_fullmixtures}), while larger $\beta$ values hierarchically cluster together tissue compartments around fewer T1/T2 values. 
Decompositions at a moderate value $\beta=10^{-3}$  indicates a  single component for WM, whereas the boundaries of GM and CSF can be decomposed into few additional compartments with long relaxation properties. 


\subsubsection{Comparison to the baselines}
We compare the performances of the \MCMRF\ and MC-MRF baselines for WM and GM estimation. 
We used $\beta=10^{-3}, \alpha=0.8$ parameters for \MCMRF. For visualisation outcomes of all tested methods were hard thresholded 
to 
exactly three 
groups:  
the T1/T2 values of top two compartments with highest energy ($\ell_2$ norm) mixture maps were reported as WM and GM. Remaining compartments with longer T1/T2 times were averaged (weighted by their mixture maps energies) and mapped to a third group.  
Given these T1/T2 estimates, the (thresholded) maps were computed by simulating a three-atom discrete dictionary and solving a small-size nonnegative least squares for~\eqref{eq:model4}.  
Figure~\ref{fig:vivo_compareMM} compares the thresholded mixture maps of the \MCMRF\ and baselines.
All methods used the LRTV reconstruction 
before demixing. 
In separate Figures~\ref{fig:vivo_compareScatter}(b,e-h) the T1/T2 values of all estimated compartments before thresholding are scatter plotted. Table~\ref{tab:t1t2vivo} compares the estimated T1/T2 values of (thresholded) WM and GM to their literature values at 1.5T, 
and Table~\ref{tab:runtime} summarises the runtimes of tested algorithms.

\begin{table}[t!]
	\centering
	\scalebox{1}{
		\begin{tabular}{ccccc}
			\toprule[0.2em]
			PVMRF & SPIJN & BayesianMRF & SG-Lasso & \MCMRF  \\	
			\midrule[0.05em]	
			\midrule[0.05em]		
			2.56& 220.91& 325.42& 	842.78 &102.30	\\
			\bottomrule[0.2em]
		\end{tabular}		}
		\caption{\footnotesize{Runtimes (in seconds) of the MC-MRF algorithms for the in-vivo experiment.}}
		\label{tab:runtime}
	\end{table}

In Figure~\ref{fig:vivo_compareMM} \MCMRF\ outperforms the baselines in terms of the visual appearance of the mixture maps, their spatial separability and correspondence to the anatomical structures of WM, GM and CSF. Further, the estimated T1/T2 values for WM and GM are within the range of literature values~(Table~\ref{tab:t1t2vivo}). The CSF relaxations underestimated the literature values and were excluded from comparisons. This issue was previously reported in all MC-MRF baselines (see e.g.~\cite{PVMRF}) due the  in-vivo pulsations that are not captured by the MRF's signal model~\cite{FISP}.  
BayesianMRF does not exploit group sparsity and results in many unclustered compartments (Figure~\ref{fig:vivo_compareScatter}(g)) and poor mixture maps visualisation. 
Baselines SPJIN and PVMRF improve this thanks to the compartment grouping, however the GM map using PVMRF is not well separated from the CSF region, and additionally in SPIJN the WM map is not well separated from the GM region (Figure~\ref{fig:vivo_compareMM}). SPJIN overestimated GM's T1/T2 and underestimated WM's T2 relaxations. BayesianMRF underestimated GM's relaxations, and PVMRF slightly overestimated GM's T2 value. 
The non-iterative PVMRF was the fastest method, followed by the \MCMRF\ which is the fastest amongst tested model-based iterative algorithms (Table~\ref{tab:runtime}). 
Notably, the (discretised) SG-Lasso implemented by FISTA at a high level of accuracy (objective tolerance=$10^{-8}$) and the longest runtime, outputs unclustered compartments (Figure~\ref{fig:vivo_compareScatter}(h)) and poor mixture maps (Figure~\ref{fig:vivo_compareMM}), highlighting the significance of an off-the-grid alternative to overcome the fundamental limitation of sparse approximation in a highly coherent discretised dictionary.

Similar comparisons 
are illustrated in Figures~\ref{fig:vivo_compareMM_SVD} 
where LRTV is replaced by SVD-MRF~\cite{SVDMRF} reconstruction i.e. the method previously adopted by all baselines. \MCMRF also outperforms baselines, but 
as can be observed, SVD-MRF produces undersampling artefacts that propagate to the demixing step and perturb the mixture maps.
SPIJN demixing favours SVD-MRF albeit outputting noisy (aliased) mixture maps indicating its high sensitivity to TSMI variations, and that the LRTV's spatial smoothing favours enforcing pixel-sparsity besides group-sparsity (as in \MCMRF) at the demixing step.


%





\section{Discussions}
\label{sec:discuss}
We introduced a novel off-the-grid approach to address the non-scalability of the dictionary-based MC-MRF baselines. 
We observed that a voxel-sparsity alone (BayesianMRF) 
results in inferior demixing performance compared to the group-sparse models (PVMRF, SPIJN) that cluster the entire image into few compartments. 
The proposed \MCMRF\ improves upon both approaches by simultaneously promoting both sparsity types via SGTV regularisation. This regularisation provides the flexibility of promoting a desired level of spatial sparsity (i.e. certain level of pixel purity) within the mixture maps of the estimated sparse compartments. 
In our simulations (Section~\ref{sec:diri_expe}) we also observed that \MCMRF\ was more robust than other baselines for separating less pixel-pure (more challenging) mixture distributions. 
Further, the \MCMRF\ was able to separate the WM, GM and CSF regions of healthy brain \emph{in-vivo} more precise than the baselines. Estimated T1/T2s for the WM and GM were consistent with the literature.  
The WM region was separated in a single compartment whereas the boundaries of GM and CSF (pre-thresholded) were decomposed into few additional compartments with long relaxation properties (also reported in~\cite{SPIJN}). 
Further in-vivo validations are required to confirm repeatably of the results and their usage for clinical applications. 

A T1/T2-encoding MRF sequence was used in our experiments. 
Encoding more parameters could potentially separate more complex e.g. pathology-related compartments. 
It will be interesting to examine potentials of this work in applications encoding larger number of parameters e.g.~\cite{MRF-perfusion,MRF_T1T2diff}, where dictionary-based gridding could create a major precision vs. storage bottleneck to the MC-MRF framework. 
Also, current implementation of \MCMRF\ uses the L-BFGS quasi Newton method which is an accurate but slow nonlinear fitting solver. Stochastic gradient methods~\cite{bottouSGD,kingma2014adam, GPIS,duchi2011adaptive} could be an interesting way forward to accelerate the \MCMRF's computations. 

Previous works~\cite{BayesianPVMRF, PVMRF, SPIJN} reported great sensitivity (e.g. in terms of noise amplification in mixture maps) to TSMIs' noise and under-sampling artefacts. For this we replaced SVD-MRF by a spatiotemporally regularised model-based reconstruction LRTV for enhancing demixing results. 
While further numerical advances for MRF reconstruction will benefit the current work, we believe that future works combining tasks of reconstruction and mixture separation could be more efficient way forward 
(e.g. see multi-task compressed sensing examples~\cite{ji2008multitask,TIPHSI, golbabaee2010distributed,duarte2012framework}) in order to numerically tackle shorter and aggressively under-sampled acquisition protocols.


\section{Conclusion}
\label{sec:conclusions}
We introduced a novel off-the-grid approach for the MC-MRF problem. The proposed \MCMRF algorithm is an accurate and importantly a scalable alternative to the MC-MRF baselines because its does not rely on fine-gridded multi-parametric MRF dictionaries. The method was theoretically described and its basic feasibility was demonstrated and compared to other baselines in simulations and in-vivo healthy brain measurements. 


%

%
%

\ifCLASSOPTIONcaptionsoff
  \newpage
\fi

\bibliographystyle{IEEEtran}
{\footnotesize{
\bibliography{mybiblio_al}
}}


\newpage
\onecolumn
\begin{center}
	\textbf{\large Supplementary Materials }\\
\end{center}
\setcounter{equation}{0}
\setcounter{figure}{0}
\setcounter{table}{0}
\setcounter{section}{0}
\setcounter{page}{1}
\makeatletter
\renewcommand{\theequation}{S\arabic{equation}}
\renewcommand{\thefigure}{S\arabic{figure}}
\renewcommand{\thesection}{S\Roman{section}}

\section{Results and discussions (supplementary)}
\subsection{Network embedding of the Bloch responses}

\begin{figure}[h!]
	\centering
	\scalebox{.95}{
	\begin{minipage}{\linewidth}
		\centering		
\includegraphics[width=.5\linewidth]{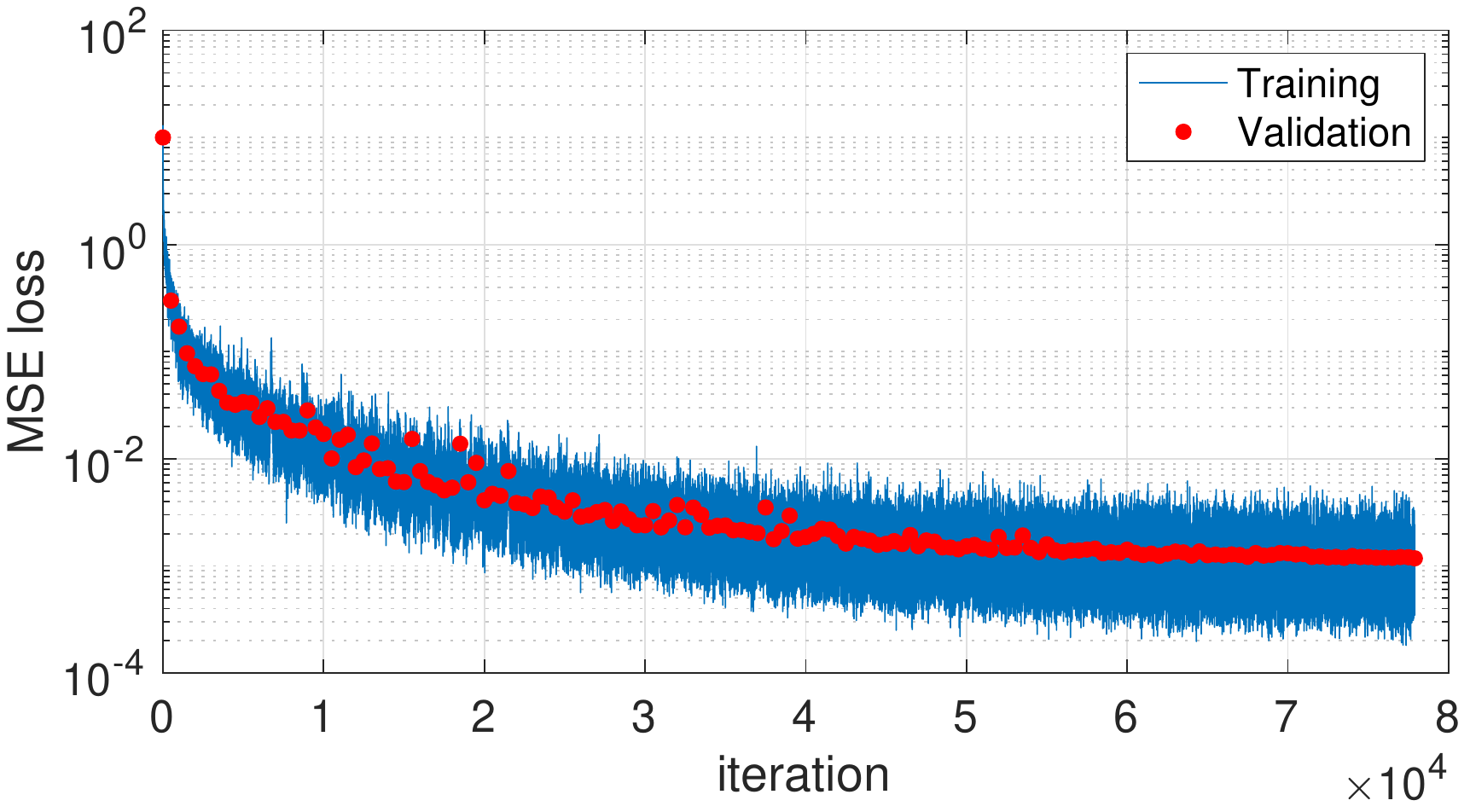}
\end{minipage}}
\caption{{Training and validation MSE losses during training the neural network model to approximate the Bloch responses. 
}\label{fig:trainingcurve} }
\end{figure}

\subsection{The numerical Brainweb phantom experiment}

\begin{figure*}[h!]
	\centering
	\scalebox{.95}{
	\begin{minipage}{\linewidth}
		\centering	
		\hspace{.8cm}WM \hspace{1.5cm} diff. WM	\hspace{1.5cm} GM \hspace{1.5cm} diff. GM \hspace{1.8cm} CSF\hspace{1.8cm} diff. CSF\vspace{.1cm}\\
\begin{turn}{90} \quad\qquad Full \end{turn}
\includegraphics[width=.15\linewidth]{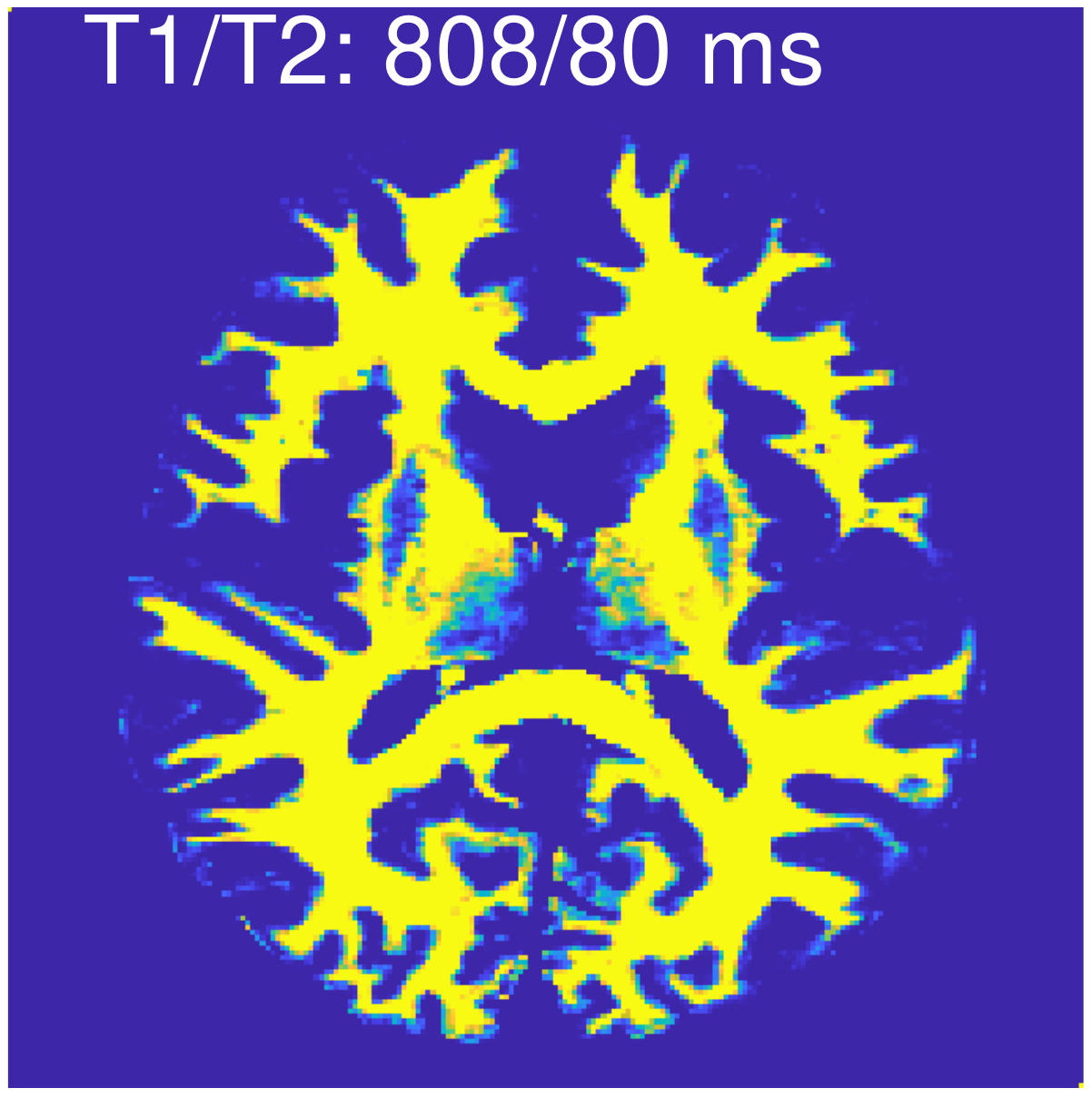}\hspace{-.1cm}
\includegraphics[width=.15\linewidth]{./figs/expe3_cut_im_50db_diff_1351.pdf}\hspace{.1cm}
\includegraphics[width=.15\linewidth]{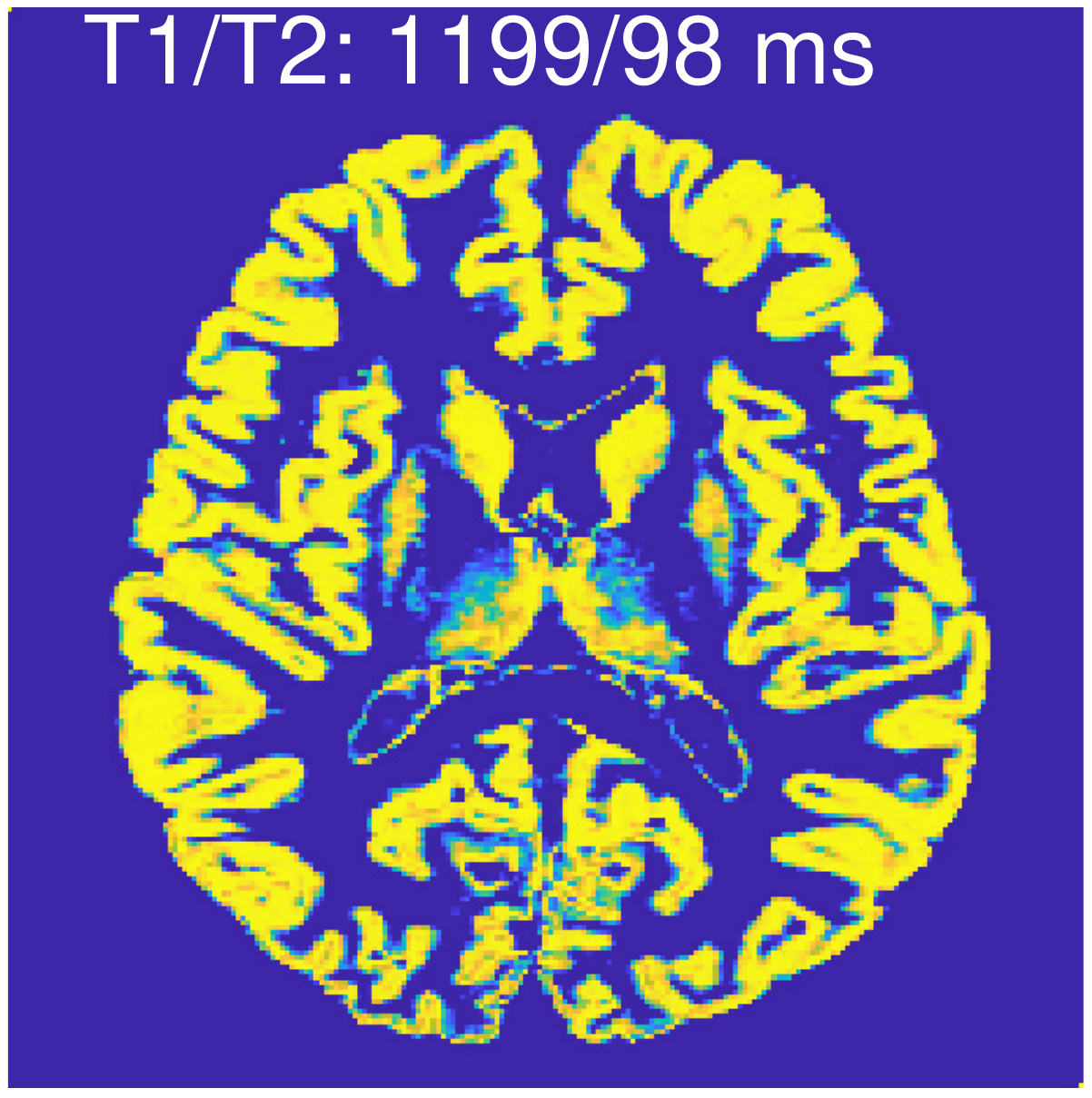}\hspace{-.1cm}
\includegraphics[width=.15\linewidth]{./figs/expe3_cut_im_50db_diff_1352.pdf}\hspace{.1cm}
\includegraphics[width=.15\linewidth]{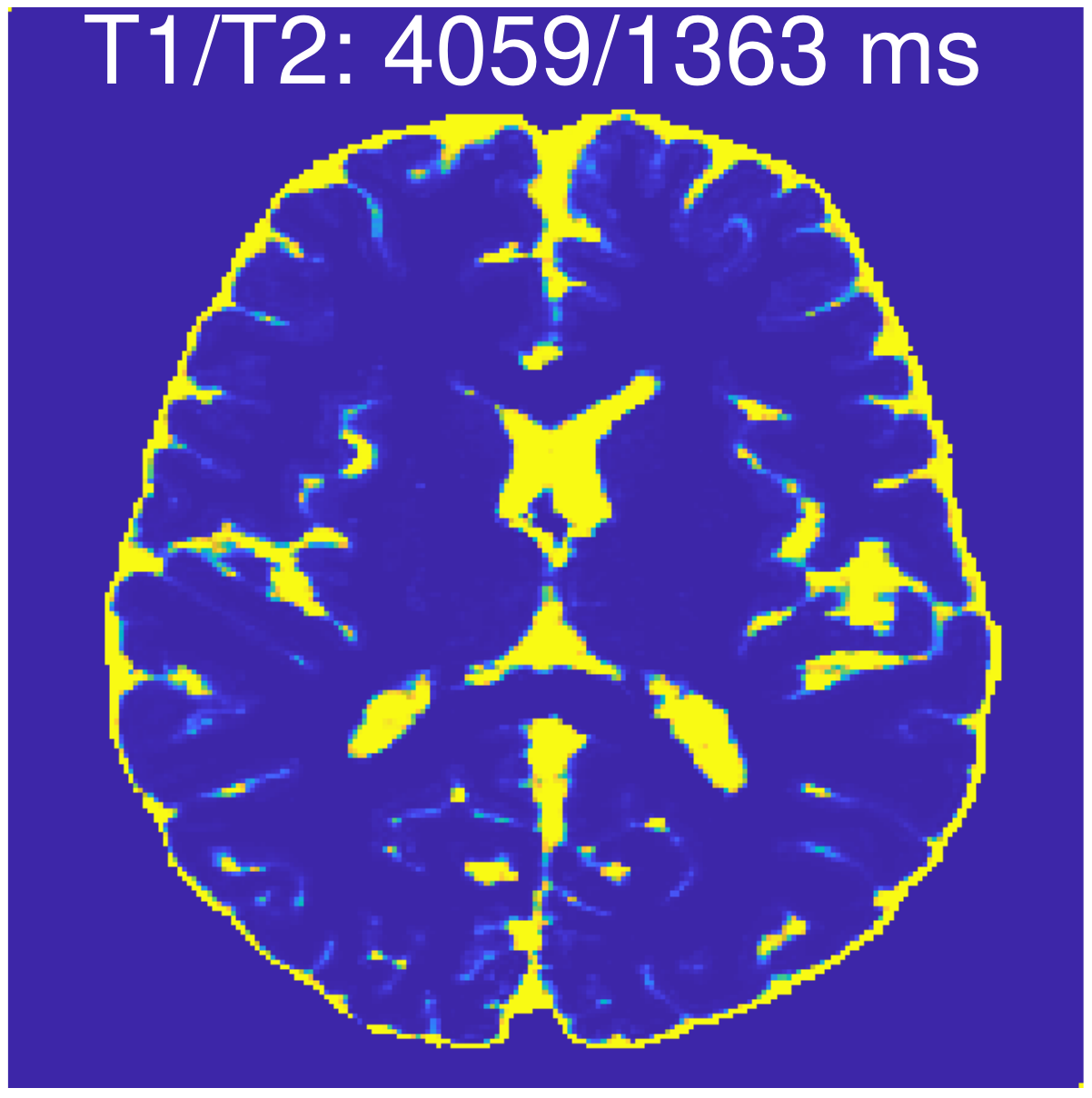}\hspace{-.1cm}
\includegraphics[width=.15\linewidth]{./figs/expe3_cut_im_50db_diff_1353.pdf}
\\
\begin{turn}{90} \quad\qquad LRTV \end{turn}
\includegraphics[width=.15\linewidth]{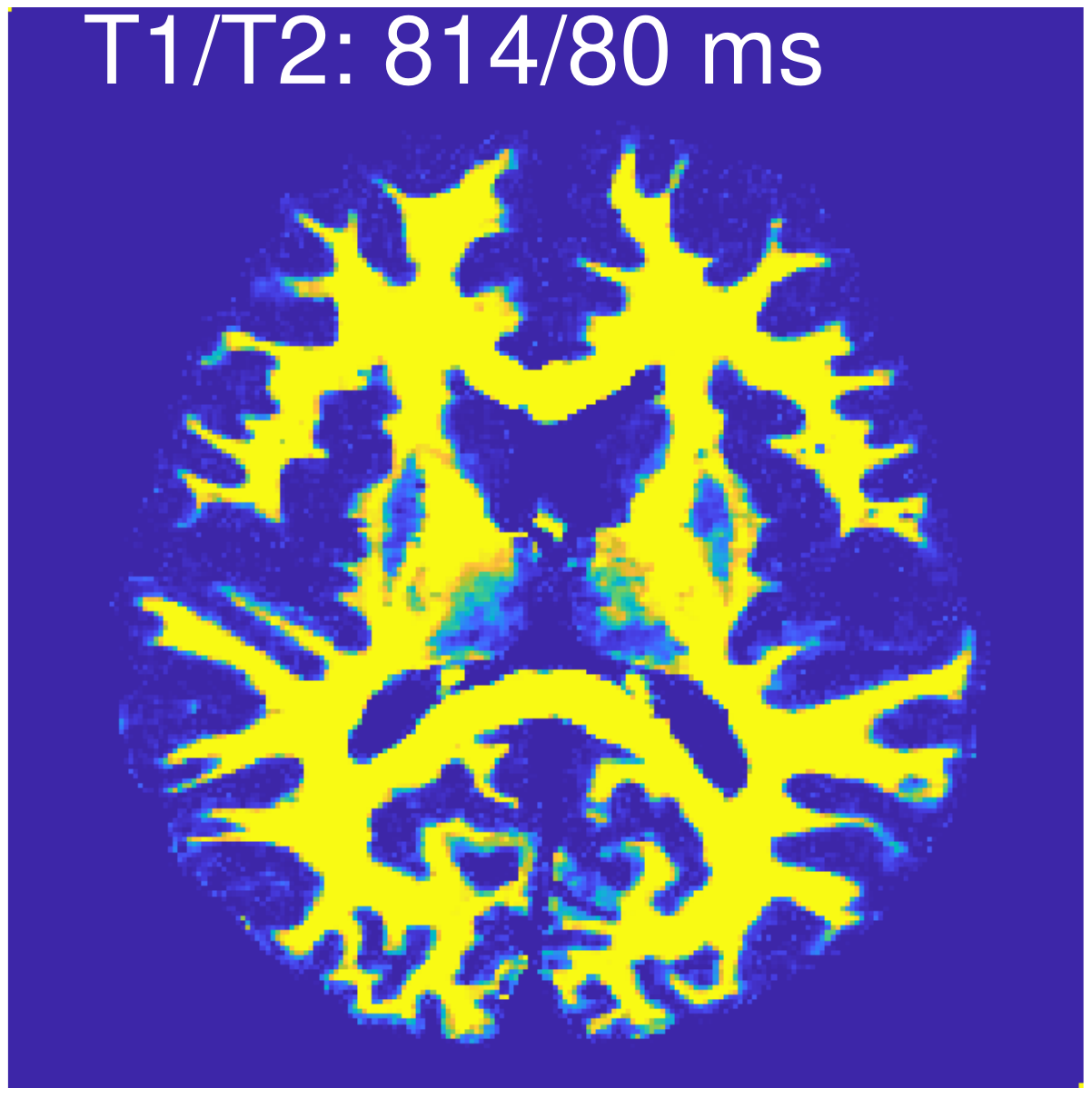}\hspace{-.1cm}
\includegraphics[width=.15\linewidth]{./figs/expe3_cut_im_50db_diff_1151.pdf}\hspace{.1cm}
\includegraphics[width=.15\linewidth]{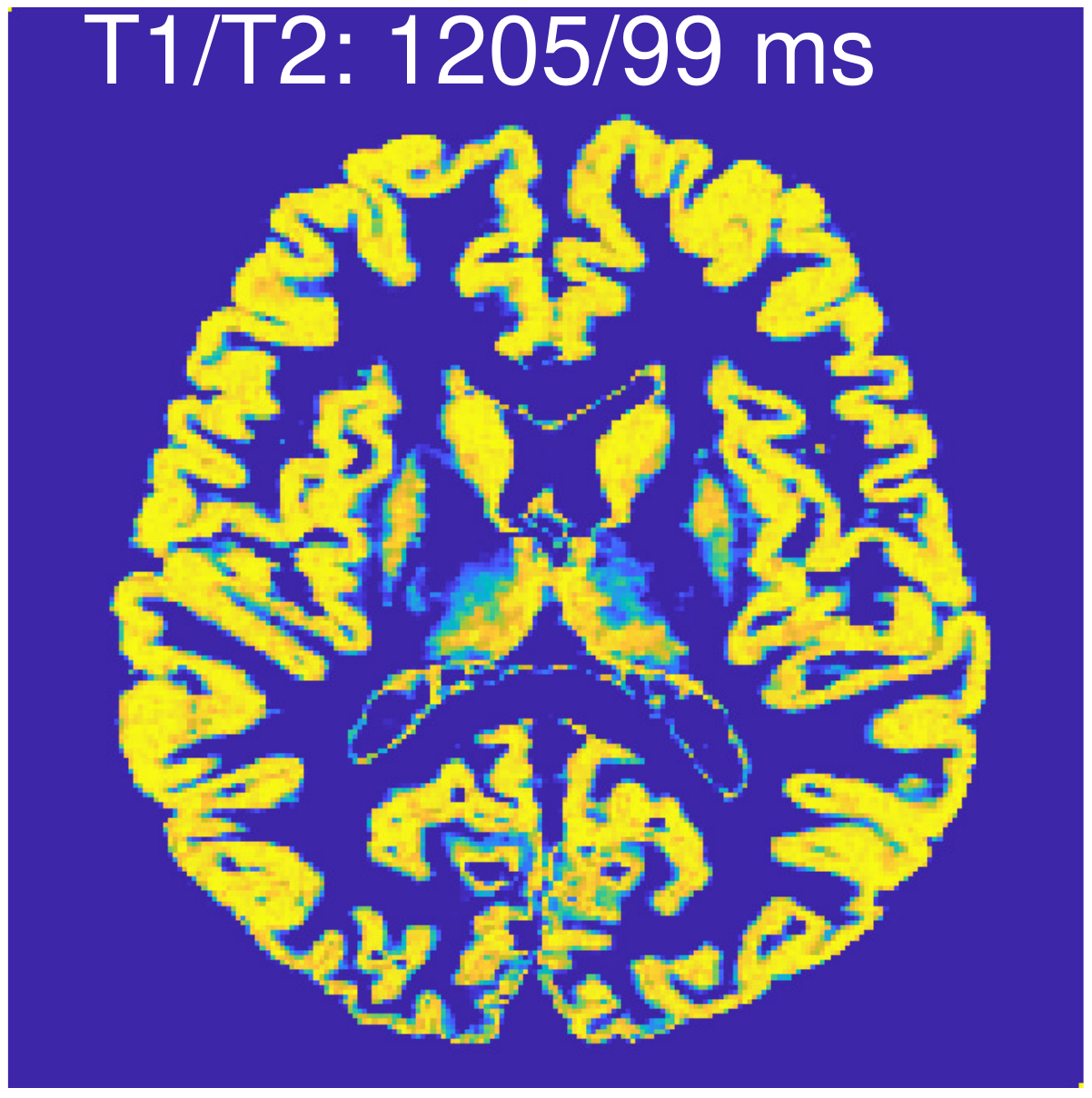}\hspace{-.1cm}
\includegraphics[width=.15\linewidth]{./figs/expe3_cut_im_50db_diff_1152.pdf}\hspace{.1cm}
\includegraphics[width=.15\linewidth]{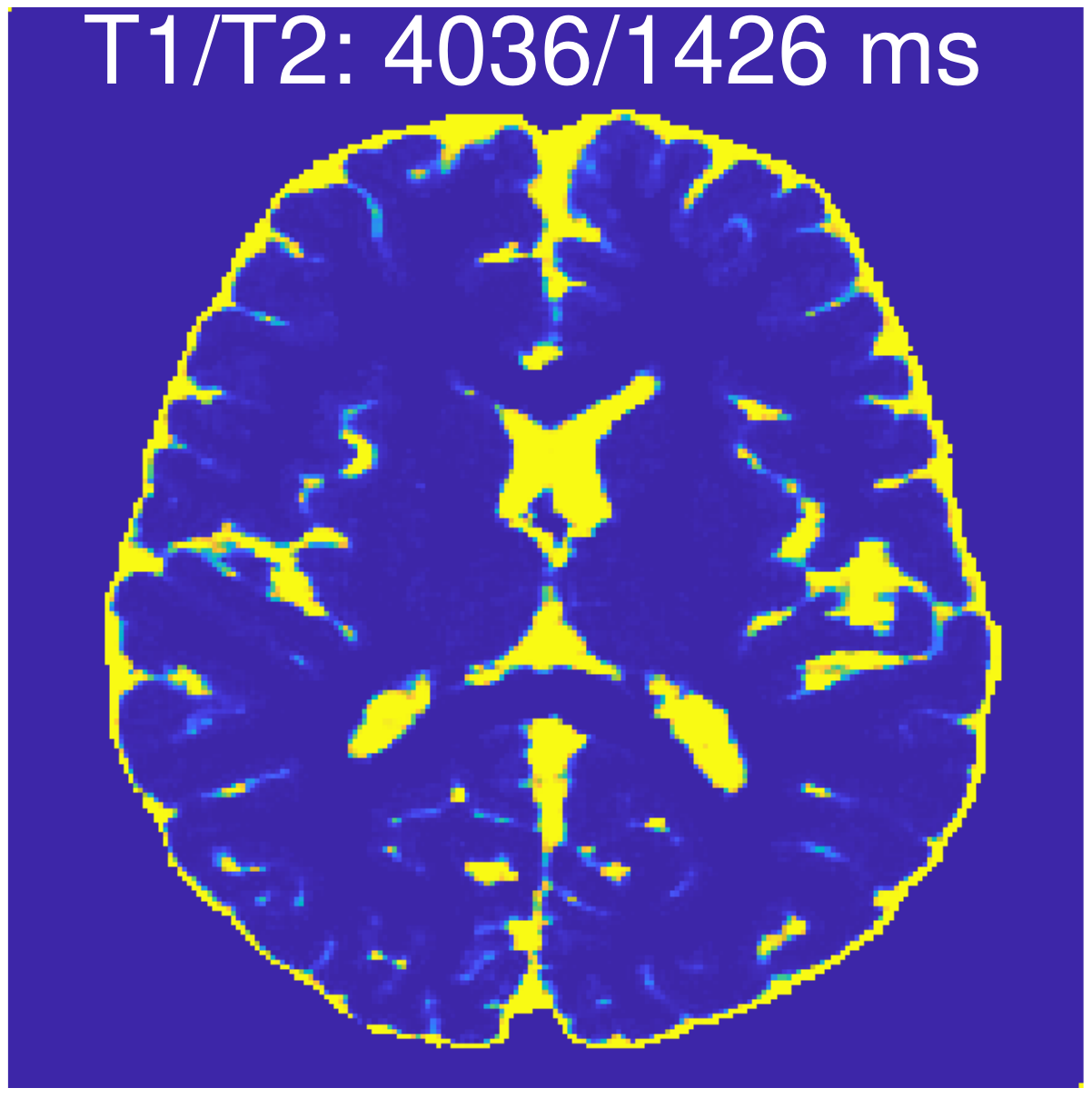}\hspace{-.1cm}
\includegraphics[width=.15\linewidth]{./figs/expe3_cut_im_50db_diff_1153.pdf}
\\
\begin{turn}{90} \qquad SVD-MRF \end{turn}
\includegraphics[width=.15\linewidth]{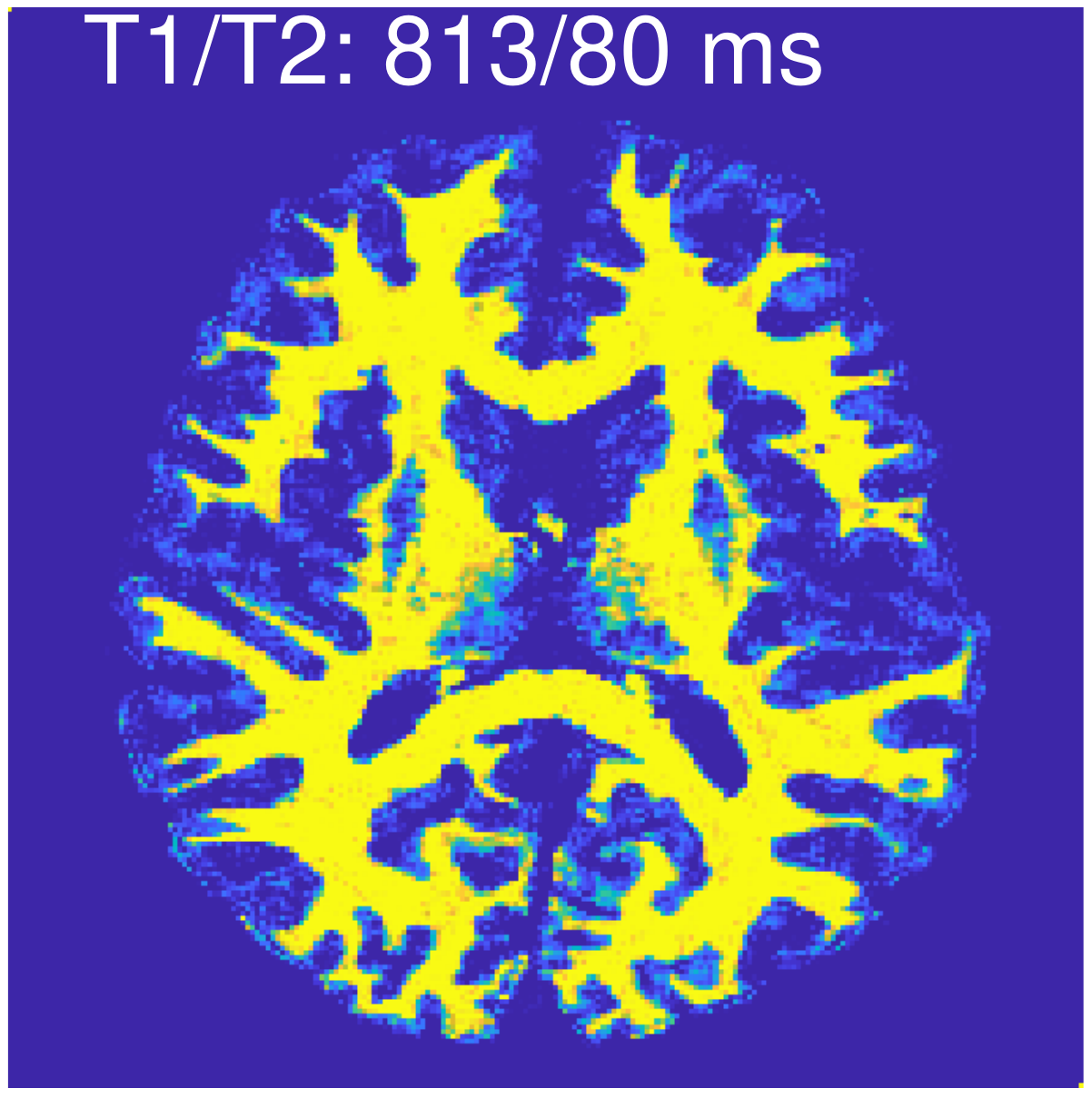}\hspace{-.1cm}
\includegraphics[width=.15\linewidth]{./figs/expe3_cut_im_50db_diff_1251.pdf}\hspace{.1cm}
\includegraphics[width=.15\linewidth]{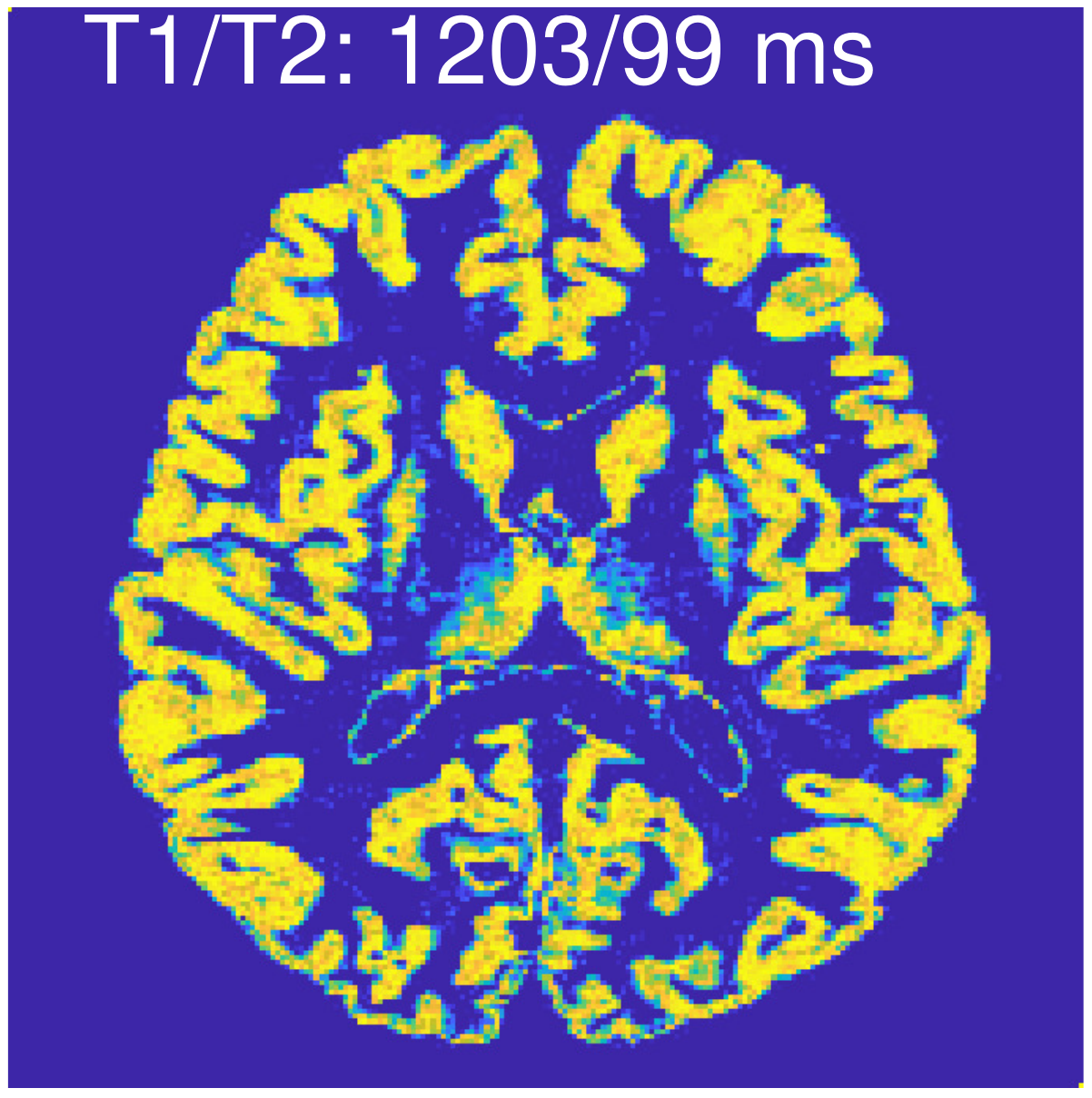}\hspace{-.1cm}
\includegraphics[width=.15\linewidth]{./figs/expe3_cut_im_50db_diff_1252.pdf}\hspace{.1cm}
\includegraphics[width=.15\linewidth]{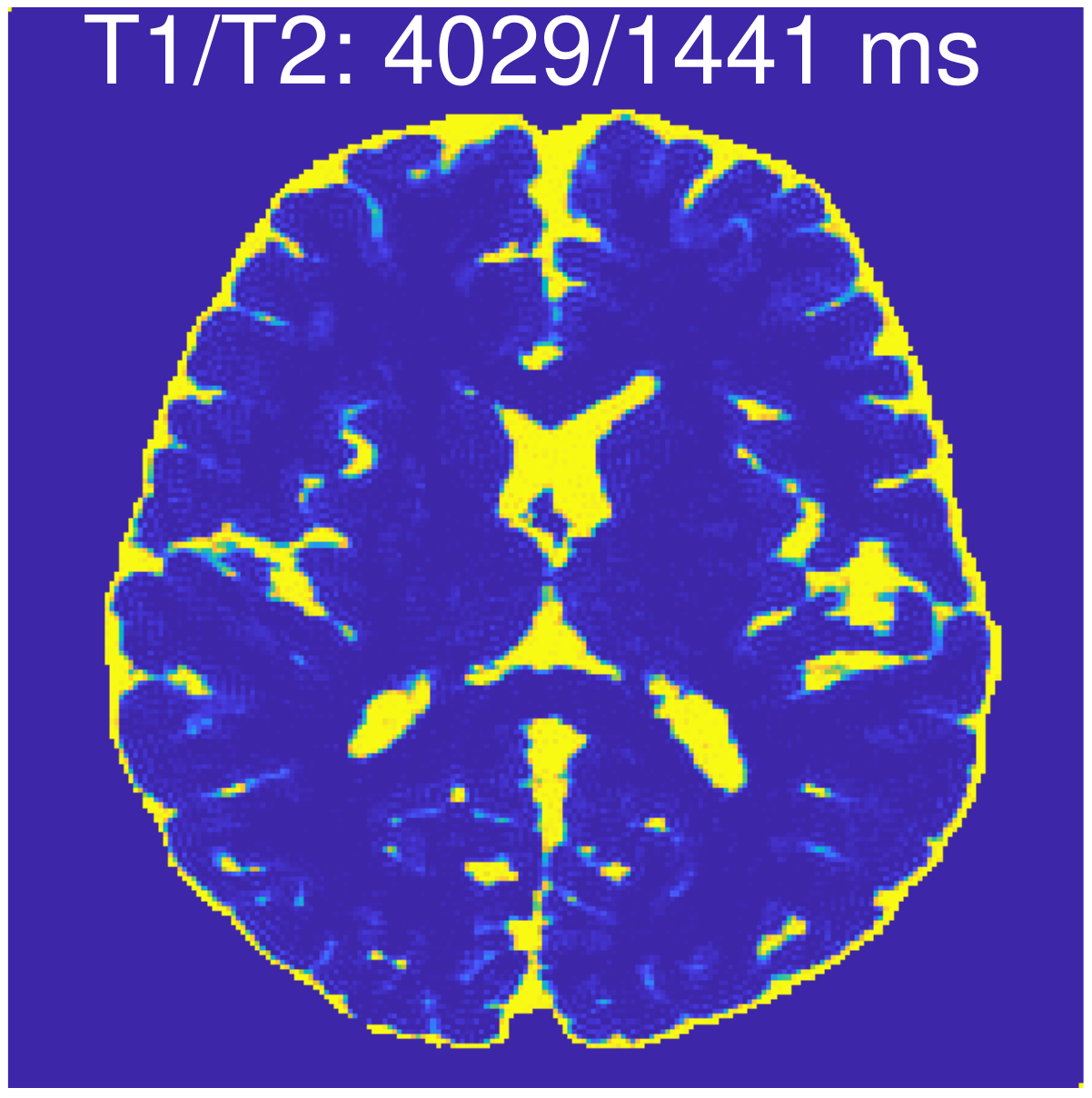}\hspace{-.1cm}
\includegraphics[width=.15\linewidth]{./figs/expe3_cut_im_50db_diff_1253.pdf}
\\
\hspace{.2cm}
\includegraphics[width=.13\linewidth]{./figs/colorbarMaps_c.pdf}\hspace{.3cm}
\includegraphics[width=.13\linewidth]{./figs/colorbardiffMaps.pdf}\hspace{.3cm}
\includegraphics[width=.13\linewidth]{./figs/colorbarMaps_c.pdf}\hspace{.3cm}
\includegraphics[width=.13\linewidth]{./figs/colorbardiffMaps.pdf}\hspace{.3cm}
\includegraphics[width=.13\linewidth]{./figs/colorbarMaps_c.pdf}\hspace{.3cm}
\includegraphics[width=.13\linewidth]{./figs/colorbardiffMaps.pdf}
\end{minipage}}
\caption{\footnotesize{The estimated T1/T2 values, mixture maps and their differences with the ground truth for the WM, GM and CSF compartments of the simulated brain phantom using \MCMRF. TSMIs were either un-compressed (Full) or subsampled using compressed sensing and reconstructed with the LRTV and SVD-MRF schemes before mixture separation. 
}\label{fig:brainweb_SGBlasso_main} }
\end{figure*}

\newpage

\subsection{SVD-MRF reconstruction prior to mixture separation (in-vivo experiment)}

\begin{figure}[h!]
	\centering
	\scalebox{.7}{
	\begin{minipage}{\linewidth}
		\centering		
\includegraphics[width=.19\linewidth]{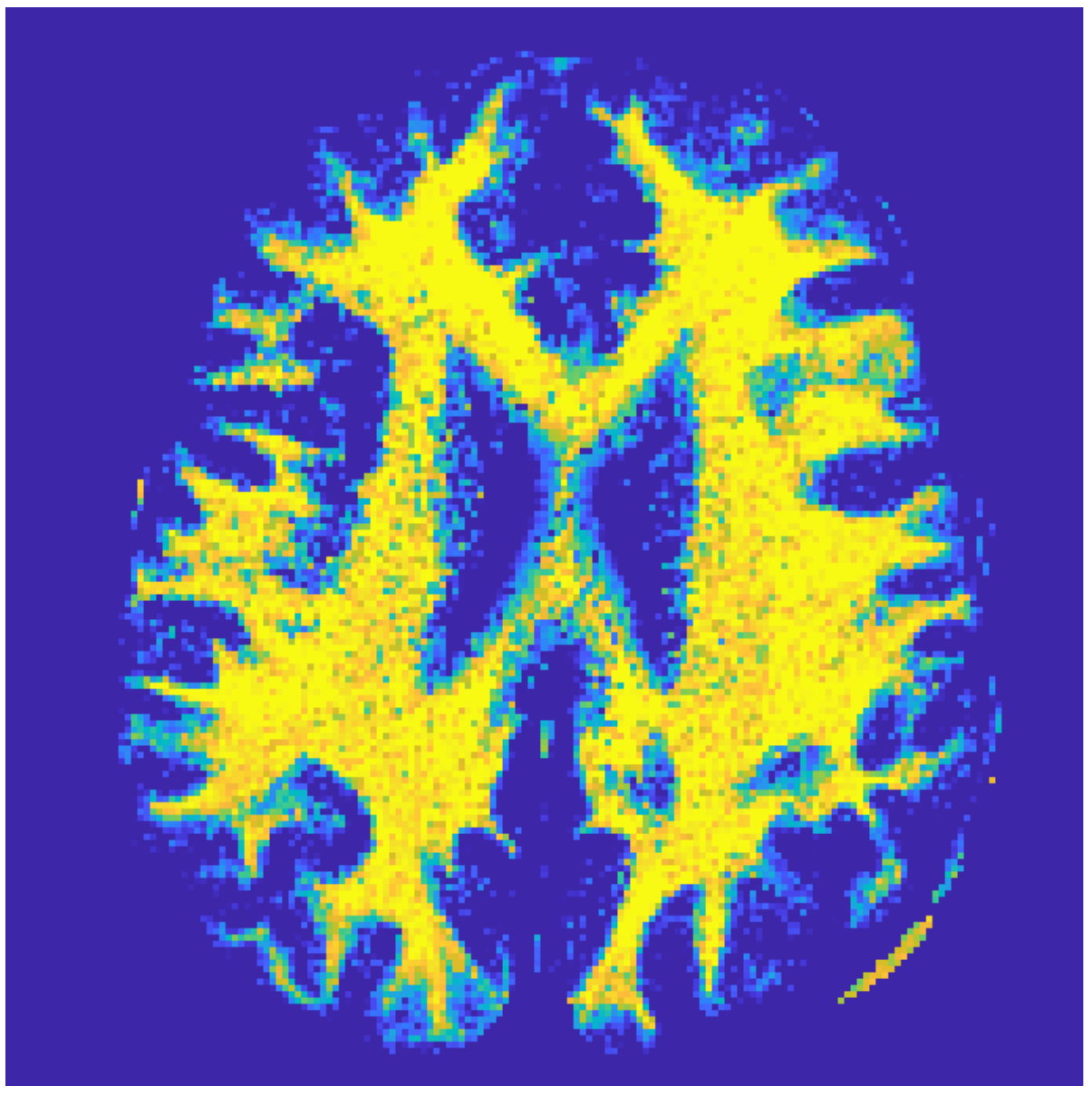}
\includegraphics[width=.19\linewidth]{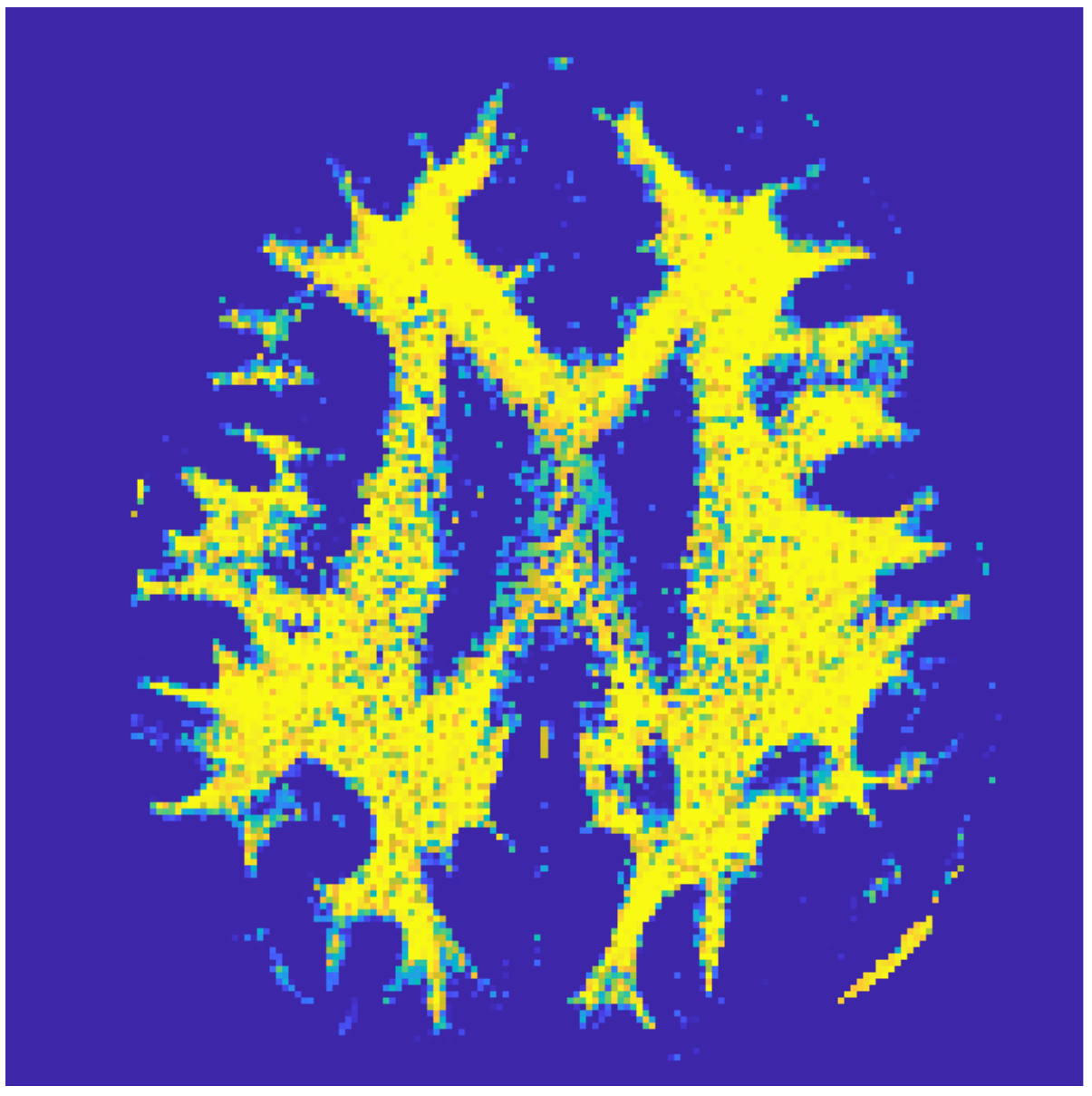}
\includegraphics[width=.19\linewidth]{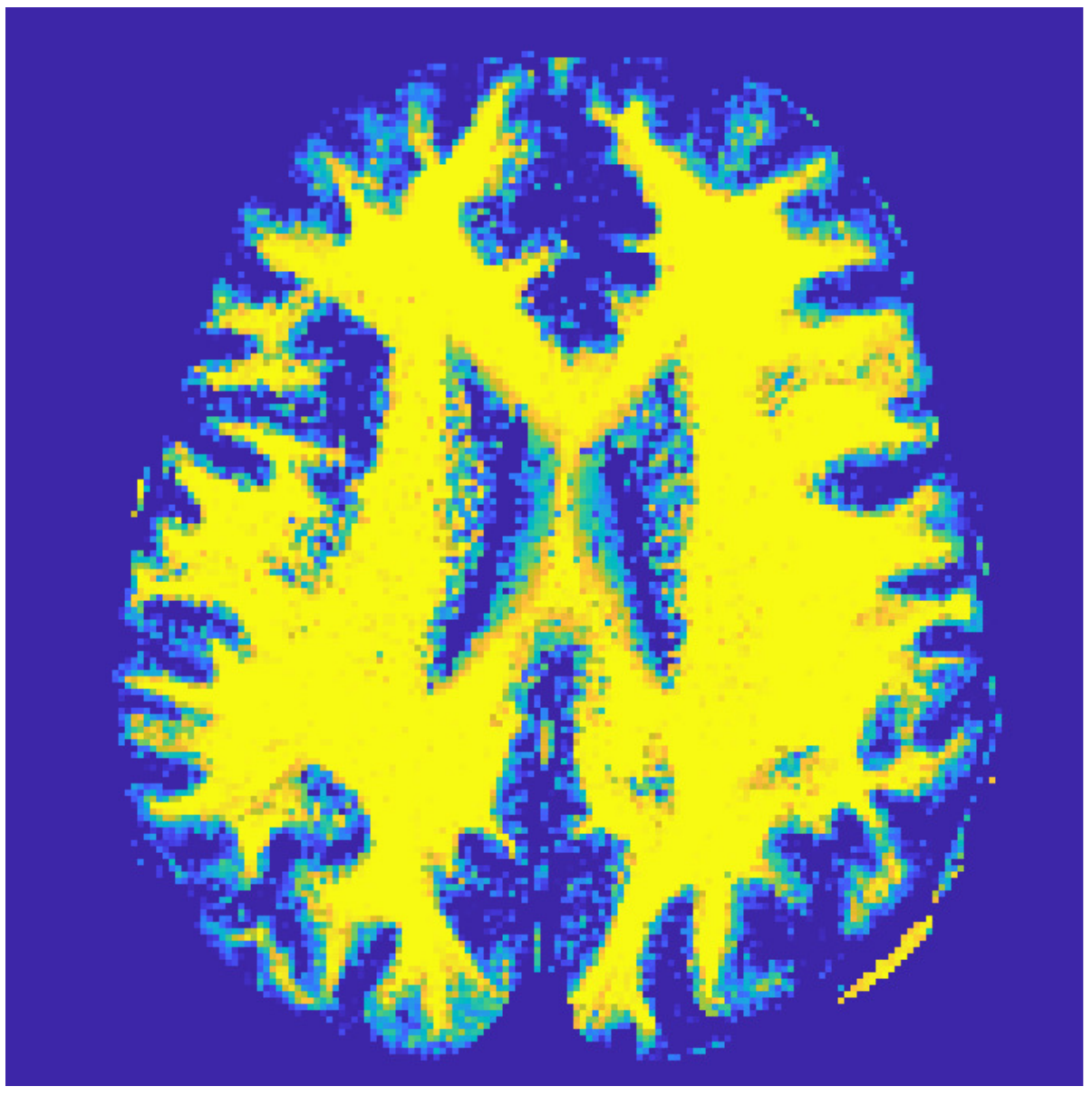}
\includegraphics[width=.19\linewidth]{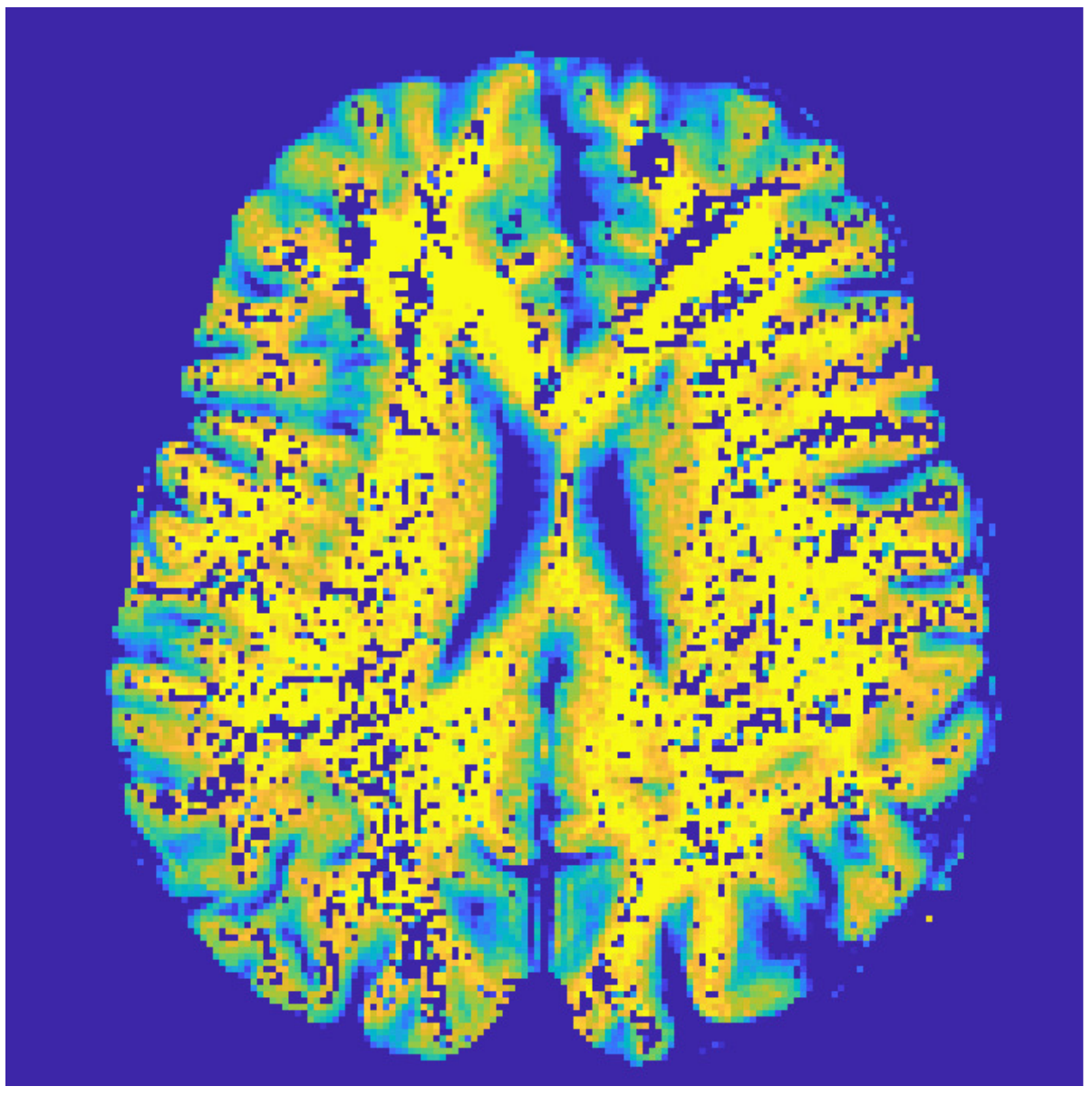}
\includegraphics[width=.19\linewidth]{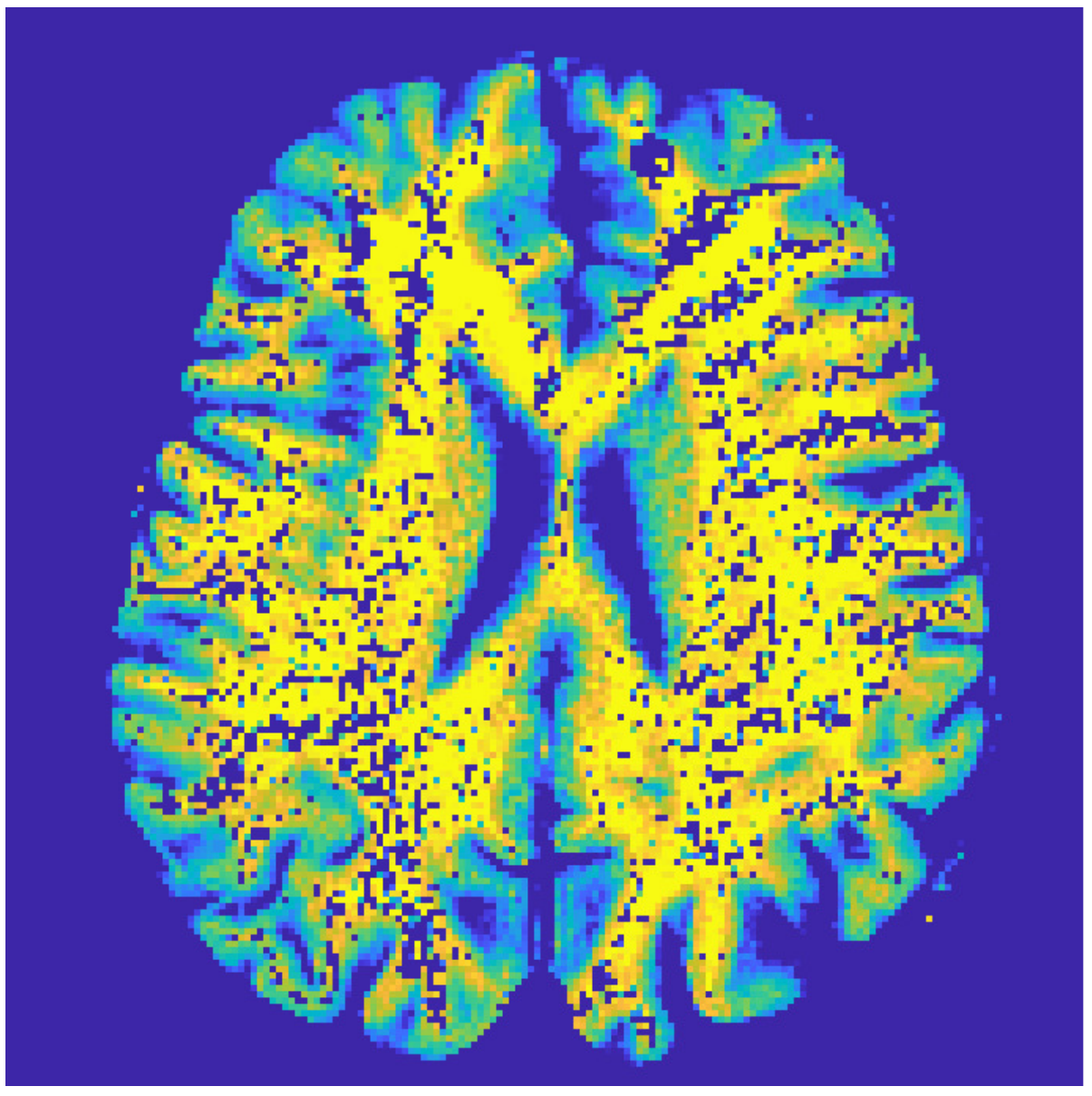}
\\
\includegraphics[width=.19\linewidth]{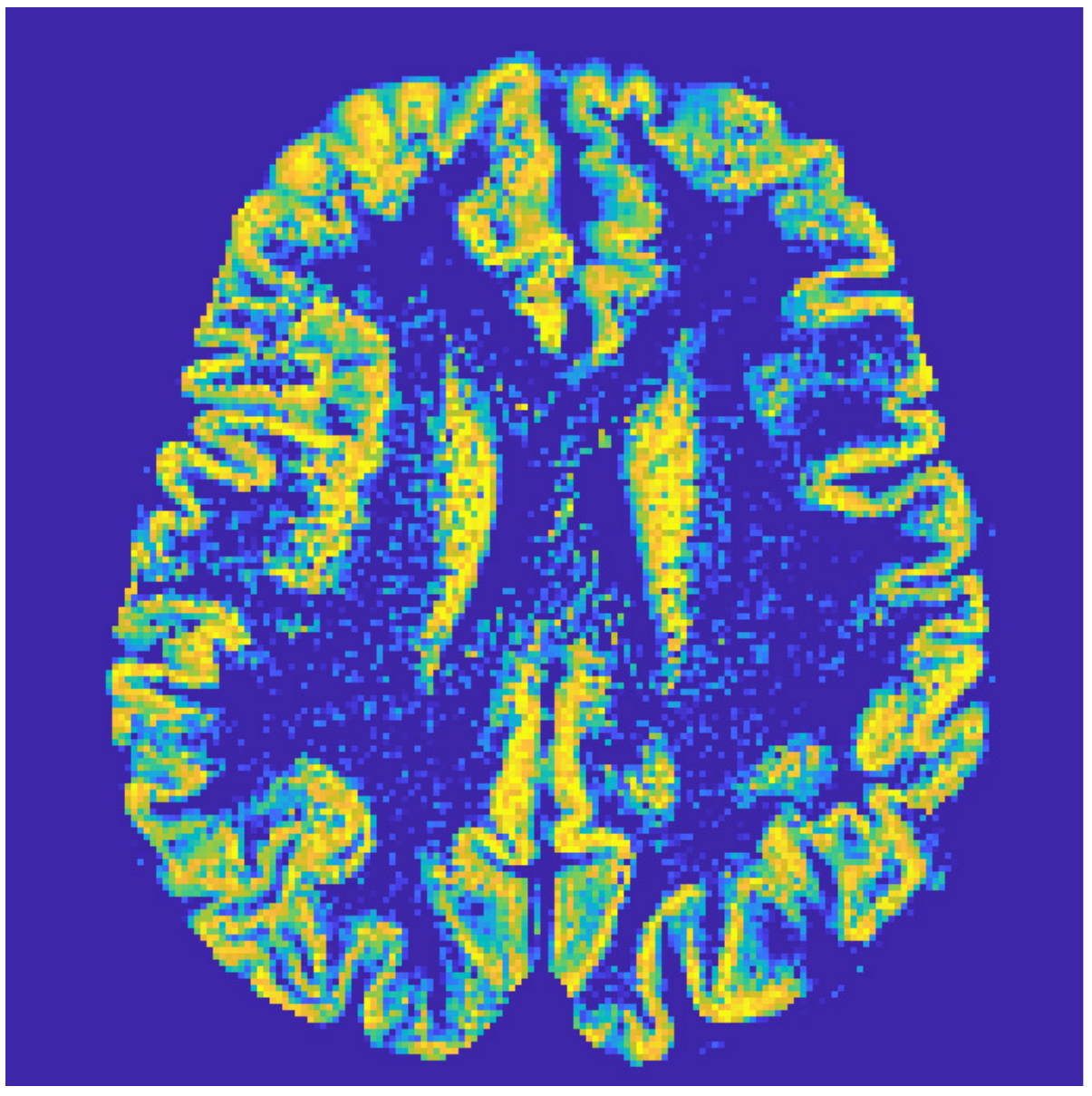}
\includegraphics[width=.19\linewidth]{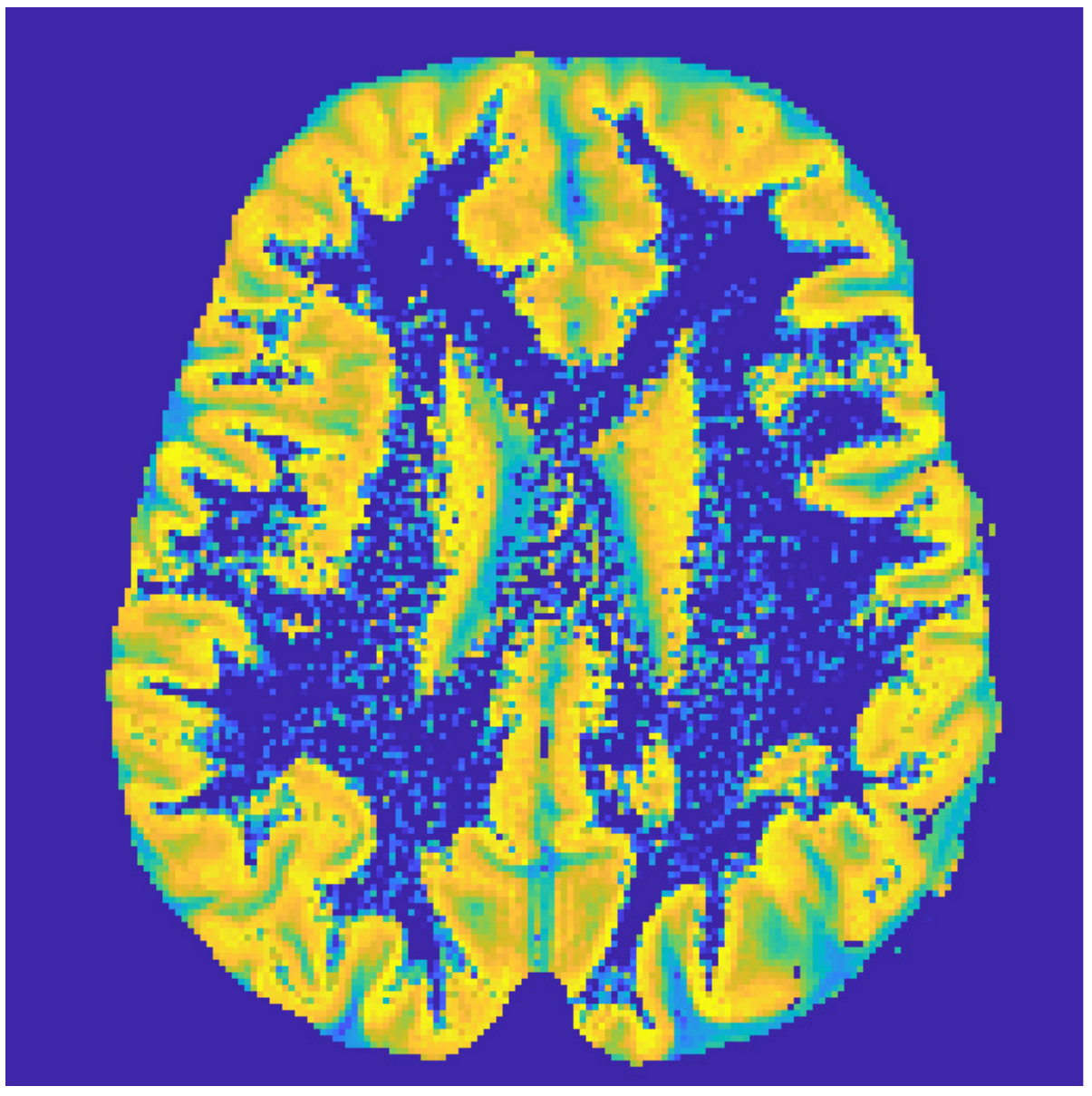}
\includegraphics[width=.19\linewidth]{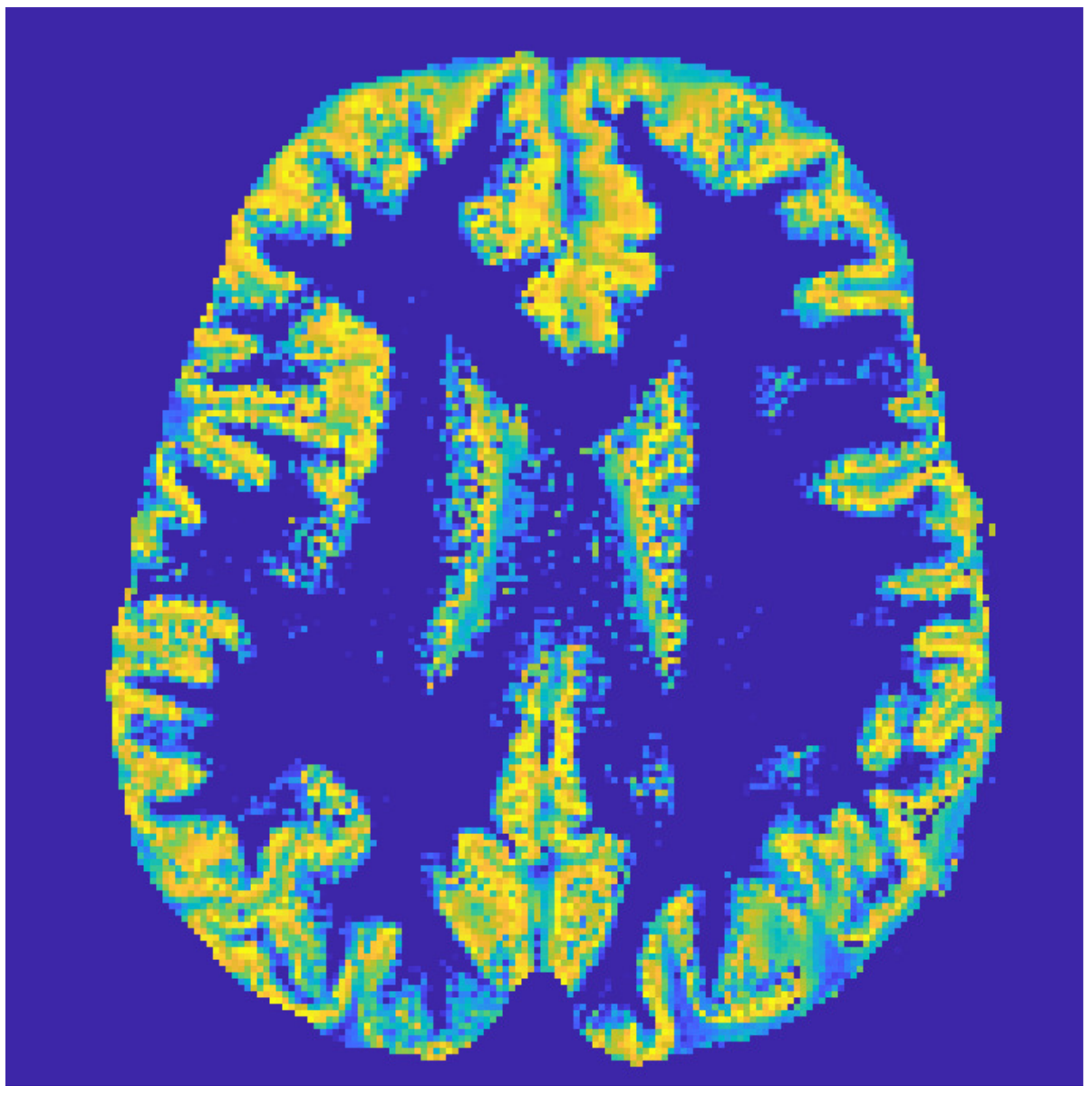}
\includegraphics[width=.19\linewidth]{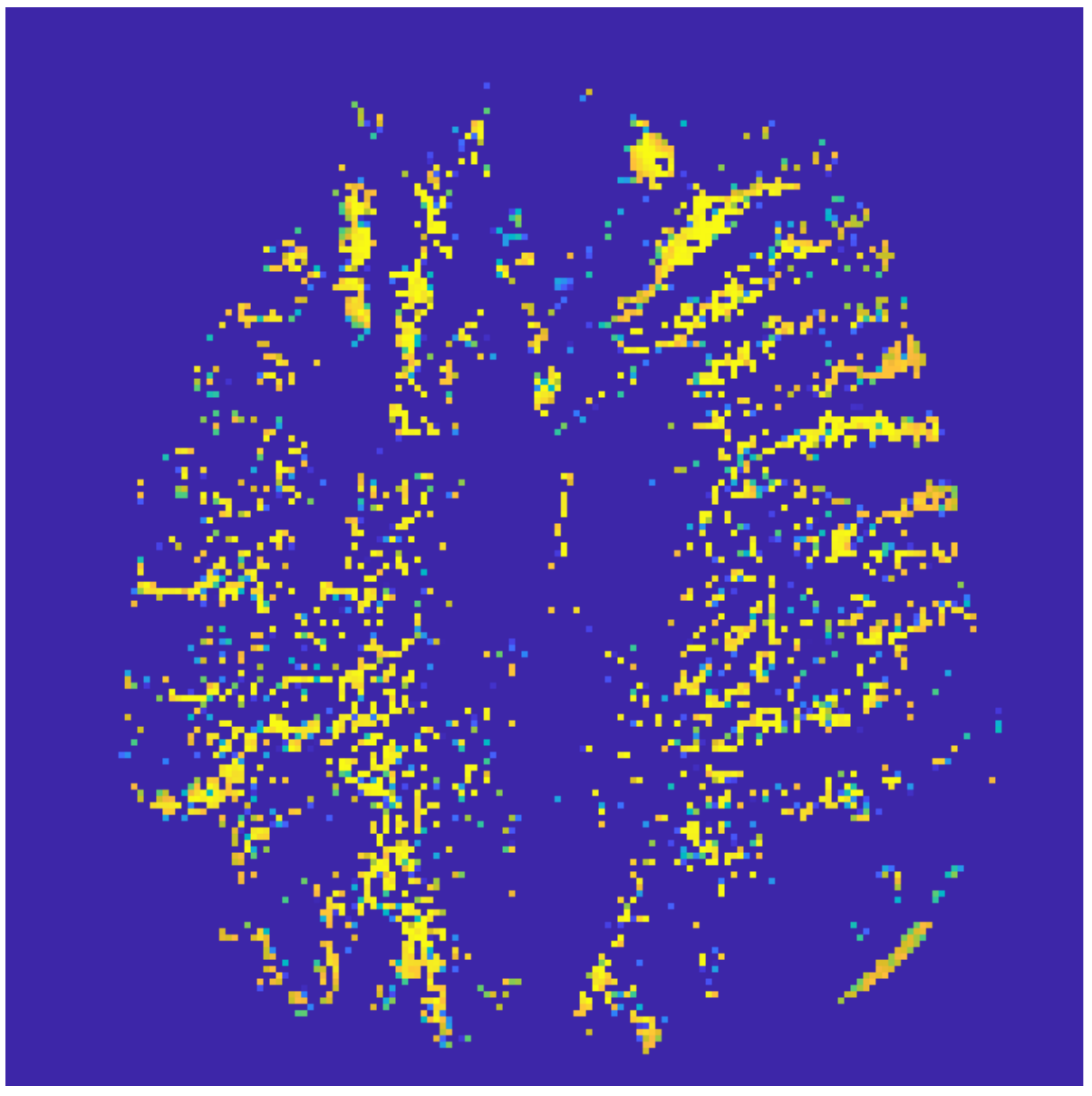}
\includegraphics[width=.19\linewidth]{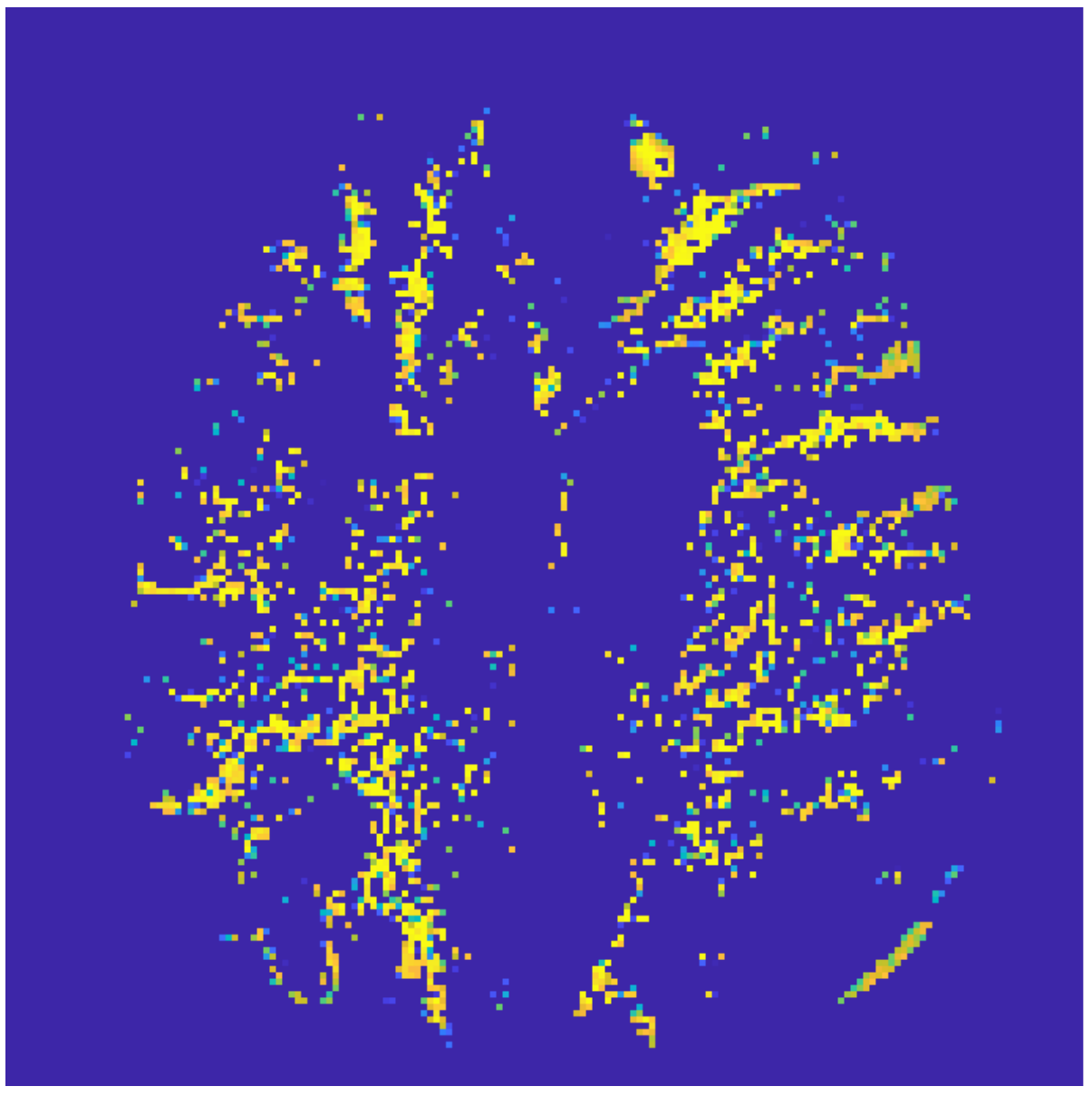}
\\
\includegraphics[width=.19\linewidth]{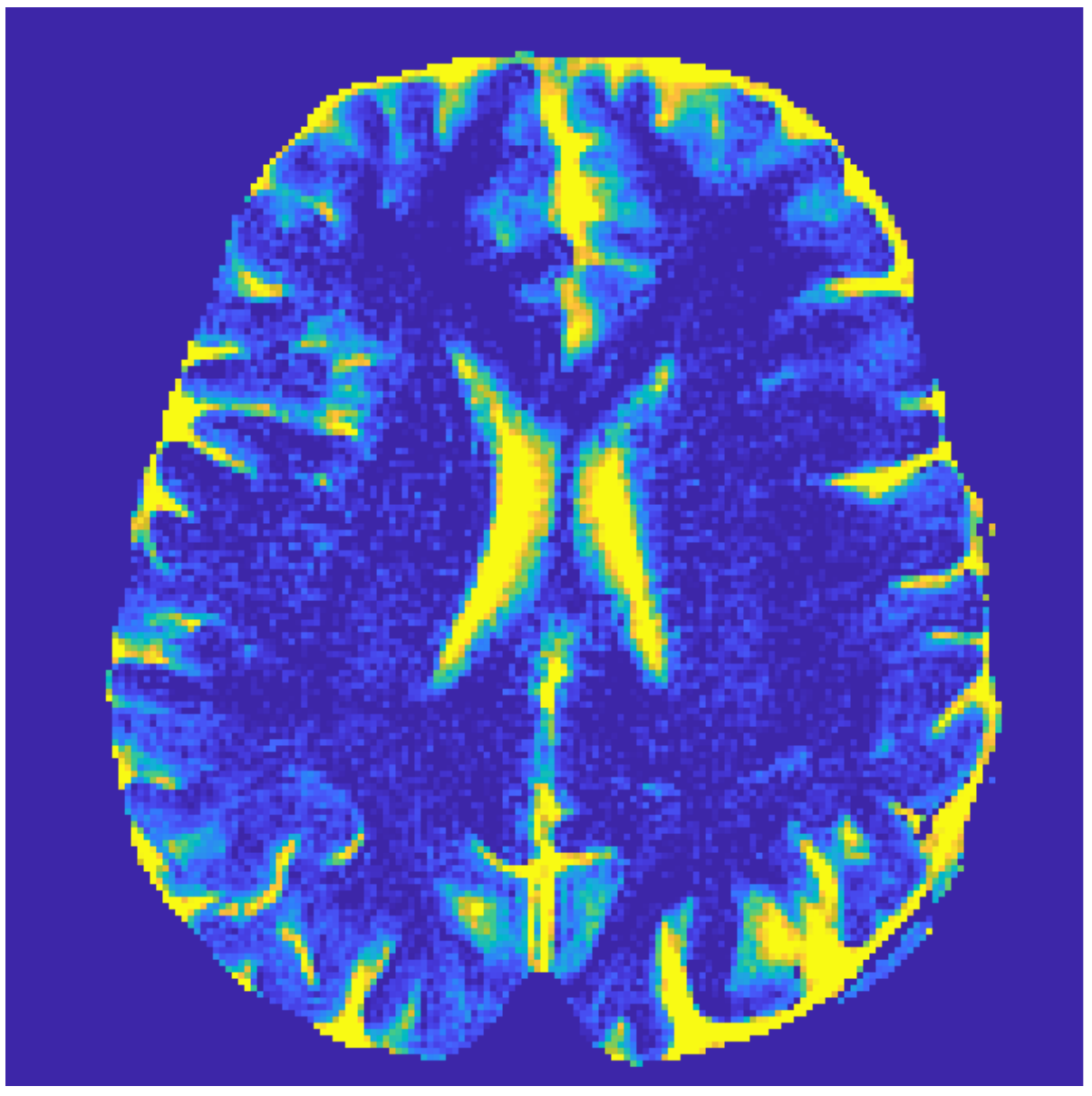}
\includegraphics[width=.19\linewidth]{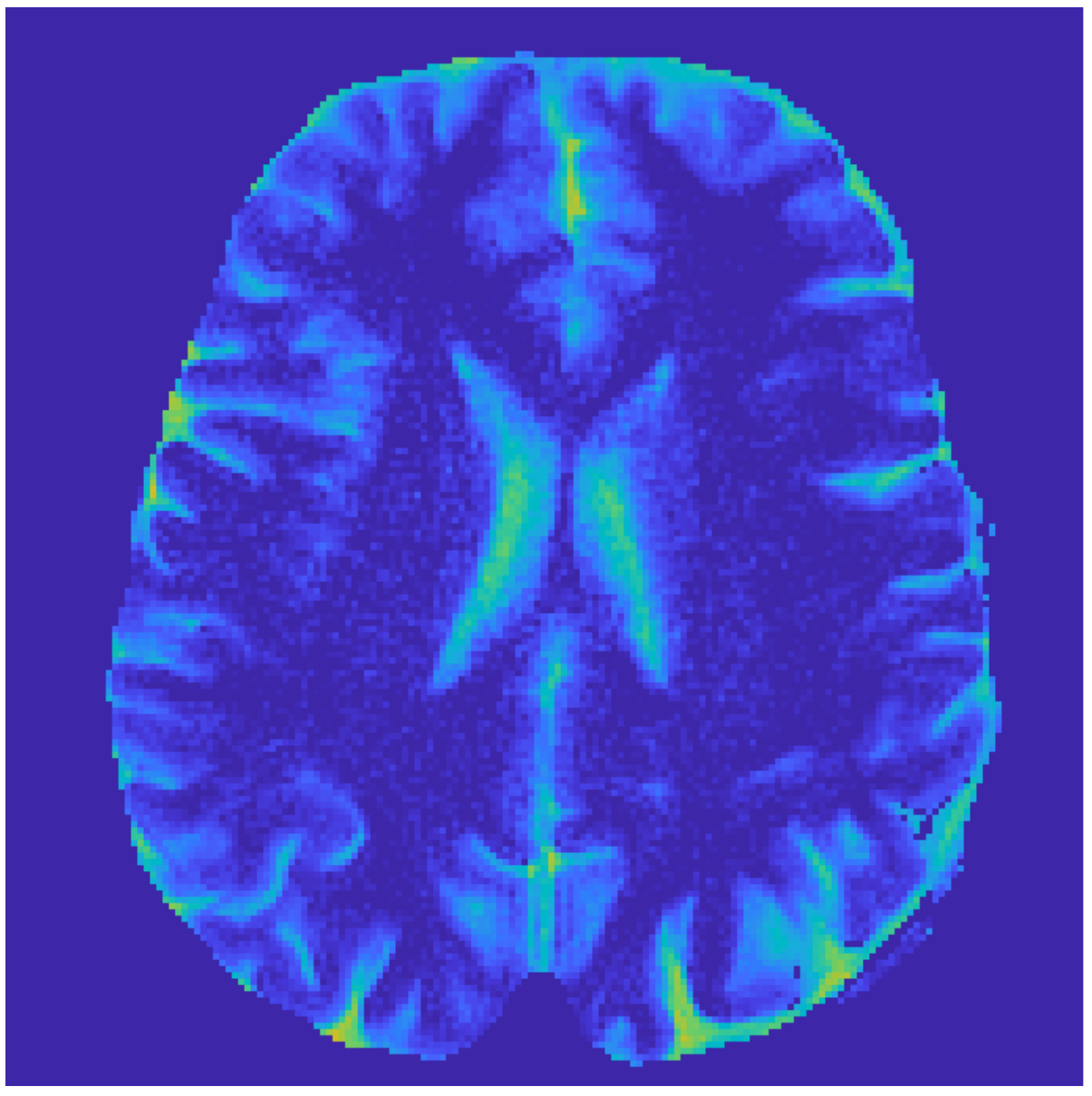}
\includegraphics[width=.19\linewidth]{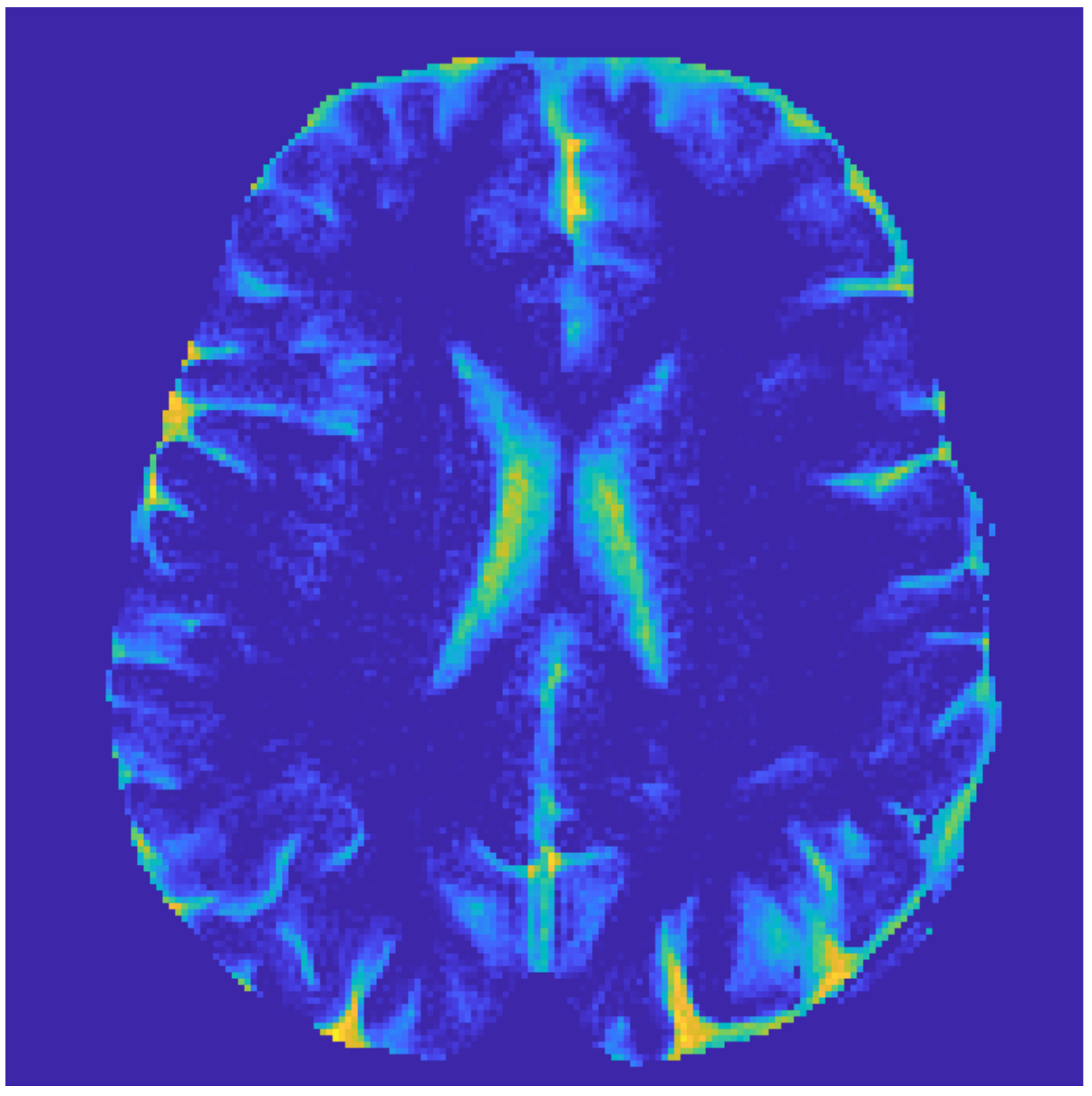}
\includegraphics[width=.19\linewidth]{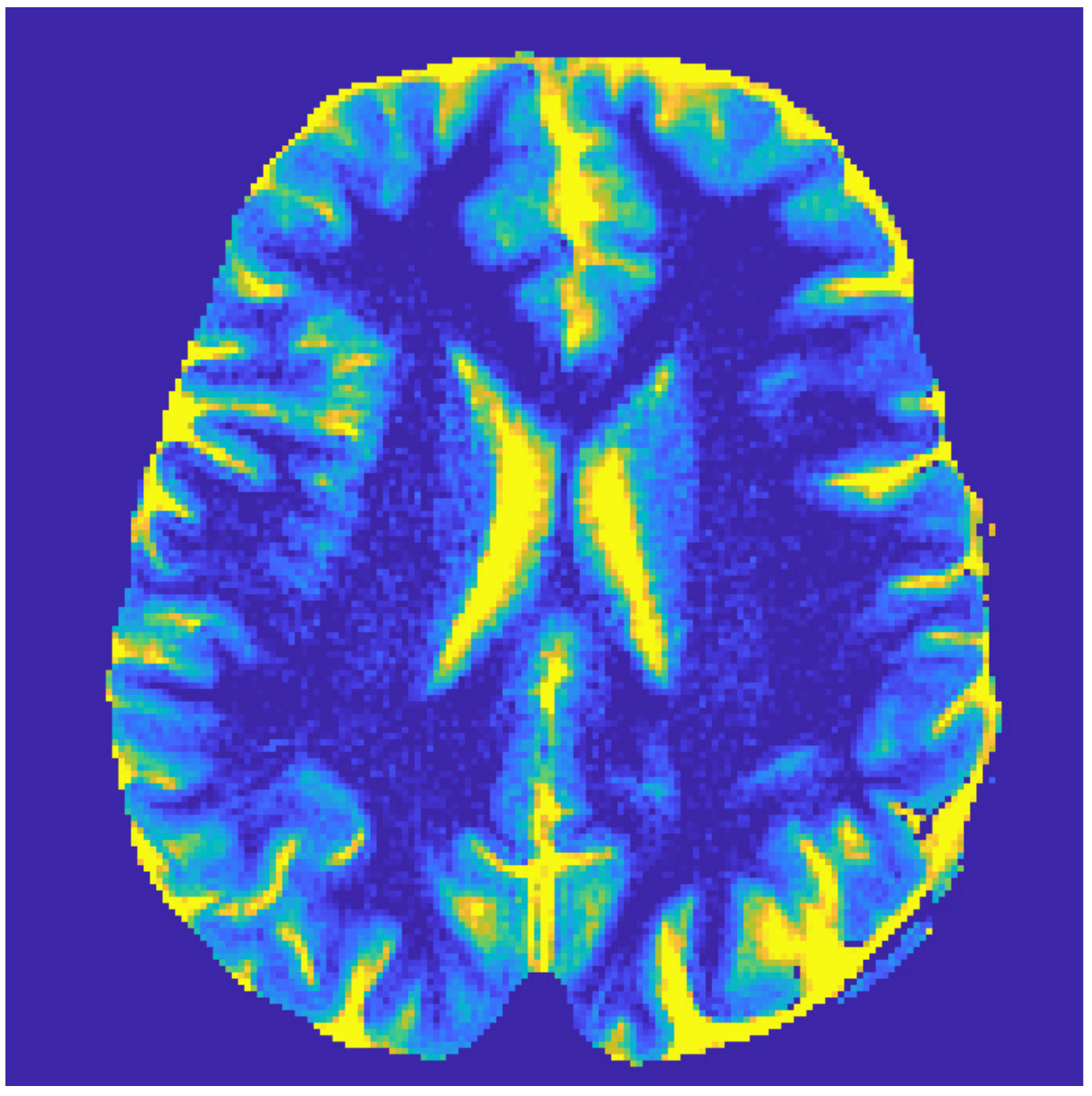}
\includegraphics[width=.19\linewidth]{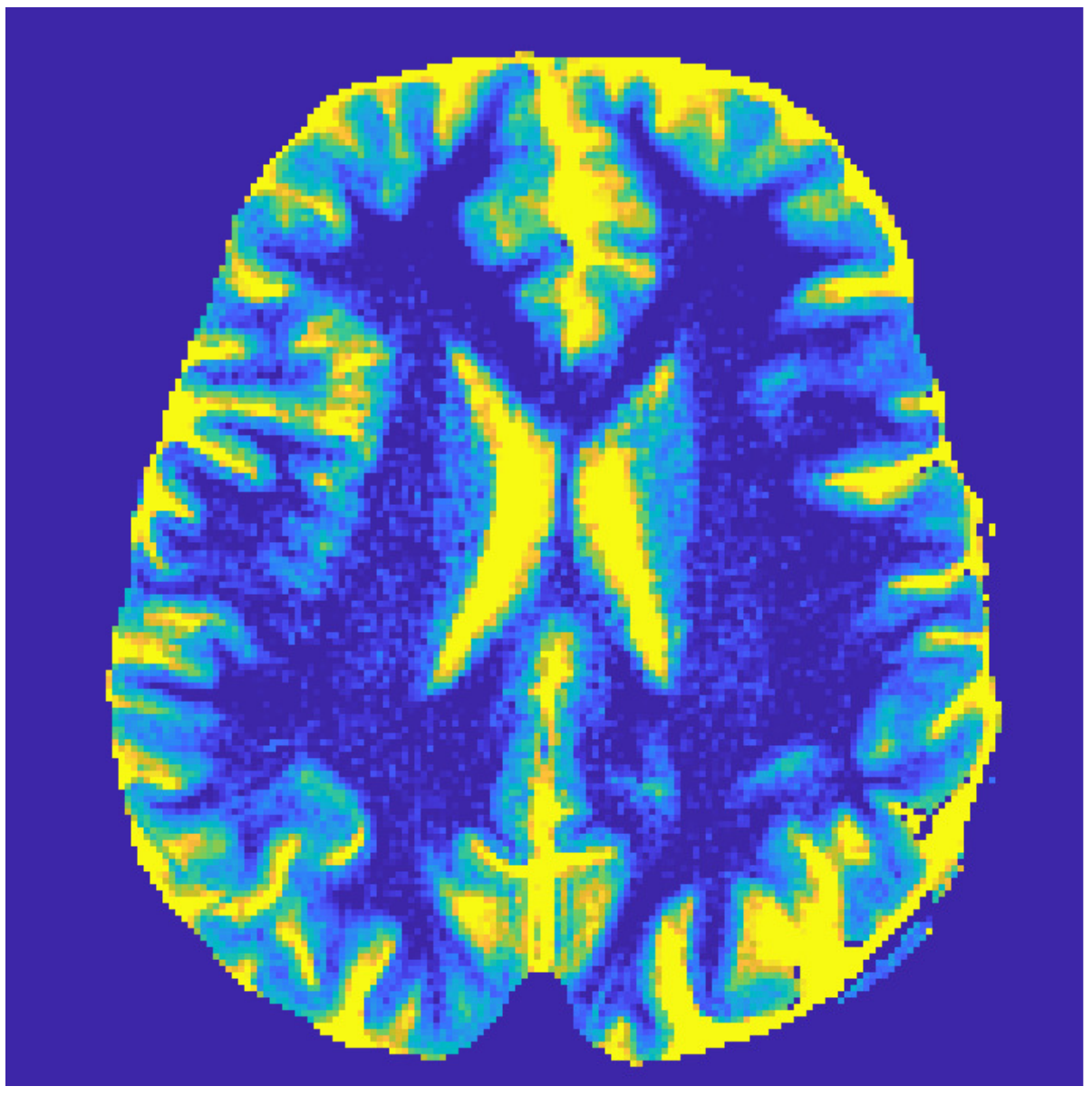}
\\
\MCMRF \hspace{2cm}PVMRF \hspace{2.2cm}SPIJN \hspace{2cm}BayesianMRF \hspace{2cm}SG-Lasso
\end{minipage}}
\caption{{Estimated mixture maps (margins are cropped) of the WM, GM and a CSF related compartment for in-vivo brain using different MC-MRF algorithms. 
Methods used SVD-MRF reconstructions before mixture separation. 
}\label{fig:vivo_compareMM_SVD} }
\vspace{-.2cm}
\end{figure}

\begin{table*}[h!]
	\centering
	\scalebox{1}{
		\begin{tabular}{c|cccc|cccc}
			\toprule[0.2em]
			\multicolumn{1}{c}{}&\multicolumn{4}{c}{T1 (ms)}&\multicolumn{4}{c}{T2 (ms)}  \\	
			\midrule[0.1em]		
			Tissue & \textbf{\MCMRF} & PVMRF & SPIJN & BayesianMRF & \textbf{\MCMRF} & PVMRF & SPIJN & BayesianMRF \\

			\midrule[0.1em]		
				\midrule[0.1em]					
			WM & 	788 	& 785	& 914 	& 	821  	  &77		& 80 		& 92  	& 77\\
			GM  &      1183 	& 990       & 1262	& 	874 	  & 104 	& 91		& 130	& 82 \\
			\bottomrule[0.2em]
		\end{tabular}		}
		\caption{Estimated T1/T2 values for in-vivo WM and GM compartments using MC-MRF algorithms. Methods used SVD-MRF reconstructions before mixture separation.
}
\label{tab:t1t2vivo_supp}
\end{table*}

\newpage
\subsection{Numerical box phantom experiment}

\begin{figure*}[h!]
	\centering	
	\scalebox{1}{
	\begin{minipage}{\linewidth}
		\centering	
\includegraphics[width=.19\linewidth]{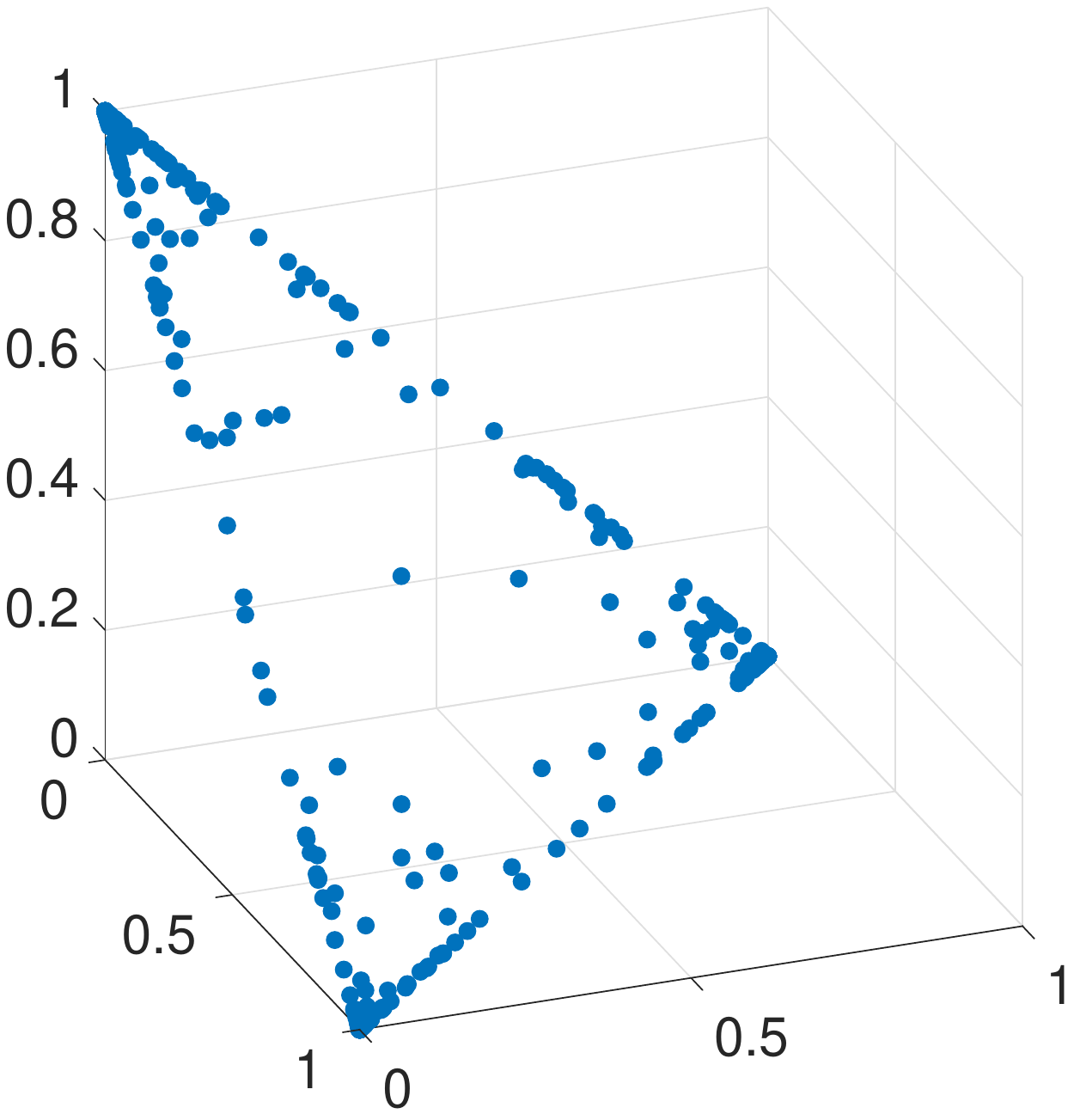}
\includegraphics[width=.19\linewidth]{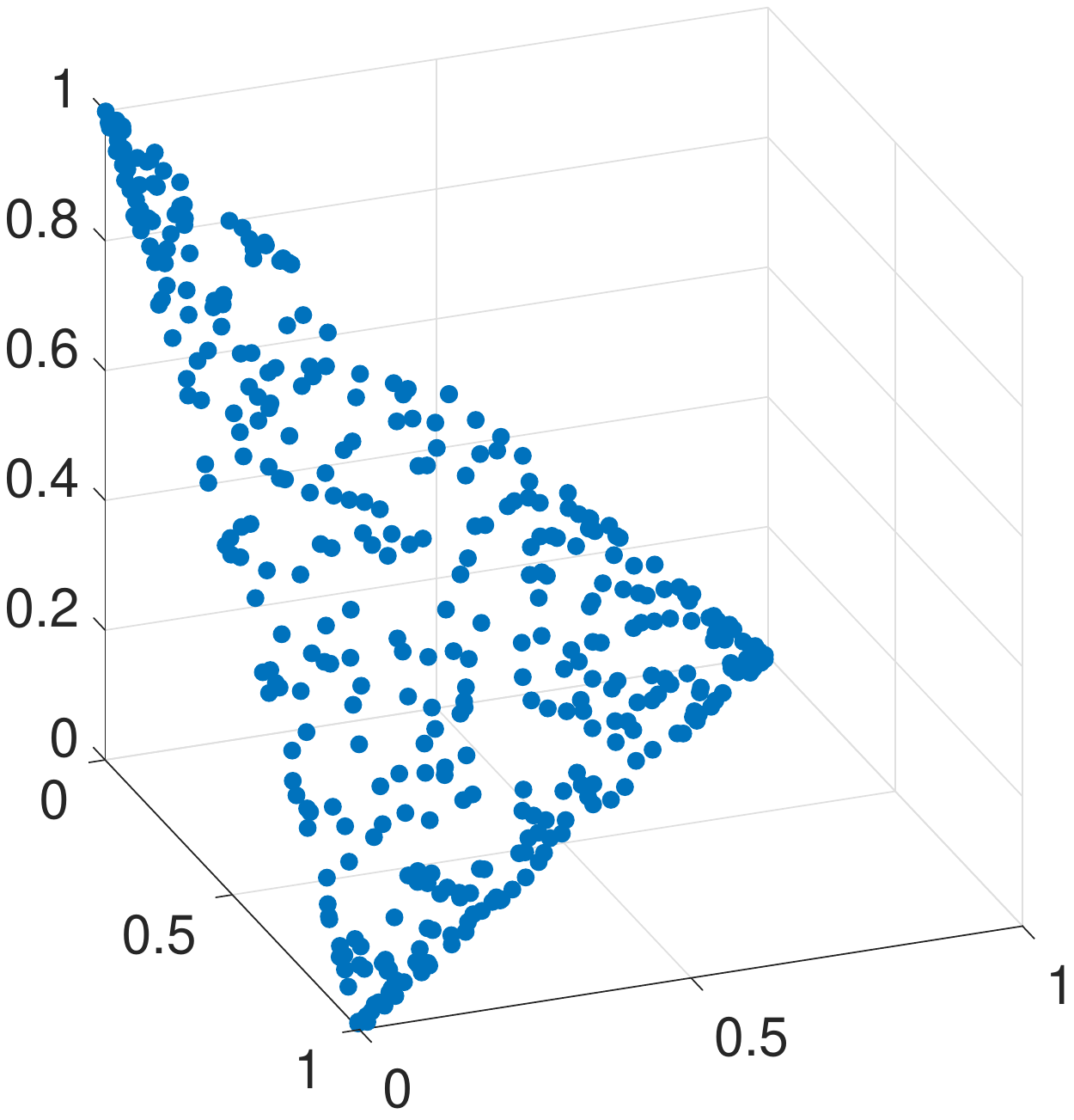}
\includegraphics[width=.19\linewidth]{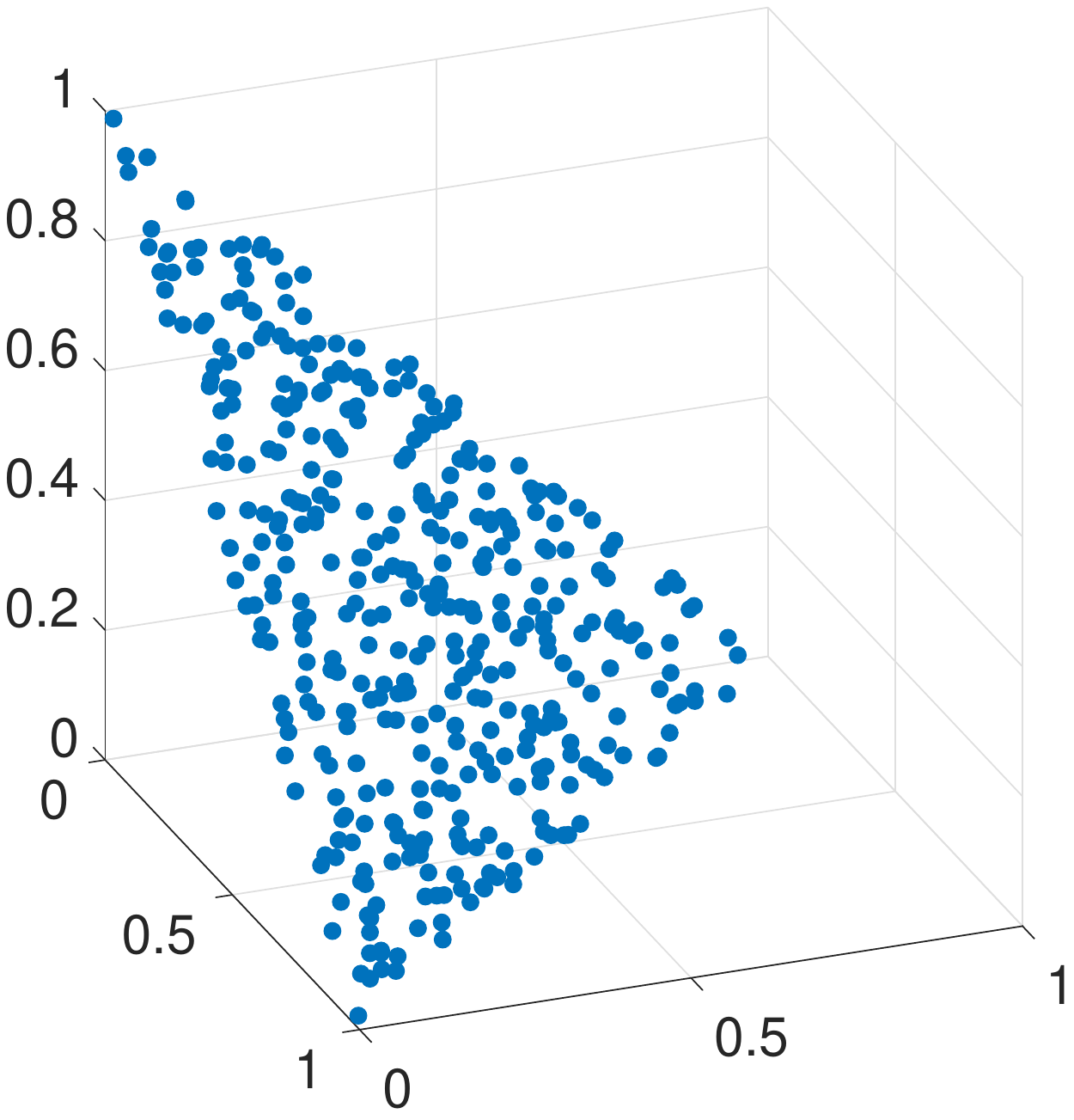}
\includegraphics[width=.19\linewidth]{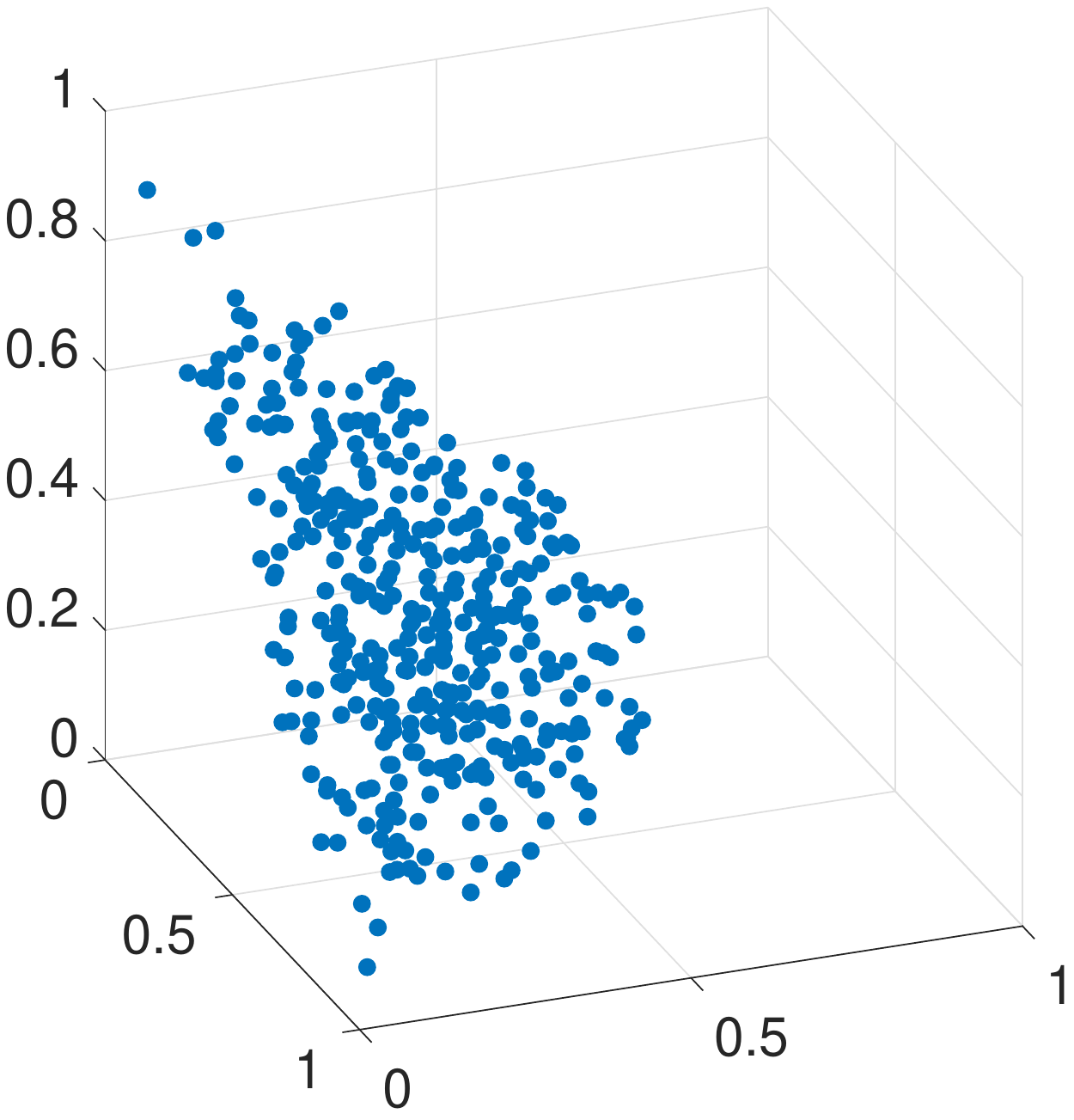}
\includegraphics[width=.19\linewidth]{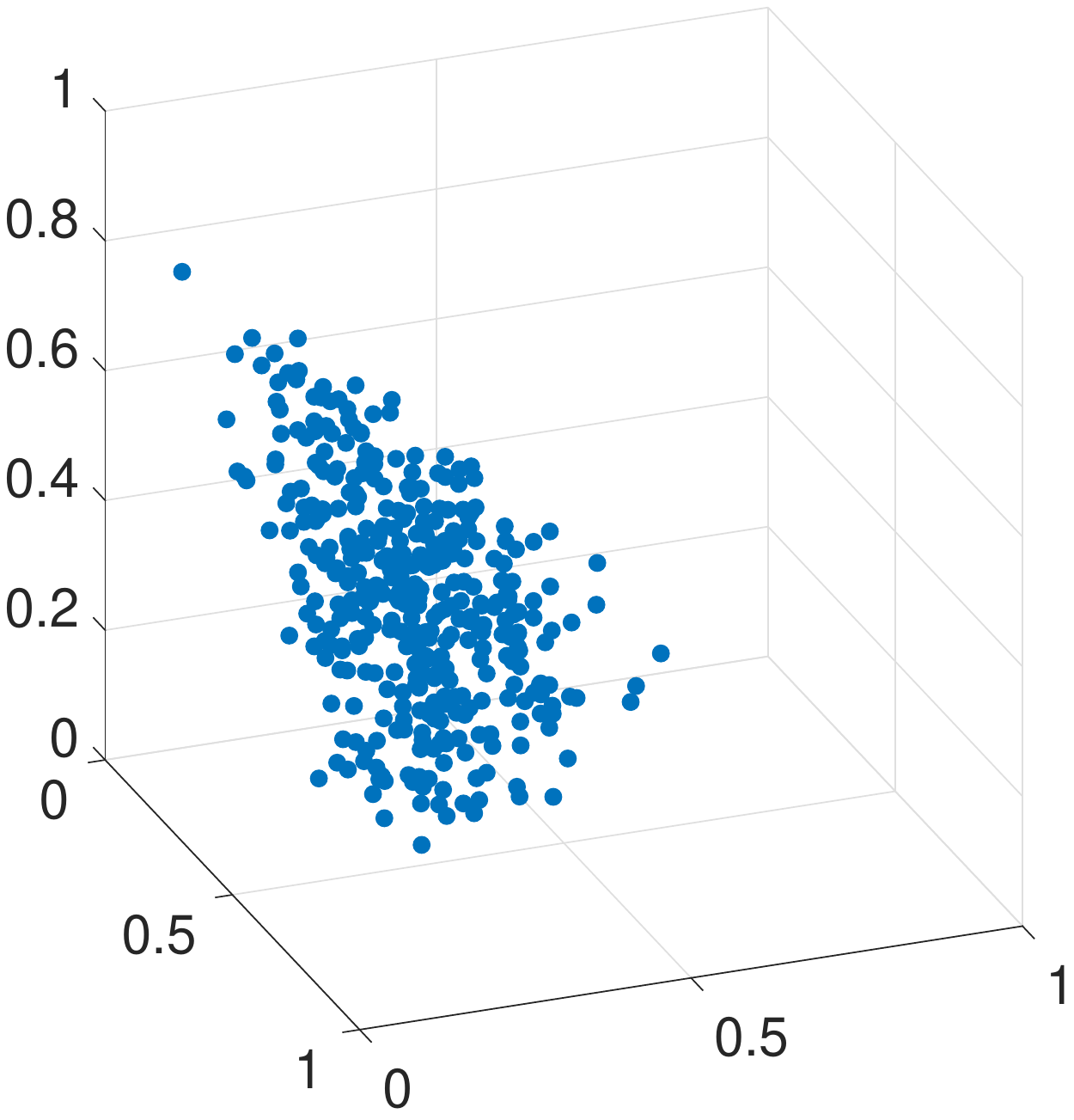}
\\
$a=0.1$ \hspace{2.5cm}$a=0.5$ \hspace{2.5cm}$a=1$ \hspace{2.5cm}$a=2$ \hspace{2.5cm}$a=4$ 
\\
\vspace{.2cm}
(a) Scatter plot of the pixels' mixture weights across the three compartments. \vspace{.3cm}
\\
\begin{turn}{90} \,1st compartment \end{turn}
\includegraphics[width=.15\linewidth]{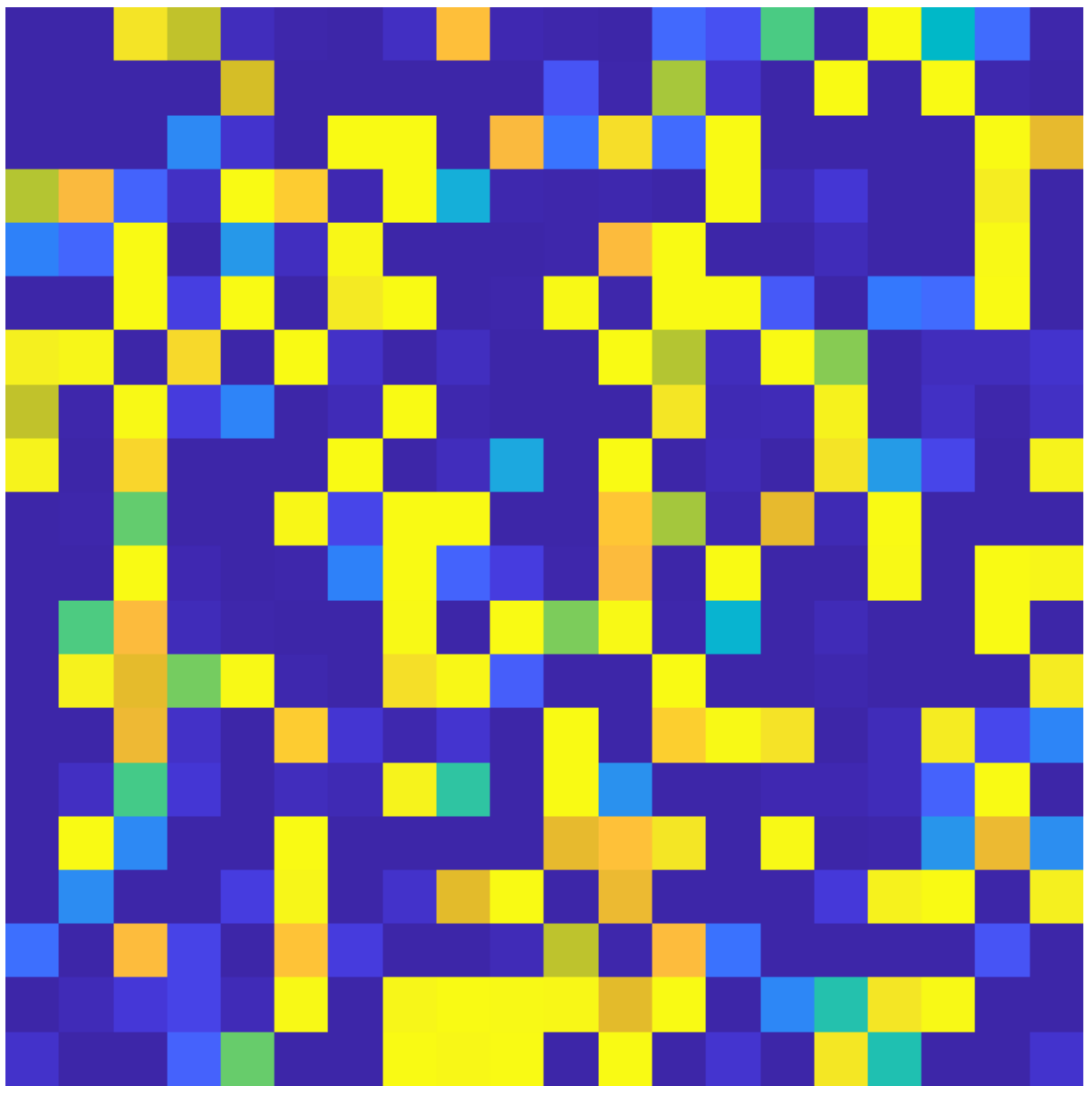}\hspace{.3cm}
\includegraphics[width=.15\linewidth]{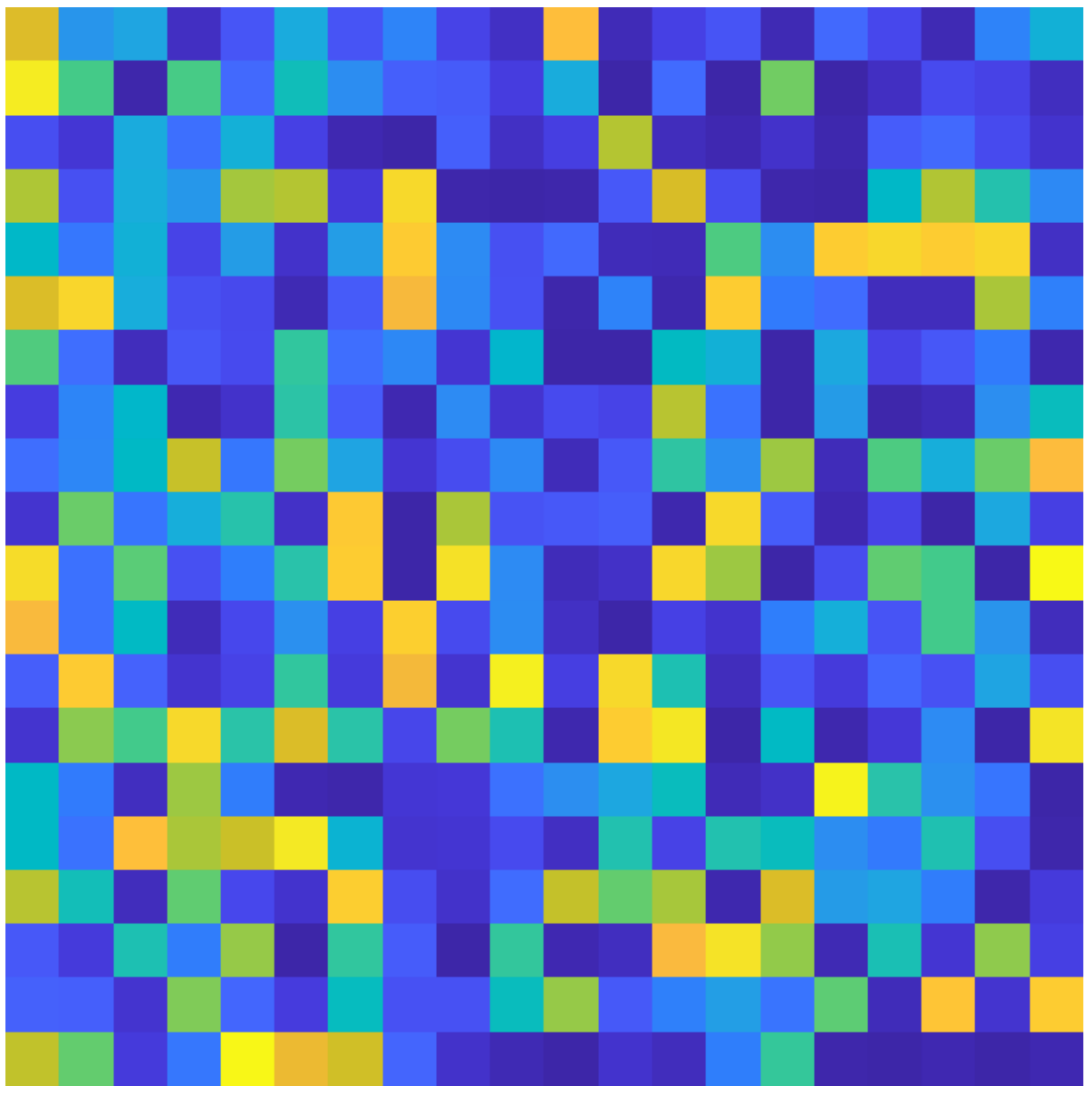}\hspace{.3cm}
\includegraphics[width=.15\linewidth]{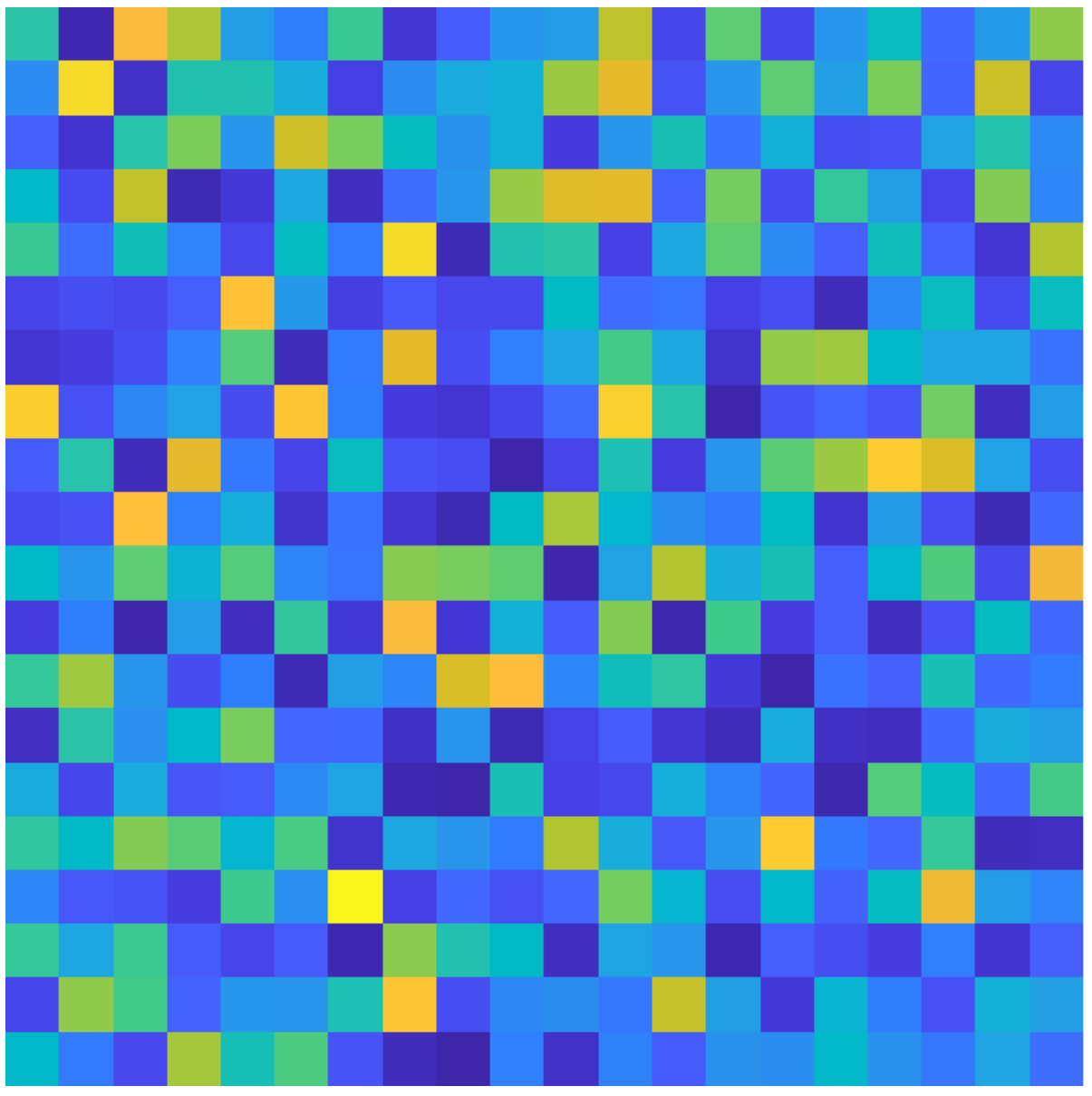}\hspace{.3cm}
\includegraphics[width=.15\linewidth]{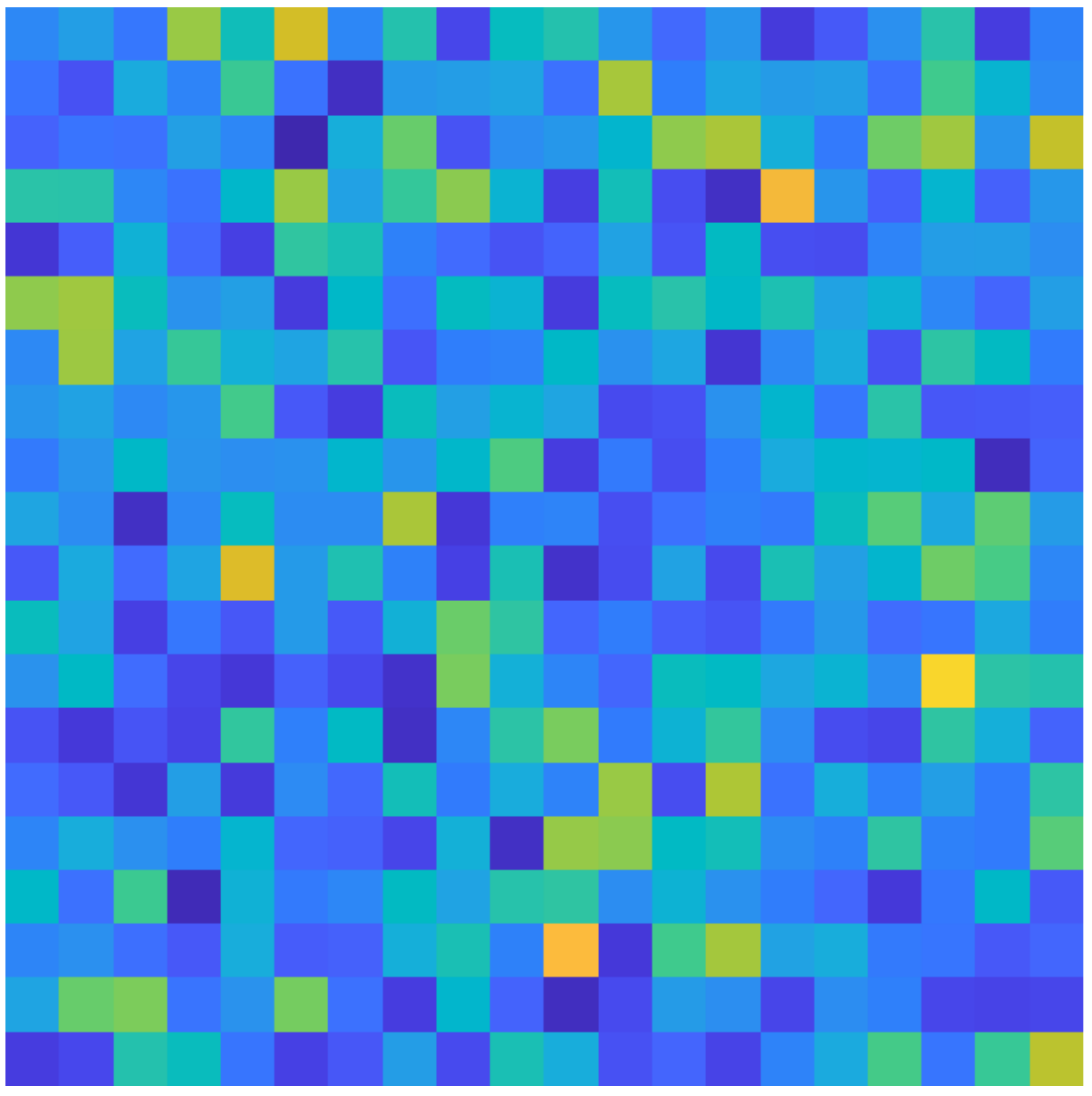}\hspace{.3cm}
\includegraphics[width=.15\linewidth]{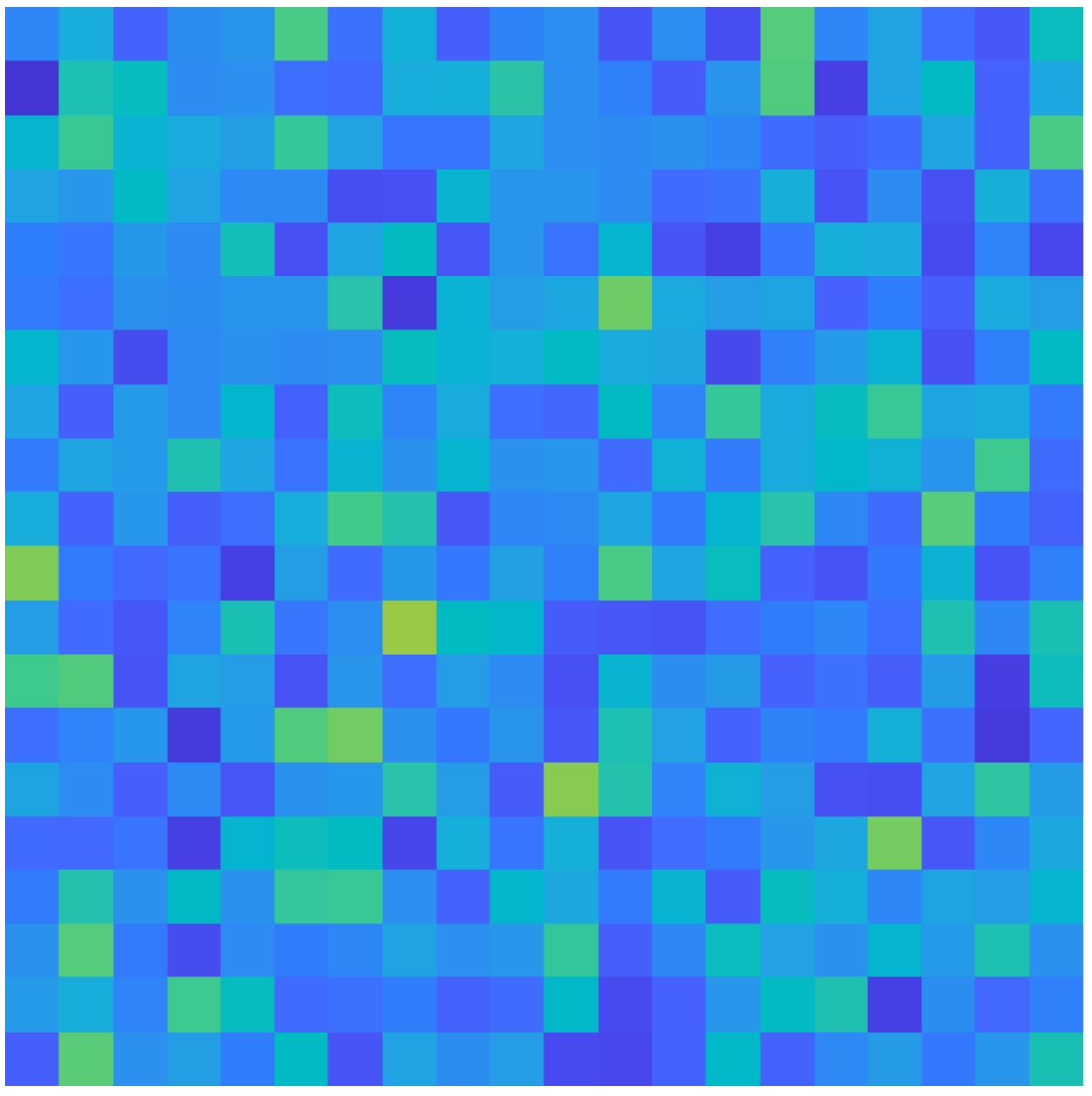}
\begin{turn}{90} \includegraphics[width=.13\linewidth]{./figs/colorbarMaps_c.pdf}\end{turn}
\\
\begin{turn}{90} 2nd compartment \end{turn}
\includegraphics[width=.15\linewidth]{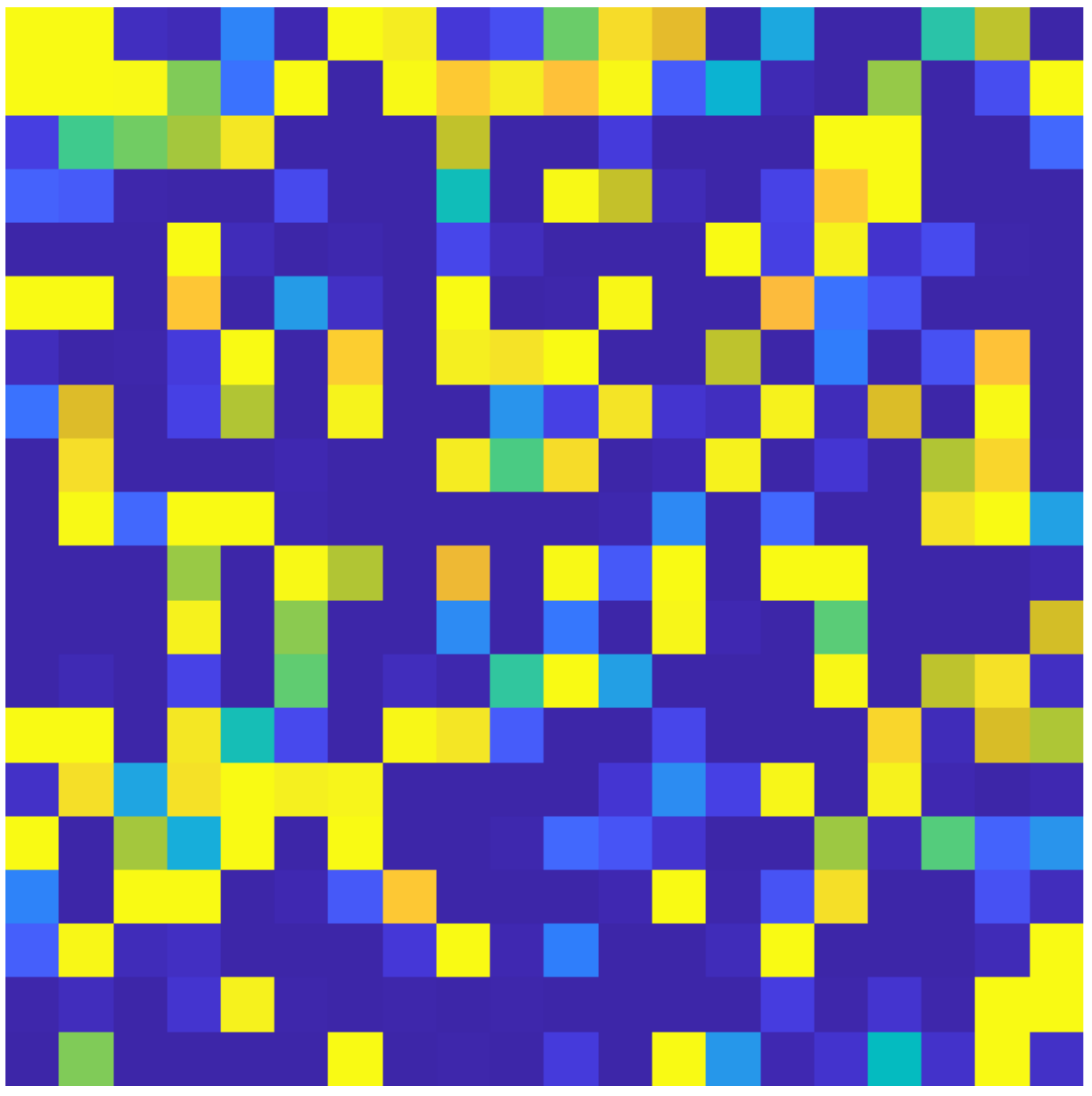}\hspace{.3cm}
\includegraphics[width=.15\linewidth]{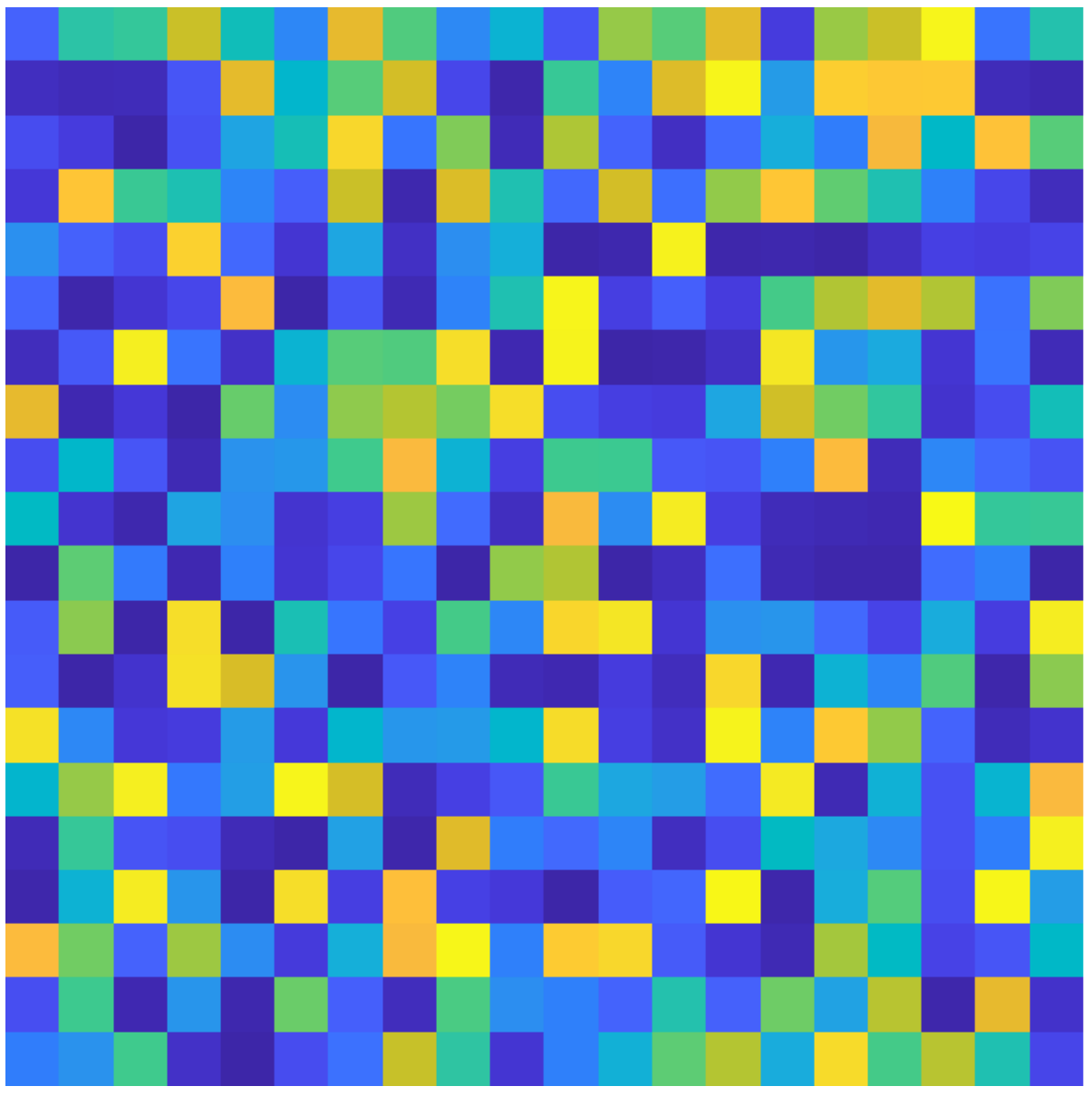}\hspace{.3cm}
\includegraphics[width=.15\linewidth]{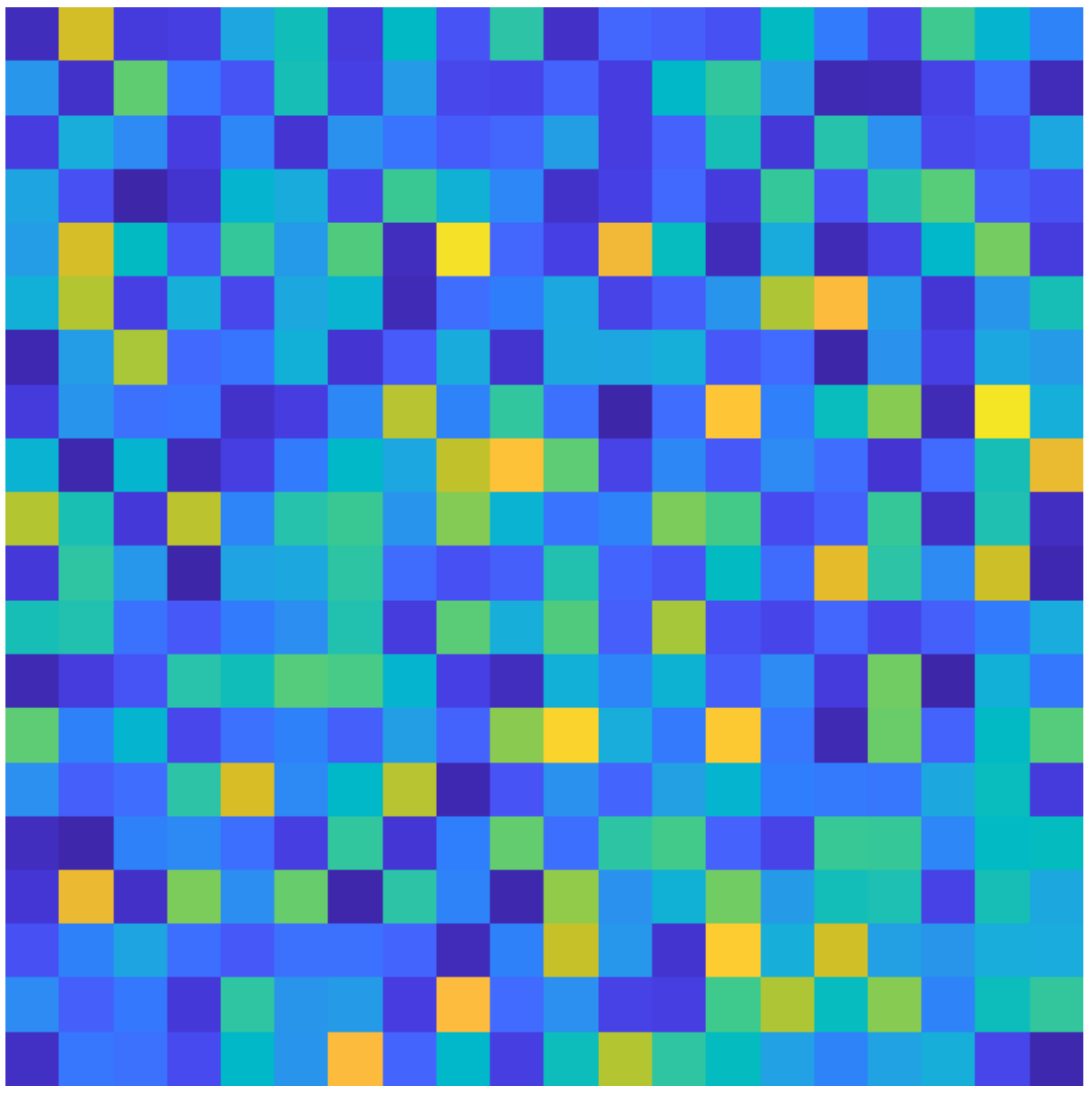}\hspace{.3cm}
\includegraphics[width=.15\linewidth]{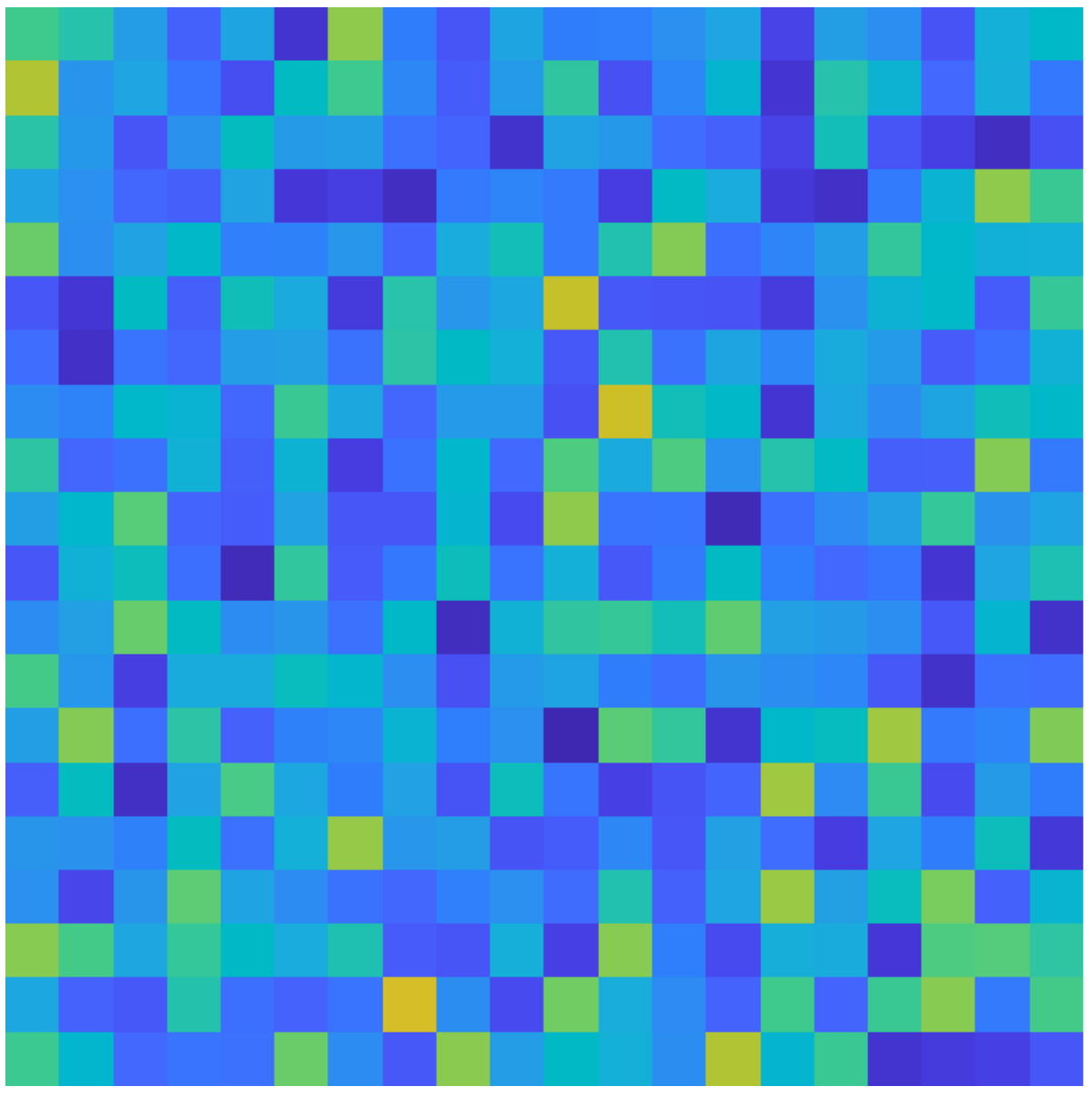}\hspace{.3cm}
\includegraphics[width=.15\linewidth]{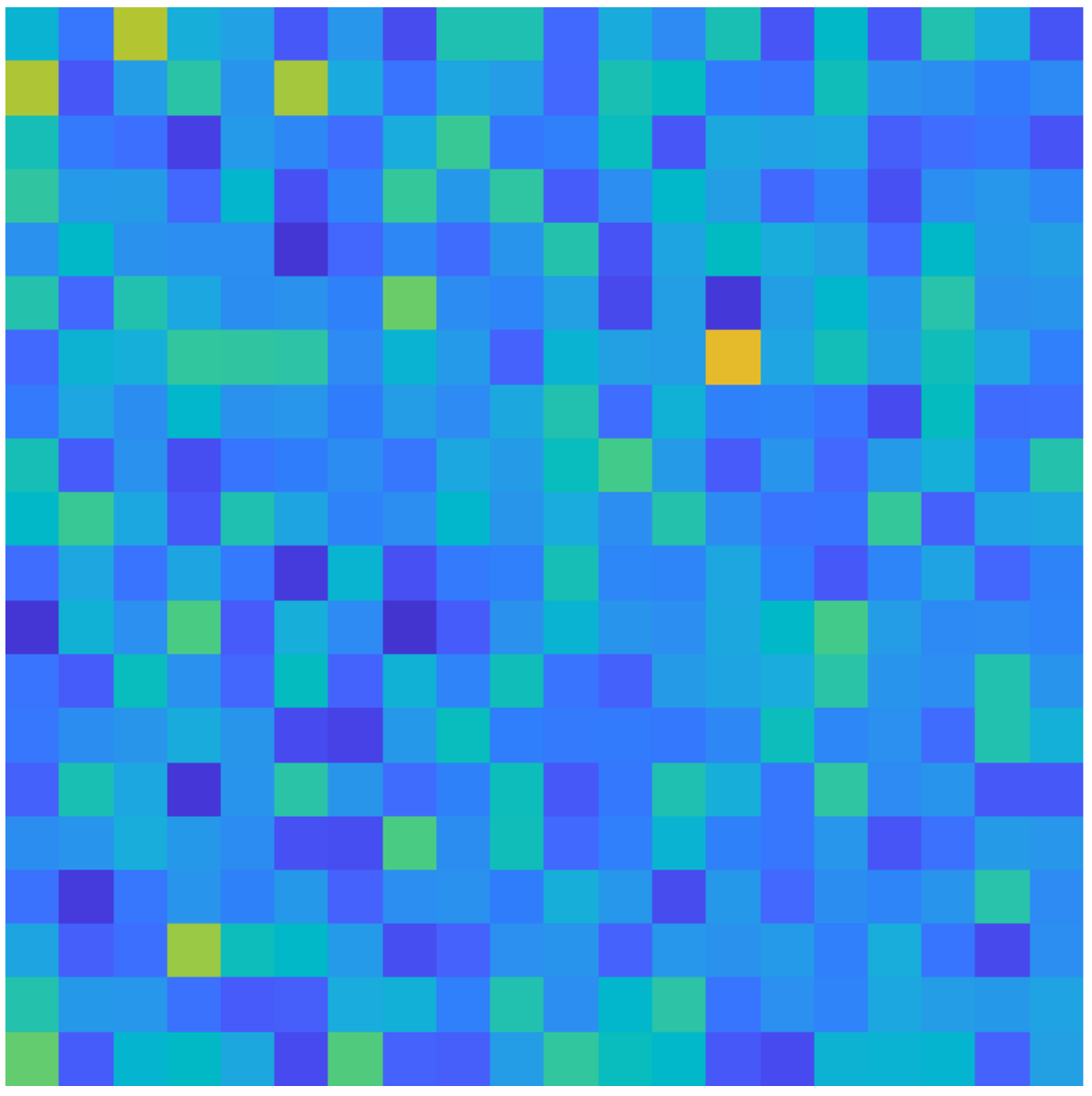}
\begin{turn}{90} \includegraphics[width=.13\linewidth]{./figs/colorbarMaps_c.pdf}\end{turn}
\\
\begin{turn}{90}  3rd compartment \end{turn}
\includegraphics[width=.15\linewidth]{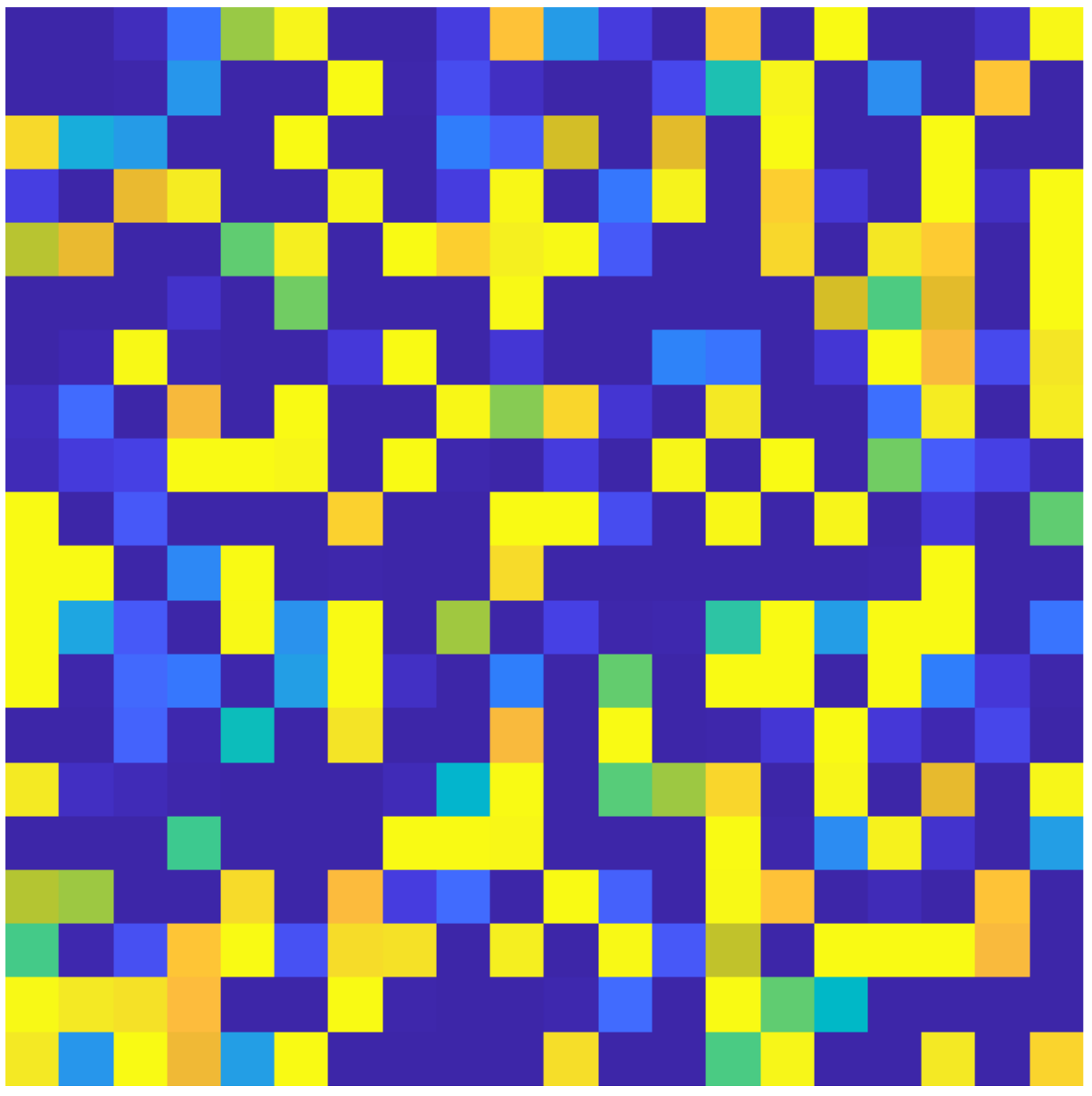}\hspace{.3cm}
\includegraphics[width=.15\linewidth]{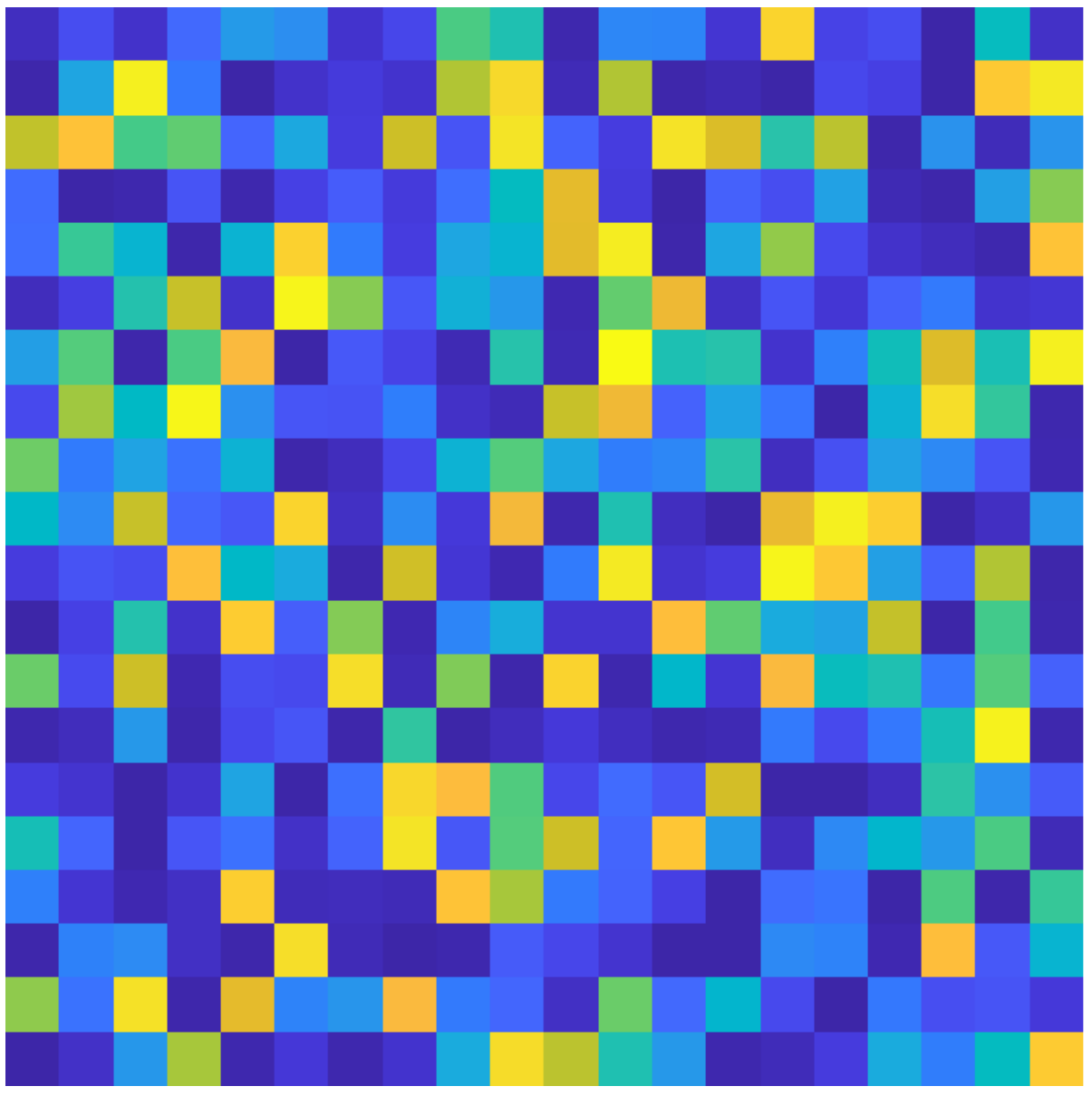}\hspace{.3cm}
\includegraphics[width=.15\linewidth]{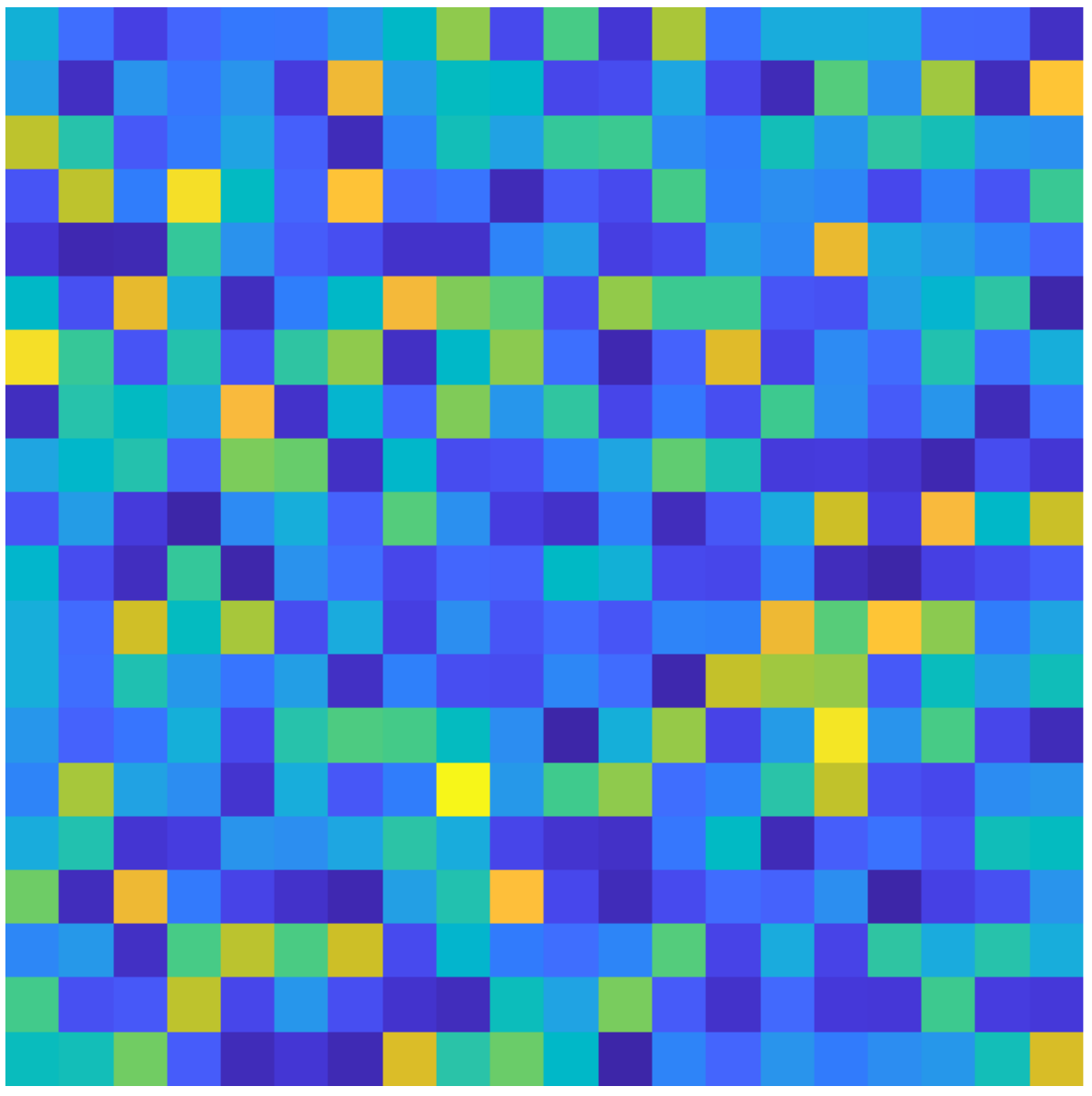}\hspace{.3cm}
\includegraphics[width=.15\linewidth]{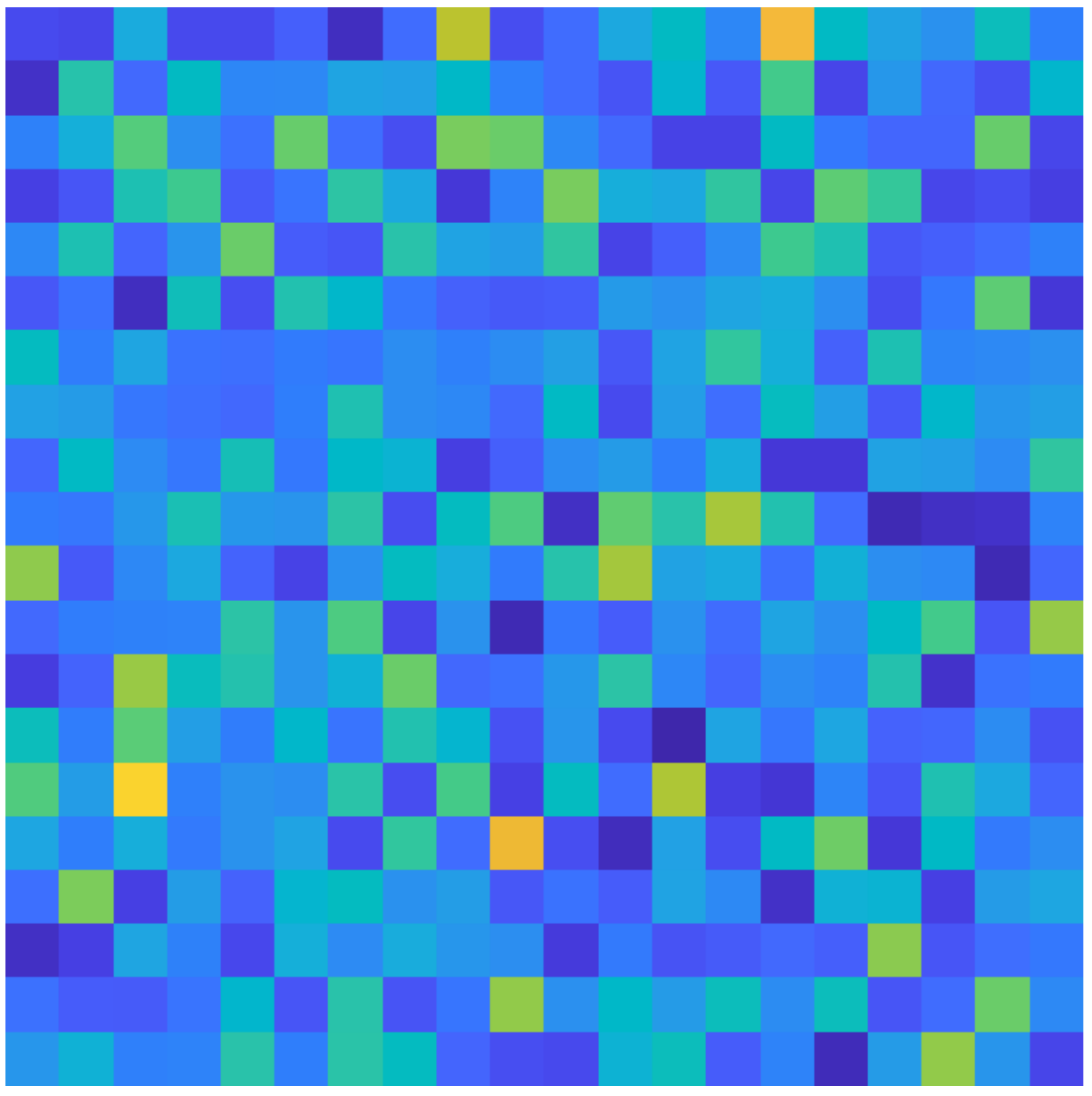}\hspace{.3cm}
\includegraphics[width=.15\linewidth]{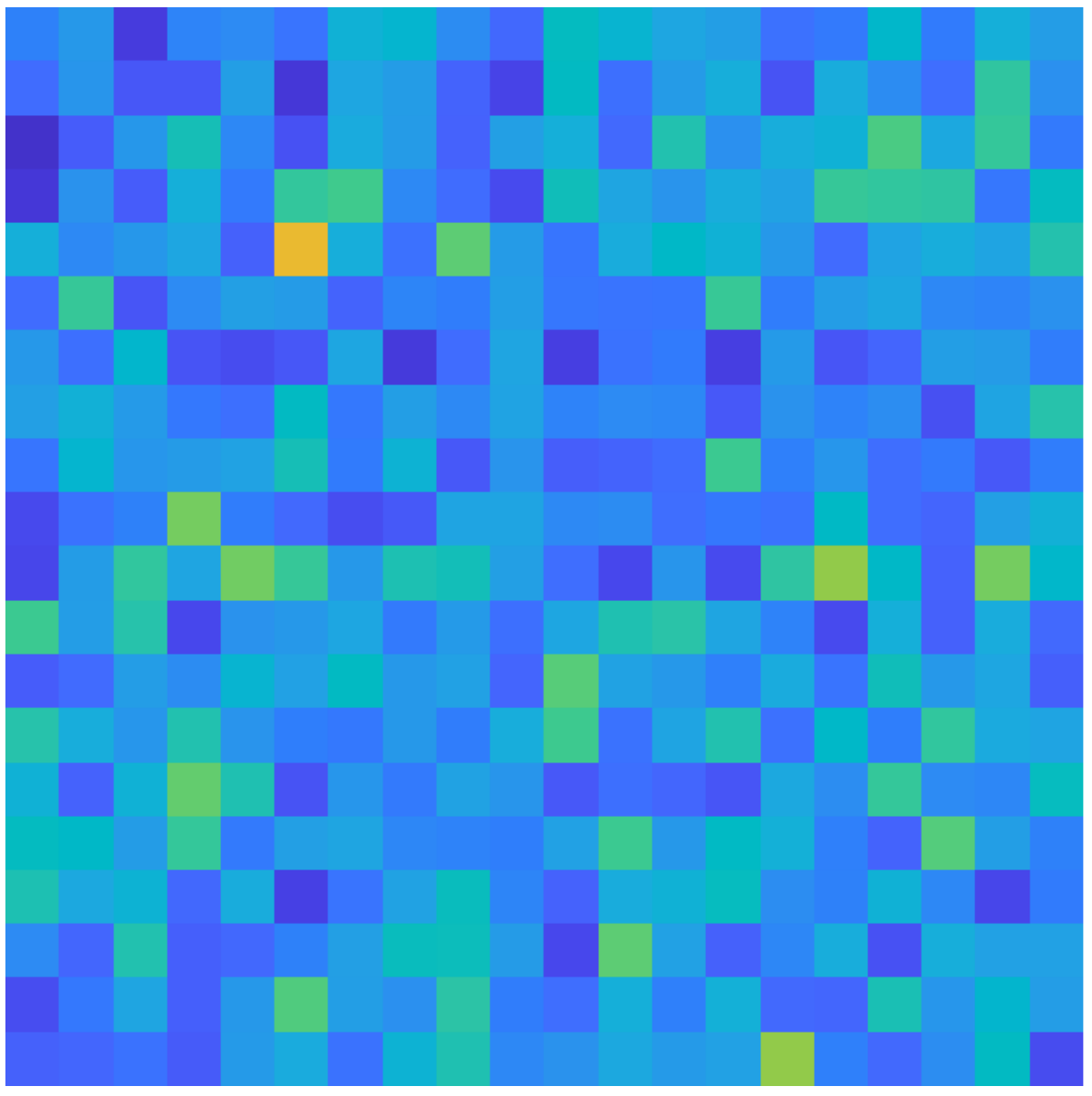}
\begin{turn}{90} \includegraphics[width=.13\linewidth]{./figs/colorbarMaps_c.pdf}\end{turn}
\\
$a=0.1$ \hspace{2cm}$a=0.5$ \hspace{2cm}$a=1$ \hspace{2cm}$a=2$ \hspace{2cm}$a=4$ 
\\
\vspace{.2cm}
(b) Mixture maps of the three compartments. 
\caption{Exemplar mixture weights used for simulating the $20\times20$ pixel phantom from three compartments: (a) distribution of the pixels' mixture weights, (b) per-compartment mixture maps. Mixture weights are drawn at random from dirichlet distributions with different parameters $a$: large $a$ values result in highly mixed pixels that are more difficult to separate, whereas small $a$ values lead to more pixel-pure mixture maps that are easier to demix. 
}\label{fig:diri_maps} 
\end{minipage}}
\end{figure*}

\newpage

\section{Derivation of Algorithm \ref{alg:FW}}
\label{sec:deriv}

We follow the presentation of \cite{denoyelle2019sliding} where the sliding Frank-Wolfe was presented for scalar-valued measures.
The Frank-Wolfe algorithm seeks to solve minimisation problems of the form 
$$
\min_{m\in \Kk} F(m)
$$
where $f:\Xx\to \RR$ is a continuously differentiable function defined on some Banach space $\Xx$, and $\Kk \subseteq \Xx$ is a bounded convex  set. To  derive our algorithm, we follow \cite{denoyelle2019sliding} by rewriting \eqref{eq:spglasso} as the minimisation of a differentiable function over a bounded convex set. Note that  the optimal measure satisfies
$$
\frac12 \norm{\Phi \meas - X}_F^2 +  \alpha \norm{\meas}_\beta  \leq \norm{X}_F^2/2
$$
Recalling the definition of $\norm{\cdot}_\beta$, we can write $ \alpha \norm{\meas}_\beta  = \lambda_1 \abs{\meas}_1+ \lambda_2 \abs{\meas}_2$ where $\lambda_1 = \alpha (1-\beta)$ and $\lambda_2 = \alpha \beta \sqrt{v}$.
So in particular, $\abs{\meas}_{1} \leq \norm{X}_F^2/\pa{2\lambda_1}$. We can therefore equivalently write this as
\begin{align*}
\min_{\meas\in\Mm_+(\Tspace;\RR^p), t_1,t_2\in\RR} F(t_1,t_2,\meas) \eqdef \lambda_1 t_1 + \lambda_2 t_2 + \frac12 \norm{\Phi\meas - y}^2
\\
\qquad \text{subject to } (t_1,t_2,\meas)\in \Kk
\end{align*}
where $\Kk\eqdef \enscond{(t_1,t_2,\meas)}{\abs{\meas}_1 \leq t_1,  \abs{\meas}_{2}\le t_2,  t_2 \leq t_1 \leq \frac{\norm{X}_F^2}{2\lambda_1}}$.
The differential of $F$ is a bounded linear operator on $\RR\times \RR \times \Mm_+(\Tspace;\RR^p)$. Writing $s \eqdef (t_1,t_2,\meas)$ and $s' \eqdef (t_1',t_2',\meas')$, we have  
$$
[\mathrm{d}F(s)]s' = \lambda_1 t_1' + \lambda_2 t_2' + \int [\Phi^*( \Phi \meas - y)] (\theta) \mathrm{d}\meas'(\theta)
$$
The Frank-Wolfe algorithm is an iterative algorithm consisting of the following two steps: Denoting the $k$th iterate by $x^k$, do
\begin{enumerate}
\item $
s^{k} \in \argmin_{s\in\Kk} \mathrm{d}F(x^k)(s)
$
\item $x^{k+1} = (1-\gamma_k) s^{k} +\gamma_k x_k$
\end{enumerate}
For the first step, we can restrict the minimisation to the the extremal points of the convex set $\Kk$, which are of the form $(\norm{M}_1, \norm{M}_2, M \delta_\theta)$ where $\theta\in \Tspace$ and $M\in \RR_+^v$. Therefore, writing $x^k = (t_1,t_2,\meas^k)$, we are led to solve
\begin{equation}\label{eq:tosolve}
(\theta,M) \in \argmin_{\theta\in\Tspace, M\in \RR^p_+}   \lambda_1 \norm{M}_1 + \lambda_2 \norm{M}_2 - \dotp{M}{\eta^k(\theta) }, 
\end{equation}
where $ \eta^k\eqdef -\Phi^*( \Phi \meas^k - y)$.
Let the optimal points be $M = (m_j)_{j=1}^v$ and $\theta$, then the optimality condition reads
$$
-\eta^k(\theta)_j + \lambda_2 \frac{m_j}{\norm{M}_2} + \lambda_1 = 0, \qquad \text{ if } m_j >0
$$
$$
\eta^k(\theta)_j \in [0, \lambda_1], \qquad \text{ if } m_j =0
$$
i.e. we have 
$$
  \frac{M}{\norm{M}_2} = \frac{1}{\lambda_2} \pa{  \eta^k(\theta) - \lambda_1}_+
$$
Plugging this back into \eqref{eq:tosolve} and simplifying, we are left with
\begin{align}
\theta \in  \argmax_{\theta\in\Tspace} 
     \sum_{i=1}^p \pa{  \eta^k(\theta)_i - \lambda_1}_+^2
\end{align}
The second step of the algorithm can be replaced by any procedure which improves the objective value, which leads to the optimization problems in lines 8 and 9 of Algorithm \ref{alg:FW}.

\end{document}